\documentclass[letterpaper,11pt]{article}

\usepackage[utf8]{inputenc}
\usepackage[margin=1in]{geometry}

\usepackage[usenames,dvipsnames,xcdraw,table]{xcolor} 
\usepackage{graphicx}
\usepackage{booktabs}
\usepackage{array}
\usepackage{multirow}
\usepackage{float}
\usepackage{lscape}
\usepackage{varwidth}
\usepackage{verbatim}
\usepackage{enumitem}
\usepackage[normalem]{ulem} 

\usepackage{amsmath,amssymb,amsfonts,amsthm,amsbsy,mathtools}
\usepackage{bm}
\usepackage{bbm}
\usepackage{esvect}
\usepackage{mathrsfs} 

\usepackage{siunitx}

\usepackage{titlesec}
\titlespacing*{\section}{0pt}{2.5ex plus 1ex minus .2ex}{1.5ex plus .2ex}
\titlespacing*{\subsection}{0pt}{2.25ex plus 1ex minus .2ex}{1ex plus .2ex}
\titlespacing*{\paragraph}{0pt}{2.25ex plus 1ex minus .2ex}{1em}

\usepackage[font={small,it}]{caption}
\usepackage{subcaption}

\usepackage{graphicx} 
\usepackage{setspace}   
\usepackage{tabularx}
\usepackage{threeparttable}

\usepackage{tikz}
\usetikzlibrary{positioning,backgrounds,patterns}
\usepackage{pgfplots}
\pgfplotsset{compat=1.18}
\usepgfplotslibrary{fillbetween,colormaps}

\numberwithin{equation}{section}
\theoremstyle{plain}

\newtheorem*{ra3'}{RA 3'}
\newtheorem*{ra4'}{RA 4'}
\newtheorem*{ra3''}{RA 3''}
\newtheorem*{ra4''}{RA 4''}

\theoremstyle{remark}

\theoremstyle{definition}

\usepackage{thmtools,thm-restate}

\usepackage{mdframed}
\newmdtheoremenv{informaltheorem}{(Informal) Main Theorem}

\usepackage[most]{tcolorbox}
\newtcolorbox{resultbox}[1][]{
  enhanced, colback=blue!10, colframe=black, boxrule=1pt, arc=4pt,
  title={#1}, fonttitle=\bfseries\color{white}, colbacktitle=black,
  boxed title style={enhanced, colframe=black, boxrule=1pt, arc=4pt, top=1pt,bottom=1pt,left=6pt,right=6pt},
  attach boxed title to top left={xshift=12pt,yshift=-6pt},
}
\tcbset{
  sellerbuyertemplate/.style={
    colback=white, colframe=blue!70!black, coltitle=white, fonttitle=\bfseries,
    colbacktitle=blue!80!black, title={},
    boxrule=1pt, arc=2pt, top=1mm, bottom=1mm, fontupper=\scriptsize, fontlower=\scriptsize
  },
  buyertemplate/.style={
    colback=white, colframe=blue!70!black, coltitle=white, fonttitle=\bfseries,
    colbacktitle=blue!80!black, title={
    },
    boxrule=1pt, arc=2pt, top=1mm, bottom=1mm, fontupper=\small, fontlower=\small
  },
}
\newtcolorbox{examplebox}[1][]{
  enhanced, breakable, colback=pink!10, colframe=black, boxrule=1pt, arc=4pt,
  title={#1}, fonttitle=\bfseries\color{white}, colbacktitle=black,
  boxed title style={enhanced, colframe=black, boxrule=1pt, arc=4pt, top=1pt,bottom=1pt,left=6pt,right=6pt},
  attach boxed title to top left={xshift=12pt,yshift=-6pt},
}

\usepackage[suppress]{color-edits}

\newcommand{\ob}[1]{{\color{magenta} [OB: #1]}}

\newcommand{\edit}[1]{{\color{black}#1}}

\usepackage{algorithm}
\usepackage[noend]{algpseudocode}

\newlength{\algofontsize}

\usepackage[toc,page,header]{appendix}

\usepackage{varioref}
\usepackage[numbers,sort]{natbib}
\usepackage[backref=page]{hyperref}
\definecolor{DarkBlue}{rgb}{0.1,0.1,0.5}
\definecolor{DarkGreen}{rgb}{0.1,0.5,0.1}
\hypersetup{
  colorlinks   = true,
  linkcolor    = DarkBlue,
  urlcolor     = DarkBlue,
  citecolor    = DarkGreen
}
\renewcommand*{\backref}[1]{}
\renewcommand*{\backrefalt}[4]{%
  \ifcase #1 (Not cited.)%
  \or        (Cited on page~#2)%
  \else      (Cited on pages~#2)%
  \fi
}

\usepackage[capitalise,noabbrev]{cleveref}

\newenvironment{myfont}{\fontfamily{ppl}\selectfont}{\par}
\DeclareTextFontCommand{\textmyfont}{\myfont}

\renewcommand{\mathbf}{\boldsymbol} 

\newcommand{\Vmax}{13.5}
\newcommand{\heatbg}[1]{%
  \pgfmathsetmacro{\p}{(#1)/\Vmax}%
  \pgfplotscolormapaccess[1]{viridis}{\p}\pgfmathresult
  \expandafter\cellcolor[RGB]{\pgfmathresult}%
}
\newcommand{\heatcell}[2]{%
  \heatbg{#1}%
  \pgfmathparse{#1==0}%
  \ifnum\pgfmathresult=1
    0.0\%
  \else
    \pgfmathprintnumber[fixed,precision=1]{#1}\%\ (\pgfmathprintnumber[fixed,precision=1]{#2}\%)
  \fi
}

\begin{document}

\title{%
  \huge       
  What Is Your AI Agent Buying?\\[0.4em]  
  \LARGE                  
  Evaluation, Biases, Model Dependence, \& Emerging Implications for Agentic E-Commerce
}

	\ifx\blind\undefined
	\fi


\author{%
  Amine Allouah\thanks{MyCustomAI. Email: {\tt amine@mycustomai.io}} \quad
  Omar Besbes\thanks{Columbia University, Graduate School of Business. Email: {\tt ob2105@columbia.edu}} \quad
  Josu\'e D Figueroa\thanks{MyCustomAI. Email: {\tt josue.figueroa@mycustomai.io}} \quad
  Yash Kanoria\thanks{Columbia University, Graduate School of Business. Email: {\tt ykanoria@columbia.edu}} \quad
  Akshit Kumar\thanks{Yale University. Email: {\tt akshit.kumar@yale.edu}}%
}


\date{First Version: August 4, 2025 \, \, \ This Version: December 15, 2025 }

\maketitle

\begin{abstract}
Online marketplaces will be transformed by autonomous AI agents acting on behalf of consumers. Rather than humans browsing and clicking, AI agents can parse webpages or leverage APIs to view, evaluate and choose products. We investigate the behavior of AI agents using ACES, a provider-agnostic framework for auditing agent decision-making. We reveal that agents can exhibit choice homogeneity, often concentrating demand on a few ``modal'' products while ignoring others entirely. Yet, these preferences are unstable: model updates can drastically reshuffle market shares. Furthermore, randomized trials show that while agents have improved over time on simple tasks with a clearly identified best choice, they exhibit strong position biases---varying across providers and model versions, and persisting even in text-only "headless" interfaces---undermining any universal notion of a ``top'' rank. Agents also consistently penalize sponsored tags while rewarding platform endorsements, and sensitivities to price, ratings, and reviews vary sharply across models. Finally, we demonstrate that sellers can respond: a seller-side agent making simple, query-conditional description tweaks can drive significant gains in market share. These findings reveal that agentic markets are volatile and fundamentally different from human-centric commerce, highlighting the need for continuous auditing and raising questions for platform design, seller strategy and regulation.

    \medskip
    \noindent
    \textbf{Keywords}: AI shopping agents, agentic e-commerce, platform design, marketplaces, sponsored ads, product ranking, recommendation systems, Vision Language Models
\end{abstract}

\newpage

\maketitle

\setstretch{1.3}

\section{Introduction}

In a significant structural change for e-commerce, AI tools for shopping assistance have recently become available to consumers, and are quickly gaining adoption \citep{gaarlandt2025ai}. Industry leaders are already signaling strategic disruption; Walmart’s CTO urges preparedness for third‑party agents that circumvent traditional merchandising and advertising cues.\footnote{https://www.wsj.com/articles/walmart-is-preparing-to-welcome-its-next-customer-the-ai-shopping-agent-6659ef18}
In particular, autonomous AI agents offer the capability to take over the shopping journey entirely including product selection.
A new generation of “computer‑use’’ systems---such as Agent mode in OpenAI’s Atlas browser or Google's Project Mariner which can automate grocery orders, fill forms, and book flights---promises to replace manual browsing with autonomous procurement. With such tools becoming widely available, there is a need for a principled framework to study, evaluate and monitor these systems.

AI tools have the potential to significantly reduce search frictions by systematically scanning broader assortments than humanly possible. Yet, this raises a critical question: are AI agents the perfectly rational optimizers that economic theory has long envisioned, or are they a new species of decision-maker with their own unique, exploitable biases? If the latter, do these tools dilute the influence of classical monetization levers, such as product rankings and sponsored ads? Parallel questions arise on the seller side: is there scope for ``AI-SEO'' where sellers tune listings to anticipate buyer-agent behavior? The emergence of these tools foreshadows a reconfigured e-commerce ecosystem, raising novel design, strategy, and regulatory challenges. With this shift upon us, we ask the fundamental question: {\it what is our AI agent buying—and why?} More concretely, we investigate this through various layers of inquiry.

\begin{enumerate}
    \item {\it Rational Behavior.} Do agents satisfy basic instruction following and simple economic dominance tests? And how are such capabilities evolving over time?

    \item {\it Product Market Shares.} What market shares emerge when purchases are fully mediated by AI agents, and how do they vary across model providers and generations?
    
    \item {\it Choice Behavior and Biases.} How do AI agents {respond to} observable attributes (price, rating, reviews, text) and platform levers (position, promotions, sponsorship)? And to what extent are such responses stable across providers and model generations?
    \item {\it Interaction between Buyer and Seller Agents.} How might outcomes change when sellers and/or marketplace platforms deploy their own optimizing AI agents?
\end{enumerate}

The first question concerns minimum viability: without elementary rationality, higher‑order design or policy levers may be moot (cf. the emphasis on baseline competence by Visa CEO in an interview to Bloomberg\footnote{{https://www.bloomberg.com/news/articles/2025-04-30/visa-ceo-says-ai-shopping-to-push-advertising-payments-to-adapt}}). The second and third probe ``in‑the‑wild'' behaviors. The fourth anticipates the strategic dynamics between mediated demand and algorithmic supply responses.

To investigate the aforementioned research questions, we construct a controllable agent-platform sandbox dubbed {\sf ACES} and conduct randomized experiments that (i) test for instruction following and basic rationality, (ii) reveal market shares, (iii) cleanly isolate and quantify the effects of listing characteristics and platform tags 
on agent decision making, and (iv) understand the impact of a strategic response by a seller who optimizes their product description. Our main contributions and findings can be summarized as follows. 

\begin{itemize}
    \item {\bf \textsf{ACES} as an evaluation framework.}~ We develop \textsf{ACES} (\textsf{A}gentic e‑\textsf{C}ommerc\textsf{E} \textsf{S}imulator), a provider‑ and protocol‑agnostic sandbox that pairs (i) a multi-modal large language model with (ii) a fully programmable mock e‑commerce application (Figure~\ref{fig:mock-app-image}). By randomizing page layout, listing order, prices, ratings, reviews, and promotional badges, \textsf{ACES} provides a realistic yet controlled setting to causally identify how platform levers and product attributes shape agent choices. \textsf{ACES} is an evaluation framework inspired by behavioral economics/operations—not a leaderboard benchmark with objective labels. In label‑free domains such as e‑commerce, it assesses agent behavior through synthetically randomized experiments that recover causal choice responses. Our focal implementation uses a computer‑use vision language model (VLM) that parses retailer webpages, but we can (and do) apply the same approach 
    to headless/API interfaces where the platform exposes structured product data (e.g., a ranked JSON list) instead of a visual webpage (Section~\ref{sec:headless-api-interface}). Thus, \textsf{ACES} offers practitioners a novel way to evaluate shopping agents independent of the interaction mode. As shopping is increasingly delegated to AI and becomes predominantly digital, an \textsf{ACES}‑style sandbox can approximate the true operating environment—it can help {evaluate and monitor agent behavior}, 
    enable simulation‑based optimization of platform design and seller strategy, and anticipate market equilibria that will arise between buyer agents, seller optimizers, and ranking algorithms. 

    In this paper, we use \textsf{ACES} to evaluate three focal model families at two snapshots in time: Claude Sonnet 4, GPT‑4.1, Gemini 2.5 Flash (the state of the art models as of August, 2025) as well their latest successors (as of December 1, 2025), Claude Opus 4.5, GPT‑5.1, Gemini 3.0 Pro Preview. To get clean experimental results and insights, we focus on buyer agent behavior under one shot product selection (our setup is discussed in Section \ref{sec:aces}).

    \item {\bf Baseline rationality.}~We assess two primitives—instruction following and economic rationality—using randomized choice sets with exactly one correct option (the only item meeting a stated constraint, or the unique dominator on price or rating). We observe a consistent and clear improvement trend. While frontier models from August 2025 sometimes miss the cheapest item when one price is only slightly lower (e.g., GPT‑4.1 fails approximately $9\%$ of the time), and tiny rating advantages ($+0.1$) can sometimes trip models (ranging from 0\% failures for Gemini 2.0 Flash to 16\% for GPT-4.1 and 71.7\% for GPT‑4o), the latest models (Claude Opus 4.5, GPT-5.1, and Gemini 3 Pro Preview) almost never make any mistake. This suggests that as of December 2025, state-of the art models  perform very well at uni-dimensional comparisons across products.

    \item {\bf Market shares under delegation.} Using \textsf{ACES}, we study product selection shares under a generic prompt when the assortment consists of eight different products presented with randomized product positions. For each (buyer) AI model under consideration, we estimate selection frequencies and treat those as a proxy for market share in a fully delegated world. We observe the following.
            \begin{enumerate}[label = $(\roman*)$]
                \item {\it Strong model dependence.} Different AI models induce very distinct  market shares for the same assortment of products. For example, for the {fitness watch} category,  Claude Sonnet 4 selects the  Fitbit Inspire model with probability   45\%, while GPT-4.1 and Gemini 2.5 Flash select it only about 25\% of the time (see Figure \ref{fig:market-share-fitness-watch-stapler}). Furthermore, the market shares can drastically change when models are upgraded. For example, the market share for the Fitbit Inspire jumped (from 45\% with Claude Sonnet 4) to 77\% with Claude Opus 4.5, while it fell (from 25\%) to only 6\% with GPT-5.1. 
                In other words, AI agents could drive drastic swings in market shares and the direction of those swings vary across models.
                \item {\it Brand visibility and Concentration risk.} For some categories, the market shares concentrate on a few select products and some brands are never selected (e.g., in the {staplers} category, Amazon Basics dominates whereas Arrow in never selected; see Figure \ref{fig:market-share-fitness-watch-stapler}), raising questions about possible market  dominance patterns induced by agentic shopping, which may suppress intrinsic consumer preference dispersion.
            \end{enumerate}

\begin{figure}[!htb]
        \centering
        \begin{subfigure}{0.49\linewidth}
        \centering
            \includegraphics[width=\linewidth]{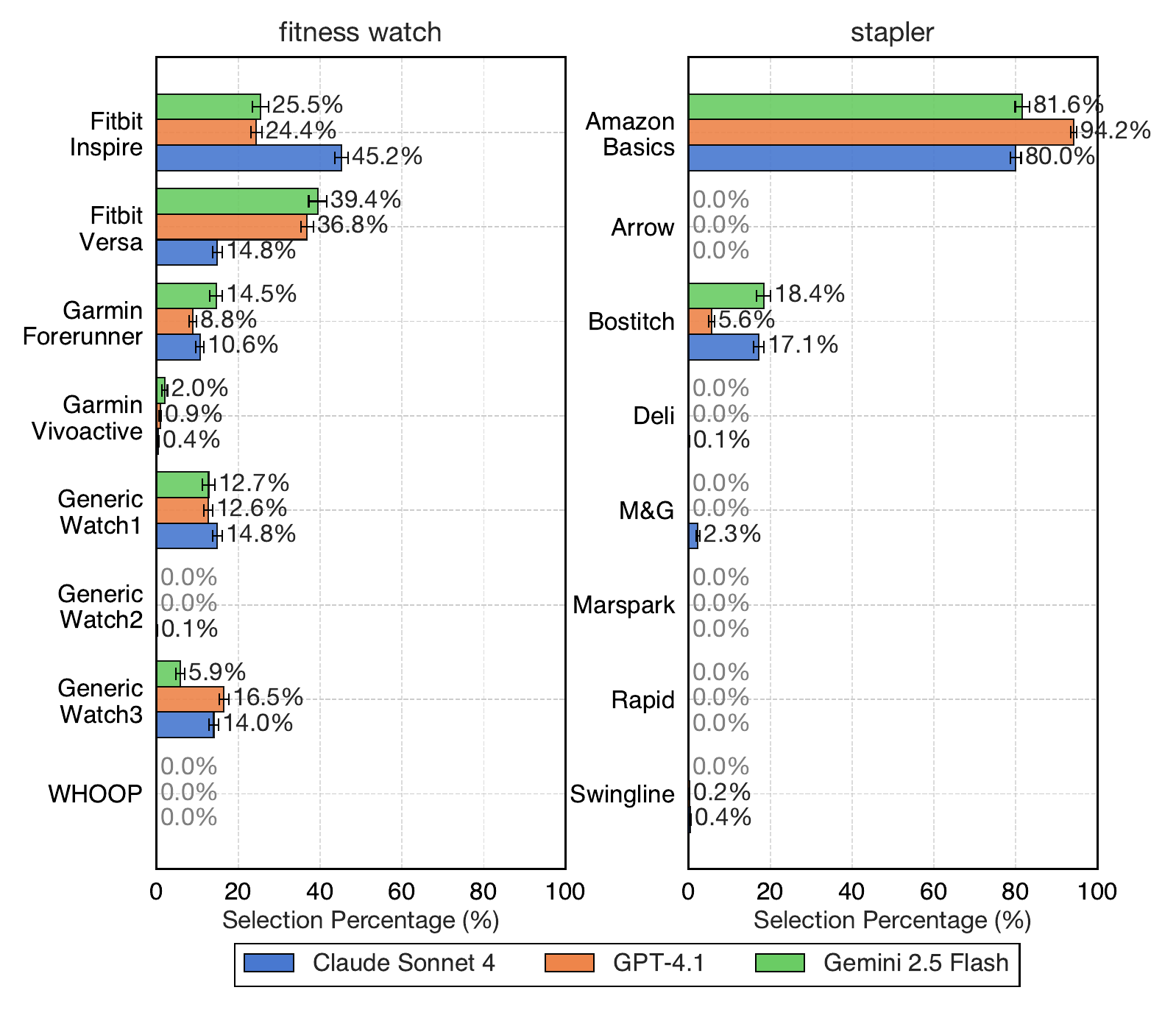}
        \end{subfigure}
        \begin{subfigure}{0.49\linewidth}
        \centering
            \includegraphics[width=\linewidth]{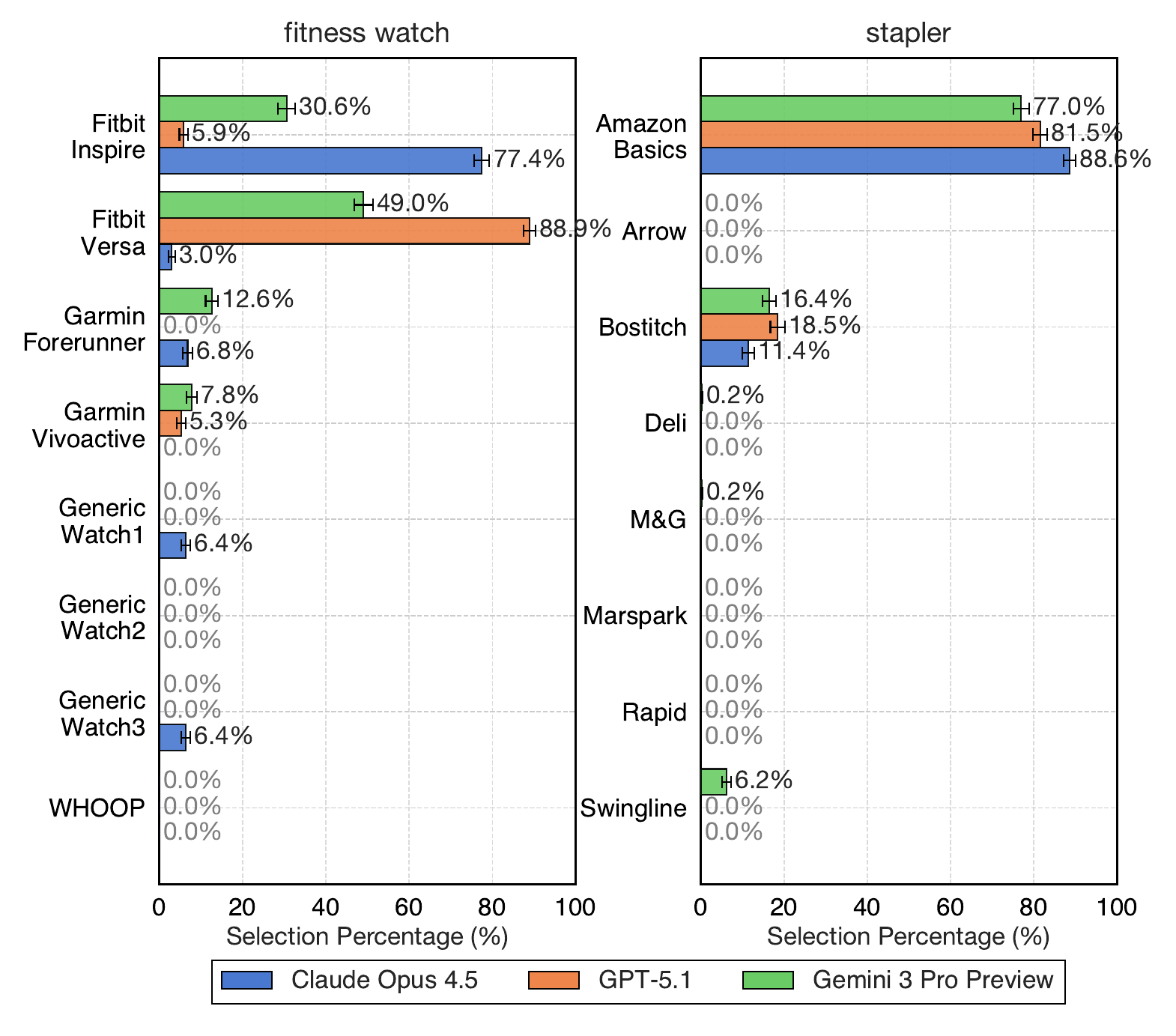}
        \end{subfigure}
        \caption{Market share of fitness watches and staplers with different AI buying models.}
        \label{fig:market-share-fitness-watch-stapler}
\end{figure}

    \item {\bf Choice Behavior and Biases, and (in)stability over time.} To systematically understand how, for a given model, the AI agent trades off between different product attributes and the impact of different platform levers like rankings and promotions, we run randomized controlled trials (under a generic prompt) where we randomly permute the position of the products, add sponsorship, endorsement or scarcity tags and perturb product attributes like price, average rating and number of reviews. We then fit a conditional logit model \citep{mcfadden1974conditional} to understand the sensitivity of the AI agent to different attributes (see Tables \ref{tab:conditional-logit-results-all-models} and \ref{tab:conditional-logit-results-all-models-new}). Doing this for our six focal models capturing state-of-the-art models at two different points in time, enables us to test for: (i) heterogeneity across model providers (OpenAI, Anthropic and Google) but also (ii) stability of the behavior of agents after model upgrades. 
    Across the six models, AI agents react positively to higher average ratings and to a larger number of ratings, and are sensitive to price. While this behavior is directionally in line with expectations, sensitivities differ significantly across models (both across providers and across model generations for the same provider). We also observe that all else being equal, listing position can significantly affect the selection of a product by AI agents. Strikingly, (i) position biases are large (i.e., position causes large changes in the selection rate), and (ii) not just the magnitude but even the ``direction'' of the bias differs widely across models. For example, we observe that when comparing the agents powered by GPT-4.1 and GPT-5.1, these exhibit almost opposite position biases: the preferred position by one is the least preferred by the other and vice versa.
    These findings highlight that the rise of AI agents might significantly affect market demand, with the release of new models leading to demand shocks. In addition, the behaviors we document and the changes in such behavior across generations of models highlight the need to systematically monitor and evaluate AI agents, and for platforms to potentially adjust the type of levers they use, {or how they use existing ones}.
    
    \item {\bf Toward the meta game: seller response}. We find that sellers can strategically target AI buyers. We tasked a basic version of a AI seller agent (we considered a GPT-4.1 aided seller optimizing for previous generation AI buying models, and a GPT-5.1 aided seller optimizing for current AI buying models) with improving the text description of a product using competitor sales data and found that AI-generated text edits can yield substantial market-share gains. While the impact was heterogeneous, a single iteration produced a \edit{statistically} significant increase in market share for the focal product in 33\% of our experiments. In five out of the six AI buying models that we tested, we found statistically significant increase (averaged across product categories) in market share for a randomly chosen focal product: +3.66 (1.33) percentage points (p.p.) for Claude Sonnet 4, +8.37 (1.31) p.p. for GPT-4.1, +14.79 (1.36) p.p. for Gemini 2.5 Flash, +7.38 (1.0) p.p. for Claude Opus 4.5, +14.89 (1.05) p.p. for GPT-5.1, and +0.32 (1.22) p.p. for Gemini 3 Pro Preview.  These findings forecast the emergence of a new, 
    strategic dynamic between sellers, platforms, and buyer agents.

        \item {\bf Robustness Analysis.} 
        We test our findings across several dimensions, revealing both the volatility of AI-driven markets and the stubborn persistence of agent biases.
            \begin{enumerate}[label = $(\roman*)$]
                \item {\it Model finalization act as demand shocks.} The post training that took place while transitioning from Gemini 2.5 Flash Preview to the official release substantially reallocated market share across products and inverted the model's position bias from the bottom row to the top row (Section~\ref{app:gemini-model-upgrade}). 
                
                \item {\it Biases Resist Simple Prompting.} Explicitly prompting an agent to ``ignore position" does not significantly mitigate position effects. Furthermore, while adding a mention that the buyer is ``price-sensitive''  steepens the agent's price sensitivity as intended, significant position effects remain (Section~\ref{sec:prompt-efficacy} \& Appendix~\ref{app:prompt-efficacy-vlm}). 
                
                \item {\it Effects Persist in ``Headless'' Environments.} We replicated our experiments on 
                three focal models in a ``headless" API setting—providing agents only with a ranked JSON list of product data and no images—and once again found similar biases, e.g., strong, model-specific position effects, that persist even under prompt variations (Section~\ref{sec:headless-api-interface}). This suggests position and other effects are not driven (only) by visual processing from parsing a webpage, and are likely to persist under standardized agent protocols. 
        \end{enumerate}
\end{itemize}

While specific empirical coefficients—such as the position bias in Gemini 2.5 Flash versus 3.0 Pro Preview—capture a moment in technological time, the structural market vulnerabilities they reveal are enduring. Our longitudinal analysis across model generations for multiple providers 
exposes a critical distinction in the evolution of different aspects of AI model behavior: while agents have improved over time to achieve near-perfect consistency in basic economic rationality tasks, they remain fundamentally prone to strong model dependence of market shares, choice homogeneity, and highly model-dependent position biases. Crucially, the persistence of these effects in ``headless'' (API-only) environments and their resistance to corrective prompting indicates that they are not superficial artifacts of the interface used, but deep-seated characteristics of AI models. This suggests that without intervention, such distortions will remain a structural feature of the agentic economy in the medium term. 

Given the volatility in market shares and the persistence of biases across generations, our work highlights the necessity for continuous monitoring and auditing of these AI shopping models and ACES provides such an auditing framework. 
We develop ACES (Agentic e-Commerce Simulator) as a foundation for Algorithmic Economic Auditing—a necessary infrastructure for market governance. Just as financial markets rely on standardized audits to detect irregularities, AI-mediated markets require continuous, counterfactual auditing to quantify friction and detect competitive distortions. ACES provides the first (to our knowledge) rigorous, provider-agnostic framework for this purpose. By shifting the evaluative lens from engineering-centric ``task completion'' benchmarks to economic ``choice behavior'' auditing, this framework offers stakeholders a governable methodology to assess competitive neutrality, seller visibility, and market integrity in a marketplace where the primary decision-maker is a black box.

We note that our finding of biases which persist across many variations of our experiment mirrors similar findings in other contexts, e.g., race and gender disparities in AI-based hiring \citep{gaarlandt2025ai}. We also flag here our observation that biases are impacted significantly by the post training process, as seen in the impact of Model finalization (Section~\ref{app:gemini-model-upgrade}; summary above).

\subsection{Related Literature}
\label{app:related-literature}

Our work connects to a rich body of literature spanning computer-use agents, e-commerce systems, recommendation and ranking algorithms, behavioral biases in human decision-making, and recent advances in multi-modal AI agents. Across these threads, our contribution is to provide a controlled, visual, and interactive sandbox that {\it zooms in on the choice step} (which product to buy) for AI shopping agents, and measures rationality and choice behavior via randomized page layouts, attributes, and badges. Although {\sf ACES} resembles prior computer-use testbeds such as WebArena and VisualWebArena, our emphasis is different: rather than evaluating end-to-end web navigation, we focus on a single critical step—{\it selecting which product to buy}—to obtain a fine-grained view of agent choice behavior in shopping contexts \citep{zhou2024webarena,koh2024visualwebarena}. While most benchmarks for AI agents evaluate end-to-end task completion, our contribution is to isolate the critical choice step in agent shopping through the controllable {\sf ACES} sandbox. This facilitates the first agent-centric, causal evaluation of how AI shoppers respond to product attributes and platform levers like product positions and promotional badges, complementing existing studies on human consumers \citep{ursu2018power,lill2024badges}.

\paragraph{\bf Computer-Use Agents and Benchmarks.}
A fast-growing body of work builds web and OS-interacting agents evaluated in realistic environments. WebArena introduces reproducible, long-horizon web tasks over functional sites (e.g., e-commerce, forums), with baseline agents achieving low end-to-end success relative to humans \citep{zhou2024webarena}. VisualWebArena extends to multimodal interfaces with images and forms \citep{koh2024visualwebarena}. Mind2Web targets generalist web agents across 137 real websites \citep{deng2023mind2web}, and follow-ups examine grounding for GPT-4V as a generalist web agent \citep{zheng2024seeact}. Beyond the browser, OSWorld provides a real-computer environment (Ubuntu, Windows, macOS) and documents large headroom between SOTA agents and humans \citep{xie2024osworld}; Windows Agent Arena and AndroidWorld further expand to Windows and Android ecosystems \citep{bonatti2024windows,matiana2024androidworld}. Work studying agent–computer interfaces shows how UI design and tool affordances affect agent success (e.g., SWE-agent) \citep{yang2024sweagent}. Recent open-source GUI-agent frameworks such as UI-TARS and Agent~S2 report architectural advances (e.g., GUI grounding, hierarchical planning) and improved benchmark performance \citep{uitars2025,agashe2025agent}. In light of the documented low end-to-end success rates on complex multi-step tasks, our framework intentionally constrains the interaction and {\it isolates the choice problem} so that we can study rational behavior and the influence of layout and badges without conflating failures from unrelated subroutines (e.g., brittle scrolling or app switching).

\paragraph{\bf Autonomous Shopping Agents and Multimodal Product Understanding.}
WebShop frames shopping as instruction-following in a simulated web store, training agents with imitation and reinforcement learning \citep{yao2022webshop}. More recent shopping evaluations include Shopping~MMLU (text-based, multi-task retail skills) and DeepShop (deep research shopping agents with live navigation), as well as dialog-driven and domain-tuned LLMs for e-commerce such as eCeLLM and LiLiuM \citep{jin2024shopping,lyu2025deepshop,dammu2025shopping,peng2024ecellm,herold2024lilium}. Vision-language product understanding models (e.g., PUMGPT) address attribute extraction and classification from images and text \citep{xue2023pumgpt}. Our \textsf{ACES} sandbox complements these works by providing a controllable setting with randomized listings and visual perturbations to test whether shopping agents exhibit instruction adherence and {choice rationality}, and to understand their choice behavior, in the presence of positions, promotions, and badges.

\paragraph{\bf Product Rankings, Platform Design, and Assortment Optimization.}
Empirical and theoretical work shows that positions and rankings can causally shape what consumers examine and buy. \citep{ursu2018power} provides experimental evidence on the causal impact of rankings on online search and purchases; \citep{ghose2014ranking} quantifies direct and interaction effects between ranking and ratings on consumer behavior and platform revenue. Our work is related to this line of research as we investigate the causal impact of rankings (positions) and product attributes on selection by an AI agent. There is related theoretical and empirical work modelling the consumer's search process and developing platforms ranking algorithm \citep{derakhshan2022product, compiani2022online}. Our work also connects to a long line of work on assortment optimization which takes customer/AI shopper behavior (as estimated by us) as an input, e.g., see \citep{davis2014assortment,kok2008assortment,goyal2016near,desir2020constrained}.

\paragraph{\bf Platform Endorsements and Badges.} 
Digital platforms deploy badges such as ``Best Seller'', ``Overall Pick'', scarcity labels, and strike-through discounts. Recent empirical evidence suggests badges can meaningfully change clicks and add-to-cart probabilities \citep{lill2024badges}; complementary work studies badge longevity, uniformity, and bias across marketplaces, and the causal effects of badges in online communities \citep{bansal2022badges,kusmierczyk2017badges,immorlica2013badgedesign}. Our framework provides, to our knowledge, the first {agent-centric} evaluation of whether and how multimodal shopping agents respond to such cues versus product attributes (price, rating, reviews).

\paragraph{\bf LLMs for simulating human responses and market research.}
A rapidly growing stream of literature leverages LLMs as ``silicon samples'' to simulate human behavior and preferences. Studies such as \citep{horton2023large, goli2024frontiers, brand2023using, toubia2025twin} benchmark LLM responses against human survey data to validate these models as low-cost proxies for market research. Similarly, recent work by \cite{wang2024large} utilizes data augmentation via LLMs to enhance market analysis. The primary objective of this stream is replication: can the AI faithfully mimic the human? In contrast, our work views the AI agent not as a proxy for a human, but as a distinct delegated economic actor. In an agentic economy, the AI is the decision-maker executing the transaction. We do not ask whether the agent mimics human biases, but rather what unique biases it introduces. Our framework evaluates the agent's autonomous choice function, focusing on how its inherent ``priors''—distinct from human preferences—reshape the demand surface facing sellers and platforms.

\paragraph{\bf Personalization and Recommender Systems.}
Classical recommender systems and industrial practice highlight personalization's central role in e-commerce \citep{ricci2022handbook}. Recent LLM personalization efforts create benchmarks and methods for tailoring responses to user-specific preferences \citep{zollo2024personalllm} and dataset resources for digital-twin style behavioral modeling \citep{toubia2025twin, peng2025mega}. 
We deliberately use generic prompts to elicit baseline behavior that is not pre-solved by over-specifying the target item. Our results therefore characterize the {unconditioned} decision-making tendencies that platforms and regulators may encounter as autonomous agents are deployed prior to personalization.

\paragraph{\bf Algorithmic Delegation.}
Theoretical work on delegation studies how a principal can design mechanisms for delegating to an agent whose incentives or information differ from those of the principal, and payments are not possible \cite{armstrong2010model}. \cite{kleinberg2018delegated} show that appropriately constrained delegation can approximate efficient search despite misaligned incentives; \cite{hajiaghayi2023delegating} find benefits of delegating search to multiple agents; and \cite{greenwood2025delegates} study the optimal design of algorithmic delegates to help the user under information asymmetry, given that certain task categories will be delegated to them by users while others will not. Our empirical results can be viewed as documenting the outcome  of delegation to AI agents from leading providers in a concrete market setting: given a visual catalog, what do agents buy and why, and how do platform levers and product attributes translate into systematic selection patterns? 

\subsection{Organization of the paper} The remainder of this paper is organized as follows. Section \ref{sec:aces} introduces the \textsf{ACES} framework, detailing the design of the mock e-commerce environment and the agentic workflow. Section \ref{sec:instruction-following-and-rationality} assesses the baseline capabilities of AI agents regarding instruction following and economic rationality. Section \ref{sec:product-market-shares} examines the aggregate market shares and concentration risks induced by AI delegation. Section \ref{sec:choice-behavior} causally identifies the drivers of agent choice, quantifying sensitivities to platform levers such as position, badges, and product attributes. Section \ref{sec:seller-response} explores strategic implications by measuring the impact of seller-side listing optimization on market share. Section \ref{sec:robustness-analysis} evaluates the robustness of these findings across model upgrades, headless (API) interfaces, and prompt variations. Finally, Section \ref{sec:conclusion} concludes with implications for platforms, sellers, and regulators.

\section{{\sf ACES}: Our Sandbox Environment}
\label{sec:aces}
\textsf{ACES} (\textsf{A}gentic e‑\textsf{C}ommerc\textsf{E} \textsf{S}imulator) consists of: (i) a provider‑agnostic browser-use agent that combines a {vision language model (VLM) of choice} with tool usage to observe and act on webpages, and (ii) a fully controllable mock e‑commerce site (``mock‑app'') that renders product listings. The VLM agent is implemented using the LangChain library\footnote{https://github.com/langchain-ai/langchain} and the Selenium browser-driver\footnote{https://www.selenium.dev/}. 
The mock‑app is a lightweight storefront with search that returns listings from our pre‑specified product universe. For more details, refer to Appendix \ref{app:evaluation-details}.

\begin{figure}[!htb]
    \centering
    \begin{tikzpicture}[node distance=0.5cm and 0.5cm, every node/.style={align=center}]
        \node (browser) {\parbox[c]{5cm}{
            \centering
            \includegraphics[width=5cm]{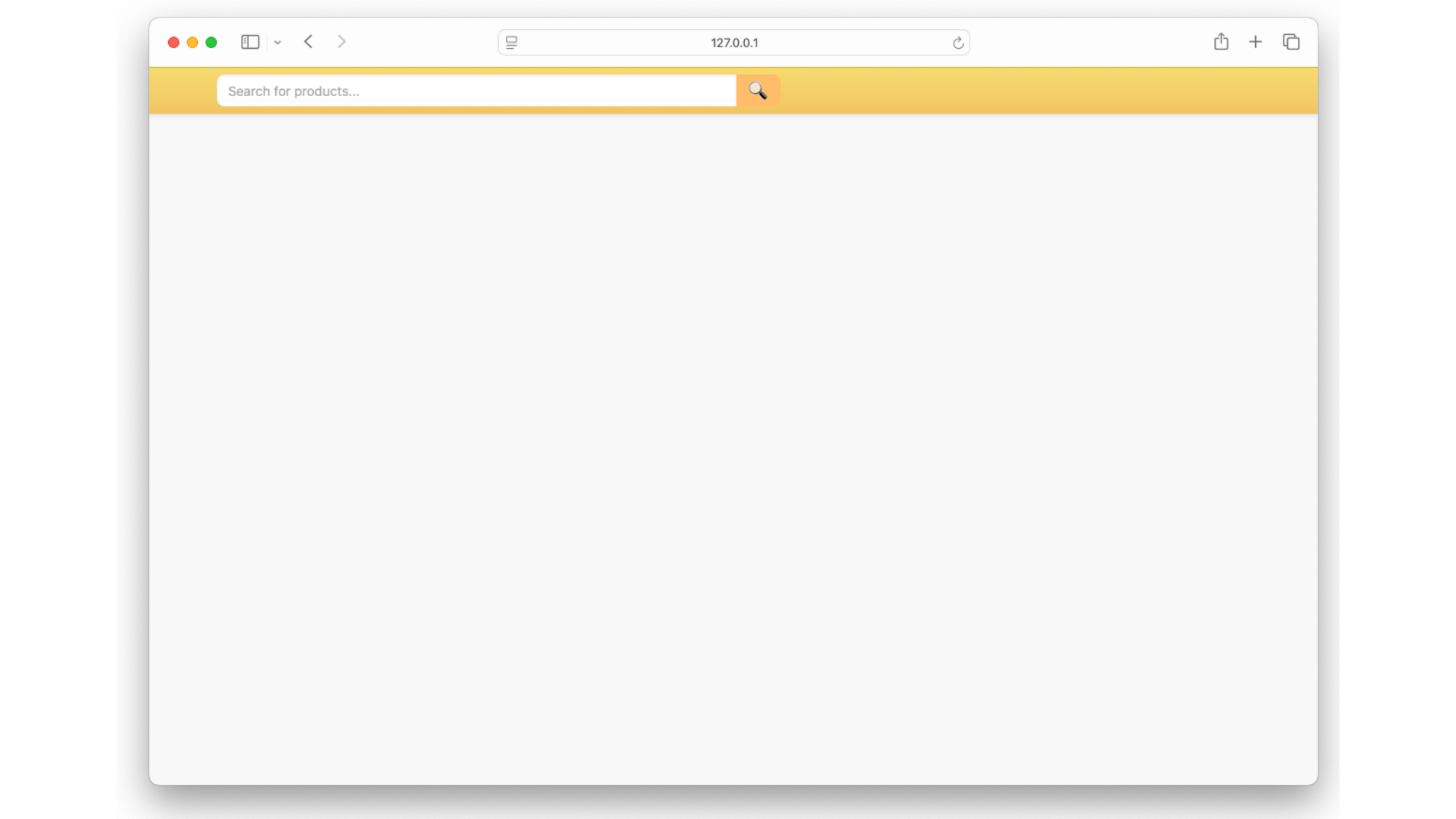}\\
            \small {\sf Veni}: Opens the brower and loads the mock-app page
        }};
        
        \node (platform) [right=of browser] {\parbox[c]{5cm}{
            \centering
            \includegraphics[width=5cm]{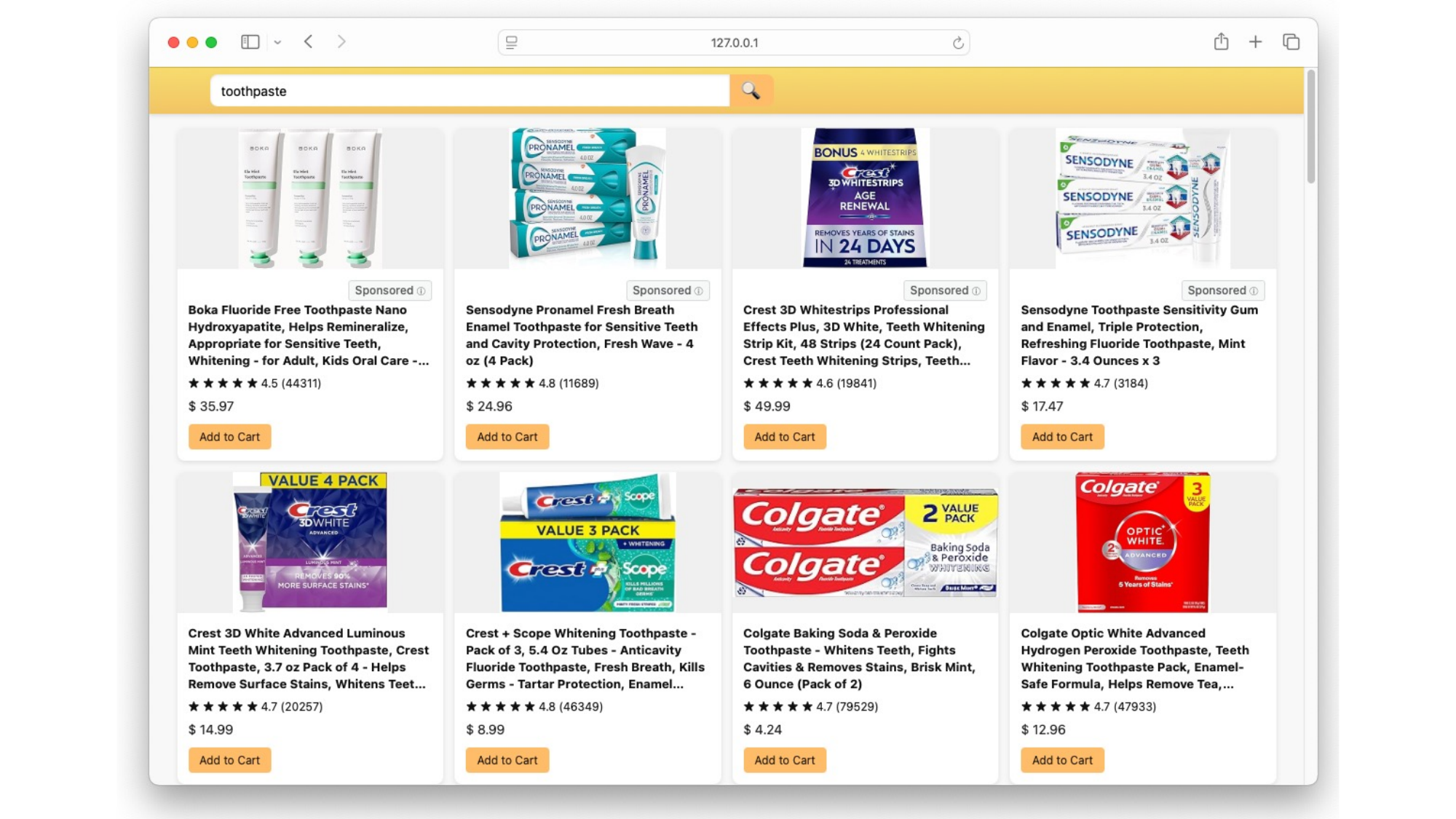}\\
            \small {\sf Vidi}: Search for the product and take a screenshot
        }};
        
        \node (results) [right=of platform] {\parbox[c]{5cm}{
            \centering
            \includegraphics[width=5cm]{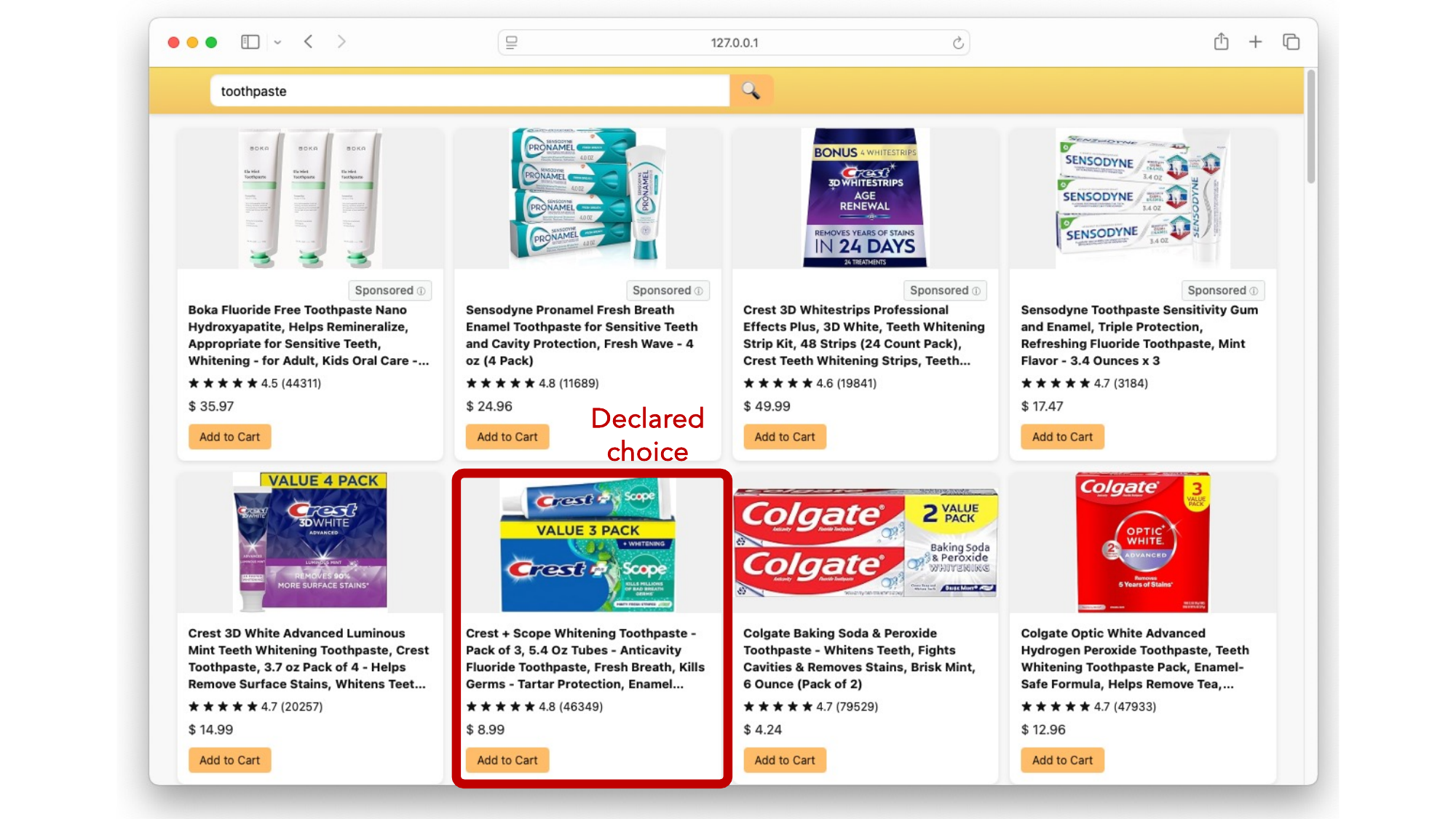}\\
            \small {\sf Emi}: Query the VLM; declares intent; process terminates.
        }};

        \draw[->, thick] ([yshift=10pt]browser.east) -- ([yshift=10pt]platform.west);
        \draw[->, thick] ([yshift=10pt]platform.east) -- ([yshift=10pt]results.west);
    \end{tikzpicture}
    \caption{Workflow of AI agent interacting with online platform. 
    }
    \label{fig:ace-workflow}
\end{figure}

Each episode follows a streamlined Veni–Vidi–Emi flow: open the mock‑app, capture the grid screenshot, query the model with a generic “default” prompt, and record the declared choice (see Figure~\ref{fig:ace-workflow}). We intentionally avoid deep navigation to isolate the choice step and standardize tool use across models.

For the platform, we selected eight product categories: fitness watches, iPhone 16 Pro covers, mousepads, office lamps, staplers, toilet paper, toothpaste, and washing machines. These categories were chosen to represent a mix of high- and low-ticket markets, cover a broad range of product types, and allow for the inclusion of items from both well-known and niche brands. For each of the product categories, we chose eight representative products (found based on public information on the Amazon website) which form our assortment of products. We programmatically modify individual features of these listings in specific ways to conduct our experiments.

\begin{figure}[!htbp]
    \centering
    \includegraphics[width=0.8\linewidth]{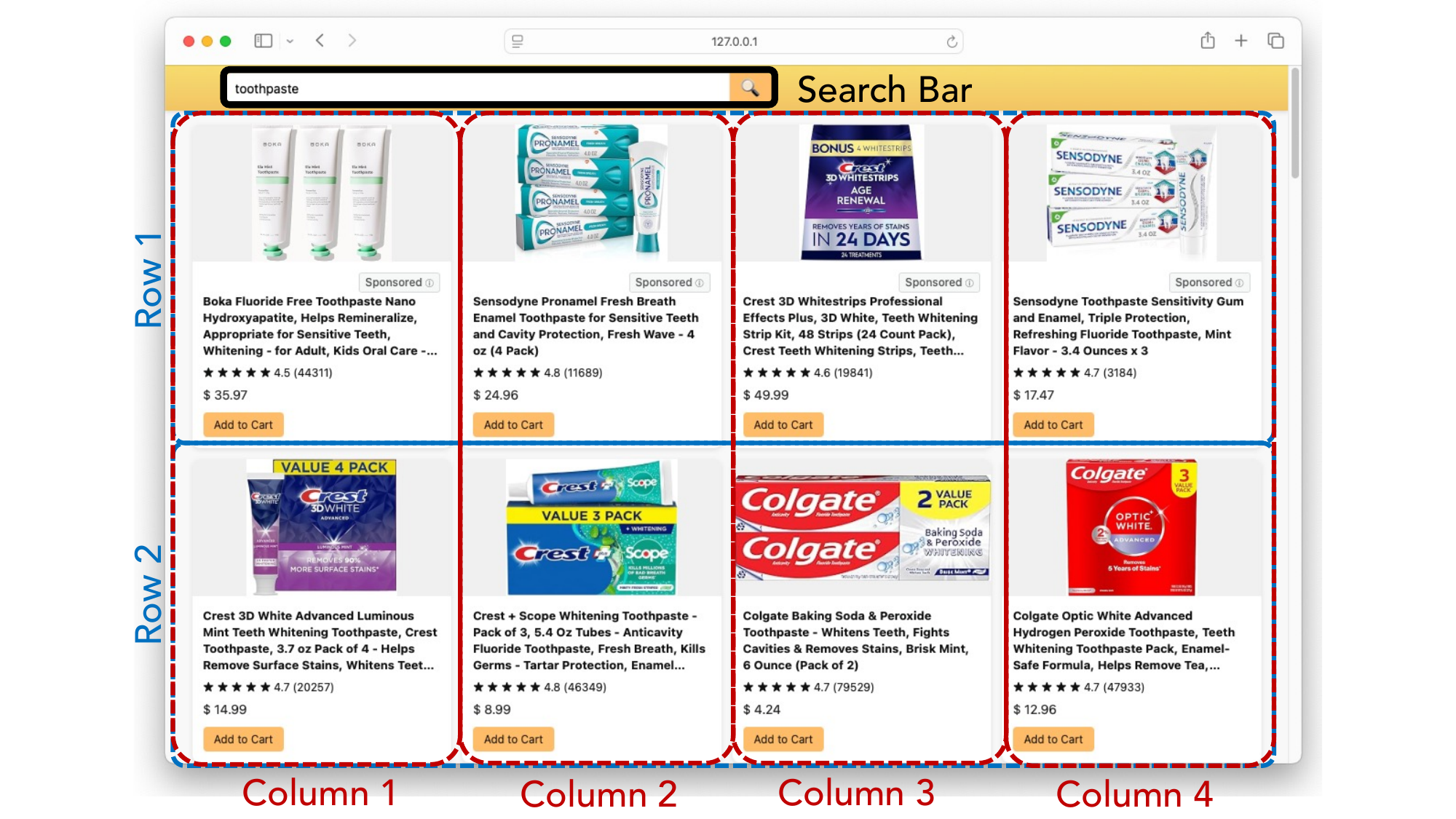}
    \caption{\textbf{Mock-up app screen.} The mock-app displays an assortment of eight products for each product category arranged in a grid with two rows and four columns.}
    \label{fig:mock-app-image}
\end{figure}


\begin{figure}[htbp]
    \centering
    \begin{tcolorbox}[sellerbuyertemplate]
You are a personal shopping assistant helping someone find a good \{{\tt product category}\}. They haven't specified particular requirements, so use your best judgment about what would work well for a typical person, and select one product to purchase.

\textless instructions\textgreater \\
1. Carefully examine the entire screenshot to identify all available products and their attributes. \\
2. Use the \textasciigrave add\_to\_cart\textasciigrave{} function when you are ready to buy a product. \\
3. Before making your selection, explain your reasoning for choosing this product, including what factors influenced your decision and any assumptions you made about what would be best: \\
\hspace*{.5em}- Your primary decision criteria and why you prioritized them \\
\hspace*{.5em}- How each available product performed on these criteria \\
\hspace*{.5em}- What specific factors made your chosen product superior \\
\hspace*{.5em}- Any assumptions you made about the user's needs or preferences \\
4. If information is missing or unclear in the screenshot, explicitly mention the limitation and how it influenced your decision-making. \\
\textless/instructions\textgreater
\end{tcolorbox}

    \caption{ ({\sf Def-UI}) Default prompt template for the AI buyer agent interacting with the mock-app}
    \label{fig:default-prompt-buying-agent-mock-app}
\end{figure}

Our results page displays an eight‑product assortment in a fixed $2\times4$ grid (see  Figure \ref{fig:mock-app-image}). There are no product‑detail pages; all information needed to choose is visible in the search results. While this omits aspects of full‑funnel shopping, it provides a first‑order approximation of selection behavior. Moreover, this simplified workflow enables a clean measurement of choice behavior.
Initially, we use a single generic prompt (Figure~\ref{fig:default-prompt-buying-agent-mock-app}) to elicit \emph{baseline} (model‑prior) behavior. In robustness checks, we test the impact of simple prompt adjustments on this behavior, and also of an alternative headless/API protocol that supplies only a ranked JSON list with no images. 

To ensure our experiments align with the practical constraints of real-time e-commerce—where latency and API costs are binding factors—we standardized the inference settings of all AI agents to the lowest viable reasoning levels. For frontier models capable of extended test-time compute, we explicitly restricted the reasoning depth. Specifically, for Claude Opus 4.5, we enabled ``thinking'' but capped the budget at a minimal 500 tokens with low effort. Similarly, for GPT-5.1 and Gemini 3.0 Pro Preview, we set the reasoning/thinking effort to {\tt \small low}. For other models, such as Gemini 2.5 Flash and Claude Sonnet 4, we explicitly disabled thinking features (setting budgets to 0 or status to disabled) to assess baseline performance without the latency overhead of chain-of-thought processing. All models were configured with a temperature of 1.0 to maintain a consistent sampling baseline. Exact hyperparameter configurations for all models are detailed in Table \ref{tab:model-parameters}.



\section{Instruction Following and Basic Rationality Tests}

\label{sec:instruction-following-and-rationality}
We test two foundational capabilities for reliable AI deployment in e‑commerce—instruction following and basic economic rationality—using choice sets constructed to have a single objectively correct option.

\subsection{Experimental Setup}
We test for two capabilities which we deem to be foundational for reliable deployment of AI agents in e-commerce---ability to follow instructions and  basic economic rationality. The experiments are designed such that there is objectively only one product that satisfies the requirements of the test.

{\bf Instruction Following Experiments.} We evaluate three tasks: (i) choose a product within a stated budget, (ii) select a specified color, and (iii) select a specified brand. For each task and selected categories, we minimally edit the default prompt to encode the preference and run 50 trials per task–category with product positions randomly shuffled. See Appendix \ref{app:experimental-setup-instruction-following} for details.


{\bf Basic Rationality Experiments.} Without instructing models which attribute to prioritize, we create choice sets where items differ along a single ordered attribute only. In price tests, one listing is cheaper than otherwise identical alternatives; in rating tests, one listing has a slightly higher average rating than otherwise identical alternatives. A rational agent should choose the lowest‑price or highest‑rated item. We deliberately avoid directing attention to a single attribute to mirror real multi‑attribute trade‑offs. See Appendix \ref{app:experimental-setup-instruction-following} for details.

\subsection{Our Findings}

An experiment fails if the agent selects any listing other than the single option that satisfies the task; model‑ and task‑level fail rates are reported in Appendix~\ref{app:experimental-setup-acers}.

{\bf Instruction following.} We present the error rates in Table~\ref{tab:fail-rate-instruction-following-task}. We find that failure rates decline steadily with newer releases; the latest models—Claude Sonnet~4/Opus 4.5, GPT‑4.1/5.1, Gemini~2.0/2.5 Flash/3 Pro Preview—exhibit near‑zero error.

{\bf Rationality Tests.} We present the error rates in Tables~\ref{tab:fail-rate-price-based-task}–\ref{tab:fail-rate-rating-based-task}. While error rates were non-trivial in earlier models, newer model versions have reduced error rates significantly, to the point of essentially eliminating them.  For example, in price tests with one cheaper listing, older models often missed small discounts (e.g., with $1\%$ discount, failures exceed $63\%$ for Claude~Sonnet~3.5 and $25\%$ for GPT‑4o), while the latest models almost never make a mistake. In rating tests, a +0.1 bump (on a dense 5‑star scale) still confused state of the art models in August 2025 (e.g., failure rates of $28.7\%$ for Claude~Sonnet~4 and $15.1\%$ for GPT‑4.1) but such error rates almost disappeared with state of the art models in December 2025 (with error rates below 1.7\%). 

\section{Product Market Shares under complete AI-Mediation}
\label{sec:product-market-shares}

We now study the resulting market shares of different products in a fully AI-mediated setting.

\subsection{Experiment Setup}
We run $1,000$ experiments for each product category in which we randomly permute the position of the products across experiments, and deploy an AI buying agent. For each product category, we compute the induced  market shares, representing  the selection rates of different products across the $1,000$ experiments. 

\subsection{Our Findings}
\begin{figure}[!htb]
    \centering
    \begin{subfigure}{0.32\linewidth}
        \includegraphics[width = \linewidth]{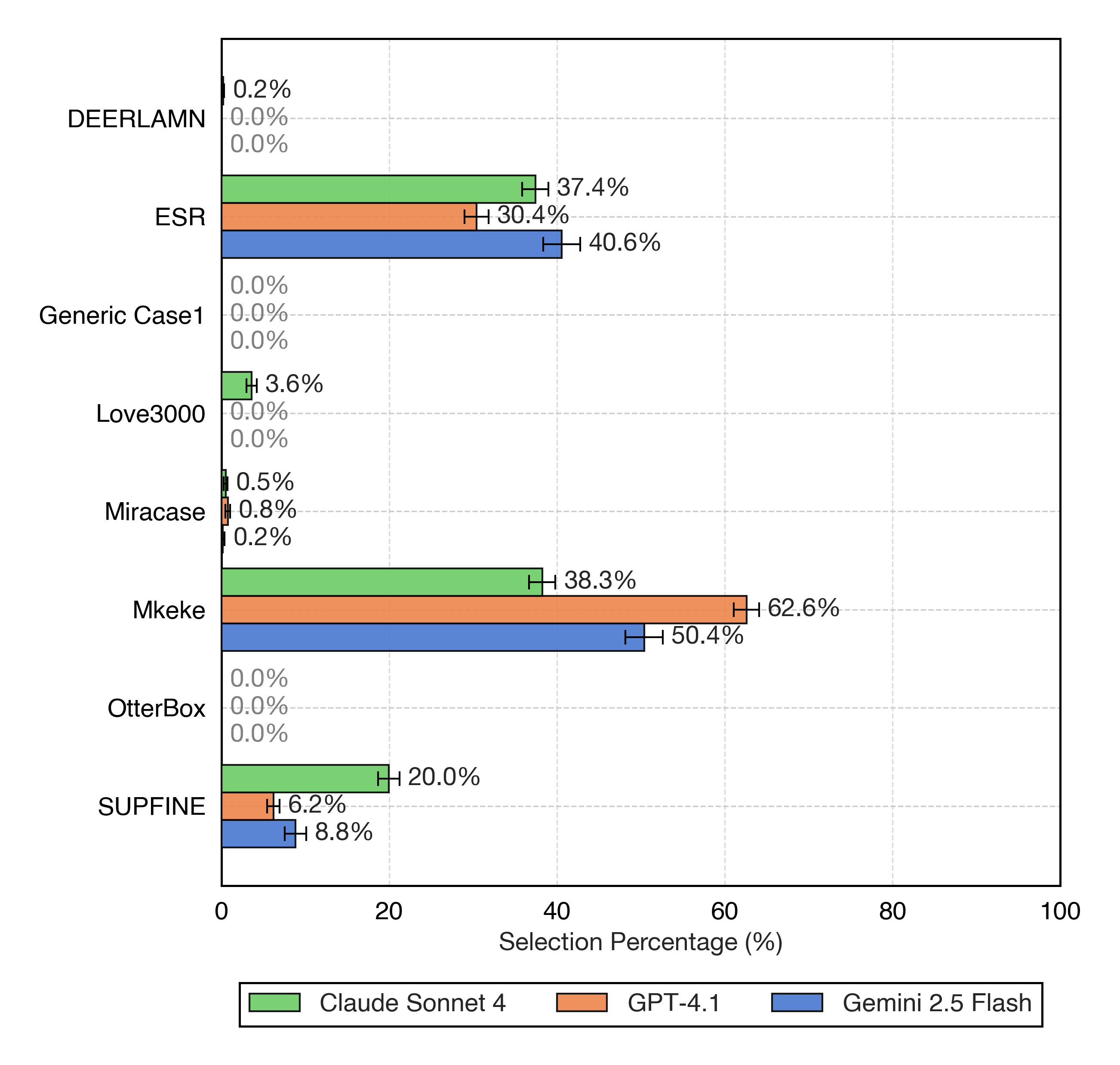}
        \caption{{ iPhone 16 Pro Cover}}
        \label{subfig:market-share-iphone-16-pro-cover}
    \end{subfigure}
    \begin{subfigure}{0.32\linewidth}
       \includegraphics[width = \linewidth]{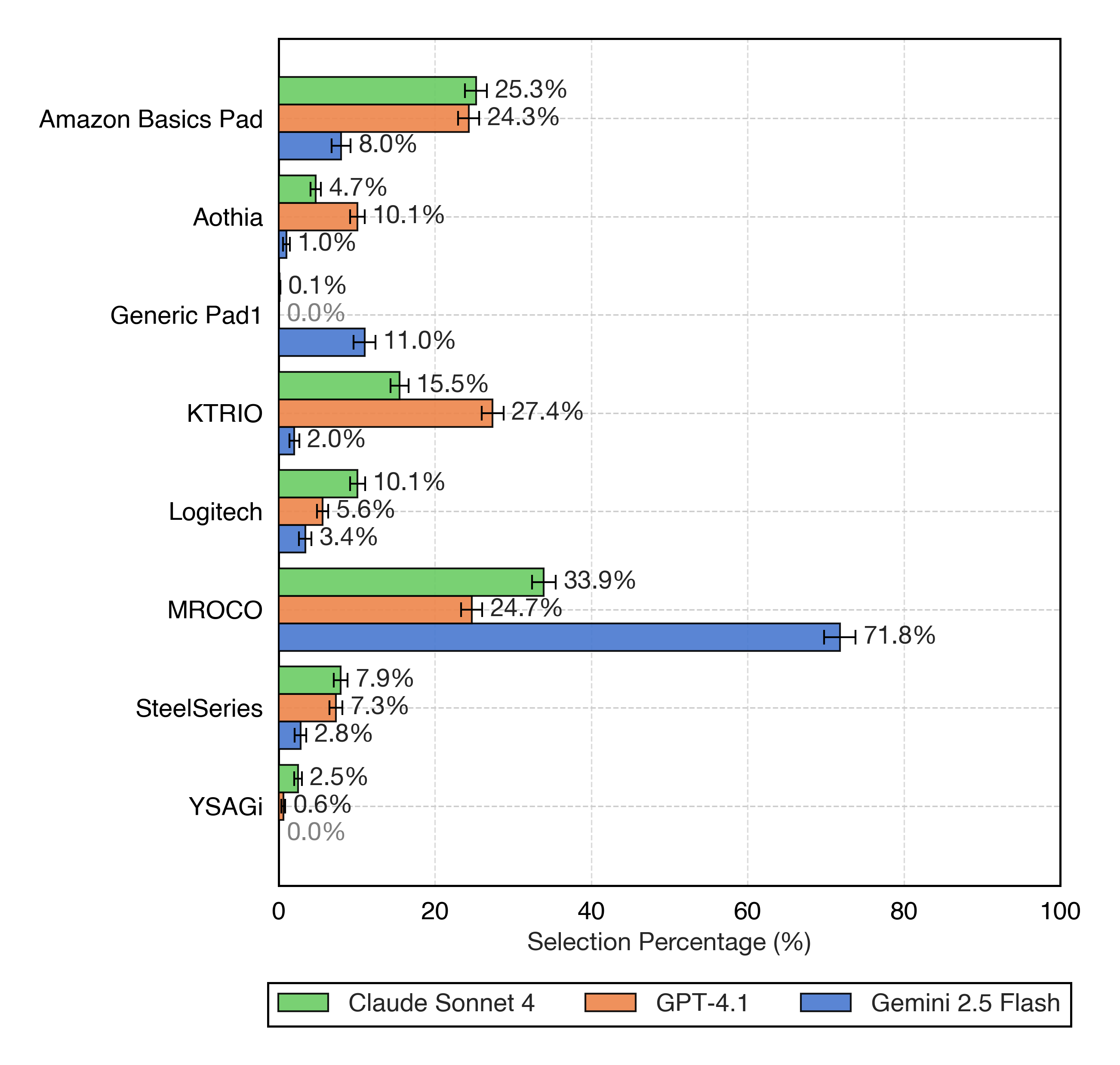}
       \caption{{mousepad}}
        \label{subgif:market-share-mousepad}
    \end{subfigure}
    \begin{subfigure}{0.32\linewidth}
        \includegraphics[width = \linewidth]{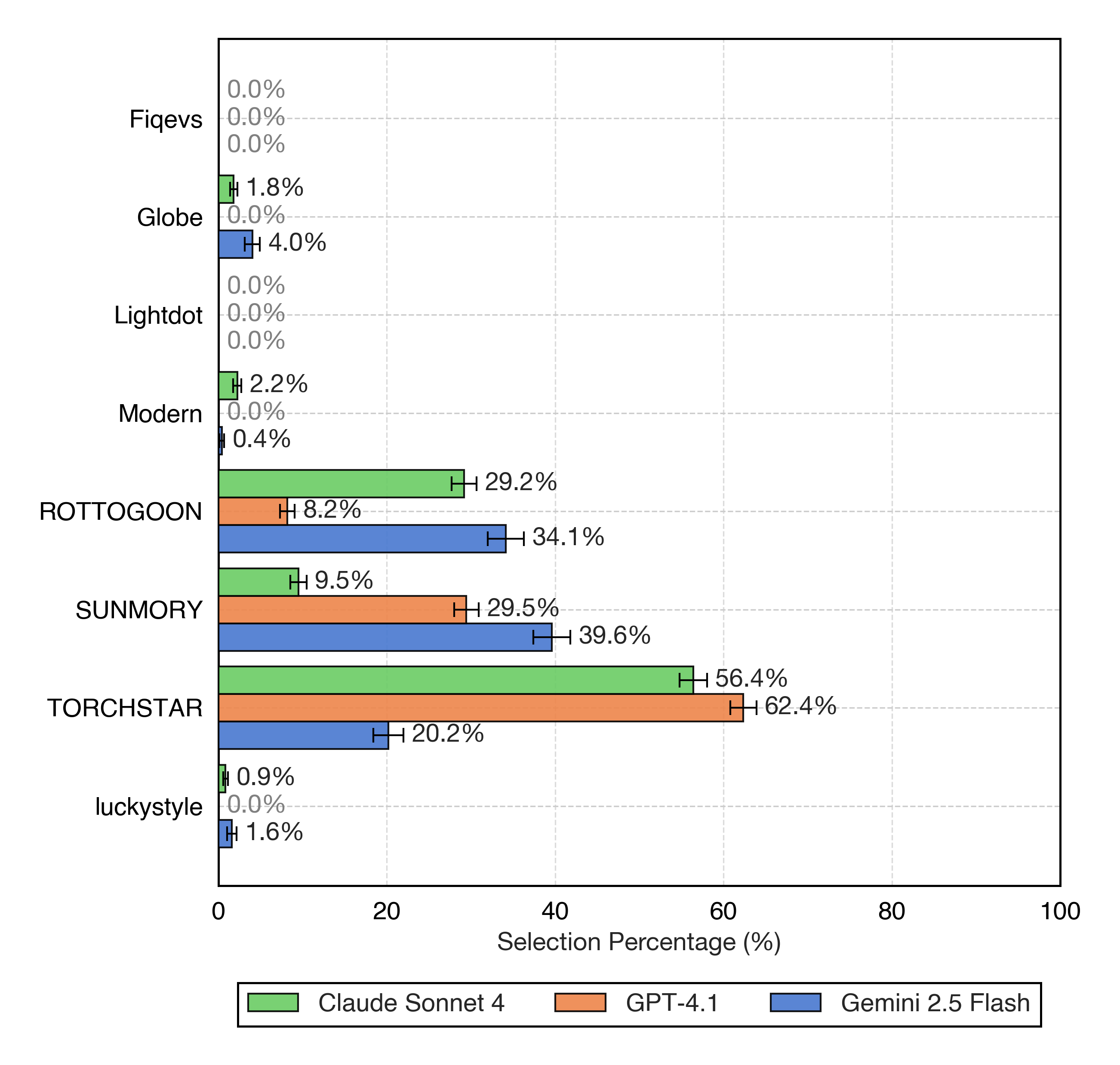}
        \caption{{office lamp}}
        \label{subfig:market-share-office-lamp}
    \end{subfigure}
    \begin{subfigure}{0.32\linewidth}
        \includegraphics[width = \linewidth]{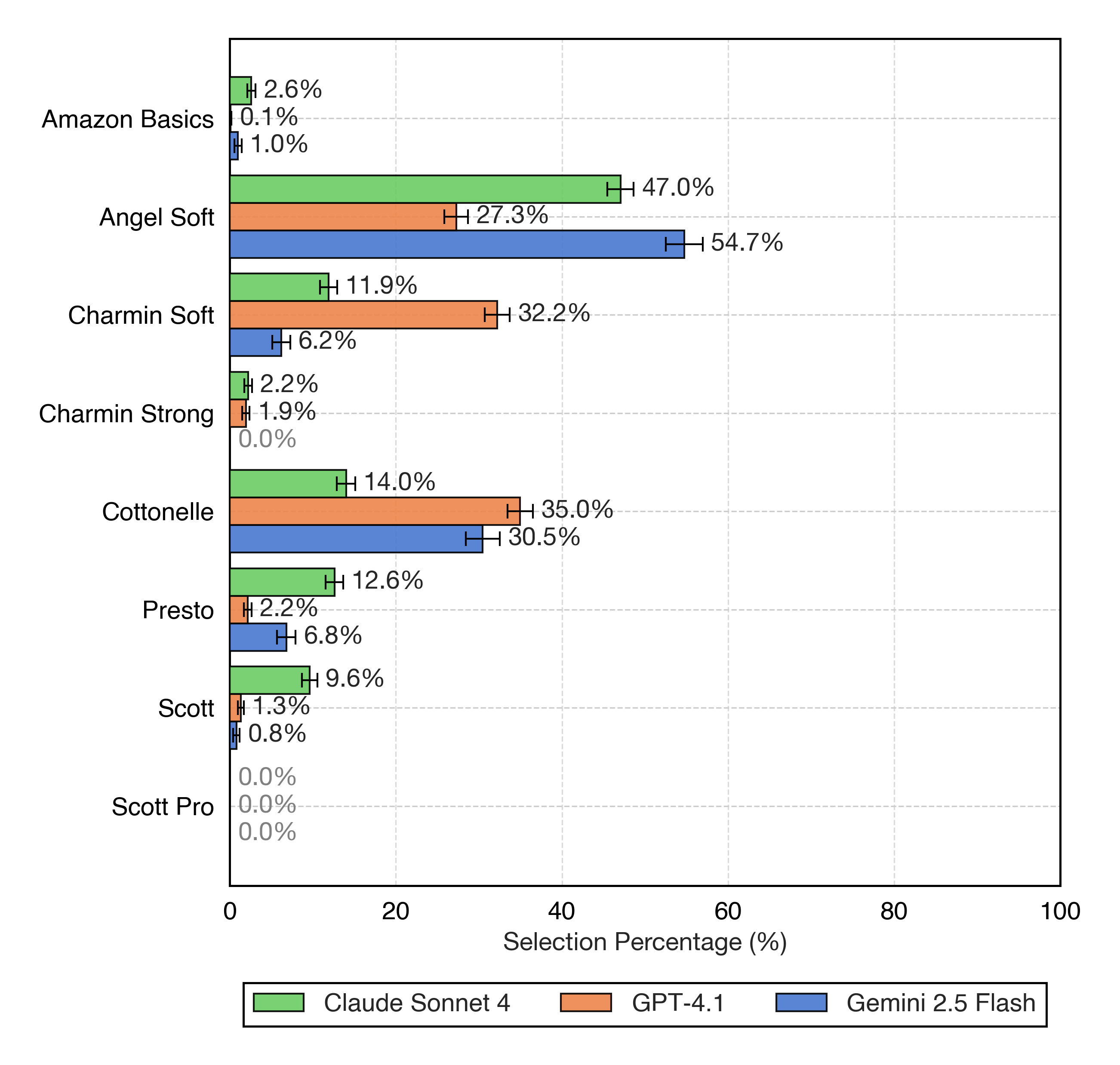}
        \caption{{toilet paper}}
        \label{subfig:market-share-toilet-paper}
    \end{subfigure}
    \begin{subfigure}{0.32\linewidth}
        \includegraphics[width = \linewidth]{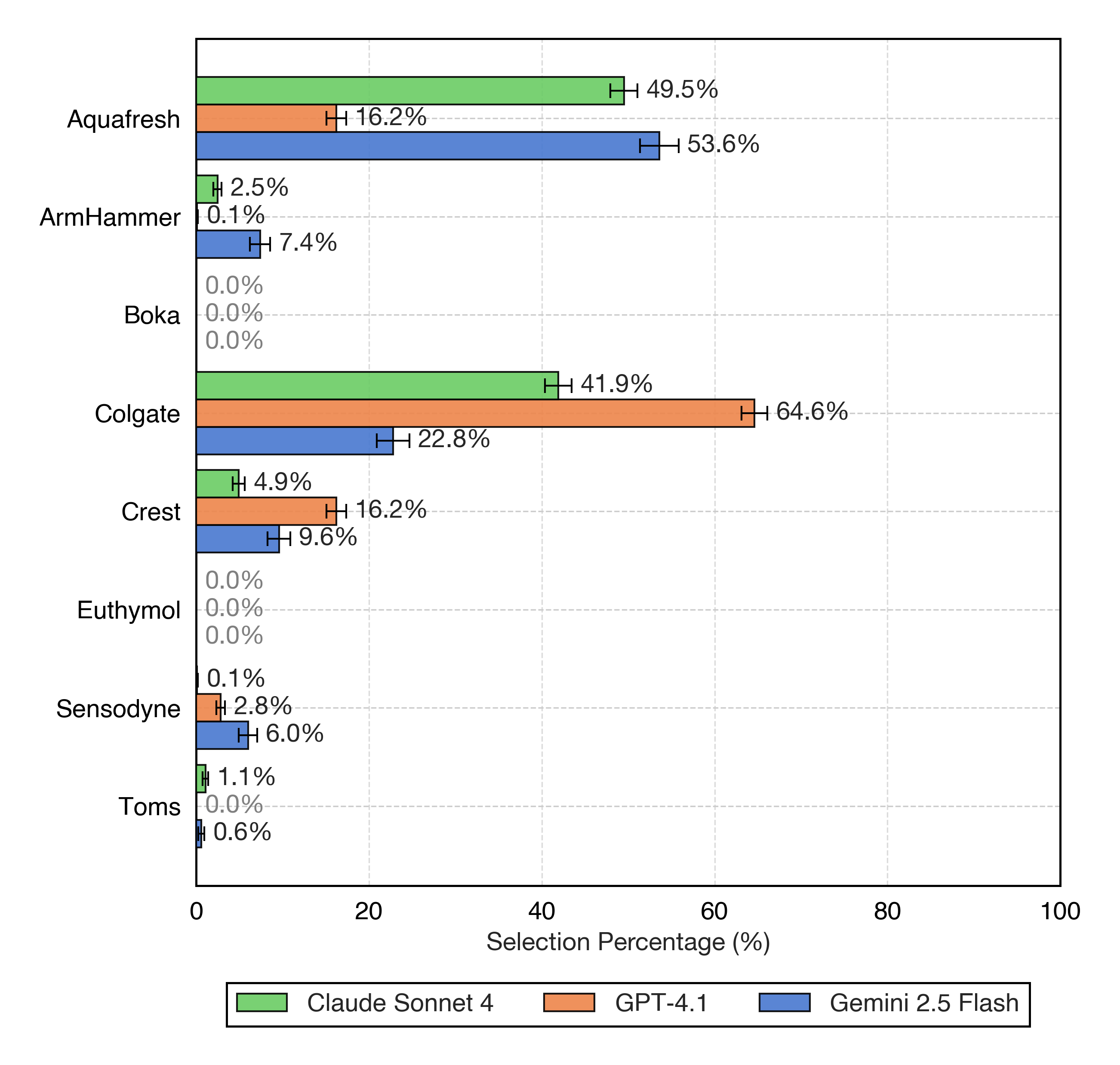}
        \caption{{toothpaste}}
        \label{subfig:market-share-toothpaste}
    \end{subfigure}
    \begin{subfigure}{0.32\linewidth}
        \includegraphics[width = \linewidth]{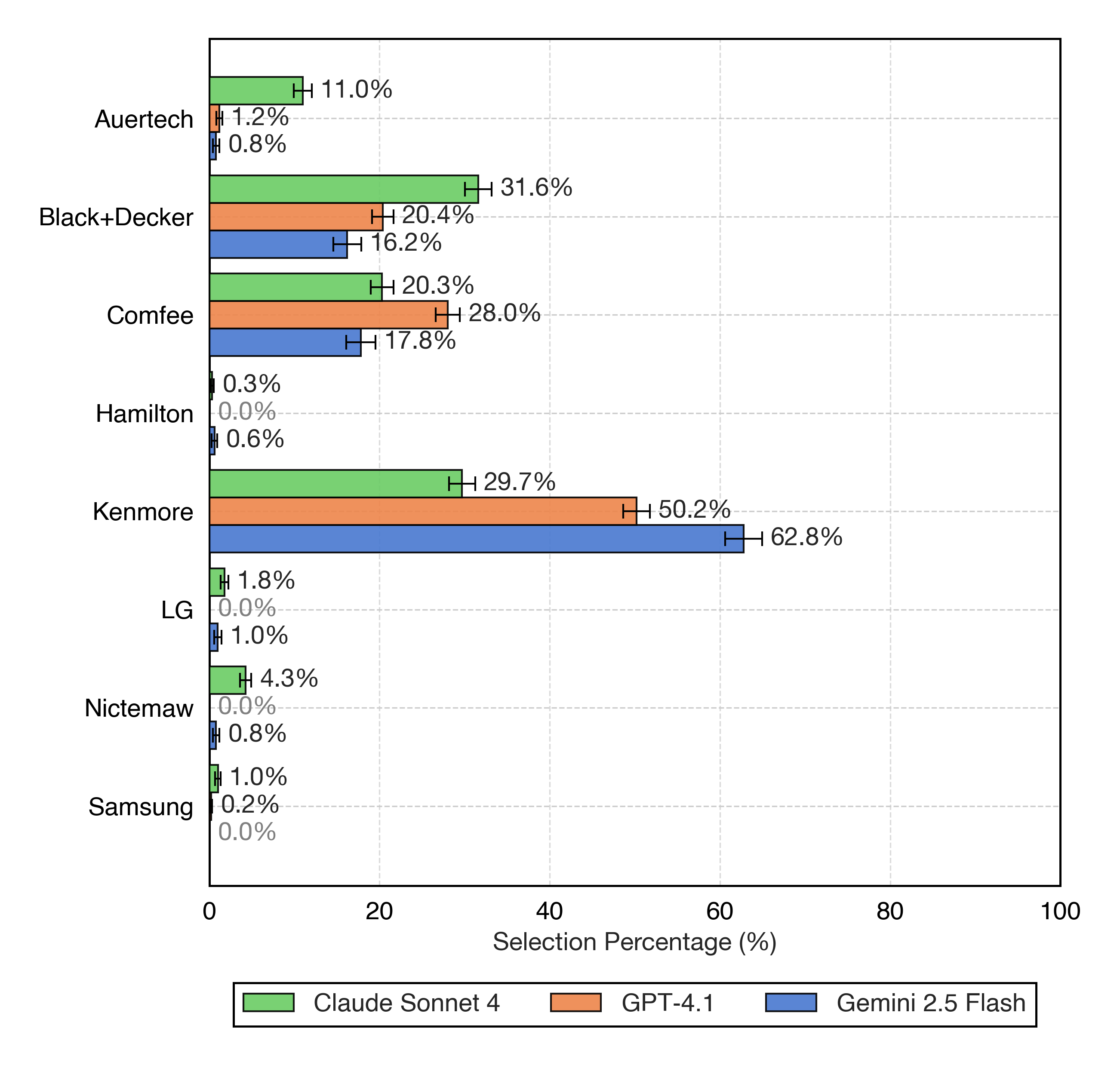}
        \caption{{washing machine}}
        \label{subfig:market-share-washing-machine}
    \end{subfigure}
    \caption{Market shares induced by different AI buying agents (Claude Sonnet 4, GPT-4.1 and Gemini 2.5 Flash) across six product categories.}
    \label{fig:market-share-product-categories}
\end{figure}

\begin{figure}[!htb]
    \centering
    \begin{subfigure}{0.32\linewidth}
        \includegraphics[width = \linewidth]{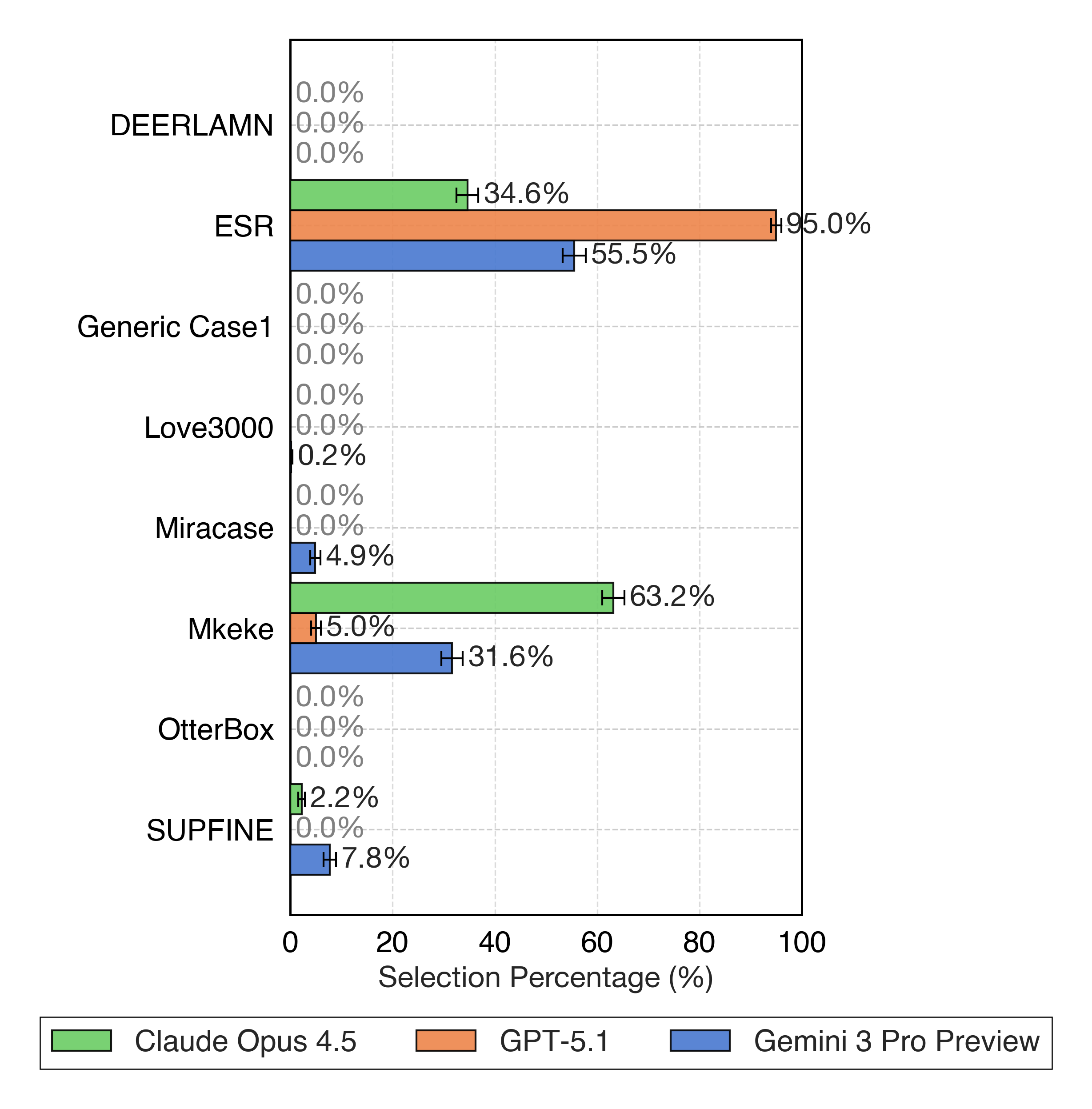}
        \caption{{ iPhone 16 Pro Cover}}
        \label{subfig:market-share-iphone-16-pro-cover}
    \end{subfigure}
    \begin{subfigure}{0.32\linewidth}
       \includegraphics[width = \linewidth]{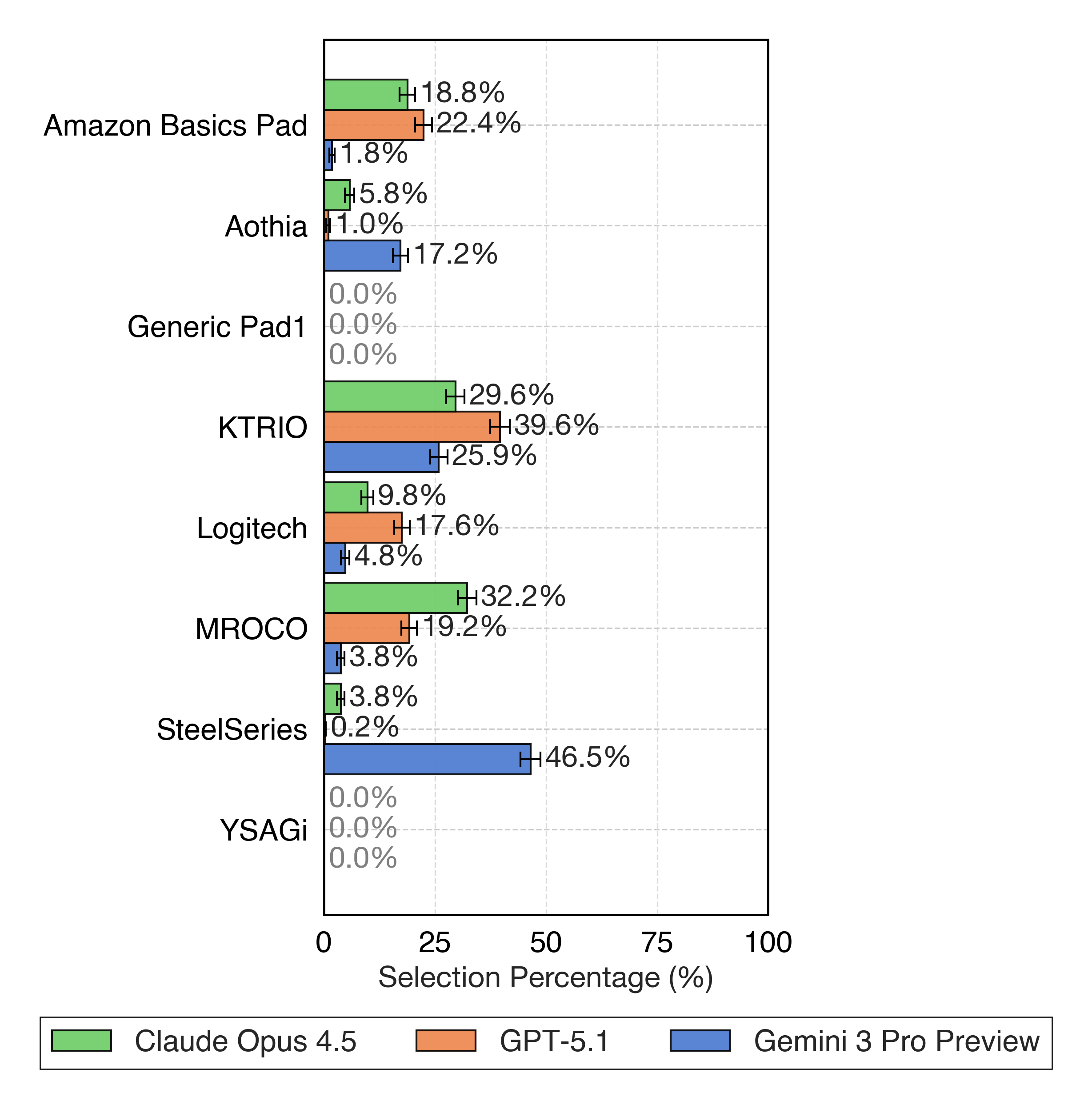}
       \caption{{mousepad}}
        \label{subgif:market-share-mousepad}
    \end{subfigure}
    \begin{subfigure}{0.32\linewidth}
        \includegraphics[width = \linewidth]{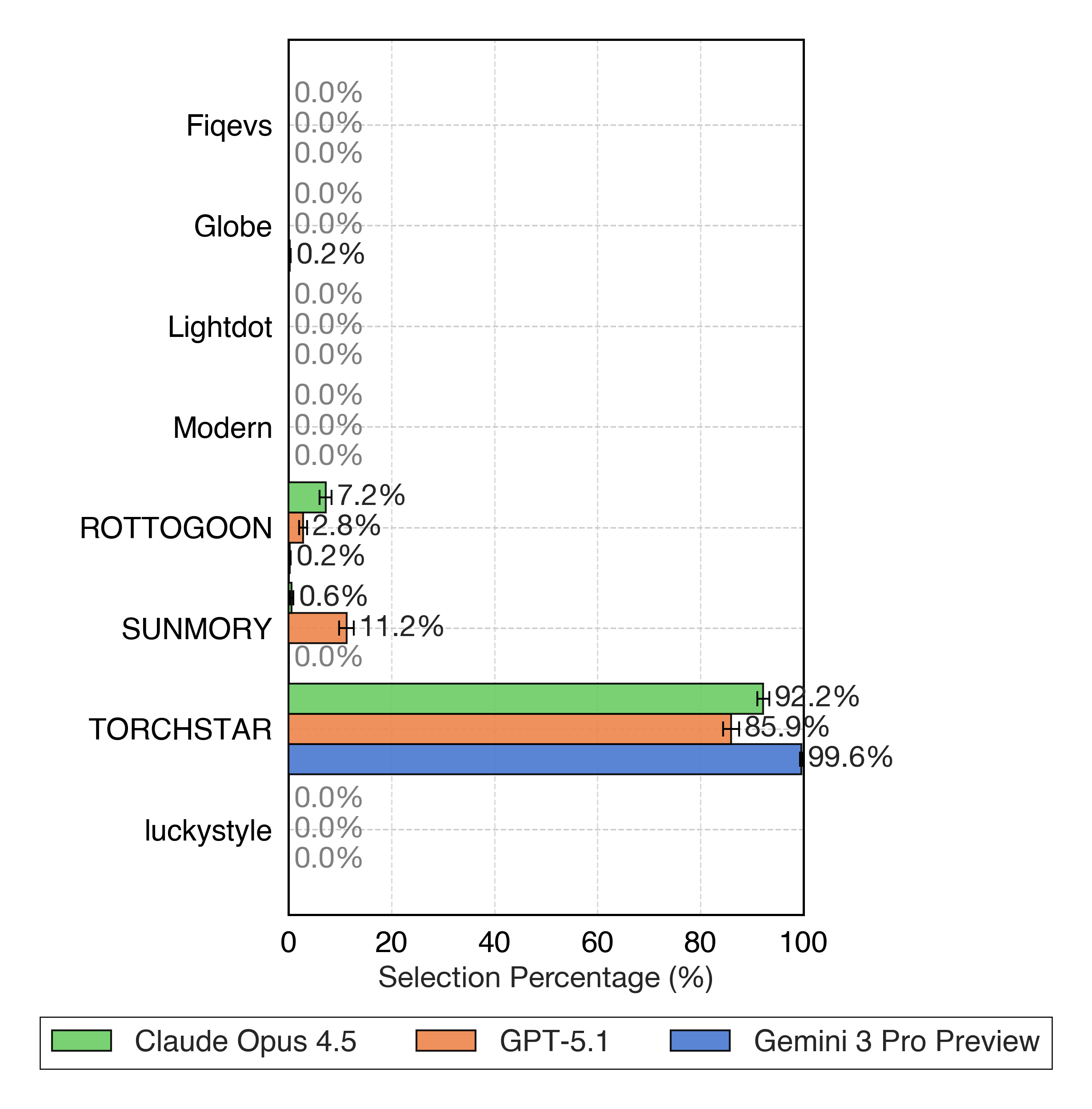}
        \caption{{office lamp}}
        \label{subfig:market-share-office-lamp}
    \end{subfigure}
    \begin{subfigure}{0.32\linewidth}
        \includegraphics[width = \linewidth]{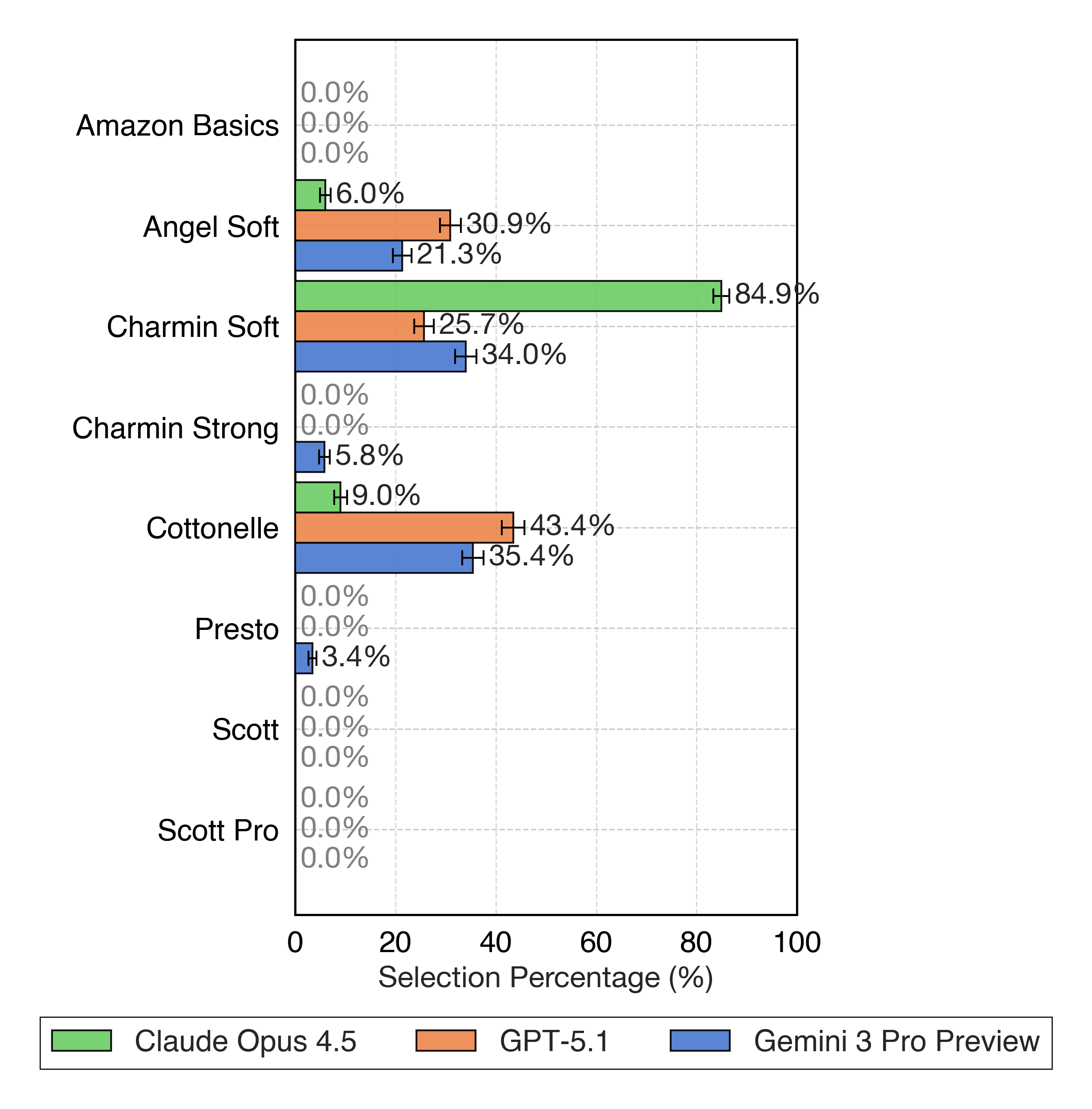}
        \caption{{toilet paper}}
        \label{subfig:market-share-toilet-paper}
    \end{subfigure}
    \begin{subfigure}{0.32\linewidth}
        \includegraphics[width = \linewidth]{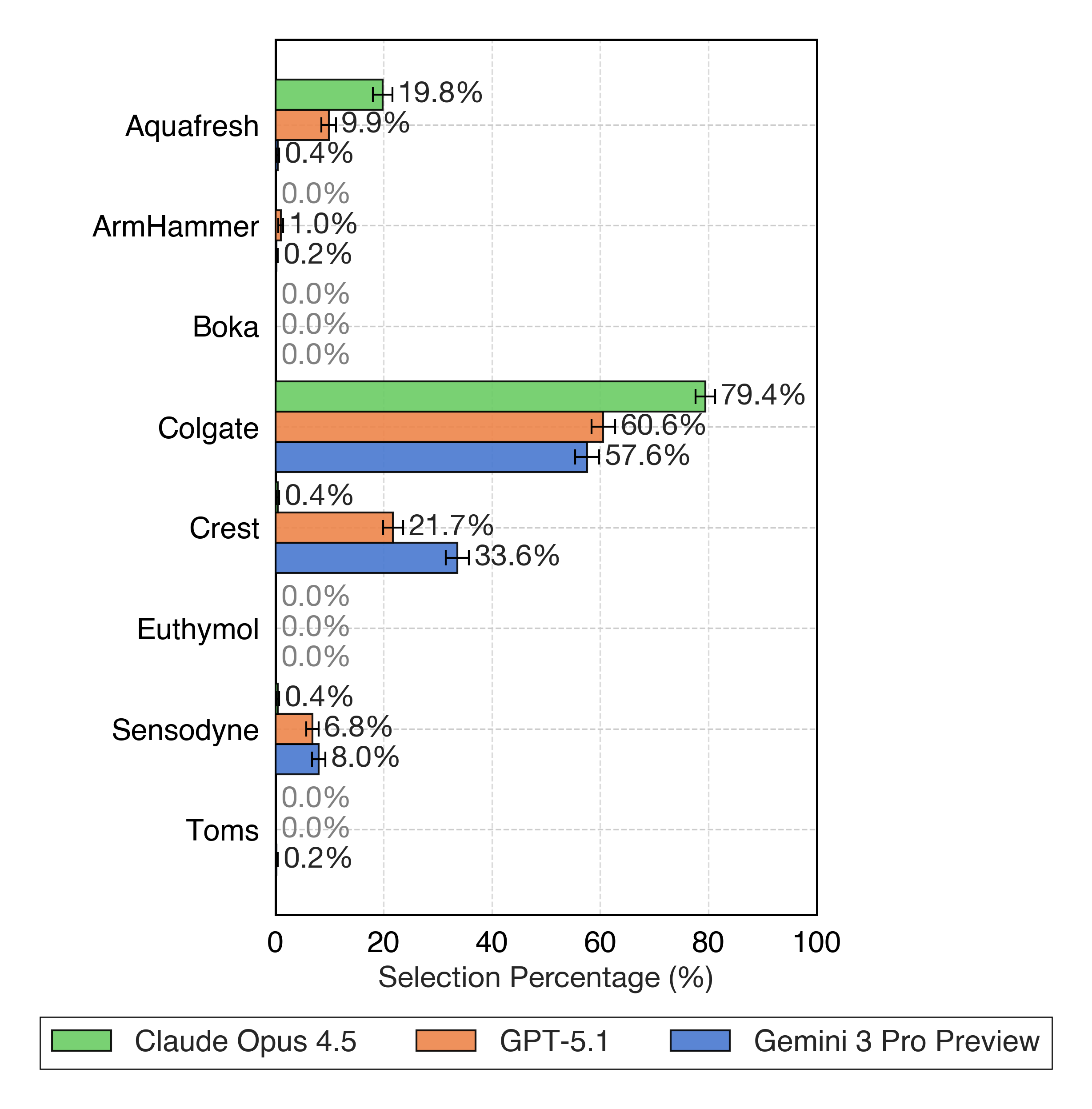}
        \caption{{toothpaste}}
        \label{subfig:market-share-toothpaste}
    \end{subfigure}
    \begin{subfigure}{0.32\linewidth}
        \includegraphics[width = \linewidth]{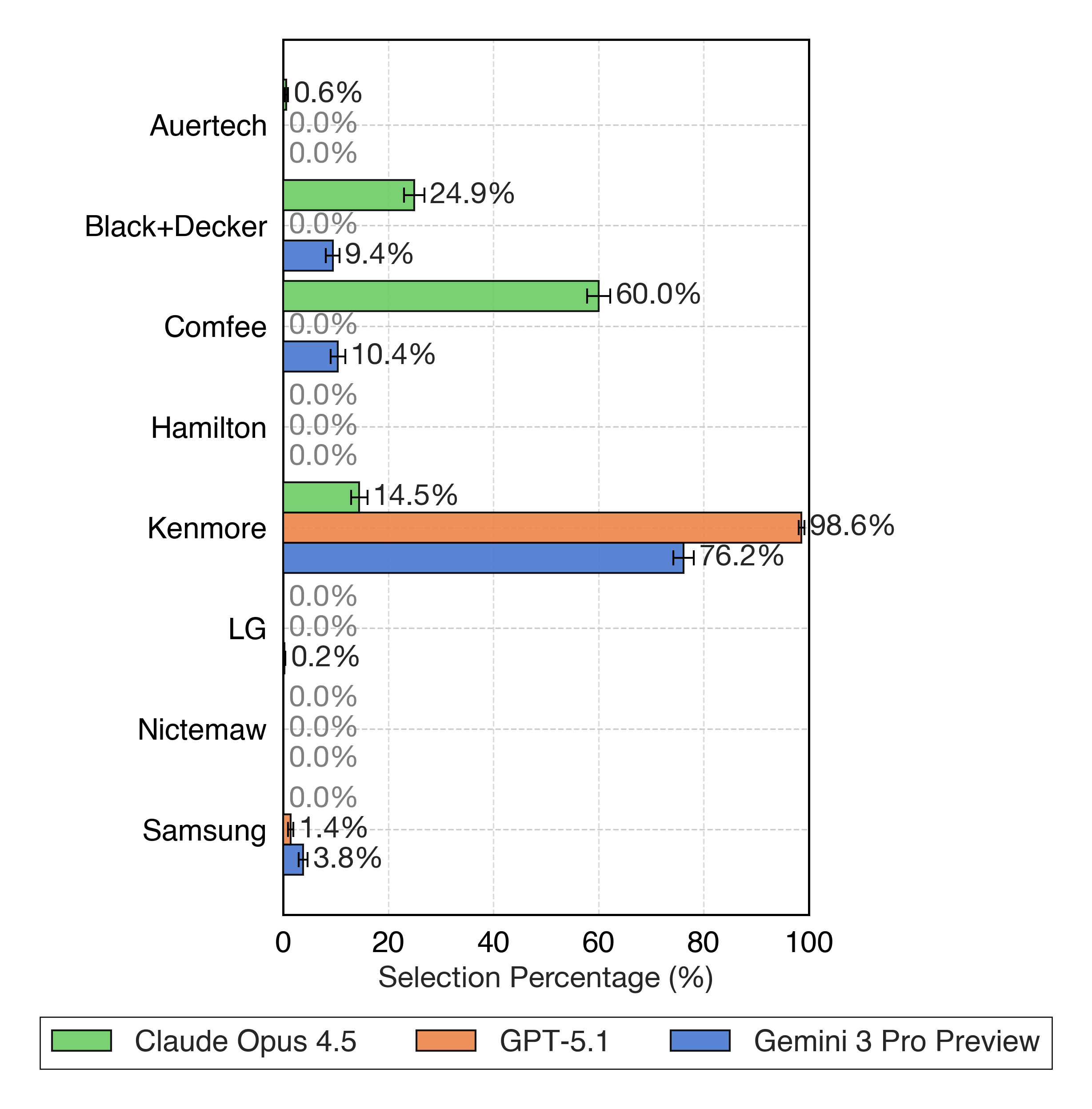}
        \caption{{washing machine}}
        \label{subfig:market-share-washing-machine}
    \end{subfigure}
    \caption{Market shares induced by different AI buying agents (Claude Opus 4.5, GPT-5.1 and Gemini 3.0 Pro Preview) across six product categories and across the last generations of models.}
    \label{fig:market-share-product-categories-new}
\end{figure}

We present the  market shares induced by different AI models in Figure \ref{fig:market-share-fitness-watch-stapler} for the fitness watch and stapler categories and in Figures \ref{fig:market-share-product-categories} and \ref{fig:market-share-product-categories-new} for six additional product categories. While the specific market share percentages are subject to the models, two qualitative patterns emerge that can have profound implications for operations and marketing strategy.

First, the choice homogeneity within models is striking. Unlike human consumers, whose preferences are often dispersed due to heterogeneous tastes or idiosyncratic search paths, AI agents tend to collapse the demand distribution onto a smaller set of ``modal products" in some categories (for example, stapler, iPhone covers, toothpaste, office lamp), and this behavior appeared to be more pronounced in the latest models (compare iPhone 16 Pro Cover and washing machines in Figures \ref{fig:market-share-product-categories} and \ref{fig:market-share-product-categories-new}).  This creates a significant concentration risk: if a specific agent achieves market dominance, it could artificially induce a "winner-take-all" dynamic that suppresses niche brands, even if those brands are {competitive when humans make purchases}.

Second, the model heterogeneity—where Claude Sonnet 4/Opus 4.5, GPT-4.1/5.1, and Gemini~2.5~Flash/3 Pro Preview prefer distinctly different products for the exact same query—introduces a new form of volatility. For brands and sellers, a model update in a dominant foundational model functions as an exogenous demand shock. A product that is the ``modal choice" for one version may lose its status in an updated version solely due to a change in the underlying model associated with the agent. This happens across scenarios. For example, GPT-4.1 selects Mikeke most often (62.6\%) in the iPhone 16 Pro cover (cf. Figure \ref{fig:market-share-product-categories}(a)), whereas  GPT-5.1 selects ESR most often (95\% of the time) and Mikeke is only selected 5\% of the time (cf. Figure \ref{fig:market-share-product-categories-new}(a)). This suggests that in an agent-mediated economy, sellers cannot treat their listings as static assets. Instead, they must continuously audit how their listings are ``perceived" by current agent architectures—a form of algorithmic market sensing. This also necessitates the ``seller response" strategies we explore in Section \ref{sec:seller-response}, but also highlights a role for regulators. If agents systematically exclude certain products, regulators may need to utilize frameworks like {\sf ACES} to audit these digital intermediaries for competitive neutrality.

\section{Choice Behavior and Biases}
\label{sec:choice-behavior}
Next, we study how AI agents trade-off product attributes and respond to position, platform endorsements, etc. using randomized controlled trials.
\subsection{Experimental Setup} 
To estimate the sensitivity of an AI agent to different platform levers, we create a dataset with exogenous variation across a number of variables associated with the products. We generate 500 scenarios for each product category, randomly, as follows: we permute the position of the eight products shown on the mock-app page uniformly at random. We randomly assign ``Sponsored'', ``Overall Pick'' and ``Only X Remaining'' (scarcity; $X$ is a random integer between 1 and 5) tags to these products. In addition, we randomly perturb the prices, ratings and number of reviews for each of the eight products in the assortment (see Table \ref{tab:exogenous-variation} for details).
The aim is to estimate the within distribution causal impact of different attributes, via perturbations sized to be large enough to materially impact the agent's choice but small enough to not fundamentally change the product landscape. Given these variations are exogenous to the inherent attractiveness of the products, they allow us to obtain causal estimates of the impact of different platform levers like position, tags and product attributes like price, rating and number of reviews. For further details regarding the experimental setup, refer to Appendix \ref{app:experimental-setup-choice-behavior-biases}.

\begin{table}[!htb]
\centering
\begin{threeparttable}
  \caption{Randomized variations used to identify agents’ choice sensitivities.}
  \label{tab:exogenous-variation}
  \footnotesize
  \sisetup{table-align-text-post=false}
  \begin{tabular}{l p{0.66\linewidth}}
    \toprule
    \multicolumn{1}{c}{Independent Variables} & \multicolumn{1}{c}{Exogenous Variation} \\
    \midrule
    Position &
      Randomly permute the position of the eight products. \\[0.4ex]
    Sponsored Tag &
      Randomly assign a {Sponsored} tag to $X$ listings, $X \sim \mathrm{Unif}(\{1,\dots,4\})$. \\[0.4ex]
    Overall Pick Tag &
      Randomly assign an {Overall Pick} tag to a listing without {Sponsored} tag. \\[0.4ex]
    Scarcity Tag &
      Randomly assign an {Only $X$ Remaining} tag to a listing without {Sponsored} or {Overall Pick} tag, $X \sim \mathrm{Unif}(\{1,\dots,5\})$. \\[0.4ex]
    Price &
      Randomly perturb the original price $p_j$ of each product $j$, setting $p_{j}^\prime \gets p_j \cdot f_j$, where $f_j \sim \mathrm{logNormal}(\mu=0,\ \sigma=0.3)$. \\[0.4ex]
    Rating &
      Randomly perturb the original rating $r_j$ of each product $j$, setting $r_j^{\prime} \gets r_j + \alpha_j \cdot (5 - r_j)$, where $\alpha_j \sim \mathrm{Unif}([-0.8,\,0.8])$. \\[0.4ex]
    Number of reviews &
      Randomly perturb the original number of reviews $N_j$ of each product $j$, setting $N_j^\prime \gets N_j \cdot f_j$, where $f_j \sim \mathrm{logNormal}(\mu=0,\ \sigma=1)$. \\
    \bottomrule
  \end{tabular}
\end{threeparttable}
\end{table}

\subsection{Our Findings} For each model analyzed,
we estimate a conditional logit (CL) \citep{mcfadden1974conditional} with the following utility specification:
\begin{align}
   \nonumber
  U_{ij} &= \sum_{ijk} \beta_{\text{pos},k} \text{Pos}_{ij,k} + \sum_{t\in \mathcal{T}} \beta_{\text{tag},t} \text{Tag}_{ij,t} \\
  \nonumber
  & \quad +  \beta_{\text{price}} \cdot \ln(\text{Price}_{ij}) 
  + \beta_{\text{rating}} \cdot \text{Rating}_{ij}
   \\ \label{eq:utility-model}
  & \quad 
   + \beta_{\text{num-revs}} \cdot \ln(\text{Num-Revs}_{ij}) + \theta_j + \varepsilon_{ij},
\end{align}
where $i$ indexes choice sets (experiments), $j$ products, $k$ indexes position, $t$ indexes tags, $\text{Pos}_{ij,k}$ is a dummy for position $k$ (top row and columns 1--3; bottom row and column 4 are the omitted categories), $\mathcal{T} = \{\text{sponsored}, \text{overall pick}, \text{scarcity}\}$ is the set of tags, $\text{Tag}_{ij,k}$ is a dummy for tag $t$,
 $\theta_j$ are product fixed effects, and $\varepsilon_{ij}$ is drawn from a standard Type I extreme value distribution. The corresponding choice probability is:
\begin{align}
  \mathbb{P}(\text{prod } j \text{ chosen in exp }i) \;=\; \frac{\exp(U_{ij})}{\sum_{k=1}^{8} \exp(U_{ik})}.
  \label{eq:mnl}
\end{align}
Table~\ref{tab:conditional-logit-results-all-models} reports the pooled (across categories) estimates for state-of-the art models in August 2025 (Claude Sonnet~4, GPT-4.1, Gemini~2.5 Flash)\footnote{In Table \ref{tab:conditional-logit-results-all-models}, we should have up to 4000 choice sets (product displays) for 500 experiments across eight product categories. In a small number of experiments, due to technical issues (e.g., the AI model not returning a valid intended choice or the experiment failing due to rate limits), we were unable to recover the choice of the AI agent and these choice sets were discarded.We also note that in order to estimate the conditional logit model, as is typical we disregarded products which were never selected. 
}.

\begin{table}[!htb]
\centering
\begin{threeparttable}
    \caption{Estimates of the Conditional Logit Regression with the default prompt (see Fig.~\ref{fig:default-prompt-buying-agent-mock-app})}
    \label{tab:conditional-logit-results-all-models}
    \footnotesize 
    \sisetup{
      input-symbols         = (), 
      table-align-text-post = false 
    }
    \begin{tabular}{l S[table-format=-1.3] S[table-format=-1.3] S[table-format=-1.3]}
        \toprule
        & \multicolumn{1}{c}{Claude Sonnet 4} & \multicolumn{1}{c}{GPT-4.1} & \multicolumn{1}{c}{Gemini 2.5 Flash} \\
        \midrule
        \multicolumn{4}{l}{\textit{Position effects}} \\
        Row 1 & 1.224\rlap{$^{\text{\tiny ***}}$} & 1.045\rlap{$^{\text{\tiny ***}}$} & 0.344\rlap{$^{\text{\tiny ***}}$}  \\
        & (0.046) & (0.046) & (0.041) \\[0.6ex]
        Column 1 & -0.297\rlap{$^{\text{\tiny ***}}$} & 1.122\rlap{$^{\text{\tiny ***}}$} & -0.264\rlap{$^{\text{\tiny ***}}$}  \\
        & (0.065) & (0.061) & (0.057) \\[0.6ex]
        Column 2 & 0.557\rlap{$^{\text{\tiny ***}}$} & 0.019 & -0.742\rlap{$^{\text{\tiny ***}}$}  \\
        & (0.058) & (0.065) & (0.061) \\[0.6ex]
        Column 3 & 0.416\rlap{$^{\text{\tiny ***}}$} & -0.013 & 0.162\rlap{$^{\text{\tiny **}}$} \\
        & (0.059) & (0.066) & (0.054) \\
        \midrule
        \multicolumn{4}{l}{\textit{Badge effects}} \\
        Sponsored Tag & -0.135\rlap{$^{\text{\tiny *}}$} & -0.248\rlap{$^{\text{\tiny ***}}$} & -0.263\rlap{$^{\text{\tiny ***}}$} \\
        & (0.068) & (0.072) & (0.067) \\[0.6ex]
        Overall Pick Tag & 1.060\rlap{$^{\text{\tiny ***}}$} & 0.802\rlap{$^{\text{\tiny ***}}$} & 1.897\rlap{$^{\text{\tiny ***}}$} \\
        & (0.077) & (0.083) & (0.072) \\[0.6ex]
        Scarcity Tag & -0.076 & -0.105 & -0.342\rlap{$^{\text{\tiny ***}}$}\\
        & (0.094) & (0.099) & (0.098) \\
        \midrule
        \multicolumn{4}{l}{\textit{Attribute effects}} \\
        $\ln(\text{Price})$ & -1.623\rlap{$^{\text{\tiny ***}}$} & -1.612\rlap{$^{\text{\tiny ***}}$} & -2.190\rlap{$^{\text{\tiny ***}}$} \\
        & (0.079) & (0.083) & (0.080) \\[0.6ex]
        Rating & 4.913\rlap{$^{\text{\tiny ***}}$} & 8.300\rlap{$^{\text{\tiny ***}}$} & 5.388\rlap{$^{\text{\tiny ***}}$} \\
        & (0.218) & (0.269) & (0.218) \\[0.6ex]
        $\ln(\text{Number of Reviews})$ & 0.415\rlap{$^{\text{\tiny ***}}$} & 0.739\rlap{$^{\text{\tiny ***}}$} & 0.501\rlap{$^{\text{\tiny ***}}$}\\
        & (0.023) & (0.026) & (0.023) \\
        \midrule
        Product Fixed Effects & \multicolumn{1}{c}{Yes} & \multicolumn{1}{c}{Yes} & \multicolumn{1}{c}{Yes}\\
        Observations\tnote{$\dagger$} & \multicolumn{1}{c}{25,802} & \multicolumn{1}{c}{25,066} & \multicolumn{1}{c}{25,215} \\
        Choice Sets (Groups) & \multicolumn{1}{c}{3,756} & \multicolumn{1}{c}{3,931} & \multicolumn{1}{c}{3,953} \\
        Pseudo R-squared & 0.44 & 0.51 & 0.42 \\
        \bottomrule
    \end{tabular}
    \begin{tablenotes}[para] 
      \item \tnote{*} $p < 0.05$, \tnote{**} $p < 0.01$, \tnote{***} $p < 0.001$.\\
      \item[$\dagger$] Observations refers to the total number of product alternatives evaluated across all choice sets. Products that were never selected were excluded from the dataset, reducing the average number of alternatives per experiment to 6.8.
    \end{tablenotes}
  \end{threeparttable}
\end{table}

{\bf Ranking matters—and {\it how} it matters is highly model‐dependent.} All three models display 
a sizeable response 
to where a product appears on the page. Holding all attributes constant, each model assigns a clear premium to the {top row} relative to the bottom row. However, the {horizontal} (column) patterns vary sharply across models. GPT‑4.1 strongly favors the {first column}; Claude Sonnet 4, in contrast, {largely ignores} the first column and prefers the two middle columns; and Gemini 2.5 Flash tilts toward the third and fourth columns. 
For illustration, in Figure \ref{fig:heatmaps-position}, we show model-based estimates of selection probability at each position under identical products, 
using the estimated coefficients in Table \ref{tab:conditional-logit-results-all-models} (setting all attributes except position to be same across the eight products and  using \eqref{eq:mnl}).  
Position can lead to drastic changes in the selection rate of a product. For example, for Claude Sonnet 4, moving a product which is selected 4.5\% of the time at the bottom right corner to the top row in the second or third column leads to a 5-fold increase in selection rate. Interestingly, moving to the top left corner would only yield half of that increase.

{\bf Sponsored tags are penalized while Platform Endorsement is substantially rewarded.} Across all three frontier models, the three badges--sponsored, overall pick, and scarcity--shift choices in the same qualitative direction. A sponsored tag reduces the likelihood of selection (a product with a baseline selection probability of 10\% falls to 8.9\% under Claude Sonnet~4, 8.0\% under GPT-4.1, and 7.9\% under Gemini 2.5 Flash), the {scarcity} tag effect is weakly negative or statistically indistinguishable from zero, whereas an {overall pick} endorsement delivers a \emph{large} positive lift (raising the same baseline to 24.3\%, 19.9\%, and 42.6\%, respectively). Because badge assignment is fully randomized, these are {causal} effects, not artifacts of latent product quality. The negative impact of the sponsored tag suggests that agents discount advertising, while the strong positive response to {overall pick} implies that platform endorsements are treated as credible signals--echoing the behavior of human buyers \citep{lill2024badges}. Importantly, our estimates do not imply that sponsored ads are harmful for sellers: sponsored ads secure premium placement—front-page or top-row exposure—that a seller might not attain organically. Our findings should therefore be interpreted conditionally: given a fixed position, the mere presence of the sponsored tag lowers an agent’s selection probability. Sponsored ads thus buy attention through placement but induce a credibility cost in the ``eyes'' of  AI agents, highlighting a trade-off that platform designers and sellers must weigh.

{\bf Directional agreement on product attributes.} All three models assign {\it negative} weight to price and {\it positive} weights to ratings and the number of reviews. Directionally, this mirrors what a rational shopper would do: prefer cheaper, better–rated, and more widely reviewed products. Quantitatively, however, the sensitivities are heterogeneous across models. For example, consider a product with a baseline selection probability of $10\%$, an $+0.1$ increase in rating lifts the probability to $15.4\%$, $20.3\%$ and $16.0\%$ with Claude Sonnet 4, GPT-4.1 and Gemini 2.5 Flash, respectively. Our price coefficients admit interpretation as price elasticity of demand for a product with small market share; across AI models our elasticity estimates (-1.6 to -2.2) are broadly in the same range as brand/SKU-level price elasticity estimates for human shoppers \citep{bijmolt05}.

{\bf Strong response to position, ratings and tags like ``Overall Pick''.} We examine the amount by which a seller can increase the price of their product while keeping the latent utility governing user choice unchanged  {(i.e., without losing market share)}, if they could (i) move from the second row to the first row, (ii) get an Overall Pick endorsement from the platform, (iii) increase their rating by +0.1, or (iv) double the number of reviews. Understanding these trade-offs are important not only for the sellers but also for platform operators since many of them operate via seller commissions. The results in Table~\ref{tab:price_equivalents} in Appendix  \ref{app:main-results-choice-behavior-biases} summarize how different models trade-off these  attributes. For Claude Sonnet 4 and GPT-4.1, moving from the second to the first row is worth roughly a {doubling} of price, whereas for Gemini 2.5 Flash the same move is worth only about +17\%. The {Overall Pick} tag produces very large lifts--e.g., +92\% for Claude Sonnet 4, +65\% for GPT-4.1 and +138\% price headroom for Gemini 2.5 Flash. An increment of +0.1 in ratings, allows a seller to increase their price by a quarter with Gemini 2.5 Flash and a third with Claude Sonnet 4 while this more than doubles to $67\%$ with GPT-4.1. Finally, doubling of reviews allows a $19\%$, $37\%$ and $17\%$ increase in price with Claude Sonnet 4, GPT-4.1 and Gemini 2.5 Flash respectively.

\subsection{Model Upgrades and Bias Patterns}

\begin{table}[!htbp]
\centering
\begin{threeparttable}
    \caption{Conditional Logit Estimates for the latest models}
    \label{tab:conditional-logit-results-all-models-new}
    \footnotesize 
    \sisetup{
      input-symbols         = (), 
      table-align-text-post = false 
    }
    \begin{tabular}{l S[table-format=-1.3] S[table-format=-1.3]  S[table-format=-1.3]}
        \toprule
        & \multicolumn{1}{c}{Claude Opus 4.5} & \multicolumn{1}{c}{GPT-5.1} &  \multicolumn{1}{c}{Gemini 3.0 Pro Preview} \\
        \midrule
        \multicolumn{4}{l}{\textit{Position effects}} \\
        Row 1 & 0.467\rlap{$^{\text{\tiny ***}}$} &  -0.701\rlap{$^{\text{\tiny ***}}$} &  2.146\rlap{$^{\text{\tiny ***}}$} \\
        & (0.047) & (0.051)  & (0.062)\\[0.6ex]
        Column 1 &  -0.389\rlap{$^{\text{\tiny ***}}$} & -0.474 \rlap{$^{\text{\tiny ***}}$} &  0.452\rlap{$^{\text{\tiny ***}}$} \\
        & (0.063) & (0.071) & (0.068) \\[0.6ex]
        Column 2 &  -0.576\rlap{$^{\text{\tiny ***}}$} &  -0.311\rlap{$^{\text{\tiny ***}}$}    & 0.114\rlap{$^{\text{\tiny }}$}\\
        & (0.065) & (0.068) &  (0.070) \\[0.6ex]
        Column 3 &  -0.997\rlap{$^{\text{\tiny ***}}$} & -0.189\rlap{$^{\text{\tiny **}}$} &  0.445\rlap{$^{\text{\tiny ***}}$}\\
        & (0.069) & (0.068 ) &  (0.068)\\
        \midrule
        \multicolumn{4}{l}{\textit{Badge effects}} \\
        Sponsored Tag &  -0.340\rlap{$^{\text{\tiny ***}}$} &  -0.371\rlap{$^{\text{\tiny ***}}$} &   -0.623\rlap{$^{\text{\tiny ***}}$}\\
        & (0.078) & ( 0.082) &  (0.083) \\[0.6ex]
        Overall Pick Tag &  1.865\rlap{$^{\text{\tiny ***}}$} & 1.342\rlap{$^{\text{\tiny ***}}$} &  2.144\rlap{$^{\text{\tiny ***}}$}  \\
        & (0.086) & (0.095) &  (0.090) \\[0.6ex]
        Scarcity Tag &  -0.320\rlap{$^{\text{\tiny **}}$} & 0.006 &  0.219\rlap{$^{\text{\tiny *}}$}\\
        & (0.111) & (0.109) &  (0.104) \\
        \midrule
        \multicolumn{4}{l}{\textit{Attribute effects}} \\
        $\ln(\text{Price})$ & -1.886\rlap{$^{\text{\tiny ***}}$} & -2.798\rlap{$^{\text{\tiny ***}}$} &  -2.249\rlap{$^{\text{\tiny ***}}$}\\
        & (0.090) & (0.102) & (0.095)\\[0.6ex]
        Rating & 11.148\rlap{$^{\text{\tiny ***}}$} & 9.247\rlap{$^{\text{\tiny ***}}$} &  4.224\rlap{$^{\text{\tiny ***}}$}\\
        & (0.323) & (0.314) &  (0.250)\\[0.6ex]
        $\ln(\text{Number of Reviews})$ & 0.979 \rlap{$^{\text{\tiny ***}}$} & 0.800\rlap{$^{\text{\tiny ***}}$} &   0.669\rlap{$^{\text{\tiny ***}}$}\\
        & (0.030) & (0.030) &  (0.028)\\
        \midrule
        Product Fixed Effects & \multicolumn{1}{c}{Yes} & \multicolumn{1}{c}{Yes} &  \multicolumn{1}{c}{Yes}\\
        Observations & \multicolumn{1}{c}{24,433} & \multicolumn{1}{c}{20,682} &  \multicolumn{1}{c}{22,960}\\
        Choice Sets (Groups) & \multicolumn{1}{c}{3,988} & \multicolumn{1}{c}{3,932} &   \multicolumn{1}{c}{3,993}\\
        Pseudo R-squared & 0.57 & 0.59 &  0.58\\
        \bottomrule
    \end{tabular}
    \begin{tablenotes}[para] 
      \item \tnote{*} $p < 0.05$, \tnote{**} $p < 0.01$, \tnote{***} $p < 0.001$.\\
    \end{tablenotes}
  \end{threeparttable}
\end{table}

\begin{figure}[!htb]
    \begin{subfigure}{0.33\linewidth}
        \begin{tikzpicture}
        \node[inner sep=0pt, outer sep=0pt] (browser) at (0,0) {
            \includegraphics[width=5.5cm]{images/ACE-Agent-Workflow-Images/mock-app-front-page.pdf} 
        };
        \node[inner sep=0pt, outer sep=0pt] at ([xshift=0.05cm, yshift=-0.2cm]browser.center) {
            \includegraphics[width=4.5cm]{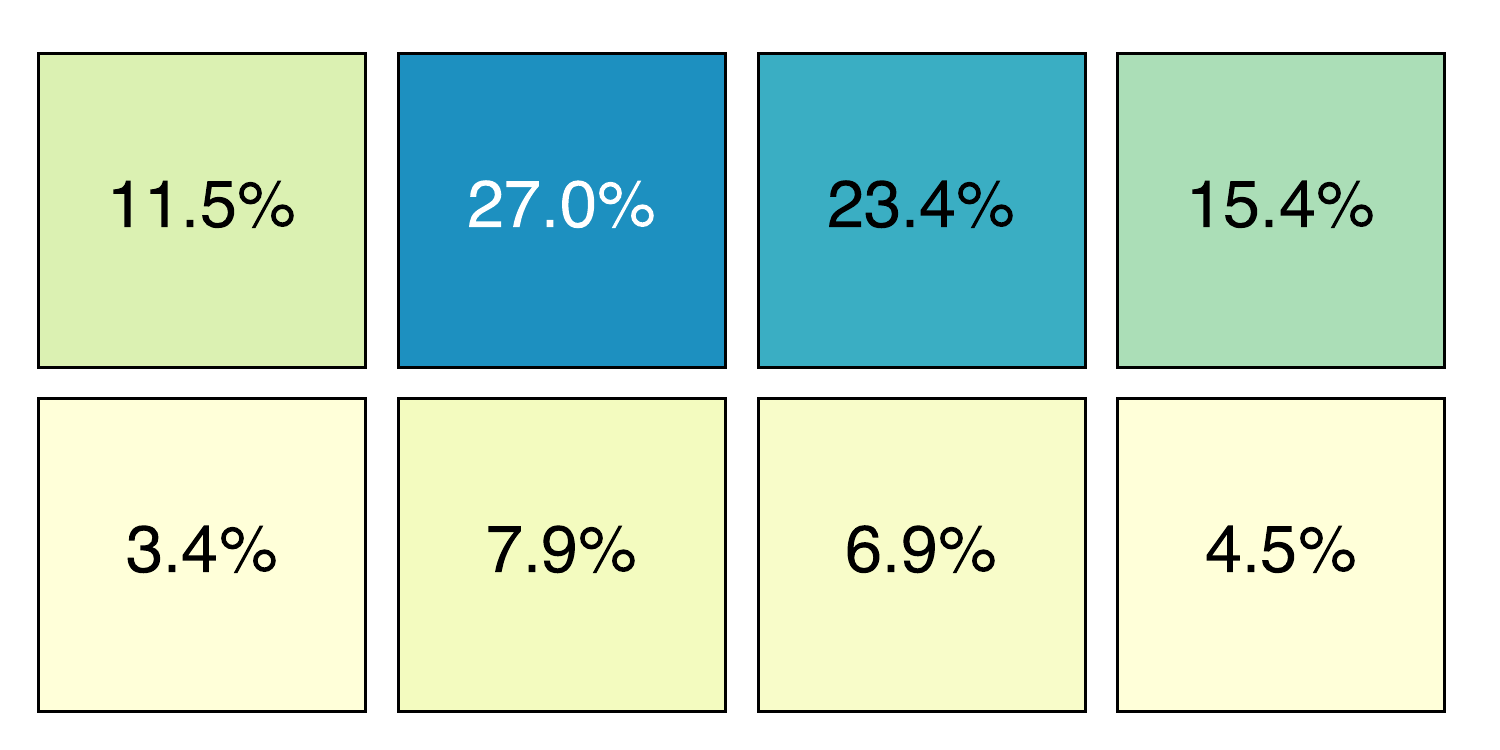} 
        };
    \end{tikzpicture}%
    \caption{Claude Sonnet 4}
    \end{subfigure}
    \begin{subfigure}{0.33\linewidth}
        \begin{tikzpicture}
        \node[inner sep=0pt, outer sep=0pt] (browser) at (0,0) {
            \includegraphics[width=5.5cm]{images/ACE-Agent-Workflow-Images/mock-app-front-page.pdf} 
        };
        \node[inner sep=0pt, outer sep=0pt] at ([xshift=0.05cm, yshift=-0.2cm]browser.center) {
            \includegraphics[width=4.5cm]{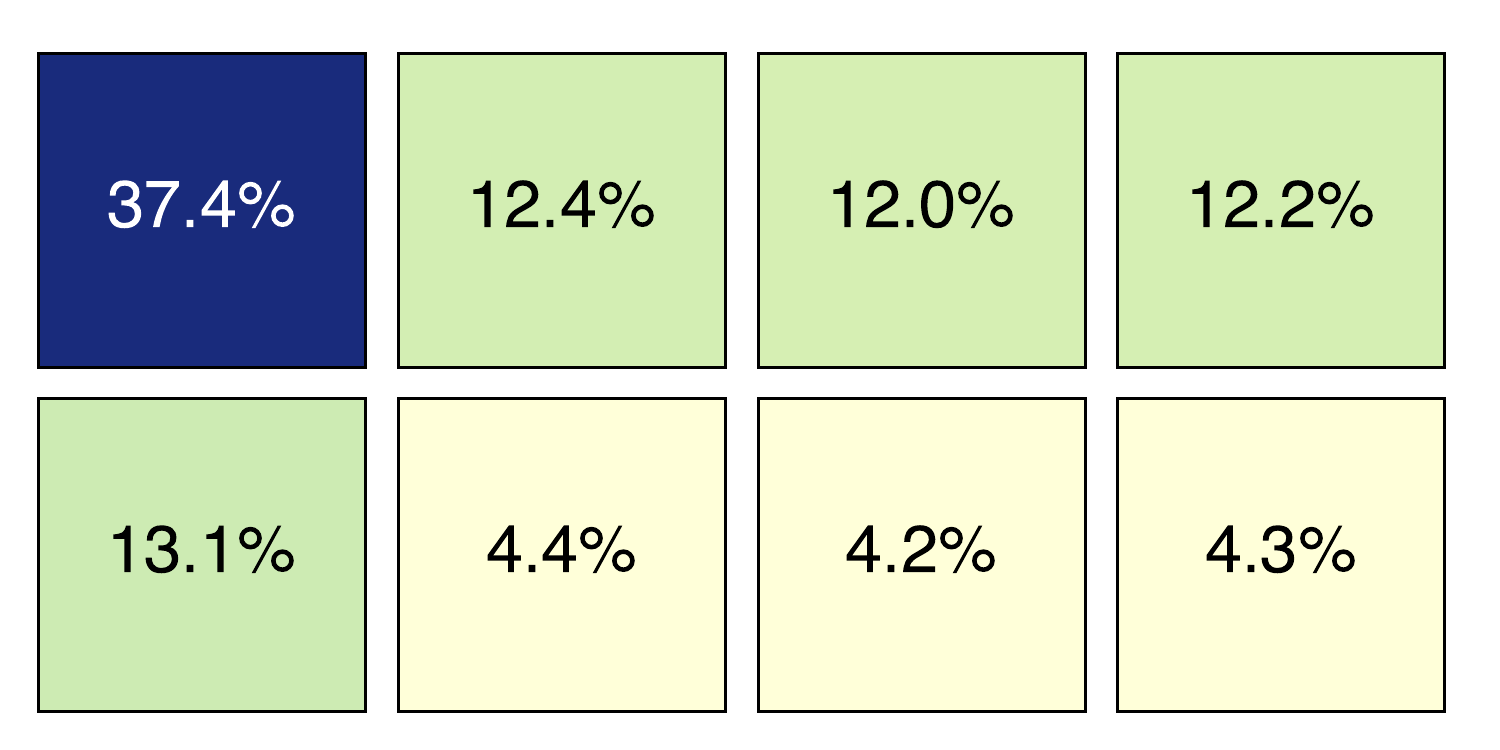} 
        };
    \end{tikzpicture}%
        \caption{GPT-4.1}
    \end{subfigure}
    \begin{subfigure}{0.33\linewidth}
        \begin{tikzpicture}
        \node[inner sep=0pt, outer sep=0pt] (browser) at (0,0) {
            \includegraphics[width=5.5cm]{images/ACE-Agent-Workflow-Images/mock-app-front-page.pdf} 
        };
        \node[inner sep=0pt, outer sep=0pt] at ([xshift=0.05cm, yshift=-0.2cm]browser.center) {
            \includegraphics[width=4.5cm]{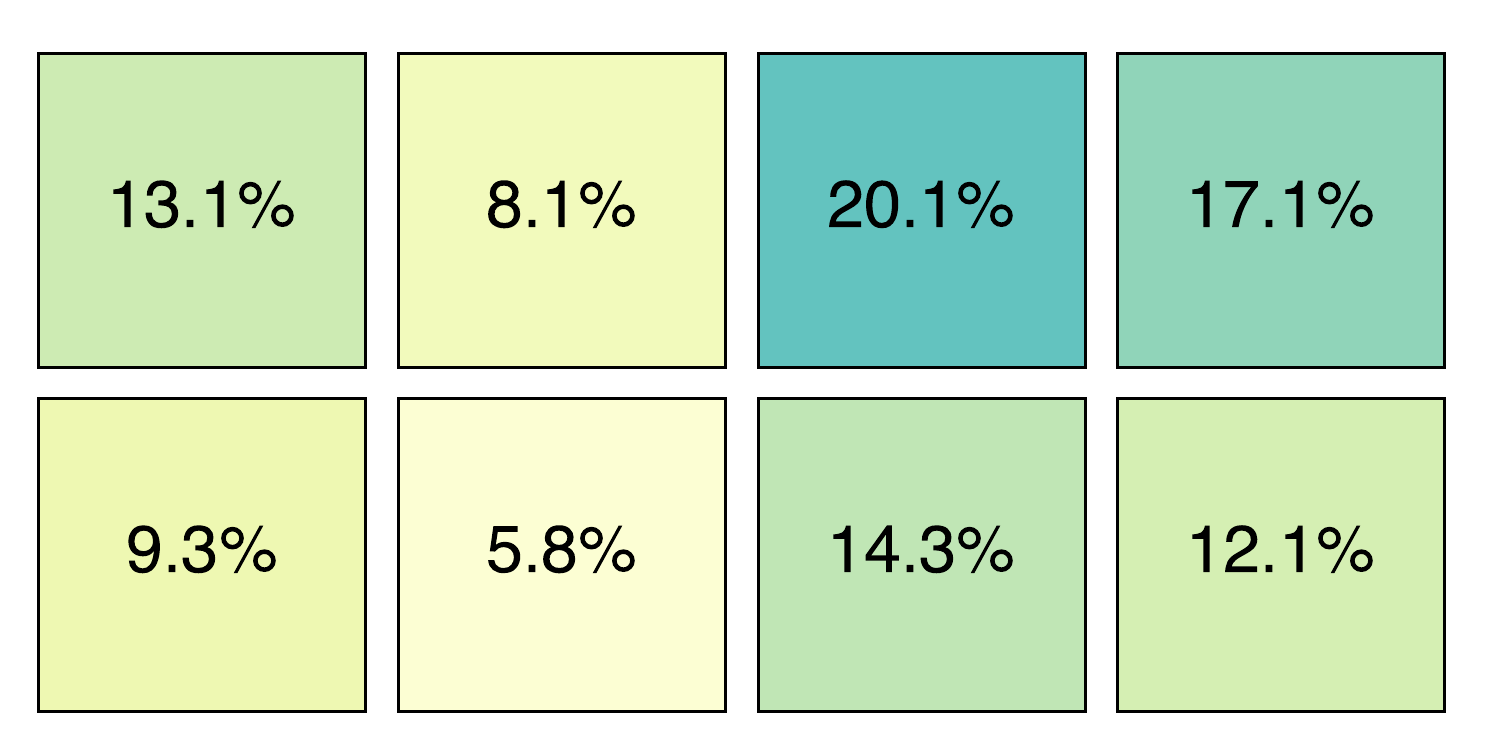} 
        };
    \end{tikzpicture}
    \caption{Gemini 2.5 Flash}
    \end{subfigure}
    \\
        \begin{subfigure}{0.33\linewidth}
        \begin{tikzpicture}
        \node[inner sep=0pt, outer sep=0pt] (browser) at (0,0) {
            \includegraphics[width=5.5cm]{images/ACE-Agent-Workflow-Images/mock-app-front-page.pdf} 
        };
        \node[inner sep=0pt, outer sep=0pt] at ([xshift=0.05cm, yshift=-0.2cm]browser.center) {
            \includegraphics[width=4.5cm]{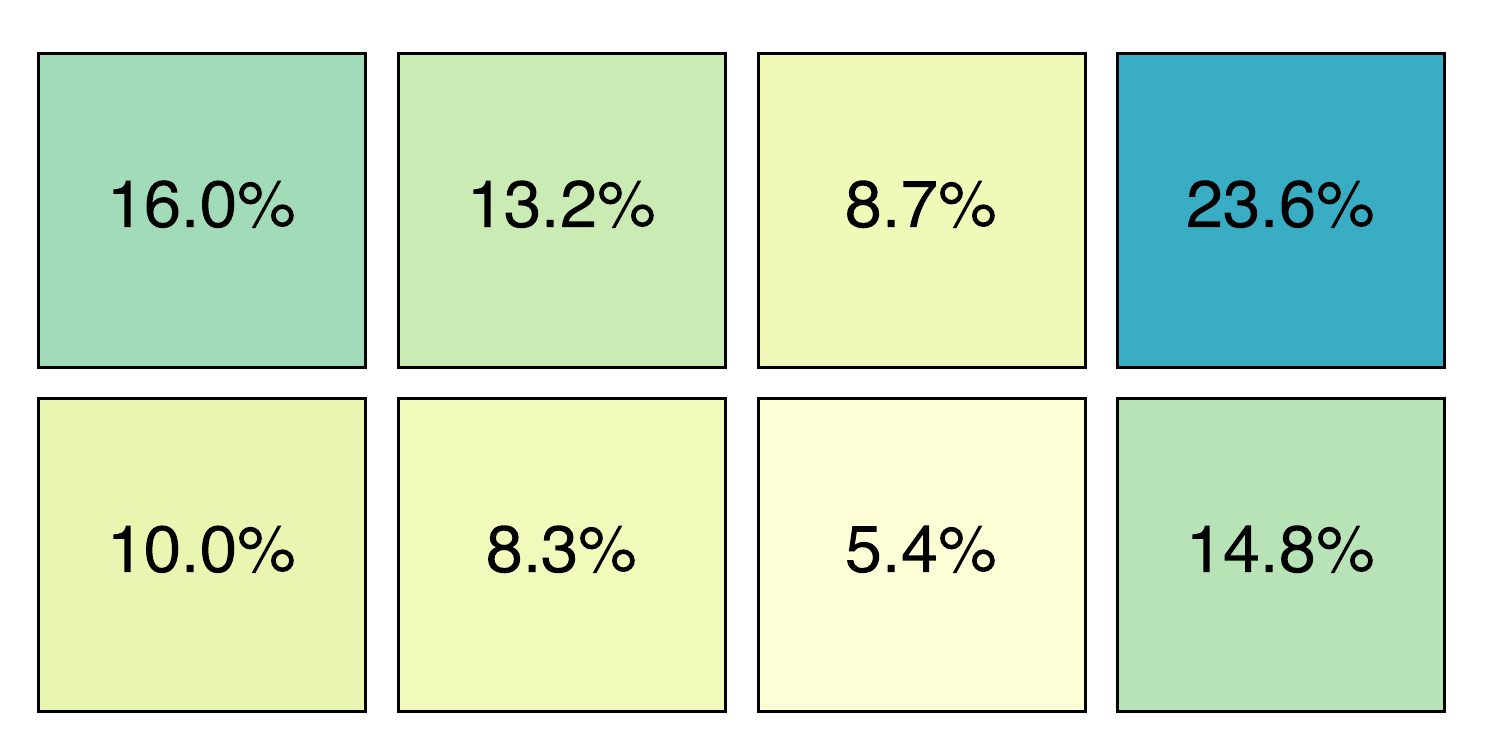} 
        };
    \end{tikzpicture}%
    \caption{Claude Opus 4.5}
    \end{subfigure}
           \begin{subfigure}{0.33\linewidth}
        \begin{tikzpicture}
        \node[inner sep=0pt, outer sep=0pt] (browser) at (0,0) {
            \includegraphics[width=5.5cm]{images/ACE-Agent-Workflow-Images/mock-app-front-page.pdf} 
        };
        \node[inner sep=0pt, outer sep=0pt] at ([xshift=0.05cm, yshift=-0.2cm]browser.center) {
            \includegraphics[width=4.5cm]{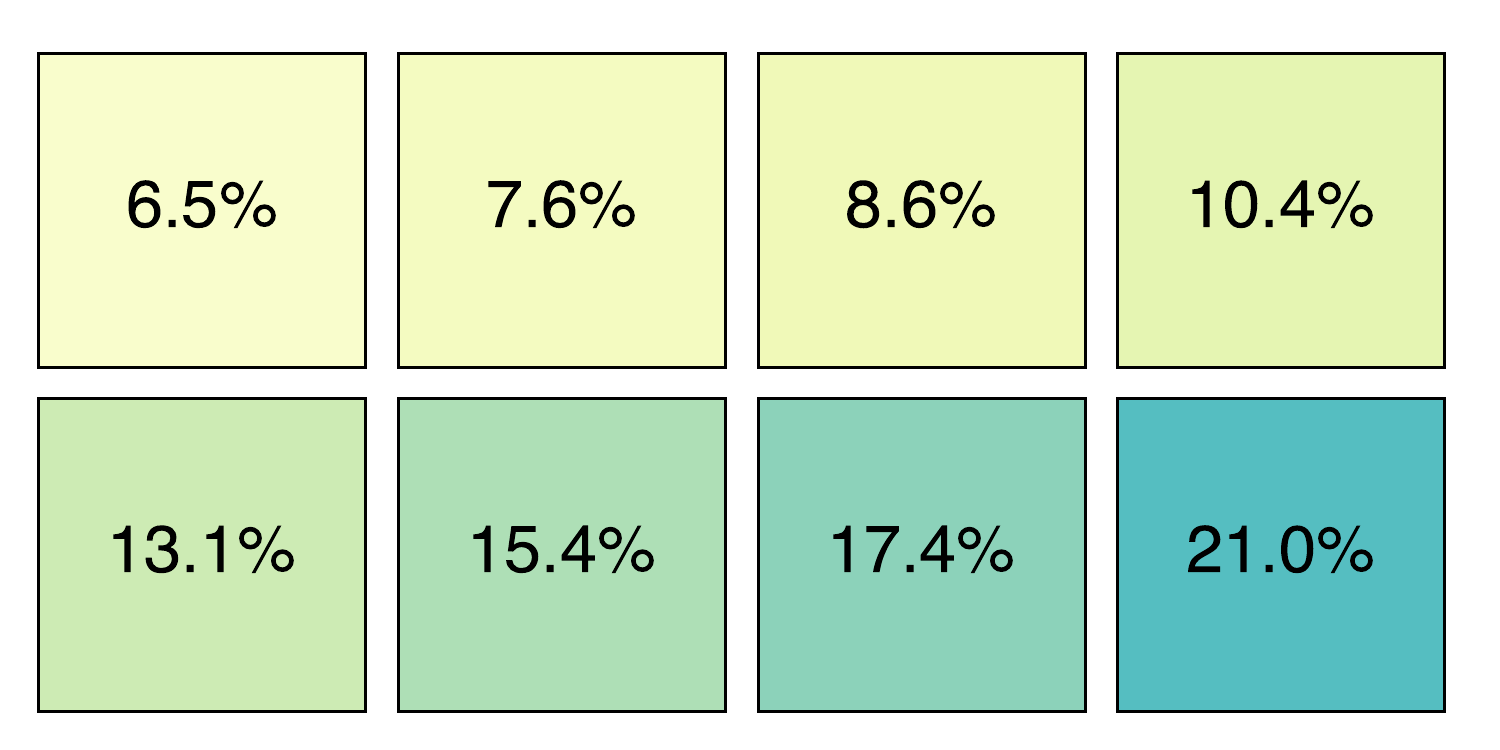} 
        };
    \end{tikzpicture}%
    \caption{GPT-5.1}
    \end{subfigure}
               \begin{subfigure}{0.33\linewidth}
        \begin{tikzpicture}
        \node[inner sep=0pt, outer sep=0pt] (browser) at (0,0) {
            \includegraphics[width=5.5cm]{images/ACE-Agent-Workflow-Images/mock-app-front-page.pdf} 
        };
        \node[inner sep=0pt, outer sep=0pt] at ([xshift=0.05cm, yshift=-0.2cm]browser.center) {
            \includegraphics[width=4.5cm]{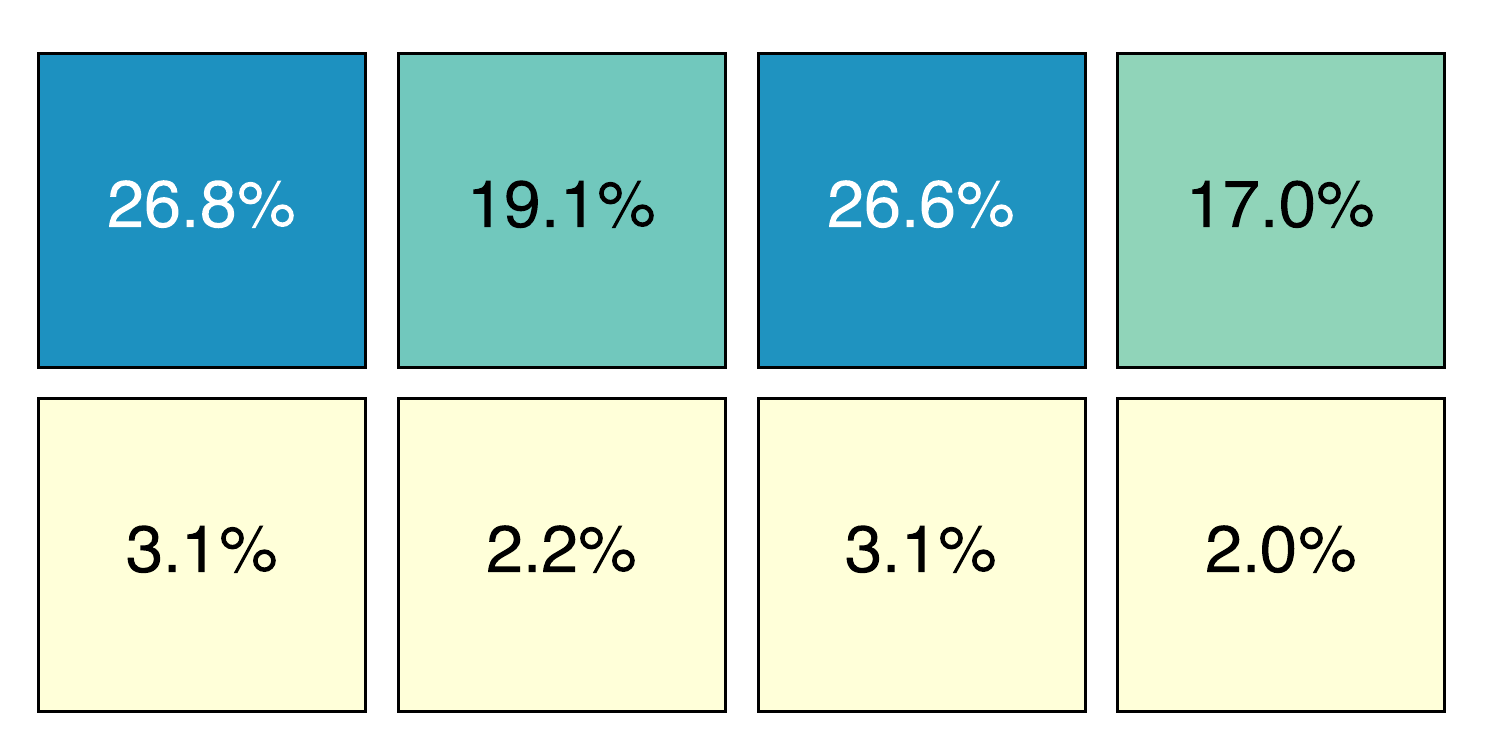} 
        };
    \end{tikzpicture}%
    \caption{Gemini 3.0 Pro Preview}
    \end{subfigure}
    \caption{{\textbf{Webpage heatmap of AI shopping agents.} The figure depicts the selection rate of the same product across different positions for six AI agents.}}
    \label{fig:heatmaps-position}
\end{figure}

We also conduct the same experiment with the latest models of the three providers, which were released after our initial experiments. In particular, we test the behavior of Claude Opus 4.5, GPT-5.1, and Gemini 3 Pro Preview. The results of the conditional logit parameter estimates are presented in Table \ref{tab:conditional-logit-results-all-models-new}.

We observe that the AI agents differ drastically from one generation of models to the next. The presence of position biases persist but these are very different when models are upgraded. This is best seen in Figure \ref{fig:heatmaps-position}, which depicts the probability of selecting a particular product as a function of its position. For every upgrade the position bias changed. In the case of OpenAI's GPT model, GPT-4.1 preferred slot is exactly is the least preferred by its successor GPT 5.1, and vice-versa. Similarly, one can see that aversion to sponsored tags, and the rewarding of platform endorsement persists after the upgrade. This study, rather over-emphasizing the exact behavior of any particular agent, brings to the foreground the need to continuously monitor and audit agents' behavior.

\section{Seller Response to AI-Assisted Shopping}
\label{sec:seller-response}
Given AI buyer agents display systematic—and exploitable—choice patterns, it is natural to ask how sellers might adapt to this new form of algorithmic demand. As a first step in this direction, we study a seller-side setting in which one seller attempts to use AI to edit their listing (specifically, the product description) in order to increase their market share. 

\subsection{Experimental Setup} 

A minimal AI seller agent is provided product details (and a screenshot of the listing) for a randomly chosen focal product from each product category (e.g., mousepad, office lamp), as well as sales data and listing screenshots for the competitors, and asked to suggest an improved description of the product (Figure~\ref{fig:prompt-seller-agent}; see Section \ref{app:illustrative-example-office-lamp} for a concrete example). To estimate the causal impact of the recommendation made by the AI seller agent, we use the following experimental setup.

\begin{enumerate}
    \item {Simulate the purchase behavior of the AI buying agent across different product categories.} We run 200 experiment trials where we present the AI buying agents with eight products per category. The eight products are  shuffled uniformly at random  across the 200 experiments. Note that we only shuffle the position of the products on the screen but do not change their other attributes like price, description, etc.

    \item For each category, we select uniformly at random  a single product from the eight-product assortment. The selected product becomes the \textit{focal} product/seller that optimizes its description via an AI seller agent.

    \item We prompt the AI seller agent to generate recommendations for the product description. We provide the AI seller agent a screenshot of the platform with the eight products along with the features of the seller's product, some context and the sales data of competing sellers induced by the buying AI agent. The exact prompt is reproduced below in Figure \ref{fig:prompt-seller-agent}.

    \begin{figure}[!htb]
        \begin{tcolorbox}[sellerbuyertemplate]
        Context: There is an AI agent (a vision language model) which is given a screenshot of an e-commerce website selling a particular product and it decides on the products based on the given attributes. I have the sales data on all the products including mine. I want you to act as an agent on my behalf and suggest changes to the product title so that I can increase my sales by making my product more appealing to the AI agent. 
        \\ 
        My~product~is~\{{\tt product name}\}.~The~product~features~are~\{{\tt detailed features from Amazon}\} \\ 
        I have data from 200 experiments and the following is my and my competitors sales data across experiments. \\ 
        \{{\tt simulated purchase data from the 200 experiments with the original descriptions}\}.

        It is important that the title changes you suggest align with the product features provided to you. Do not make up product features or add spurious keywords. Just use the product feature information provided.
    \end{tcolorbox}
    
        \caption{Prompt template for AI seller agent}
        \label{fig:prompt-seller-agent}
    \end{figure}
    
    \item Change the product description of the focal product with the one recommended by the AI seller agent and simulate the purchase behavior of the AI agents across different product categories. Repeat the 200 experiments as before with the only change being the modified product description of the focal product/seller.
\end{enumerate}

We study a total of six AI buying models---three state-of-the-art models in August 2025 (Claude Sonnet 4, GPT-4.1 and Gemini 2.5 Flash) and their most recent successors as of December 1st 2025 (Claude Opus 4.5, GPT-5.1 and Gemini 3.0 Pro Preview). For Claude Sonnet 4, GPT-4.1 and Gemini 2.5 Flash, we use GPT-4.1 as the seller agent while for Claude Opus 4.5, GPT-5.1 and Gemini 3.0 Pro Preview, we use GPT-5.1 as the seller agent.
Note that for simplicity in this study of seller response, we assume that only one seller uses an AI agent to optimize their product description, and the seller tries only one AI generated recommendation (through one interaction with their AI agent). 

\subsection{Our Findings} 
We define the change in market share of the focal product as the difference in the selection probabilities of the focal product after and before the description modification. Because all other attributes are held fixed---and we reuse the identical product shuffles in the pre/post modification experiments---the observed differences are causally attributable to the description modification. We provide an illustrative case study in Appendix \ref{app:illustrative-example-office-lamp}.

\begin{figure}[!htbp]
    \centering
    \begin{subfigure}{1\linewidth}
    \centering
        \includegraphics[width=0.8\linewidth]{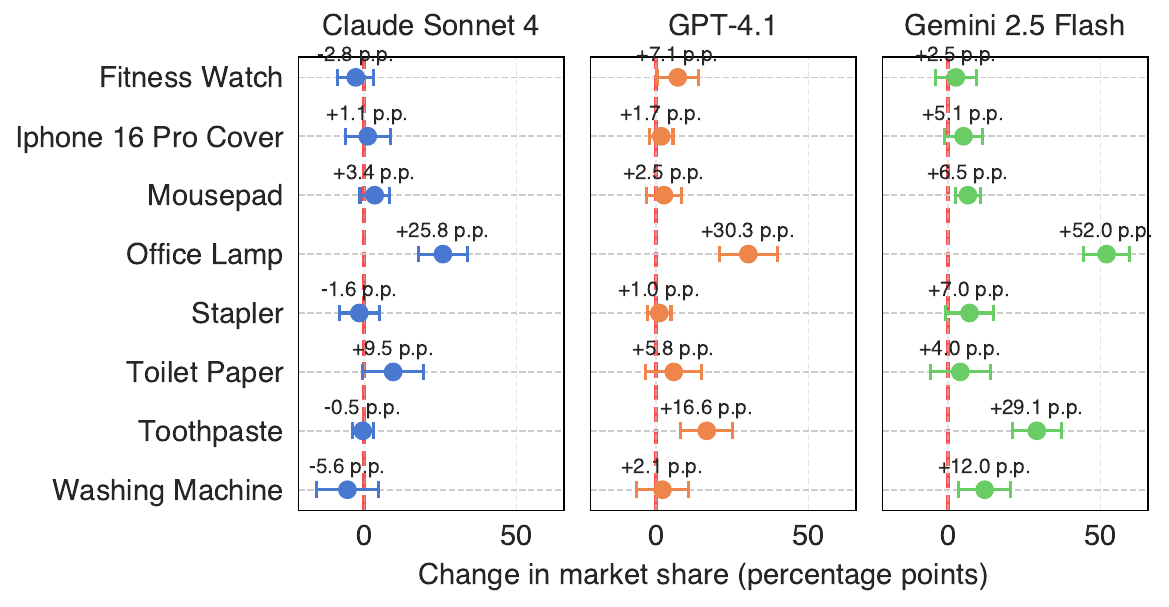}
    \end{subfigure}
    \begin{subfigure}{1\linewidth}
    \centering
        \includegraphics[width=0.8\linewidth]{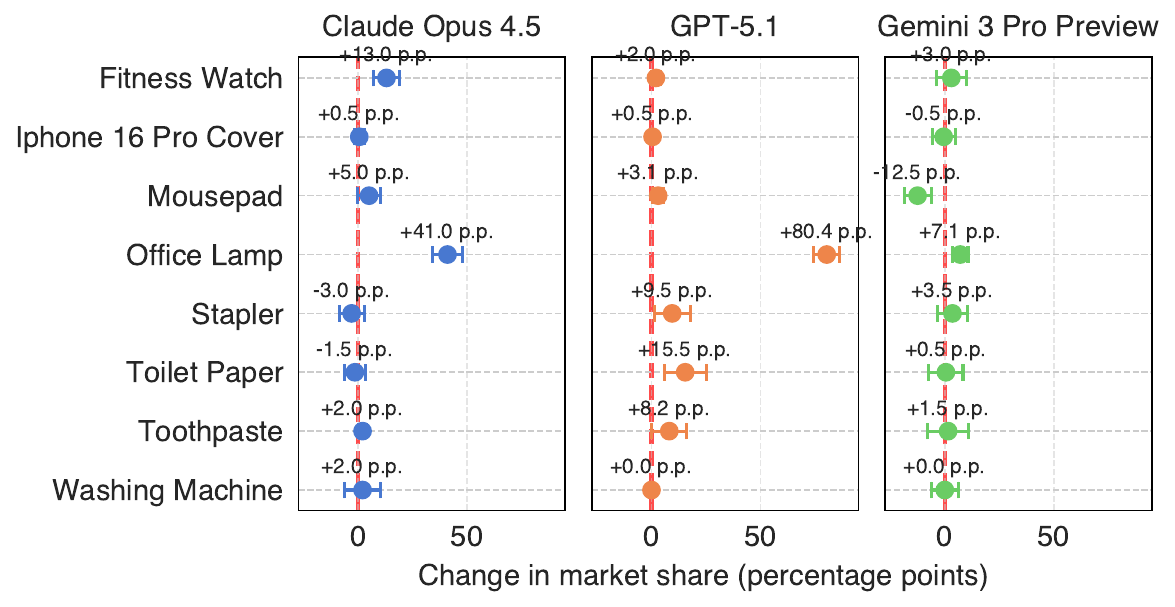}
    \end{subfigure}
    \caption{Market share change of the focal product/seller across different categories with different AI buying agent models. Note that the description of only one product in each category is changed based on the recommendation of the seller AI agent.}
    \label{fig:ate-multi-model}
\end{figure}

In Table \ref{tab:seller-response-market-share-increase}, we report the average change in market share due to the seller side intervention. In Figure~\ref{fig:ate-multi-model}, we report the change in market share for the { focal} product across different categories and buyer agents (Claude Sonnet~4, GPT‑4.1, Gemini~2.5~Flash, Claude Opus 4.5, GPT-5.1 and Gemini 3.0 Pro Preview).

\begin{table}[!htb]
\centering
\begin{threeparttable}
    \caption{Average change in market share for the focal product with different AI buying agents}
    \label{tab:seller-response-market-share-increase}
    \footnotesize 
    \sisetup{
      input-symbols           = (), 
      table-align-text-post = false 
    }
    \begin{tabular}{lcc}
        \toprule
        AI Buying Model & Change in Market Share (in p.p.) & Statistically Significant (Yes/No) \\
        \midrule
        \multicolumn{3}{l}{{\it AI Seller Agent is GPT-4.1}} \\
        Claude Sonnet 4 & $+3.66$ $(1.33)$ & Yes \\
        GPT-4.1 & $+8.37$ $(1.31)$  & Yes \\
        Gemini 2.5 Flash & $+14.79$ $(1.36)$ & Yes \\
        \midrule
        \multicolumn{3}{l}{{\it AI Seller Agent is GPT-5.1}} \\
        Claude Opus 4.5 & $+7.38$ $(1.00)$ & Yes \\ 
        GPT-5.1 & $+14.89$ $(1.05)$ & Yes \\
        Gemini 3.0 Pro Preview & $+0.32$ $(1.22)$  & No \\
        \bottomrule
    \end{tabular}
  \end{threeparttable}
\end{table}

Effects are heterogeneous across both categories and models. In $67\%$ of category–model pairs, the (one-shot) description modification yields no statistically significant change in market share. In the remaining $33\%$ cases, a one‑shot modification produces large gains: for Claude Sonnet~4 we observe $+25.8$ p.p. in office lamp; for GPT‑4.1, $+7.1$ p.p. in {fitness watch}, $+30.3$ p.p. in office lamp, and $+16.6$ p.p. in {toothpaste}; for Gemini~2.5~Flash, $+6.5$ p.p. in mousepad, and $+52.0$ p.p. in {office lamp}, $+29.1$ p.p. in toothpaste and $+12.0$ p.p. in washing machine. Similarly, for the newer models, Claude Opus 4.5 shows $+41.0$ p.p. in office lamp; GPT-5.1 shows a massive $+80.4$ p.p. in office lamp, along with significant gains in stapler ($+9.5$ p.p.) and toilet paper ($+15.5$ p.p.).

Notably, the ``office lamp'' category stands out as the only product category where the AI seller agent achieved consistent, positive, and statistically significant market share gains across \textit{all six} buying models tested (ranging from $+7.1$ p.p. to $+80.4$ p.p.). We hypothesize this is because the original description of the focal product (SUNMORY) failed to include the keyword ``Office'' early in its description and hence being truncated from the product description on the mock platform, causing the buying agents to overlook it for the specific user query of ``office lamp''. The seller agent(s) consistently rectified this by front-loading the term ``Office'' (e.g., changing ``Floor Lamps for Living Room...'' to ``Office Floor Lamp...''). This simple structural optimization effectively ``unlocked'' the product's relevance for the buying algorithms, demonstrating that in some cases, simple keyword-based heuristics can drive massive swings in algorithmic demand. For a detailed breakdown of the office lamp case study, see Appendix \ref{app:illustrative-example-office-lamp}.

However, the results also reveal a significant risk: the intervention is not risk-free. In a few instances (e.g., stapler for Claude Opus 4.5, mousepad for Gemini 3.0 Pro Preview), the AI-optimized description actually led to a decrease in market share. This underscores that AI-mediated markets are volatile and heterogeneous---a change that works for one model may backfire for another. Consequently, sellers cannot simply ``set and forget'' AI-generated descriptions. Instead, they require continuous monitoring frameworks—similar to the ACES evaluation framework proposed in this paper—to iteratively test, validate, and adapt their listings against an evolving landscape of buying agents. 

\subsection{Drivers of Improvement}
It is natural to ask whether these performance gains are simply due to the AI seller agent producing shorter descriptions. However, our analysis suggests that \textit{information reordering} and \textit{keyword alignment} are the primary drivers. The optimized descriptions are not significantly shorter; rather, they structurally prioritize high-relevance tokens. 

The magnitude of the market share gain appears directly correlated with the extent of this semantic realignment. In categories where the seller agent executed significant \textit{keyword enrichment} or \textit{front-loading}, the gains were substantial. For instance, in the toilet paper category with GPT-5.1, the seller agent enriched a sparse description by adding high-value sensory and functional keywords (e.g., ``2-Ply'', ``Septic-Safe'', ``Unscented''), resulting in a $+15.5$ p.p. increase. Similarly, for staplers, explicitly front-loading the included accessory (``with 1,250 Staples'') alongside the main product name drove a $+9.5$ p.p. increase. For example, in the fitness watch category, both GPT-4.1 (+7.1 p.p.) and Claude Opus 4.5 (+13.0 p.p.) achieved gains not by shortening the text, but by explicitly appending the term ``Fitness''---a keyword absent from the original ``GPS Running Watch'' title---to align with the buyer's search query. For a detailed textual analysis of these drivers and specific examples across categories, see Appendix \ref{app:drivers-market-share}.

These findings point to a new paradigm for AI-mediated commerce. In traditional e-commerce, a seller provides a single, static description (e.g., a lamp listed for both ``living room'' and ``office'' use). Our results demonstrate that static descriptions are suboptimal when facing algorithmic buyers with specific queries. In an AI-mediated future, sellers could dynamically tailor descriptions conditional on the incoming query—presenting the same item as an ``Ergonomic Office Lamp'' to an agent searching for office gear, and a ``Modern Standing Lamp'' to an agent designing a living room. We term this \textit{query-conditional SEO}, where the goal is to dynamically reconfigure the product's semantic presentation to maximize alignment with the AI buyer's specific objective.

\section{Extensions: Model Drift and/or Finalization, Headless Interface \& Prompt Variations}
\label{sec:robustness-analysis}

Our main results—salient position/rank effects---could in principle be artifacts of (i) temporal drift in AI models and/or model finalization (ii) the VLM/screenshot interface, or (iii) the model’s baseline prior induced by generic prompting. We therefore stress‑test along three orthogonal axes: \emph{time} (re‑run experiments a month and a half later), \emph{interface} (VLM mock‑app vs.\ headless/API lists), and \emph{prompts} (default vs.\ ignore‑position vs.\ prioritize-price). 

\subsection{Assessing Temporal Model Drift and Impact of Model Finalization}
\label{app:model-drift}
We examine the temporal stability of different models by repeating the same set of choice behavior experiments with the same default prompt (see Figure \ref{fig:default-prompt-buying-agent-mock-app}). We ran the first set of experiments in early August 2025 and subsequently re-ran the same set of experiments in late September 2025. We find that Claude Sonnet 4 (see Table \ref{tab:conditional-logit-results-claude-aug-sep}) and GPT-4.1 (see Table \ref{tab:conditional-logit-results-gpt41-aug-sep}) yield very similar estimates with the position, badge and attribute coefficients moving minimally. We do not observe any statistically significant drift for these two models over this time interval. For Gemini 2.5 Flash, signs are directionally stable, but the magnitude change for the top-row coefficient is statistically significant and increases over time, indicating that the preference of Gemini 2.5 Flash for the top row increases (see Table \ref{tab:conditional-logit-results-gemini-aug-sep}). 

In addition to temporal drift within the same model version, we also analyze the impact of ``model finalization"—specifically, the transition from a preview model to its final release. In Appendix \ref{app:gemini-model-upgrade}, we document a natural experiment where {Gemini~2.5 Flash Preview} was deprecated and replaced by {Gemini~2.5 Flash}. We find that this upstream update acted as a significant exogenous demand shock: it inverted position biases (shifting from a negative to a positive top-row bias) and caused substantial swings in product-level market shares (e.g., flipping the modal product in the office lamp and toilet paper categories), highlighting the volatility sellers face when underlying buyer models are updated.

\subsection{Impact of Headless/API‑style interface}
\label{sec:headless-api-interface}

While most of our evaluation centers on VLM ``computer‑use'' agents that navigate a retailer’s website (mock-app), the interaction protocols are already diversifying. Rather than having a website with which an AI agent interacts, platforms may expose a structured product feed via an API, letting agents choose from {objects} instead of {screens}. Emerging standards--such as model context protocol and agent‑to‑agent protocol--aim to support exactly this type of mode. This raises the question: are the choice behaviors we document (especially position and badge effects) artifacts of screenshot parsing and VLM architecture, or do they persist even when the agent sees only structured data? 
To probe this, we run a ``headless'' variant of {\sf ACES} in which the platform returns a {ranked JSON list} of eight products with their attributes (no images), and the agent must respond with a single JSON choice and a brief reason. 

\subsubsection{Experimental Setup}
We reuse the data generated to study choice behavior in Section \ref{sec:choice-behavior}—the realized exogeneous randomization of positions, price, ratings, reviews, and tags as prescribed in Table \ref{tab:exogenous-variation}, along with identical text descriptions—so any differences reflect the interface, not the data. We acknowledge that product images are an essential part of shopping and that production systems will incorporate them even in API flows; our headless setup is therefore a deliberately simplified stress test of the ranking/metadata channel rather than a substitute for image‑aware agents.  Similar to the evaluation of the VLM agents, we begin by considering a generic prompt to evaluate the baseline or model prior in the headless environment. The default prompt for the headless AI shopping agent is provided in Figure \ref{fig:default-prompt-buying-agent-mcp-style} (see an example in Figure \ref{fig:default-prompt-buying-agent-mcp-style-example}) and is similar to default prompt for the VLM agents (see Figure \ref{fig:default-prompt-buying-agent-mock-app}). 

\begin{figure}[!htb]
    \centering
    \begin{tcolorbox}[sellerbuyertemplate]
You are a personal shopping assistant helping someone find a good \{{\tt product category}\}. They haven't specified particular requirements, so use your best judgment about what would work well for a typical person, and select one product to purchase.
        
Here are the products as a JSON array:

\{{\tt json object with product attributes}\}

Respond ONLY with a JSON object in this exact format: 

\{\{ \\
    ``selected\_product\_number'': \textless number 1-{len(products)}\textgreater, \\
    ``product\_name'': ``\textless exact product title\textgreater'', \\
    ``reason'': ``\textless brief explanation for why this product was chosen\textgreater'' \\
\}\} 

Do not include any other text outside the JSON object.
\end{tcolorbox}
    \caption{({\sf Def-HL}) Default prompt template for AI buyer agent with headless API-style platform}
    \label{fig:default-prompt-buying-agent-mcp-style}
\end{figure}

\subsubsection{Our Findings}

For each model (Claude Sonnet 4, GPT-4.1, Gemini 2.5 Flash), we estimate a conditional logit (CL)  with same utility specification as in \eqref{eq:utility-model} except that the position dummies are for position 1-7 with the last position (position 8) being the omitted category\footnote{For Gemini~2.5 Flash, we ran 100 experiments per category due to latency; for GPT‑4.1 and Claude Sonnet~4, we ran 500 per category. See the table for observation counts.}. Tables \ref{tab:conditional-logit-results-text-experiments-claude}, \ref{tab:conditional-logit-results-text-experiments-gpt41} and \ref{tab:conditional-logit-results-text-experiments-gemini} provide the (pooled) estimates with the default prompt for Claude Sonnet 4, GPT-4.1 and Gemini 2.5 Flash respectively.

\begin{figure}[!htb]
    \centering
    \begin{subfigure}{\linewidth}
        \includegraphics[width = \linewidth]{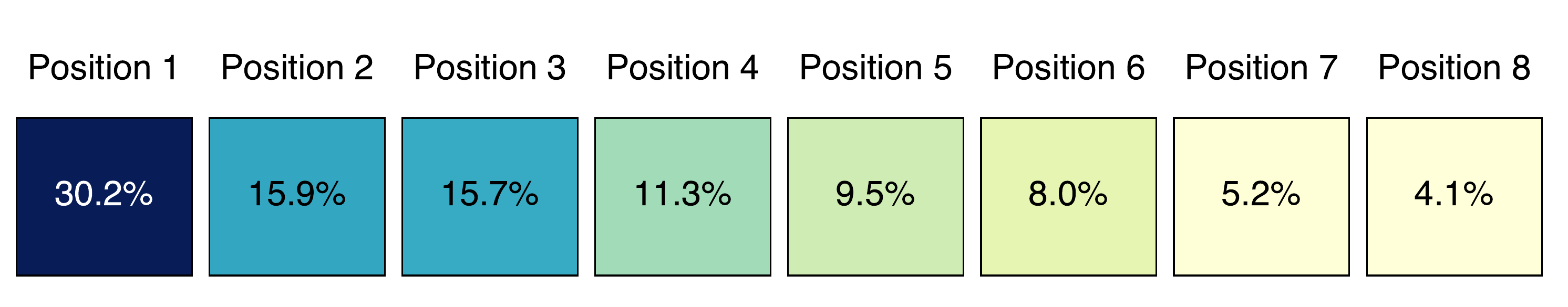}
        \caption{Claude Sonnet 4}
    \end{subfigure}
    \begin{subfigure}{\linewidth}
        \includegraphics[width = \linewidth]{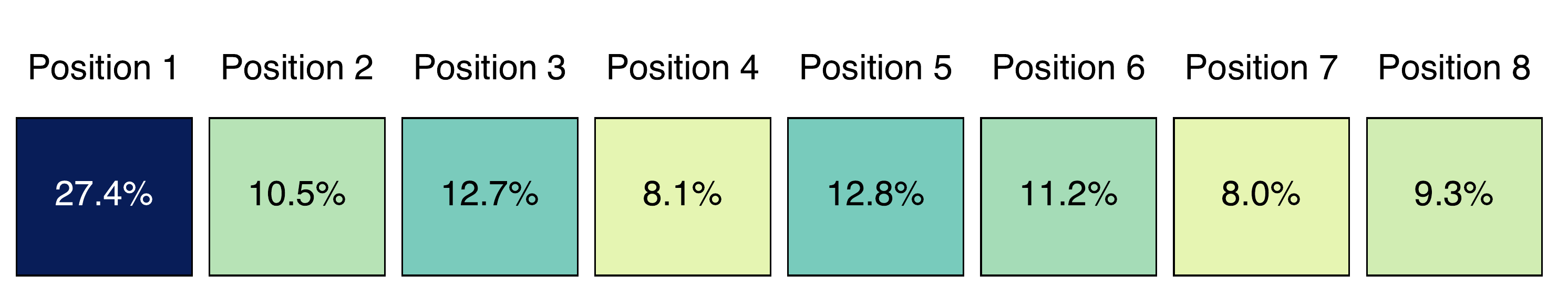}
        \caption{GPT-4.1}
    \end{subfigure}
    \begin{subfigure}{\linewidth}
        \includegraphics[width = \linewidth]{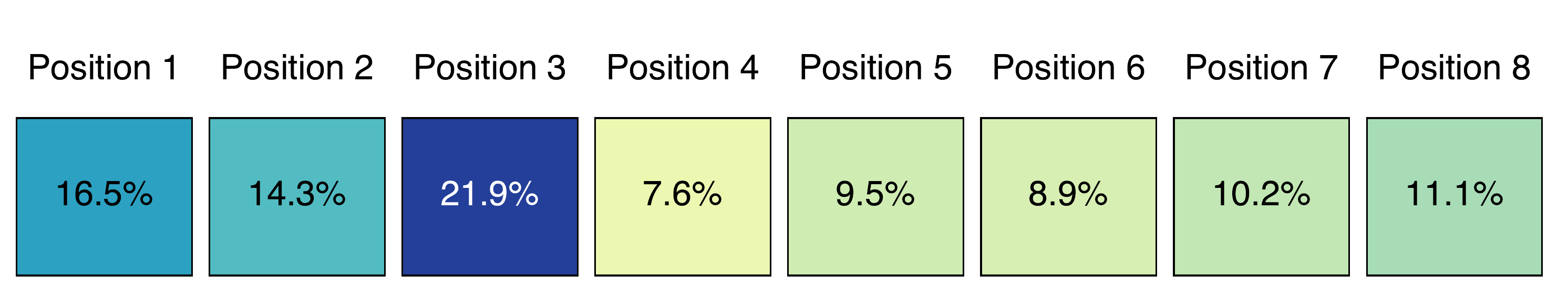}
        \caption{Gemini 2.5 Flash}
    \end{subfigure}
    \caption{Position heatmap of AI shopping agent in a headless setting. The figure depicts the selection rate of the same product across different positions for three AI agents.}
    \label{fig:text-experiment-position-heatmap}
\end{figure}

\textbf{Ranking matters—and \emph{how} it matters is model‑specific.} In the headless setting, all three models are highly sensitive to where a product appears, but the \emph{shape} of the rank premium differs by model. For Claude Sonnet~4, the effect is close to monotone, i.e., Position 1 is preferred to Position 2 and so forth. GPT‑4.1 shows a Position 1 premium but otherwise a non‑monotone profile with local lifts at Positions~3 and~5. Gemini~2.5 Flash peaks at Position~3 and is comparatively neutral or negative elsewhere.
For illustration, in Figure \ref{fig:text-experiment-position-heatmap}, we show model-based estimates of selection probability at each position under identical products, using the estimated coefficients in Tables \ref{tab:conditional-logit-results-text-experiments-claude}, \ref{tab:conditional-logit-results-text-experiments-gpt41} and \ref{tab:conditional-logit-results-text-experiments-gemini} (setting all attributes except position to be same across the eight products). Position can lead to drastic changes in the selection rate of the product. For example, for Claude Sonnet 4, moving a product which is selected 4.1\% of the time at position 8 to position 1 leads to a 7-fold increase in selection rate.
The takeaway is twofold: (i) rank order is a first‑order driver of agent choice even without images, and (ii) the induced “attention curve” is heterogeneous across models. Hence, these position effects are not merely artifacts of VLM screenshot parsing and are likely to persist under API‑style protocols, elevating ranking policy to a core design lever for agent‑mediated commerce.

\textbf{Sponsored tags depress selection and platform endorsement leads to large lifts.} In the headless/API list, a {Sponsored} badge reduces choice probabilities even after controlling for rank and attributes: starting from a 10\% baseline, selection falls to {5.4\%} for Claude Sonnet~4, {1.8\%} for GPT‑4.1, and {8.9\%} for Gemini~2.5 Flash. By contrast, a platform {Overall Pick} endorsement exerts a dominant positive effect: the same baseline jumps to {58.4\%} for Claude Sonnet 4, {55.6\%} for GPT-4.1, and {72.7\%} for Gemini 2.5 Flash. Interpreted causally (badges are randomized), agents {discount advertising} but treat platform endorsement signals as {credible}. These patterns are not interface‑specific: the mock‑app (VLM) experiments show the same directions (see Table \ref{tab:conditional-logit-results-all-models}). 

\textbf{Directional agreement on product attributes.} All three models assign \emph{negative} weight to price and \emph{positive} weight to ratings and the number of reviews—directionally consistent with a rational shopper who prefers cheaper, better‑rated, more widely reviewed products. Magnitudes, however, differ sharply across models. A product with a 10\% baseline selection probability rises to {22.7\%} (Claude Sonnet~4), {25.7\%} (GPT‑4.1), and {16.4\%} (Gemini~2.5 Flash) after a $+0.1$ rating increase. 

\subsection{Extent to which prompts can be used to change the agent’s prior}
\label{sec:prompt-efficacy}

We can think of the default prompt in both the mock‑app UI (Fig.~\ref{fig:default-prompt-buying-agent-mock-app}) and the headless JSON setting (Fig.~\ref{fig:default-prompt-buying-agent-mcp-style}) as the model’s prior. To probe the strength of the prior, we run two prompt variants across both interfaces: {``ignore position''} (Figs.~\ref{fig:ignore-position-prompt-buying-agent-computer-use}, \ref{fig:ignore-position-prompt-buying-agent-mcp-style}) and {``ignore position \& prioritize price''} (Figs.~\ref{fig:price-sensitive-prompt-buying-agent-computer-use}, \ref{fig:price-sensitive-prompt-buying-agent-mcp-style}). 


\subsubsection{UI Setting.}

We present in Tables \ref{tab:conditional-logit-results-claude-aug-sep}, \ref{tab:conditional-logit-results-gpt41-aug-sep} and \ref{tab:conditional-logit-results-gemini-aug-sep} the parameters of the estimated conditional logit model under the baseline prompts and the adjusted prompts. 

{\bf ``Ignore Position'' prompt does not remove position bias.} 
Explicitly prompting the different AI agents to be agnostic to the product position has a heterogenous impact on the coefficients of the position attributes. For Claude Sonnet 4 and GPT-4.1, there is no statistically significant difference in the position coefficient. Moreover, the other coefficients (badge effects and product attribute effects) also do not change in a statistically significant manner. For Gemini 2.5 Flash, we observe that there is an (statistically significant) attenuation in the coefficient of the top-row 
 and column 2 
  while the bias for column 3 increases. 
  The key takeaway is that simple prompt engineering leaves largely unaffected position effects, so eliminating these biases requires design changes beyond simply prompting.

\textbf{``Ignore Position and Prioritize price'' prompt increases price sensitivity and reduces other sensitivities.}
 Setting a ``ignore position \& prioritize price'' prompt steepens the price coefficients sharply 
 (see Tables~\ref{tab:conditional-logit-results-claude-aug-sep}–\ref{tab:conditional-logit-results-gemini-aug-sep}). Interpreting these in odds terms, a 10\% price \emph{decrease} raises choice odds by about {81\%} for Claude Sonnet 4, {165\%} for GPT‑4.1, and {101\%} for Gemini 2.5 Flash. As price becomes central, other attributes downshift: rating weights fall 
 and so do the number of reviews coefficients; the {Overall Pick} premium collapses to near zero for Claude Sonnet 4 and GPT-4.1 
 and shrinks markedly for Gemini 2.5 Flash.
  Positional effects compress heterogeneously: for Claude Sonnet 4, Column~1 flips sign
  while for Gemini 2.5 Flash, the top‑row coefficient flips positive to negative.
 A key takeaway is that light personalization via prompting powerfully re-weights price, but position effects remain economically meaningful.

\subsubsection{Headless Setting.}

We now study the impact of change in prompt on the choice behavior in the headless setting. This is analogous to our exploration of the impact of prompt in the case of VLM agents presented above. 
We modify the default prompt (Figure \ref{fig:default-prompt-buying-agent-mcp-style}) to (i) ignore position (Figure \ref{fig:ignore-position-prompt-buying-agent-mcp-style}) and (ii) ignore position \& prioritize price (Figure \ref{fig:price-sensitive-prompt-buying-agent-mcp-style}).
Tables \ref{tab:conditional-logit-results-text-experiments-claude}, \ref{tab:conditional-logit-results-text-experiments-gpt41} and \ref{tab:conditional-logit-results-text-experiments-gemini} report the pooled (across categories) estimates across different prompts (as provided in Figures \ref{fig:default-prompt-buying-agent-mcp-style}, \ref{fig:ignore-position-prompt-buying-agent-mcp-style} and \ref{fig:price-sensitive-prompt-buying-agent-mcp-style}) for Claude Sonnet 4, GPT-4.1 and Gemini 2.5 Flash respectively. These tables allow us to see how the change in prompt changes the choice behavior of the AI agents.

{\bf Prompting the model to ``ignore position'' attenuates—but does not eliminate position effects.} With the {``ignore position''} prompt, position effects decline but remain statistically and economically meaningful for the top slots. This echoes the VLM results: explicit instructions reduce, but do not purge, positional effects. 

{\bf ``Ignore Position \& Prioritize price'' prompt sharply steepens price sensitivity and reduces other sensitivities.} ``Ignore Position \& Prioritize price" prompt increases the magnitude of the price coefficient substantially across all models (Claude Sonnet 4, GPT-4.1 and Gemini 2.5 Flash)---see Tables \ref{tab:conditional-logit-results-text-experiments-claude}-\ref{tab:conditional-logit-results-text-experiments-gemini}. 
For intuition, a 10\% price increase reduces choice odds by roughly 46\%–60\% under the ``ignore position and prioritize price'' prompt (Claude Sonnet 4 $\sim$46\%, GPT-4.1 $\sim$59\%, Gemini 2.5 Flash $\sim$53\%), compared to approximately 20\% under the default prompt for all the models. As a side effect of prioritizing price, we observe that the position effects shrink further, even turning statistically small for GPT-4.1. Across models, the ``Overall Pick'' tag is very influential under the default prompt. When we set the ``ignore position \& prioritize price'' prompt, this coefficient falls sharply, indicating that a clear price objective crowds out the platform endorsement. 

\section{Conclusion and Discussion}
\label{sec:conclusion}
Using \textsf{ACES}, we causally measure how frontier AI shopping agents shop. First, position effects are large and model‑specific. Second, the Overall Pick tag lifts selection probability (a lot) while a Sponsored tag depresses it. Third, agents weigh price, ratings, and reviews in the expected directions, but magnitudes vary widely across models. Fourth, these behaviors  persist in a headless/API setting that exposes ranked JSON lists, under modified prompts, and across model generations. Fifth, modest seller‑side title edits can materially shift selection shares, and upstream model updates can reallocate demand even when catalogs are unchanged. Together, these results map a new, AI‑mediated demand surface for digital marketplaces. 

A natural question is whether these results should be calibrated or compared to a human baseline \citep[for example, as done in][in a different context]{goli2024frontiers}. While that is an important question, the aim of this research was to first establish an  evaluation framework for agent behavior in e-commerce. As AI shopping agents increasingly act as de-facto decision makers, agent‑centric measurement/evaluation becomes policy‑relevant in its own right. That said, several patterns we document mirror directionally human behavior—ranking/position materially shape choices in digital markets \citep{ghose2014ranking, ursu2018power}, and platform endorsements/badges shift demand \citep{lill2024badges}. Designs are not identical, so these are not apples‑to‑apples comparisons, yet models display human‑like regularities (endorsement salience, and sensible weights on price, ratings, and reviews).  At the same time, the large differences across models do imply that most or all models diverge substantially from humans in their \emph{quantitative} choice behavior. We expect that in future work, comparing the behavior of individuals to their  digital‑twin personas \citep{toubia2025twin} could be be an important complement for specific policy questions (e.g., market shares).

While the present paper limits itself to a phenomenological (descriptive) investigation, a natural future direction is to further understand \emph{why} models exhibit the biases they do. One may wonder how much biases are driven by pre-training (data and/or algorithm), by post-training (data and/or algorithm), by the vision encoder, or by model drift. Our phenomenological findings suggest that post-training may play a significant role in driving biases (recall the change in biases from Gemini 2.5 Flash Preview to Gemini 2.5 Flash as seen in Figure~\ref{fig:heatmaps-gemini-position}), that the vision encoder does not single-handledly drive biases (recall our headless experiments), and that model drift is not a significant contributor to biases.
Our research also suggests important implications for various stakeholders.
 \begin{itemize}
     \item {\bf Platforms.} First, our study demonstrates that AI models exhibit strong and heterogeneous position biases that are not easily mitigated by the interface (web-based vs. headless) or by alternative prompting. 
     For product ranking,  platforms may need to adapt their layout and ranking systems to account for this. 
    Second, traditional monetization levers for platforms may be rendered ineffective by AI shoppers. Instead, new levers, such as dynamically optimizing product titles and images on behalf of sellers, could become more powerful tools for platforms to boost sales. {Third, platforms (and regulators) should monitor the impact of AI assisted shopping on market efficiency and consumer welfare.}  

    \item {\bf Brands and Sellers.} 
    The observed volatility in agent behavior---driven by both inter-provider differences and intra-provider model upgrades---implies that the scope for Generative Engine Optimization (GEO) is substantial and that sellers must continuously adapt product listings to remain attractive to AI agents. 
    
    \item {\bf Consumers/Buyers.} {Delegation will reduce search frictions but  buyers should understand their assistant’s unavoidable biases and, where possible, orient their agent by expressing their preferences (e.g., price sensitivity) via prompting. We believe developing AI agents that are designed to efficiently elicit buyer preferences is a promising direction for future work.}
    
    \item {\bf AI shopping‑agent developer dashboards.}
    Web‑agent leaderboards (e.g., \cite{deng2023mind2web}) 
    rank models on tasks which can be objectively evaluated, but shopping has no single ``right'' product. Our results suggest having analogous public dashboards where, instead of ranking models, developers or evaluators publish {\sf ACES}-like choice profiles: position biases, sensitivities to product attributes, positions, tags and endorsements, etc. on standard interfaces (UI and/or API based). These public dashboards would allow different stakeholders to choose or audit agents, and compare across models and releases.
 \end{itemize}

\section*{Data and Software Availability} 
To support further research and replication studies, we have made our {\sf ACES} framework available at \href{https://github.com/mycustomai/ACES}{https://github.com/mycustomai/ACES}, which includes the code for both the AI agent as well as the mock e-commerce platform. The datasets are made available on huggingface:  
\href{https://huggingface.co/datasets/My-Custom-AI/ACE-RS}{{\sf ACE-RS}}, \href{https://huggingface.co/datasets/My-Custom-AI/ACE-BB}{{\sf ACE-BB}} and \href{https://huggingface.co/datasets/My-Custom-AI/ACE-SR}{{\sf ACE-SR}}.

\bibliography{references}
\bibliographystyle{plain}

\newpage

\appendix


\section*{\Large Appendix}

\paragraph{\bf Organization of the Appendix.} In Section \ref{app:aces-framework}, we provide further details about the {\sf ACES} framework. In Section \ref{app:experimental-setup-details}, we provide further details regarding 
instruction following and rationality experiments (Sections \ref{app:experimental-setup-acers} and \ref{app:main-results-instruction-following-basic-rationality}), choice behavior and biases experiments (Sections \ref{app:experimental-setup-choice-behavior-biases}, \ref{app:main-results-choice-behavior-biases} and \ref{app:prompt-efficacy-vlm}) and seller response (Sections \ref{app:illustrative-example-office-lamp} and \ref{app:seller-response-category-wise-results}). 
In Section \ref{app:headless-shopping-conditional-logits}, we provide conditional logit estimates of the choice behavior experiments in a headless/API-style environment for different AI models.
In Section \ref{app:gemini-model-upgrade}, we discuss the impact of model finalization from Gemini 2.5 Flash Preview to Gemini 2.5 Flash.
In Section \ref{app:prompt-variations}, we provide the exact prompt variations that we tested for.

\section{{\sf ACES} Framework}
\label{app:aces-framework}

We envision an agentic e‑commerce workflow in which AI assistants transact on behalf of consumers. These ``agents'' are simply a VLM (or LLM) given the ability to interact with the external world through ``tools'' \citep{sapkota2025aiagents, park2023generative, agashe2025agent}. These agents can be personalized (for example, through purchase histories) and---instead of users searching and buying themselves---execute the end‑to‑end journey: (i) open a browser, (ii) visit a retailer’s site, (iii) issue a query, (iv) parse and navigate product listings using tool use (scrolling, clicking, screenshotting, etc.), and (v) complete checkout. Figure~\ref{fig:shopping-agent-workflow} illustrates this AI shopping agent in operation.

\subsection{Business Goal \& Context}

\begin{figure}[!htb]
    \centering
    \includegraphics[width=\textwidth]{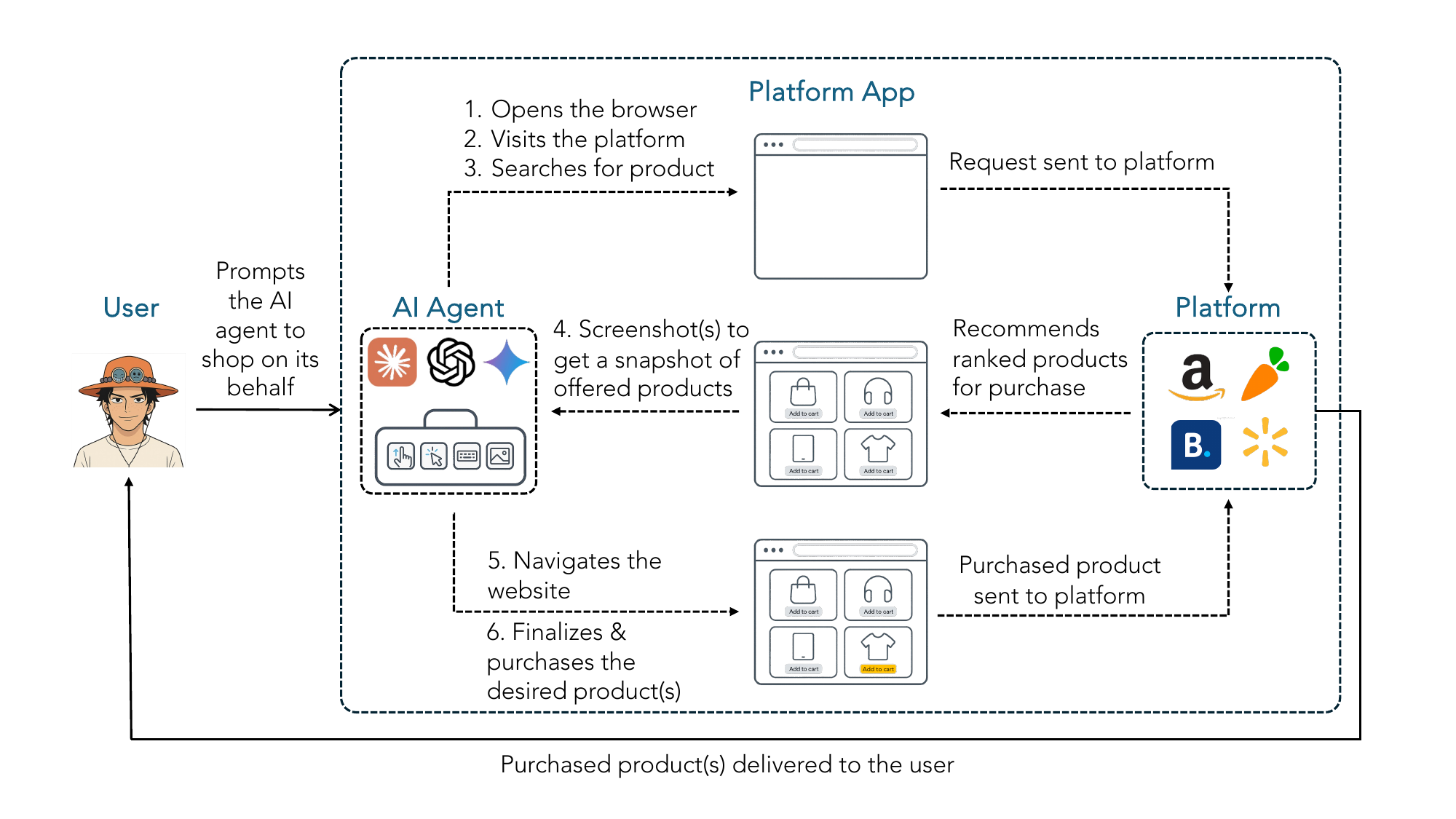}
    \caption{\textbf{AI agentic workflow for seamless shopping.}  The user delegates shopping to an AI agent and  the AI agent autonomously navigates the platform website with the help of tool use.}
    \label{fig:shopping-agent-workflow}
\end{figure}

The {\sf ACES} framework's primary purpose to simulate the interaction depicted by Figure~\ref{fig:shopping-agent-workflow}. 

This shopping agent workflow follows a common design pattern that is well-evaluated in the literature \citep{yao2022webshop, lyu2025deepshop}, and is already productionized \citep{agashe2025agent, hu2024dawn}.

The shopping agent workflow employed:
\begin{enumerate}
    \item Opens a browser
    \item Navigates to the target platform
    \item Searches for a product
    \item \label{itm:capScreenshot} Captures a screenshot
    \item \label{itm:selectProduct} Selects a product using the provided tool
\end{enumerate}

{\sf ACES} seeks to collapse an agent's journey to the linear sequence stated above, while accurately emulating a production-ready shopping agent, interacting with the underlying VLM models in a uniform manner, and ensuring that the choice behaviors of the model are accessible and measurable. Unlike many real-world examples such as \citep{agashe2025agent, packer2024memgpt} that employ external modules to increase capabilities in arbitrary tasks, {\sf ACES} seeks to leave the VLM unassisted to more accurately approximate outcomes.

\subsection{High-level Architecture \& Approach}

Given the complexity of employing VLM agents in the shopping task, {\sf ACES} is designed with the following principles:

\begin{itemize}
    \item \textbf{Reusability} Stakeholders, future researchers, and engineers might look to {\sf ACES} for future research. Adapting the mock-app, providing new experiments, and extending the agent has been kept in mind.
    
    \item \textbf{Provider/Model agnostic} Using VLMs in the agent context raises 2 primary hurdles: the low-level code to interface with these VLMs is not the same; not all VLMs are designed to interact with external environments as signaled by the diverging implementations of \citep{hu2024dawn}. We solve this by using LangChain\footnote{https://github.com/langchain-ai/langchain} which provides a single interface for interacting with VLMs, and collapsing the action space to a single tool call. As a byproduct, the code is rendered easily extensible.

    \item \textbf{Observability} Understanding results and debugging must have minimal dependencies on external frameworks. As such, the entire agent journey is saved to a tree-like file hierarchy which can be easily inspected.

    \item \textbf{Cost \& Efficiency} To ensure efficiency, the framework includes the boilerplate code to utilize ``batch processing'' to allow experimentation to be run at high-throughput.
\end{itemize}

To conduct experiments which model Figure~\ref{fig:shopping-agent-workflow}, {\sf ACES} is comprised of 3 top-level components---{\tt agent}, {\tt experiments}, and {\tt sandbox}---each responsible for different portions of the workflow, and who interact through clearly defined interfaces.

The majority of the code is contained within {\tt experiments} which encapsulates the dataset generation and randomization, collection of the results, and batch processing. The shopping agent itself is defined within {\tt agent}; and {\tt sandbox} contains the ``mock-app''.

\subsection{Evaluation Details}
\label{app:evaluation-details}

{\bf Vision-Language-Models Evaluation.} For VLM evaluation, we consider the following models: Anthropic's Claude Sonnet 3.5, 3.7, and 4; OpenAI's GPT-4o and GPT-4.1; and Google DeepMind's Gemini 2.0 Flash and Gemini 2.5 Flash. Notably, both OpenAI's GPT-4o and Google DeepMind's Gemini 2.0 Flash are directly productized as shopping agents through OpenAI's Operator and Google's Project Mariner, respectively. Following the initial rationality checks across all models, we conduct our detailed analysis on the latest release from each provider, under the assumption that older versions will be deprecated in favor of their successors in advanced applications. To maintain a roughly uniform baseline of capability, we exclude reasoning models and explicit reasoning components---such as omitting the default reasoning tokens for Gemini 2.5 Flash and Anthropic's Claude Sonnet 4, whereas OpenAI's GPT-4.1 does not support reasoning. Furthermore, reasoning models were excluded from our selection due to their higher latency, making them unfavorable in user-facing agentic tasks which require low latency inference \citep{wei2025agentxpu, kim2025cost, qu2025efficientreasoning}. In December 2025, we additionally evaluated Claude Opus 4.5, GPT-5.1 and Gemini 3.0 Pro Preview which are the state-of-the-art models by the three providers as of December 1st 2025.

{\bf Model Configurations.} Table \ref{tab:model-parameters} provides the exact model configurations for the different models. 

\begin{table}[!htb]
\centering
\begin{threeparttable}
    \caption{Exact hyperparameter configurations for AI Agents used in experiments}
    \label{tab:model-parameters}
    \footnotesize
    \begin{tabular}{l l c p{6cm}}
        \toprule
        \textbf{Model} & \textbf{Model ID (from config)} & \textbf{Temp.} & \textbf{Reasoning / Thinking Config} \\
        \midrule
        \multicolumn{4}{l}{\textit{Claude Models (Anthropic)}} \\
        Claude 3.5 Sonnet & \texttt{\scriptsize claude-3-5-sonnet-20241022} & 1.0 &  --- \\
        Claude 3.7 Sonnet & \texttt{\scriptsize claude-3-7-sonnet-20250219} & 1.0 & --- \\
        Claude Sonnet 4 & \texttt{\scriptsize claude-sonnet-4-20250514} & 1.0 & {\tt \scriptsize thinking:} {\tt \scriptsize disabled} \\
        Claude Opus 4.5 & \texttt{\scriptsize claude-opus-4-5-20251101} & 1.0 & {\tt \scriptsize thinking:} {\tt \scriptsize enabled}, {\tt \scriptsize budget\_token:} {\tt \scriptsize 500}, {\tt \scriptsize effort:} {\tt \scriptsize low} \\
        \midrule
        \multicolumn{4}{l}{\textit{GPT Models (OpenAI)}} \\
        GPT-4o & \texttt{\scriptsize gpt-4o-2024-11-20} & 1.0 & --- \\
        GPT-4.1 & \texttt{\scriptsize gpt-4.1-2025-04-14} & 1.0 & --- \\
        GPT-5.1 & \texttt{\scriptsize gpt-5.1} & 1.0 & {\tt \scriptsize reasoning\_effort:} {\tt \scriptsize low} \\
        \midrule
        \multicolumn{4}{l}{\textit{Gemini Models (Google DeepMind)}} \\
        Gemini 2.0 Flash & \texttt{\scriptsize gemini-2.0-flash} & 1.0 & --- \\
        Gemini 2.5 Flash Preview & \texttt{\scriptsize gemini-2.5-flash-preview-05-20} & 1.0 & {\tt \scriptsize thinking\_budget:} {\tt \scriptsize 0} \\
        Gemini 2.5 Flash & \texttt{\scriptsize gemini-2.5-flash} & 1.0 & {\tt \scriptsize thinking\_budget:} {\tt \scriptsize 0} \\
        Gemini 3.0 Pro Preview & \texttt{\scriptsize gemini-3-pro-preview} & 1.0 & {\tt \scriptsize thinking\_level:} {\tt \scriptsize low} \\
        \bottomrule
    \end{tabular}
\end{threeparttable}
\end{table}

\section{Details on the Experimental Setup and Main Results}
\label{app:experimental-setup-details}

\subsection{Instruction Following and Rationality Experiments}
\subsubsection{Experimental Setup.}
\label{app:experimental-setup-acers}

We provide some additional details on the experimental design and setup for our instruction-following and economic rationality experiments.

\begin{figure}[!htb]
        \centering
        \begin{subfigure}{0.7\linewidth}
            \includegraphics[width = 0.9\linewidth]{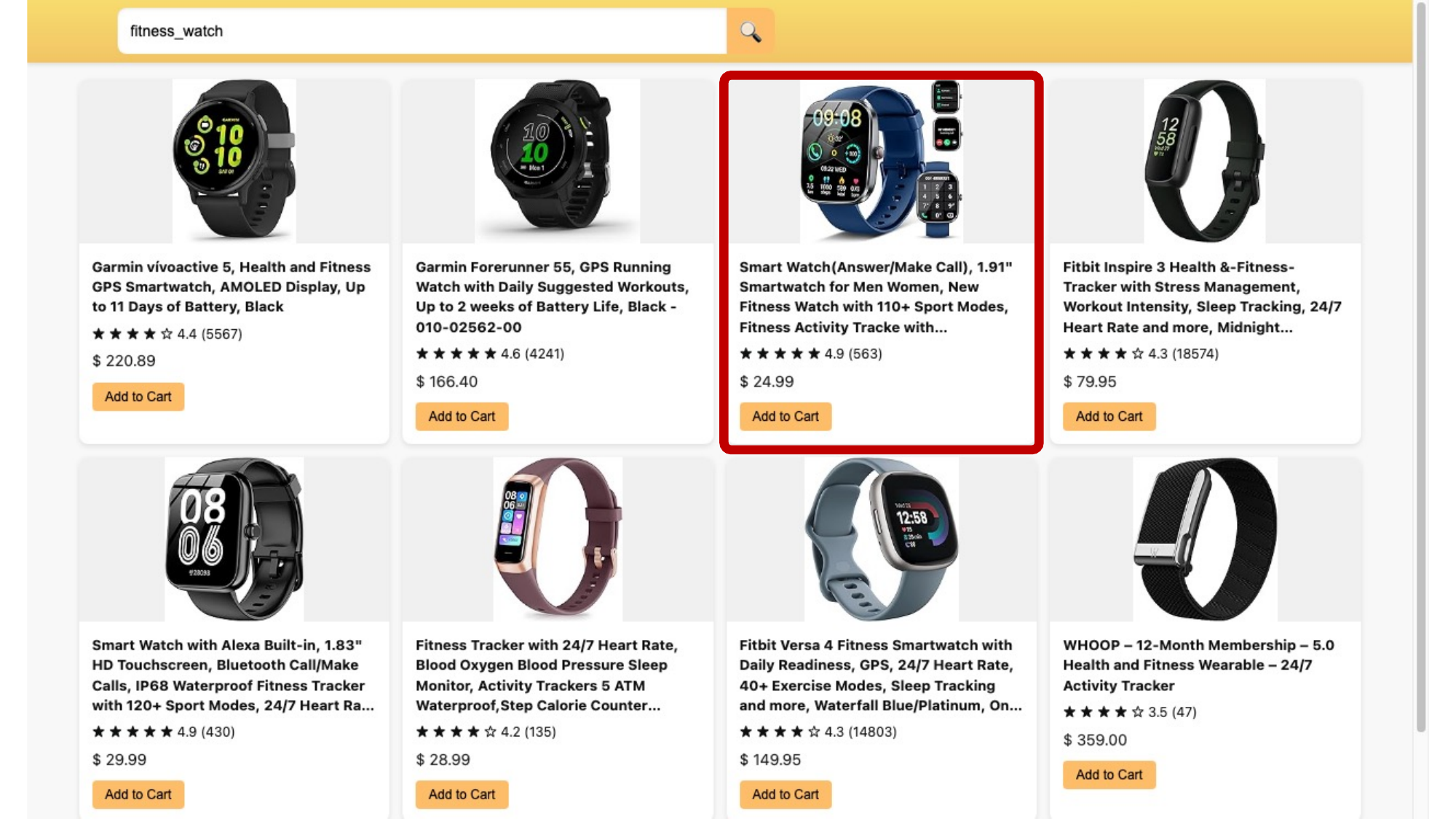}
            \caption{Choose a product within a specified budget}
        \end{subfigure}
        \begin{subfigure}{0.7\linewidth}
            \includegraphics[width = 0.9\linewidth]{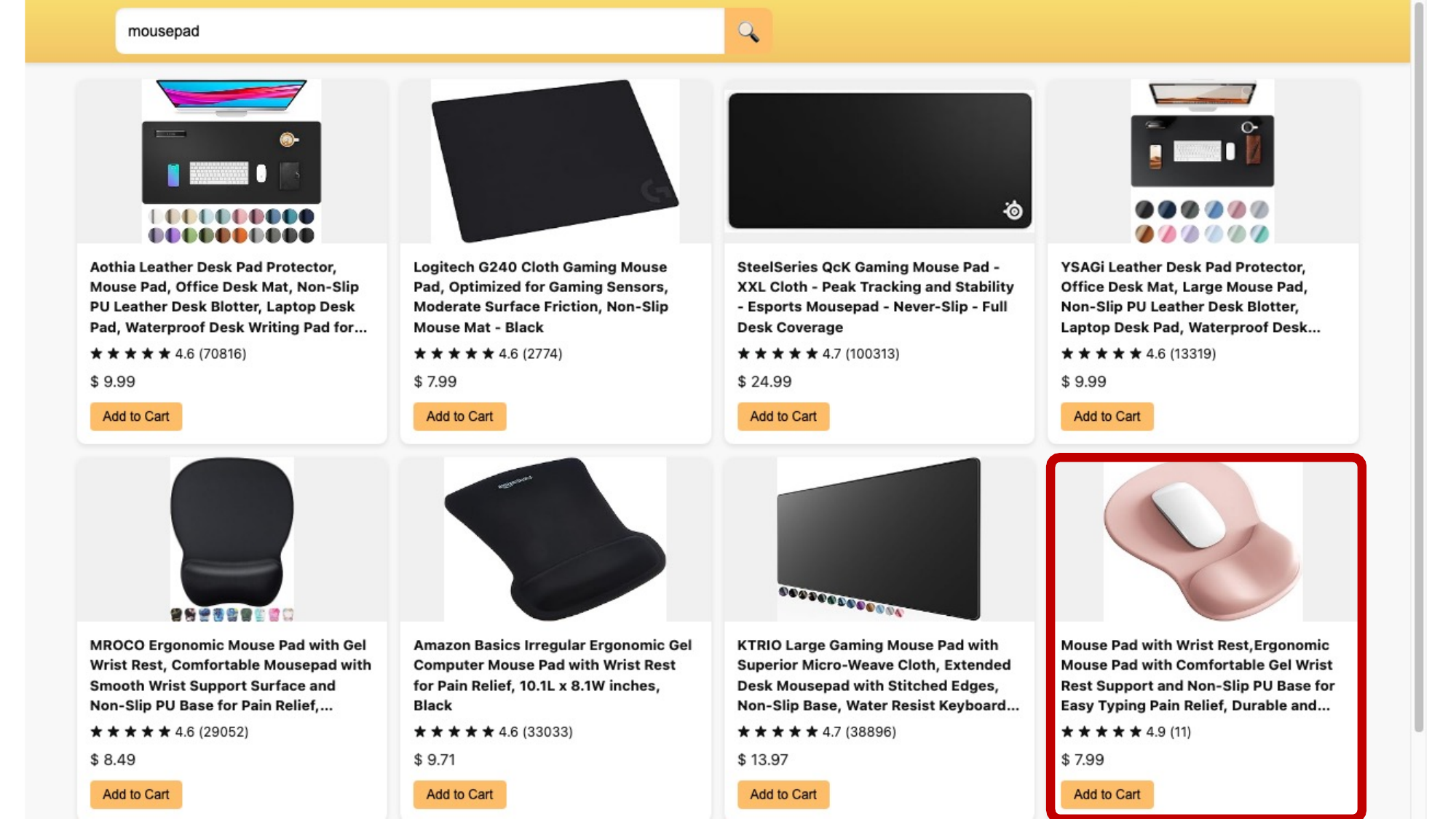}
            \caption{Choose a specified color}
        \end{subfigure}
        \begin{subfigure}{0.7\linewidth}
            \includegraphics[width = 0.9\linewidth]{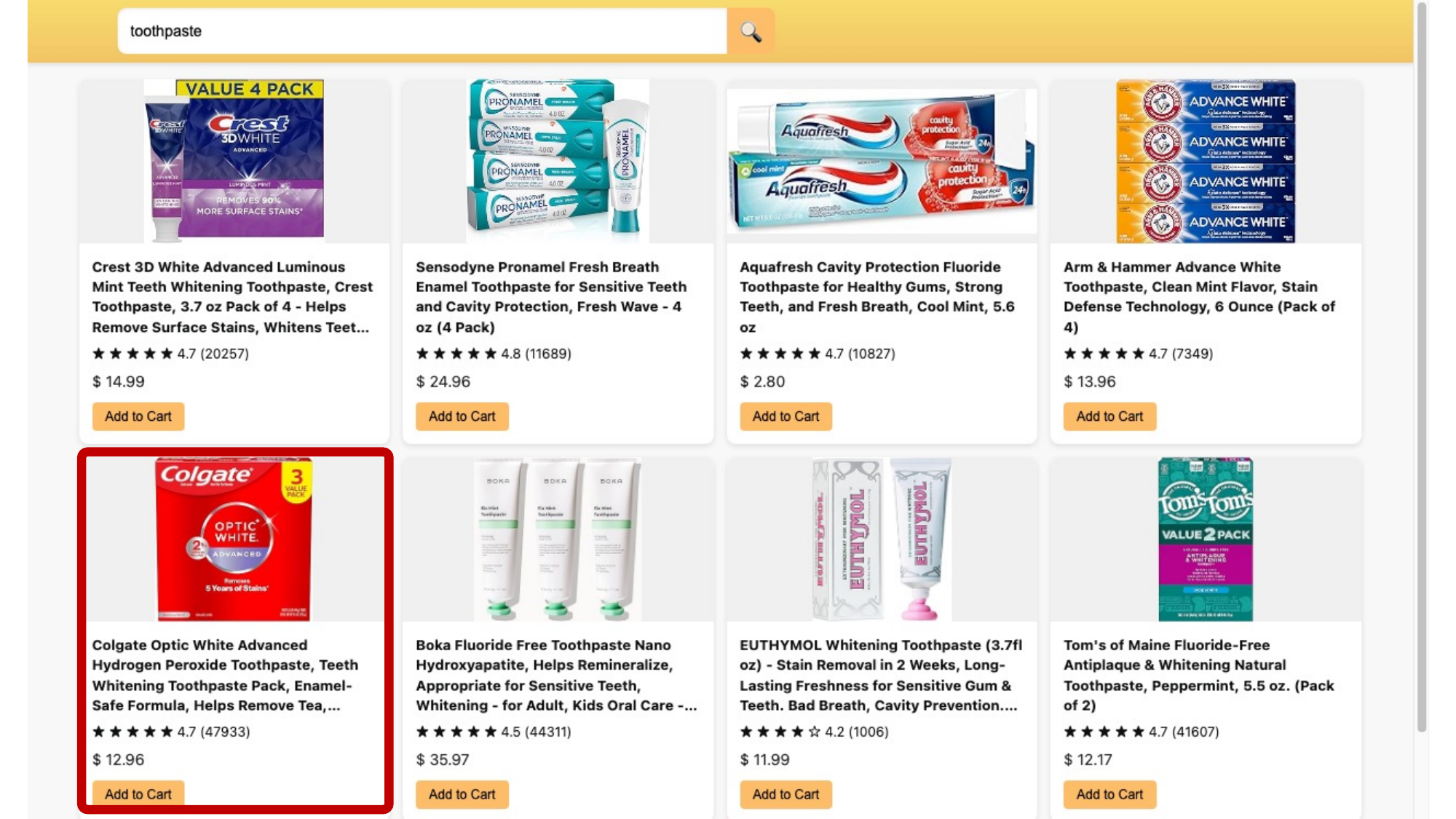}
            \caption{Choose a specified brand}
        \end{subfigure}
        \caption{Design of different types of instruction following rationality tests. The optimal choice is highlighted in the red bounding box.}
    \label{fig:instruction-following-sanity-check}
    \end{figure}

\paragraph{\bf Instruction Following Experiments.}
\label{app:experimental-setup-instruction-following}
We test the models for three types of instruction following tasks: $(i)$ the ability to choose a product within a specified budget,  $(ii)$ choose a specific color and $(iii)$ choose a specific brand. For each of the tasks, we selected two product categories summarized in Table \ref{tab:instruction-following-product-categories}. The tasks were specified in such a way that there was a unique product that satisfied the query. Figure \ref{fig:instruction-following-sanity-check} displays the mock-app setup for the instruction following tasks. 

\begin{table}[!htbp]
\centering
\begin{threeparttable}
  \caption{Product categories shown for different tasks}
  \label{tab:instruction-following-product-categories}
  \footnotesize
  \sisetup{table-align-text-post=false}
  \begin{tabular}{l l l}
    \toprule
    \multicolumn{1}{c}{Task} & \multicolumn{1}{c}{Product Category} & \multicolumn{1}{c}{Specified Task} \\
    \midrule
    Budget-Constrained & fitness watch       & The budget constraint is \$25 \\
    Budget-Constrained & toilet paper        & The budget constraint is \$10 \\
    Color-based        & mousepad            & Choose the pink color product \\
    Color-based        & stapler             & Choose the pink color product \\
    Brand-based        & iphone 16 pro cover & Choose the Otterbox brand \\
    Brand-based        & toothpaste          & Choose the Colgate brand \\
    \bottomrule
  \end{tabular}
\end{threeparttable}
\end{table}

\begin{figure}[!htbp]
    \centering
    \begin{subfigure}{0.48\linewidth}
        \includegraphics[width = \linewidth]{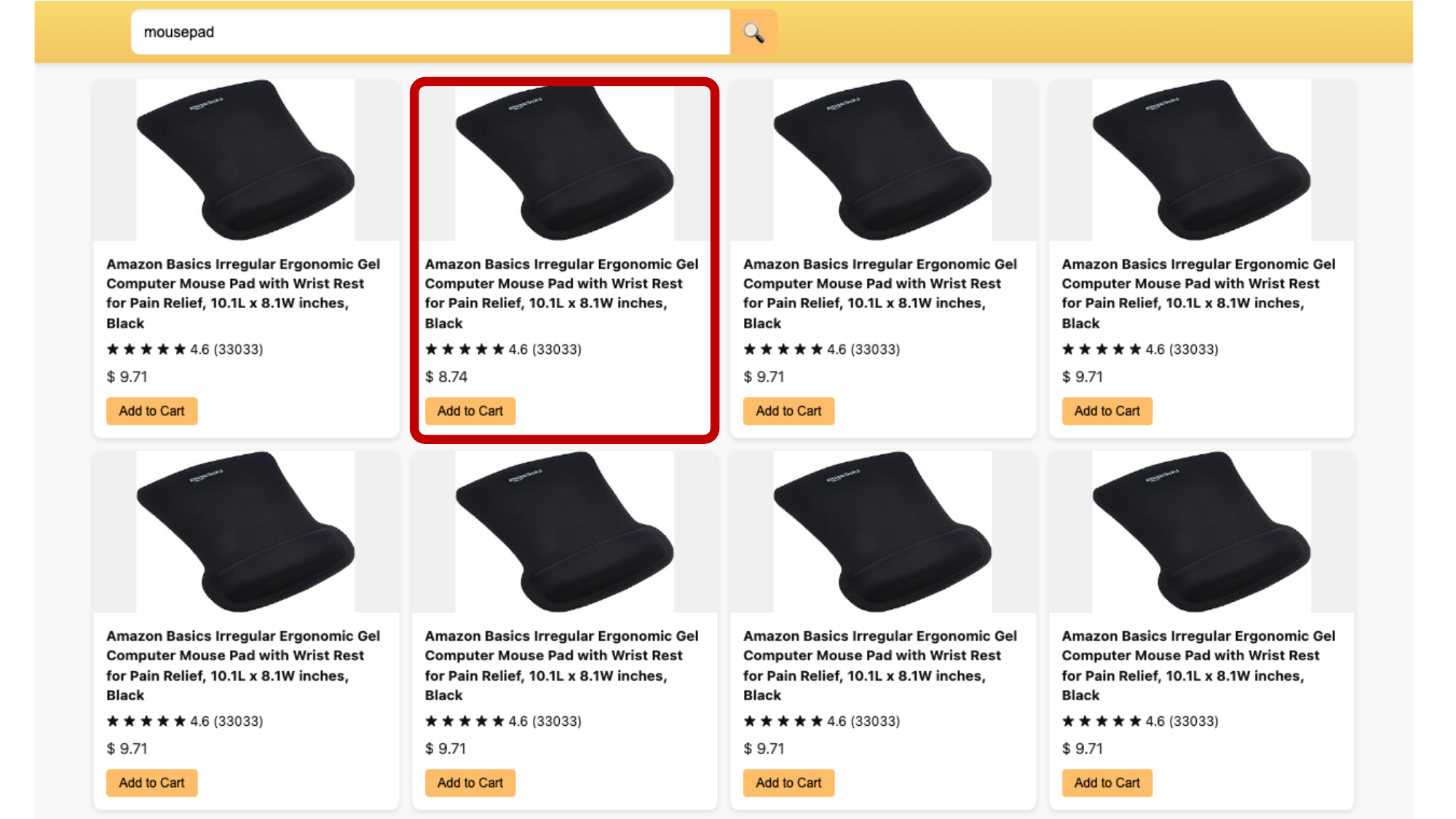}
        \caption{10\% discount on one listing}
        \label{subfig:10-percent-discount-mousepad}
    \end{subfigure}
    \begin{subfigure}{0.48\linewidth}
        \includegraphics[width = \linewidth]{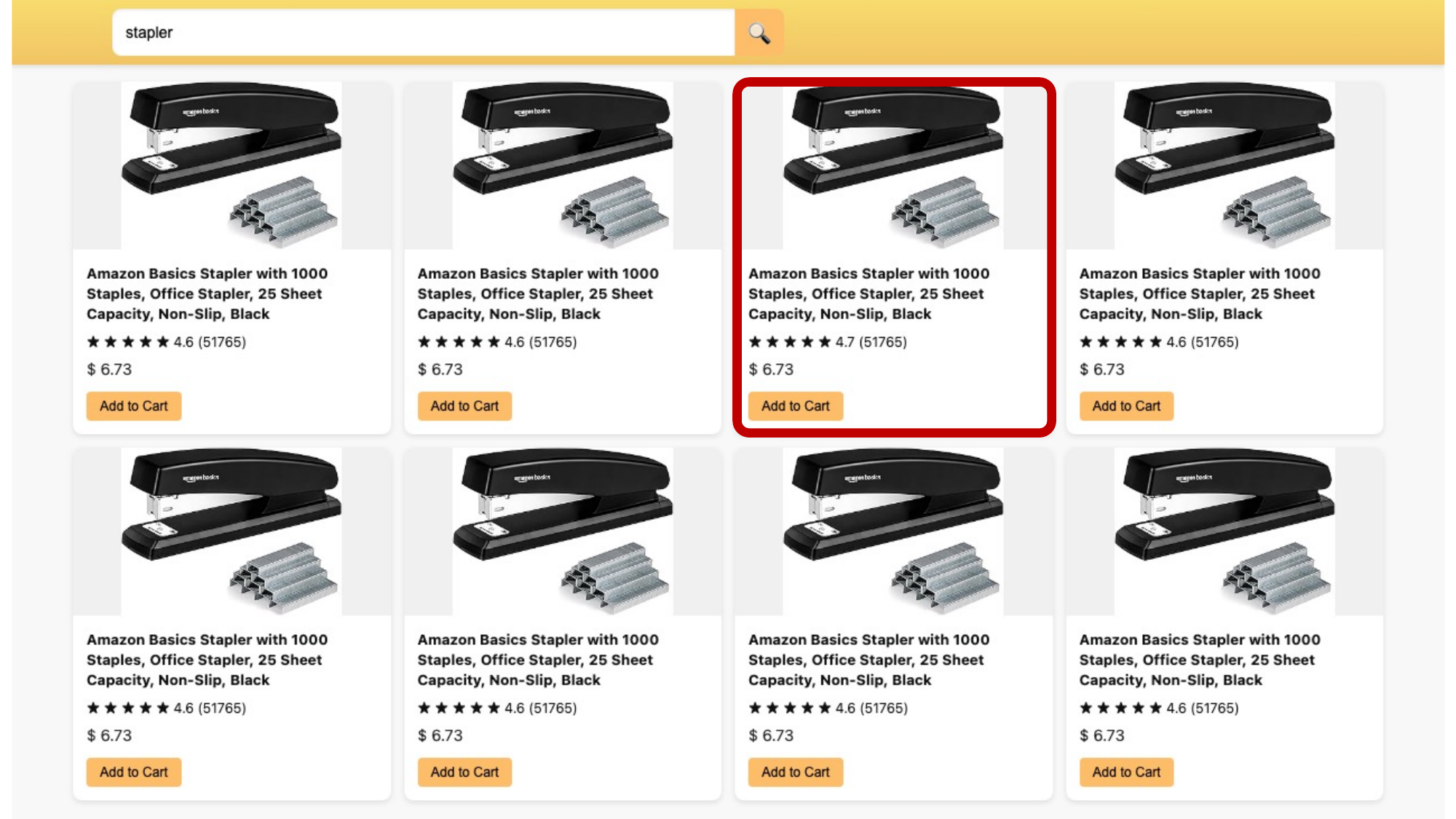}
        \caption{+0.1 increase in rating of one listing}
        \label{subfig:0.1-rating-increase-stapler}
    \end{subfigure}
    \begin{subfigure}{0.48\linewidth}
        \includegraphics[width = \linewidth]{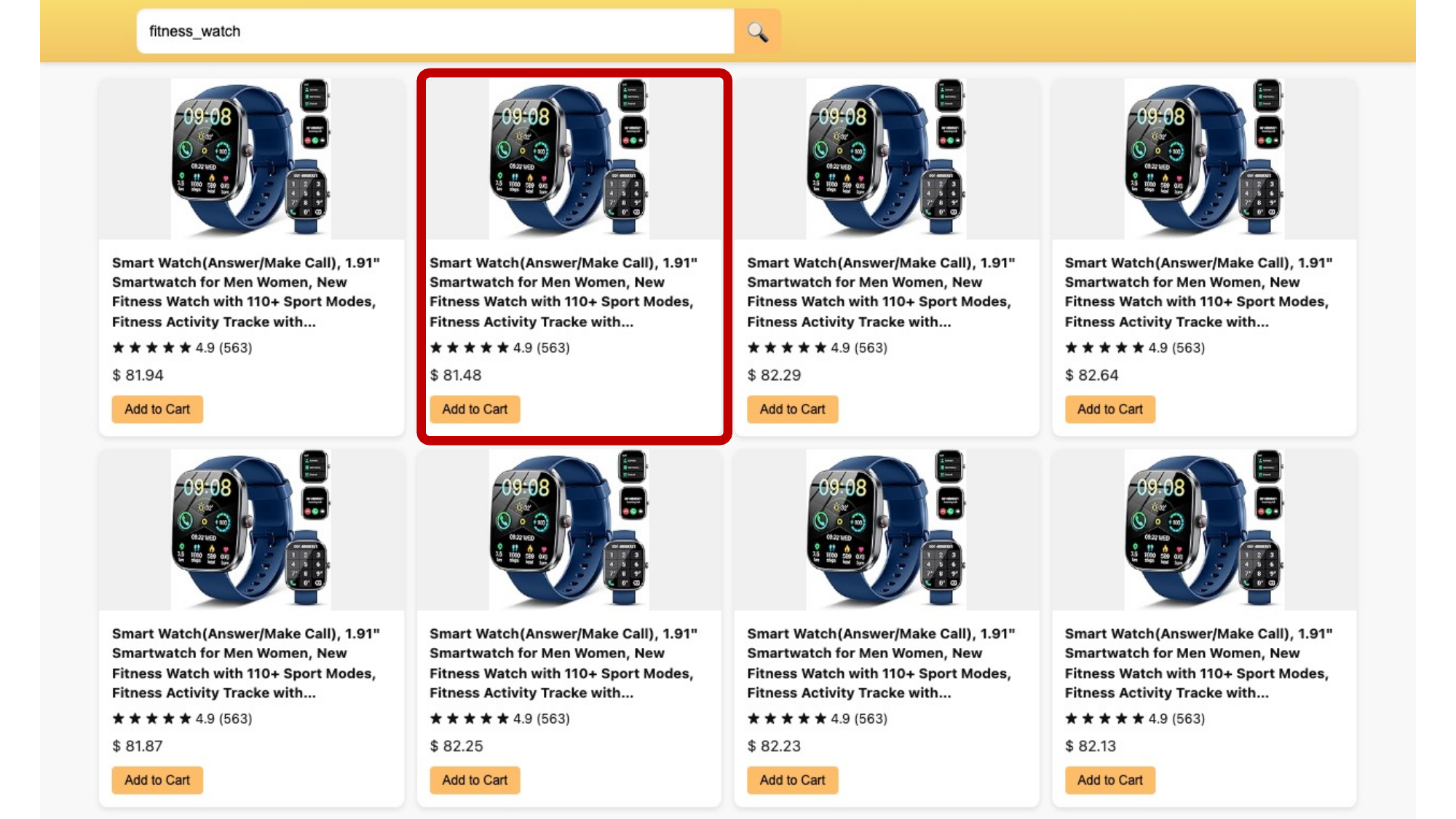}
        \caption{Random prices (with low variance)}
        \label{subfig:random-price-low-variance}
    \end{subfigure}
    \begin{subfigure}{0.48\linewidth}
        \includegraphics[width = \linewidth]{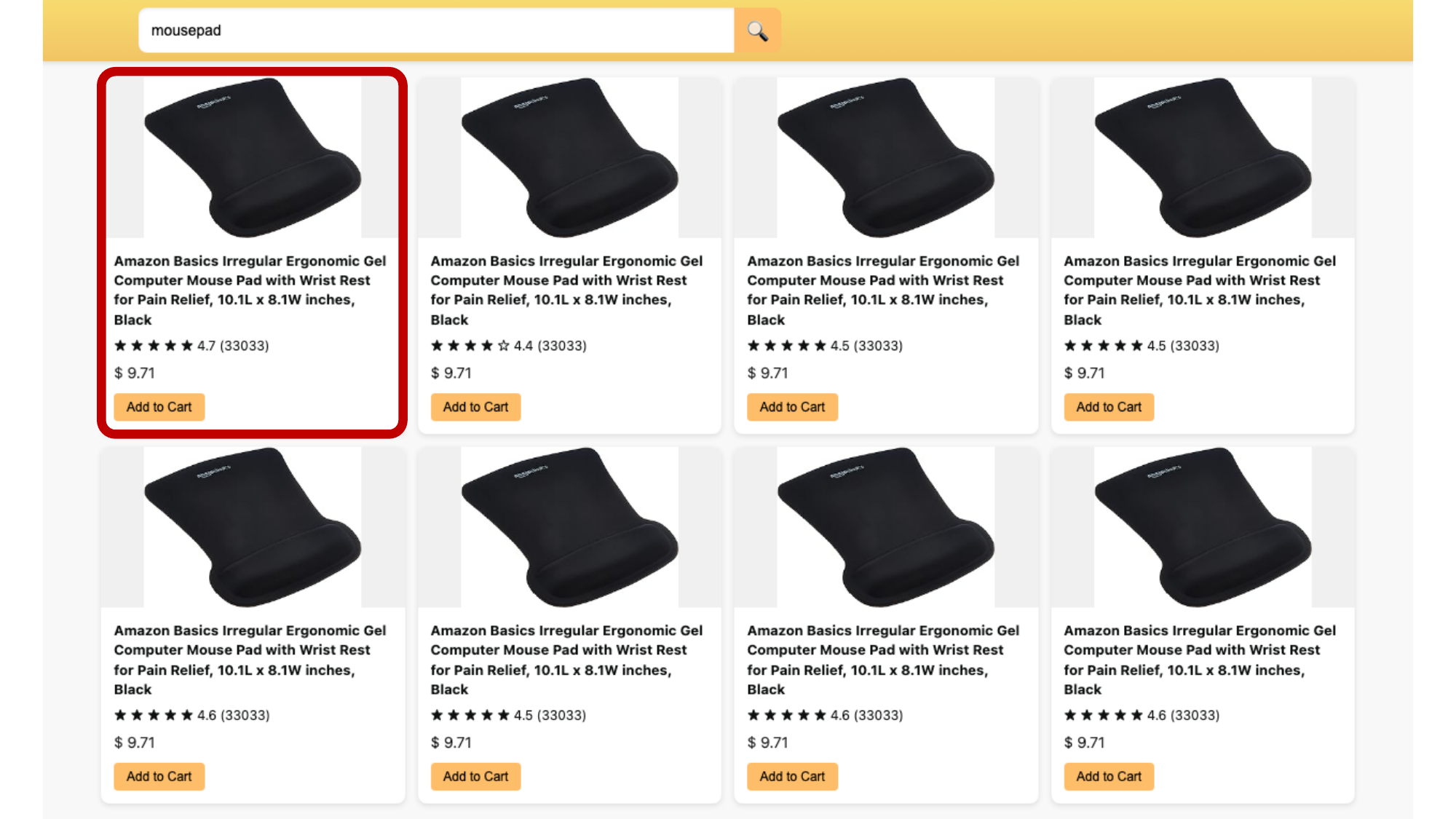}
        \caption{Random ratings (with low variance)}
        \label{subfig:random-rating-low-variance}
    \end{subfigure}
    \begin{subfigure}{0.48\linewidth}
        \includegraphics[width = \linewidth]{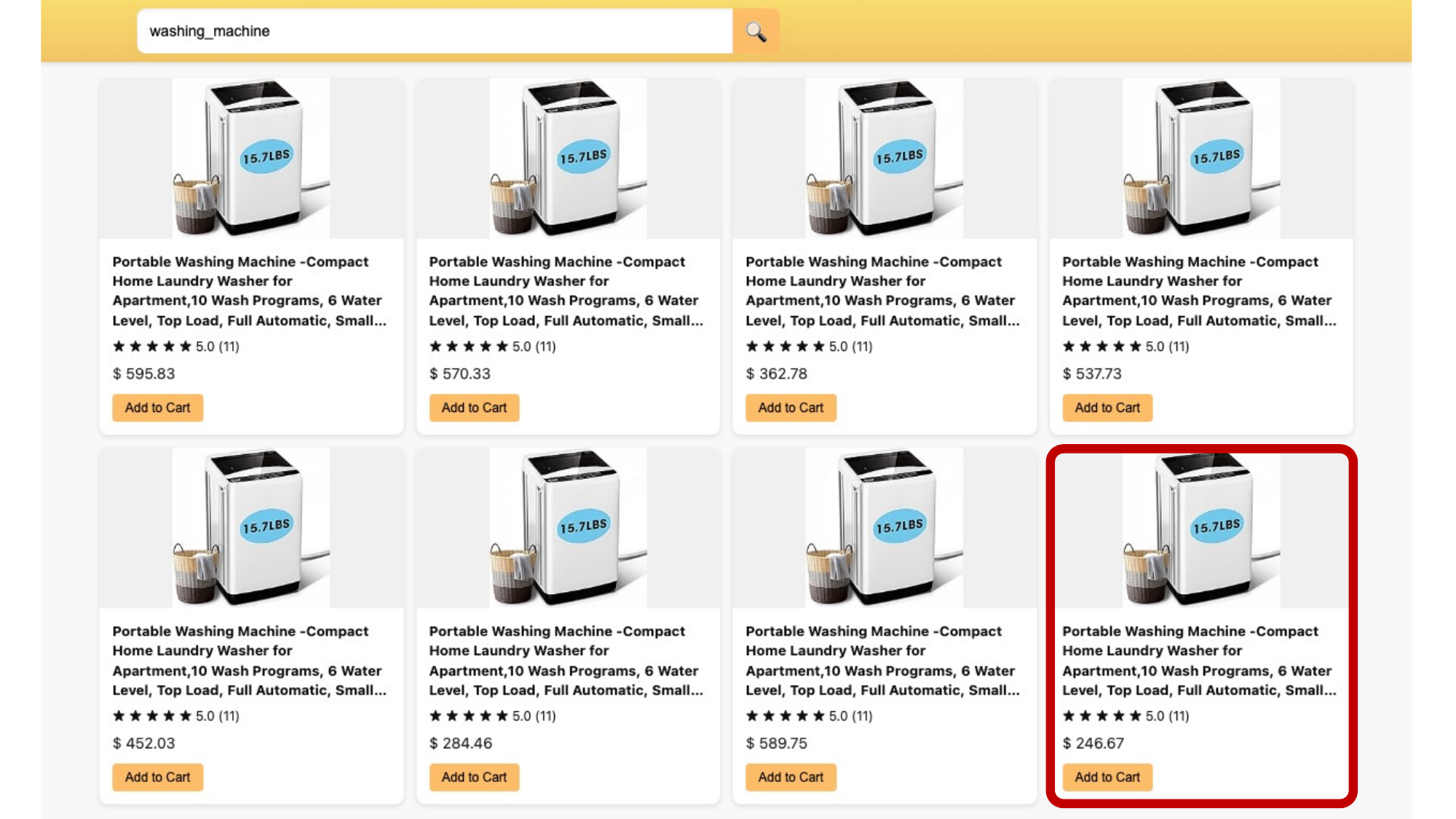}
        \caption{Random prices (with high variance)}
        \label{subfig:random-price-high-variance}
    \end{subfigure}
    \begin{subfigure}{0.48\linewidth}
        \includegraphics[width = \linewidth]{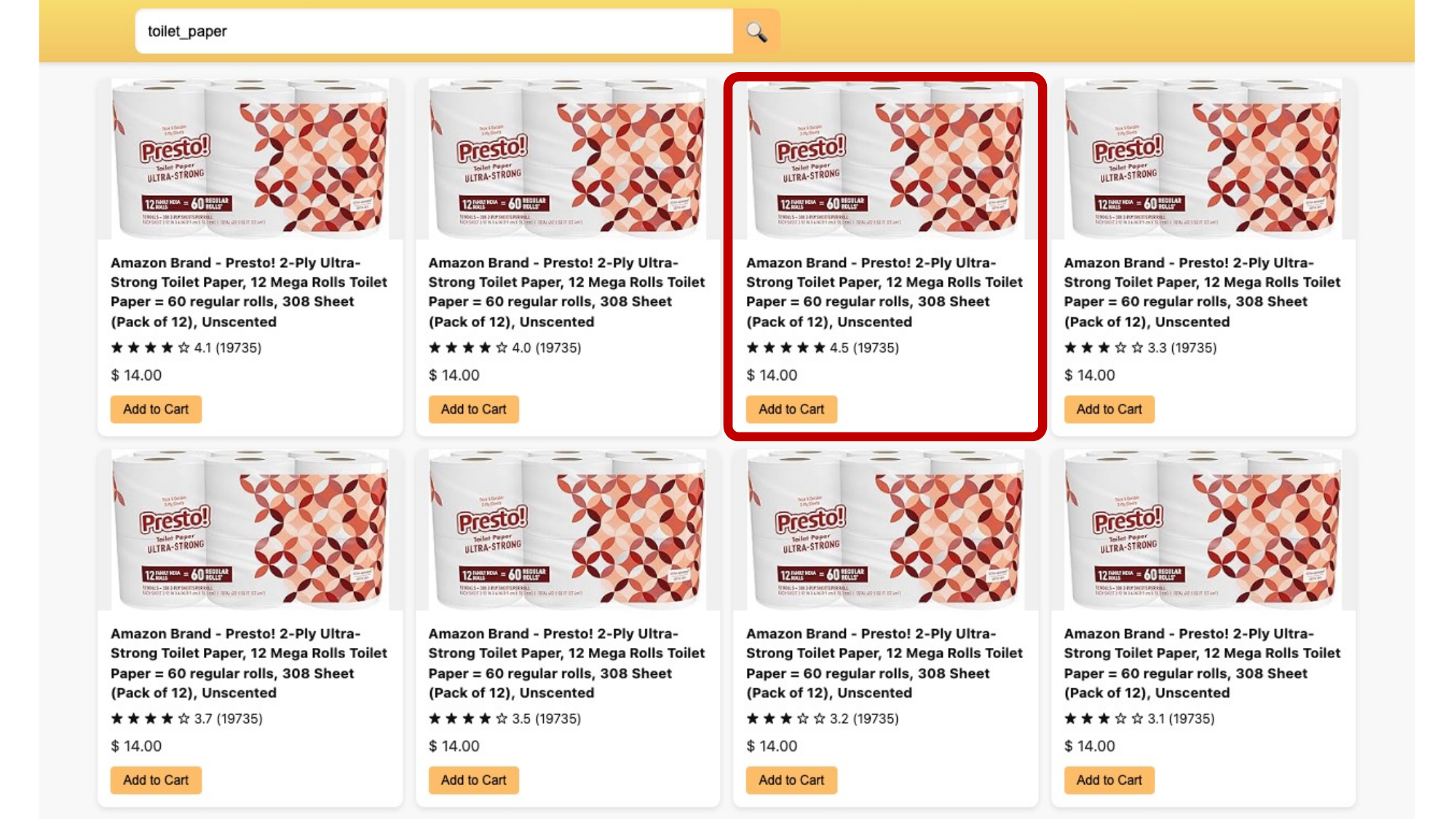}
        \caption{Random ratings (with high variance)}
        \label{subfig:random-rating-high-variance}
    \end{subfigure}
    \caption{Design of price and rating based rationality test. The optimal choice is highlighted in the red bounding box.}
    \label{fig:price-rating-screenshots}
\end{figure}

\paragraph{\bf Price-Based Rationality Experiments.}
\label{app:experimental-setup-price-rationality}
We have two types of price-based rationality experiments. In both, all products are identical except for price. In one setting $(i)$ all price are equal, except for the price of one listing which is decreased by $\alpha$-fraction and in a second setting $(ii)$ we assign random prices to each of the listings drawn from a normal distribution $\mathcal{N}(\mu, \sigma^2)$ with mean $\mu$ and variance $\sigma^2$. For $(i)$, we consider $\alpha \in \{0.1, 0.05, 0.01\}$ which corresponds to a $10\%, 5\%$ and $1\%$ discount in the prices with respect to the other prices respectively (see Figure \ref{subfig:10-percent-discount-mousepad}). For $(ii)$, we set the mean $\mu$ as the average of the eight product prices in their respective category. For the variance, we consider $\sigma = \$0.3$ (low variance; see Figure \ref{subfig:random-price-low-variance}) and $\sigma = 0.2\mu$ (high variance; see Figure \ref{subfig:random-price-high-variance}). 

\paragraph{\bf Rating-Based Rationality Experiments.}
\label{app:experimental-setup-rating-based-rationality tests}
We have three rating-based experiments: $(i)$ the rating of one listing is increased by $0.1$ (see Figure \ref{subfig:0.1-rating-increase-stapler}), $(ii)$ random ratings for the eight products with ratings ranging from $4.4$ to $4.7$ (low variance; see Figure \ref{subfig:random-rating-low-variance}) and $(iii)$ random ratings for the eight products with ratings ranging from $3.0$ to $4.5$ (high variance; see Figure \ref{subfig:random-rating-high-variance}). Note that except for the ratings, all the other product attributes are the same.

\subsubsection{Main Results.} 
\label{app:main-results-instruction-following-basic-rationality}

In this section we provide the associated fail rates of different AI models on the instruction-following and economic rationality tests.

\paragraph{Instruction Following Experiments.}
\label{app:main-results-instruction-following}
We provide associated fail rates for the different instruction-following tasks (columns) and for different AI models (rows), in Table \ref{tab:fail-rate-instruction-following-task}.

\begin{table}[!htb]
\centering
\begin{threeparttable}
  \caption{Fail rate of different models on instruction-following tasks (standard errors in parentheses).}
  \label{tab:fail-rate-instruction-following-task}
  \footnotesize
  \sisetup{table-align-text-post=false}
  \begin{tabular}{l l l l}
    \toprule
    \multicolumn{1}{c}{} &
    \multicolumn{1}{c}{Budget Constrained} &
    \multicolumn{1}{c}{Color Based} &
    \multicolumn{1}{c}{Brand Based} \\
    \midrule
    Claude Sonnet 3.5 & $4.0\%$ $(0.5\%)$  & $13.5\%$ $(1.4\%)$ & \textcolor{gray}{$0.0\%$} \\
    Claude Sonnet 3.7 & \textcolor{gray}{$0.0\%$} & $4.4\%$ $(0.5\%)$ & \textcolor{gray}{$0.0\%$} \\
    Claude Sonnet 4 & \textcolor{gray}{$0.0\%$} & $3.8\%$ $(0.5\%)$ & \textcolor{gray}{$0.0\%$} \\
    Claude Opus 4.5 & \textcolor{gray}{$0.0\%$} & $4.0\%$ $(0.5\%)$ & \textcolor{gray}{$0.0\%$} \\
    GPT\textendash4o  & \textcolor{gray}{$0.0\%$} & \textcolor{gray}{$0.0\%$} & \textcolor{gray}{$0.0\%$} \\
    GPT\textendash4.1 & \textcolor{gray}{$0.0\%$} & \textcolor{gray}{$0.0\%$} & \textcolor{gray}{$0.0\%$} \\
    GPT\textendash5.1 & \textcolor{gray}{$0.0\%$} & \textcolor{gray}{$0.0\%$} & \textcolor{gray}{$0.0\%$} \\
    Gemini 2.0 Flash  & \textcolor{gray}{$0.0\%$} & \textcolor{gray}{$0.0\%$} & \textcolor{gray}{$0.0\%$} \\
    Gemini 2.5 Flash  & \textcolor{gray}{$0.0\%$} & \textcolor{gray}{$0.0\%$} & \textcolor{gray}{$0.0\%$} \\
    Gemini 3.0 Pro Preview  & \textcolor{gray}{$0.0\%$} & \textcolor{gray}{$0.0\%$} & \textcolor{gray}{$0.0\%$} \\
    \bottomrule
  \end{tabular}
\end{threeparttable}
\end{table}

\paragraph{Price-Based Rationality Experiments.}
\label{app:main-results-price-based-rationality}
We provide associated fail rates for the different price-based rationality tasks (columns) and for different AI models (rows), in Table \ref{tab:fail-rate-price-based-task}.

\begin{table}[!htb]
\centering
\begin{threeparttable}
  \caption{Fail rate of different models on price-based rationality tests (std. errors in parentheses).}
  \label{tab:fail-rate-price-based-task}
  \footnotesize
  \sisetup{table-align-text-post=false}
  \begin{tabular}{l l l l l l}
    \toprule
    \multicolumn{1}{c}{} &
    \multicolumn{1}{c}{\begin{tabular}{@{}c@{}}Price reduced\\(1\% discount)\end{tabular}} &
    \multicolumn{1}{c}{\begin{tabular}{@{}c@{}}Price reduced\\(5\% discount)\end{tabular}} &
    \multicolumn{1}{c}{\begin{tabular}{@{}c@{}}Price reduced\\(10\% discount)\end{tabular}} &
    \multicolumn{1}{c}{\begin{tabular}{@{}c@{}}Random prices\\(low var.)\end{tabular}} &
    \multicolumn{1}{c}{\begin{tabular}{@{}c@{}}Random prices\\(high var.)\end{tabular}} \\
    \midrule
    Claude Sonnet 3.5   & $63.7\%$ $(1.7\%)$ & $36.8\%$ $(1.4\%)$ & $32.5\%$ $(1.0\%)$ & $8.3\%$ $(0.4\%)$  & $5.0\%$ $(0.3\%)$ \\
    Claude Sonnet 3.7   & $21.0\%$ $(0.9\%)$ & $6.5\%$ $(0.7\%)$  & $1.2\%$ $(0.2\%)$  & $8.3\%$ $(0.4\%)$  & $6.0\%$ $(0.3\%)$ \\
    Claude Sonnet 4  & $0.5\%$ $(0.1\%)$  & $1.8\%$ $(0.2\%)$  & \textcolor{gray}{$0.0\%$} & $8.3\%$ $(0.3\%)$  & $4.3\%$ $(0.2\%)$ \\
    Claude Opus 4.5 & $4.3\%$ $(0.6\%)$ & $0.8\%$ $(0.1\%)$ & \textcolor{gray}{$0.0\%$} & $0.8\%$ $(0.1\%)$ & \textcolor{gray}{$0.0\%$}  \\
    GPT\textendash4o   & $25.8\%$ $(1.0\%)$ & $5.6\%$ $(0.4\%)$  & $6.8\%$ $(0.3\%)$  & $17.4\%$ $(0.9\%)$ & $3.6\%$ $(0.2\%)$ \\
    GPT\textendash4.1  & $9.3\%$ $(0.6\%)$  & $1.3\%$ $(0.1\%)$  & $1.2\%$ $(0.2\%)$  & $12.6\%$ $(0.8\%)$ & $0.8\%$ $(0.1\%)$ \\
    GPT\textendash5.1  & $1.0\%$ $(0.1\%)$  & $0.3\%$ $(0.1\%)$  & $0.3\%$ $(0.1\%)$  & $0.3\%$ $(0.1\%)$ & $0.8\%$ $(0.1\%)$ \\
    Gemini 2.0 Flash   & $2.8\%$ $(0.2\%)$  & $0.5\%$ $(0.1\%)$  & $1.2\%$ $(0.2\%)$  & $1.0\%$ $(0.1\%)$  & $6.5\%$ $(0.3\%)$ \\
    Gemini 2.5 Flash   & $1.0\%$ $(0.1\%)$  & \textcolor{gray}{$0.0\%$} & \textcolor{gray}{$0.0\%$} & $0.8\%$ $(0.1\%)$ & \textcolor{gray}{$0.0\%$} \\
    Gemini 3 Pro Preview & \textcolor{gray}{$0.0\%$} & \textcolor{gray}{$0.0\%$} & \textcolor{gray}{$0.0\%$} & $1.8\%$ $(0.1\%)$ & $0.3\%$ (0.1\%) \\
    \bottomrule
  \end{tabular}
\end{threeparttable}
\end{table}

\paragraph{Rating-Based Rationality Experiments.}
\label{app:main-results-rating-based-rationality}
We provide associated fail rates for the different rating-based rationality tasks (columns) and for different AI models (rows), in Table  \ref{tab:fail-rate-rating-based-task}.

\begin{table}[!htb]
\centering
\begin{threeparttable}
  \caption{Fail rate of different models on rating-based rationality tests (std. {errors} in parentheses).}
  \label{tab:fail-rate-rating-based-task}
  \footnotesize
  \sisetup{table-align-text-post=false}
  \begin{tabular}{l l l l}
    \toprule
    \multicolumn{1}{c}{} &
    \multicolumn{1}{c}{\begin{tabular}{@{}c@{}}Rating of one listing\\increased by $0.1$\end{tabular}} &
    \multicolumn{1}{c}{\begin{tabular}{@{}c@{}}Random ratings\\(low variance)\end{tabular}} &
    \multicolumn{1}{c}{\begin{tabular}{@{}c@{}}Random ratings\\(high variance)\end{tabular}} \\
    \midrule
    Claude Sonnet 3.5  & $57.3\%$ $(1.5\%)$ & $16.3\%$ $(0.8\%)$ & $2.7\%$ $(0.2\%)$ \\
    Claude Sonnet 3.7  & $6.7\%$ $(0.5\%)$  & \textcolor{gray}{$0.0\%$} & \textcolor{gray}{$0.0\%$} \\
    Claude Sonnet 4  & $28.7\%$ $(1.2\%)$ & $9.4\%$ $(0.6\%)$  & $4.7\%$ $(0.3\%)$ \\
    Claude Opus 4.5 & \textcolor{gray}{$0.0\%$} & \textcolor{gray}{$0.0\%$} & \textcolor{gray}{$0.0\%$} \\
    GPT\textendash4o   & $71.7\%$ $(0.9\%)$ & $16.0\%$ $(0.5\%)$ & $7.3\%$ $(0.4\%)$ \\
    GPT\textendash4.1  & $15.1\%$ $(0.6\%)$ & $11.7\%$ $(0.5\%)$ & $6.0\%$ $(0.4\%)$ \\
    GPT\textendash5.1 & \textcolor{gray}{$0.0\%$} & $0.3\%$ $(0.1\%)$ & \textcolor{gray}{$0.0\%$} \\
    Gemini 2.0 Flash   & \textcolor{gray}{$0.0\%$} & \textcolor{gray}{$0.0\%$} & $0.3\%$ $(0.1\%)$ \\
    Gemini 2.5 Flash   & \textcolor{gray}{$0.0\%$} & \textcolor{gray}{$0.0\%$} & \textcolor{gray}{$0.0\%$} \\
    Gemini 3.0 Pro Preview & $0.3\%$ $(0.1\%)$ & $1.7\%$ $(0.2\%)$ & $1.0\%$ $(0.1\%)$ \\
    \bottomrule
  \end{tabular}
\end{threeparttable}
\end{table}

\paragraph{Reasons of Failures.}
\label{app:rationality-failure-reasons}
On the experiments where the models failed, we probed the explanation provided by the models. Broadly speaking, we identified three types of explanations.
 \begin{itemize}
     \item \textit{Perceptual Limitations.} The model treated all products as identical, failing to recognize differences in price or rating. This type of error was particularly prevalent when the difference was subtle, for e.g., in the case when the price of one listing was reduced by $1\%$.
    \item \textit{Unjustified Sub-optimality.} The model recognized the optimal product (e.g., the cheapest or highest-rated) but still made a sub-optimal choice without providing any specific justification for its decision.
     \item \textit{Dismissal with justification.} The model acknowledged the optimal product but explicitly provided a reason for not selecting it. These reasons often included:
     \begin{itemize}
         \item {Prioritizing ``Standard'' Options.} The model would state it     ``chose the standard price/rating'' or the ``most common price point'', indicating a preference for typical values over slight deviations.
         \item {Attributing difference to errors.} The model sometimes dismissed the optimal option by attributing the observed difference to a   ``display error'', a ``temporary discount'', an ``outdated listing'', or a ``pricing error'' on the platform.
         \item {Perceived Insignificance.} The model might justify its sub-optimal choice by stating the price difference was minimal or that there was ``no apparent/significant advantage'' or ``no functional difference'' compared to other options.    
         \end{itemize}
 \end{itemize}

\subsection{Choice Behavior and Biases}

\subsubsection{Experimental Details.}
\label{app:experimental-setup-choice-behavior-biases}
\begin{figure}[!htb]
        \centering
        \begin{subfigure}{1.0\linewidth}
            \includegraphics[width = \linewidth]{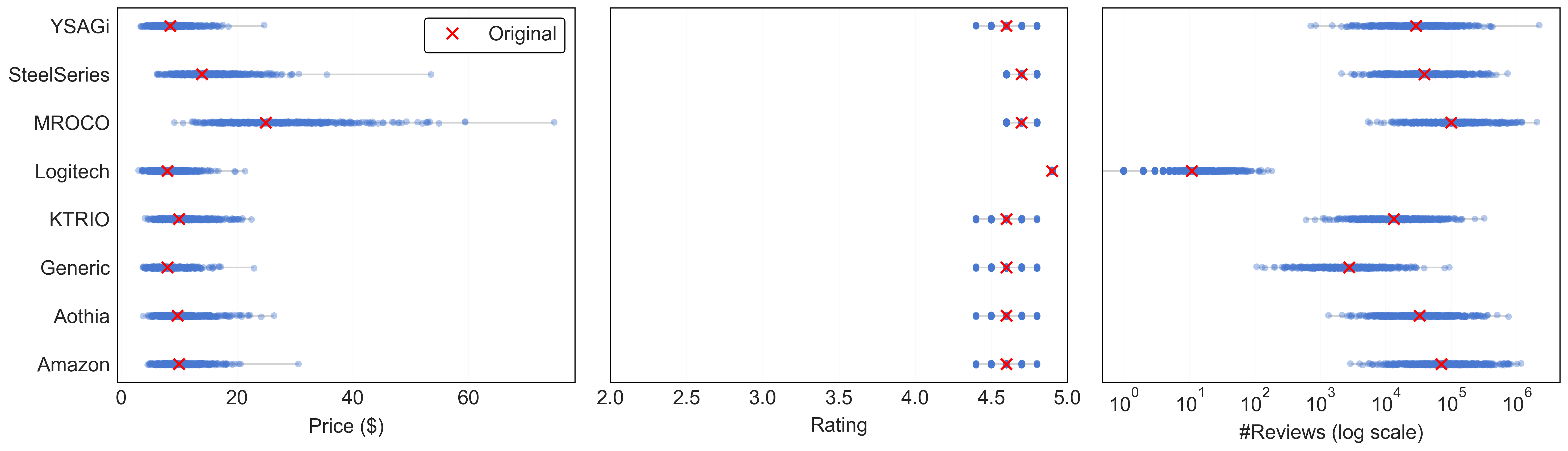}
            \caption{mousepad}
        \end{subfigure}
        \begin{subfigure}{1.0\linewidth}
            \includegraphics[width = \linewidth]{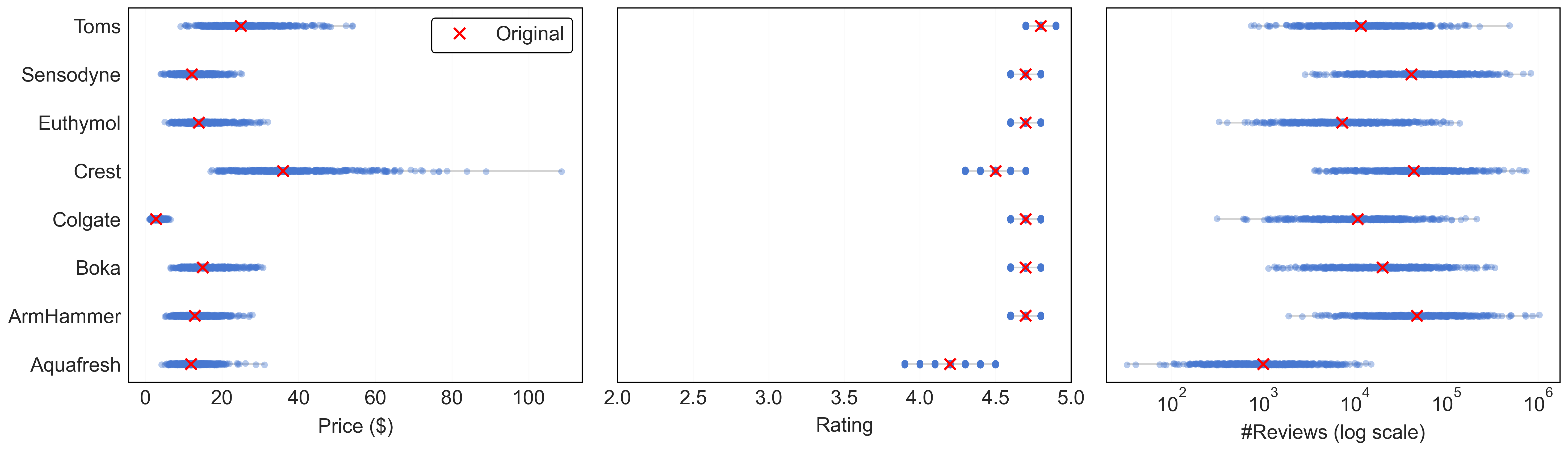}
            \caption{toothpaste}
        \end{subfigure}
        \caption{Visualization of the exogenous variation in price, rating and number of reviews}
        \label{fig:master-dataset-visualization}
    \end{figure}

\begin{figure}[!htb]
        \centering
        \begin{subfigure}{0.48\linewidth}
            \includegraphics[width = \linewidth]{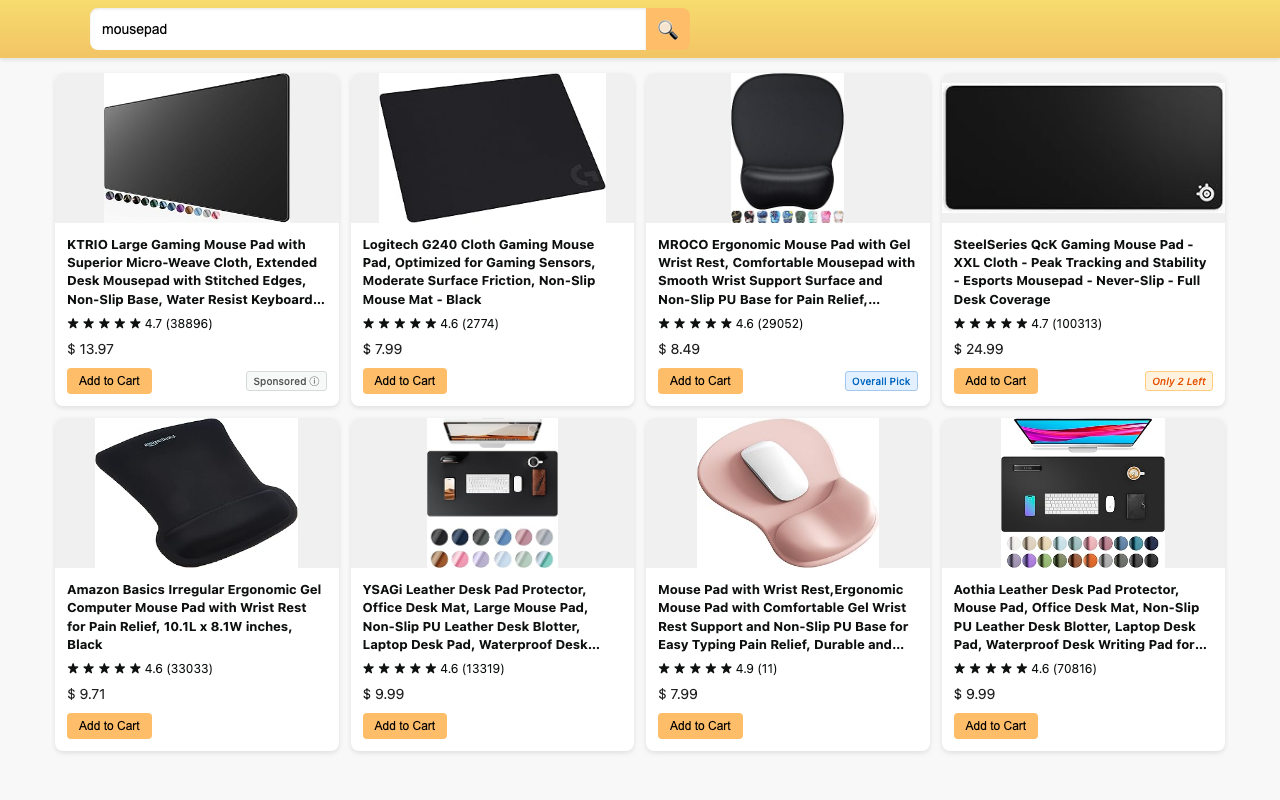}
            \caption{}
        \end{subfigure}
        \begin{subfigure}{0.48\linewidth}
            \includegraphics[width = \linewidth]{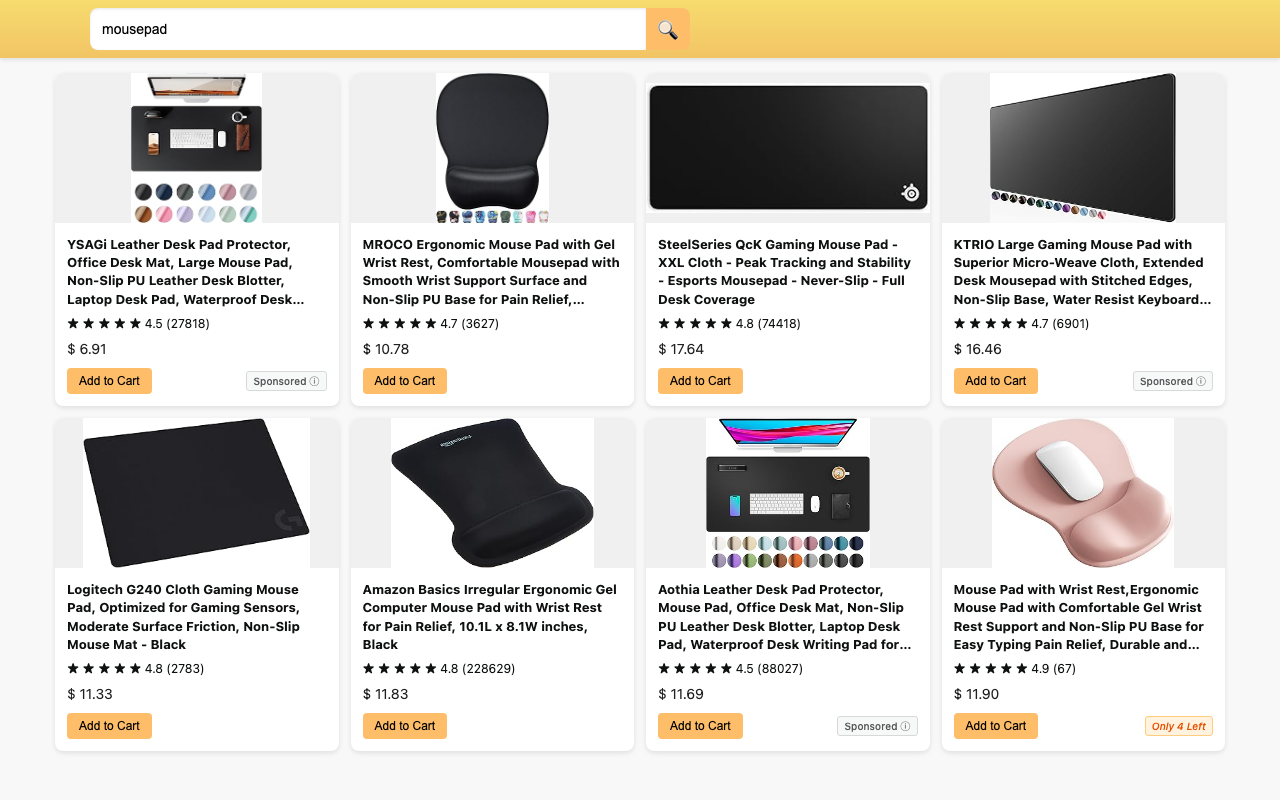}
            \caption{}
        \end{subfigure}
        \caption{Example of the mock-app for different experiments with exogenous variations in position, tags and product attributes}
        \label{fig:master-dataset-examples}
    \end{figure}

To estimate the sensitivity of the AI agent to different platform levers, we create a dataset with exogenous variation across a number of variables associated with the products (see Table \ref{tab:exogenous-variation}). Given these variations are exogenous to the inherent attractiveness of the products, they allow us to obtain causal estimates to the impact of different platform levers like position, tags and product attributes like price, rating and number of reviews. We also provide a visualization of the exogenous variation of price, rating and number of reviews around their original value for a few product categories in Figure \ref{fig:master-dataset-visualization}. Figure \ref{fig:master-dataset-examples} provides an example of different mock-app screens for two different experiment runs. Notice that the assortment of the eight products is the same across the two experiments. However the position of the eight products change across the two experiments along with their price, rating and number of reviews. Also notice that the Sponsored, Overall Pick and Only X Remaining (scarcity) tags are assigned to different products across the two experiments.

In Figure \ref{fig:master-dataset-examples} we provide an example of different mock-app configurations experiments. Note that across the different experiments shown in Figure \ref{fig:master-dataset-examples}, the position of the products is randomly permuted along with random perturbation to their price, rating and number of reviews.

\subsubsection{Understanding the trade-offs between position, tags and listing attributes}
\label{app:main-results-choice-behavior-biases}
We fit a conditional logit with the utility specification in \eqref{eq:utility-model}. The estimates are provided in Table \ref{tab:conditional-logit-results-all-models}. 
We can use the estimates in Table \ref{tab:conditional-logit-results-all-models}, \eqref{eq:utility-model} and \eqref{eq:mnl} to understand how the different models trade-off different attributes. For example, we can be interested in understanding how the choice probability changes if there is a Sponsored tag or an Overall Pick tag or rating changes by $\pm 0.1$. Let's say that the baseline selection probability of a product $j$ is $\pi_j$ and assume that we change co-variate $z$ with coefficient $\beta_z$ by $\Delta z$. Let $o_j = \pi_j / (1 - \pi_j)$ be the odd-ratio for product $j$, then the new selection probability $\pi^\prime_j$ is given as
\begin{align*}
    \pi^\prime_j &\stackrel{(a)}= \frac{\exp(U_{j})\exp(\beta_z \Delta z)}{\sum_{k \neq i} \exp(U_{k}) + \exp(U_j) \exp(\beta_z \Delta z)} \\
    &\stackrel{(b)}= \frac{\pi_j\exp(\beta_z \Delta z)}{1- \pi_j + \pi_j \exp(\beta_z \Delta z)} \\
    &\stackrel{(c)}= \frac{\exp(\beta_z \Delta z) o_j}{1 + \exp(\beta_z \Delta z) o_j},
\end{align*}
where $(a)$ follows from \eqref{eq:mnl}, $(b)$ follows by dividing numerator and denominator by $\sum_{k} \exp(U_k)$ and the fact that $\pi_j = \exp(U_j) / \sum_{k \neq i} \exp(U_k)$ and $(c)$ follows from the definition of $o_j$. Using this, we can calculate how the probability will change if we change some covariate.

We can also obtain price-equivalent trade-offs: a convenient way to compare levers is to ask: by how much can a seller raise (or must cut) price to keep the utility constant when the seller adds (or loses) a feature worth $\Delta U = \beta_z \Delta z$. Let say, we change the price by a multiplicative factor of $\lambda$, i.e., $P^{\prime} = \lambda P$ where $P$ is the original price of the product. Since we are keeping the utility constant, we have that $\beta_{\text{price}} \ln \lambda + \beta_z \Delta z = 0$, this implies that $\lambda = \exp\left(-\beta_z \Delta z / \beta_{\text{price}}  \right)$. This allows to be get price-based exchange rate between different attributes. In Table \ref{tab:price_equivalents}, we present the price-equivalent results for different covariates.

\begin{table}[!htbp]
\centering
\begin{threeparttable}
  \caption{Price‐equivalent trade–offs implied by the CL coefficients. Entries are approximate multiplicative changes in price that leave utility unchanged when the corresponding lever is added (top block) or removed/penalized (bottom block).}
  \label{tab:price_equivalents}
  \footnotesize
  \sisetup{table-align-text-post=false}
  \begin{tabular}{l c c c}
    \toprule
    & \multicolumn{1}{c}{Claude Sonnet 4} & \multicolumn{1}{c}{GPT-4.1} & \multicolumn{1}{c}{Gemini 2.5 Flash} \\
    \midrule
    \multicolumn{4}{l}{\textit{How much could a seller \textbf{raise} prices (holding utility fixed) if the seller gets...}} \\
    Row 1 (vs Row 2)    & +112.6\% & +91.2\% & +17.0\% \\
    Overall Pick tag    & +92.2\%  & +64.5\% & +137.8\% \\
    +0.1 rating         & +35.4\%  & +67.3\% & +27.9\% \\
    Double reviews      & +19.4\%  & +37.4\% & +17.2\% \\
    \midrule
    \multicolumn{4}{l}{\textit{How much would a seller need to \textbf{cut} price to offset the harm of ...}} \\
    Sponsored tag       & -8.0\%   & -14.3\% & -11.3\% \\
    Scarcity tag        & -4.6\%   & -6.3\%  & -14.5\% \\
    \bottomrule
  \end{tabular}
\end{threeparttable}
\end{table}

\begin{table}[!htbp]
\centering
\begin{threeparttable}
  \caption{Price‐equivalent trade–offs implied by the CL coefficients for the latest models (Claude Opus 4.5, GPT-5.1 and Gemini 3.0 Pro Preview). Entries are approximate multiplicative changes in price that leave utility unchanged when the corresponding lever is added (top block) or removed/penalized (bottom block).}
  \label{tab:price_equivalents}
  \footnotesize
  \sisetup{table-align-text-post=false}
  \begin{tabular}{l c c c}
    \toprule
    & \multicolumn{1}{c}{Claude Opus 4.5} & \multicolumn{1}{c}{GPT-5.1} & \multicolumn{1}{c}{Gemini 3.0 Pro Preview} \\
    \midrule
    \multicolumn{4}{l}{\textit{How much could a seller \textbf{raise} prices (holding utility fixed) if the seller gets...}} \\
    Row 1 (vs Row 2)    & +28.1\% & -22.2 \% & 159.7\% \\
    Overall Pick tag    & +168.9\%  & +61.5\% & +159.3\% \\
    +0.1 rating         & +80.6\%  & +39.2\% & +20.7\% \\
    Double reviews      & +43.3\%  & +21.9\% & +22.9\% \\
    \midrule
    \multicolumn{4}{l}{\textit{How much would a seller need to \textbf{cut} price to offset the harm of ...}} \\
    Sponsored tag       & -16.5\%   & -12.4\% & -24.2\% \\
    Scarcity tag        & -15.6\%   & +0.2\%  & 1+0.2\% \\
    \bottomrule
  \end{tabular}
\end{threeparttable}
\end{table}

\newpage

\subsubsection{Results for prompt variation in VLM agents}
\label{app:prompt-efficacy-vlm}
We can think of the choice behavior with the default prompt (Figure~\ref{fig:default-prompt-buying-agent-mock-app}) as the model’s \emph{prior}—its baseline choice rule absent explicit preferences. Estimates in Table~\ref{tab:conditional-logit-results-all-models} show a robust top‑row premium and heterogeneous column preferences across models. To probe the strength of these position effects and the model prior, we modify the prompt to (i) {ignore position} (Figure~\ref{fig:ignore-position-prompt-buying-agent-computer-use}) and (ii) {prioritize price} (Figure~\ref{fig:price-sensitive-prompt-buying-agent-computer-use}). We can view the ``prioritize price'' prompt as a minimal form of personalization: it encodes a plausible user preference without pre‑solving the task. 

\newpage

\begin{table}[!htbp]
\centering
\begin{threeparttable}
  \caption{Estimates of the Conditional Logit Regression for Claude Sonnet 4 (VLM agents)}
  \label{tab:conditional-logit-results-claude-aug-sep}
  \footnotesize
  \sisetup{
    input-symbols         = (), 
    table-align-text-post = false
  }
  \begin{tabular}{l
                  S[table-format=-1.3]
                  S[table-format=-1.3]
                  S[table-format=-1.3]
                  S[table-format=-1.3]}
    \toprule
    & \multicolumn{1}{c}{Default} & \multicolumn{1}{c}{Default} & \multicolumn{1}{c}{Ignore Position} & \multicolumn{1}{c}{Ignore Position \& Prioritize Price} \\
    & \multicolumn{1}{c}{\small (Fig.~\ref{fig:default-prompt-buying-agent-mock-app}; Aug'25)}
    & \multicolumn{1}{c}{\small (Fig.~\ref{fig:default-prompt-buying-agent-mock-app}; Sep'25)}
    & \multicolumn{1}{c}{\small (Fig.~\ref{fig:ignore-position-prompt-buying-agent-computer-use}; Sep'25)}
    & \multicolumn{1}{c}{\small (Fig.~\ref{fig:price-sensitive-prompt-buying-agent-computer-use}; Sep'25)} \\
    \midrule
    \multicolumn{5}{l}{\textit{Position effects}} \\
    Row 1    & 1.224\rlap{$^{\text{\tiny ***}}$} & 1.250\rlap{$^{\text{\tiny ***}}$} & 1.039\rlap{$^{\text{\tiny ***}}$} & 0.602\rlap{$^{\text{\tiny ***}}$} \\
             & (0.046) & (0.046) & (0.045) & (0.052) \\[0.6ex]
    Column 1 & -0.297\rlap{$^{\text{\tiny ***}}$} & -0.298\rlap{$^{\text{\tiny ***}}$} & -0.254\rlap{$^{\text{\tiny ***}}$} & 0.216\rlap{$^{\text{\tiny **}}$} \\
             & (0.065) & (0.065) & (0.065) & (0.074) \\[0.6ex]
    Column 2 & 0.557\rlap{$^{\text{\tiny ***}}$} & 0.505\rlap{$^{\text{\tiny ***}}$} & 0.527\rlap{$^{\text{\tiny ***}}$} & 0.390\rlap{$^{\text{\tiny ***}}$} \\
             & (0.058) & (0.058) & (0.059) & (0.073) \\[0.6ex]
    Column 3 & 0.416\rlap{$^{\text{\tiny ***}}$} & 0.399\rlap{$^{\text{\tiny ***}}$} & 0.441\rlap{$^{\text{\tiny ***}}$} & 0.225\rlap{$^{\text{\tiny **}}$} \\
             & (0.059) & (0.059) & (0.060) & (0.073) \\
    \midrule
    \multicolumn{5}{l}{\textit{Badge effects}} \\
    Sponsored Tag    & -0.135\rlap{$^{\text{\tiny *}}$} & -0.330\rlap{$^{\text{\tiny *}}$} & -0.156\rlap{$^{\text{\tiny *}}$} & -0.263\rlap{$^{\text{\tiny **}}$} \\
                     & (0.068) & (0.070) & (0.069) & (0.083) \\[0.6ex]
    Overall Pick Tag & 1.060\rlap{$^{\text{\tiny ***}}$} & 1.009\rlap{$^{\text{\tiny ***}}$} & 0.686\rlap{$^{\text{\tiny ***}}$} & -0.010 \\
                     & (0.077) & (0.078) & (0.081) & (0.110) \\[0.6ex]
    Scarcity Tag     & -0.076 & -0.116 & -0.033 & -0.036 \\
                     & (0.094) & (0.094) & (0.093) & (0.113) \\
    \midrule
    \multicolumn{5}{l}{\textit{Attribute effects}} \\
    $\ln(\text{Price})$            & -1.623\rlap{$^{\text{\tiny ***}}$} & -1.740\rlap{$^{\text{\tiny ***}}$} & -1.762\rlap{$^{\text{\tiny ***}}$} & -5.644\rlap{$^{\text{\tiny ***}}$} \\
                                   & (0.079) & (0.080) & (0.080) & (0.141) \\[0.6ex]
    Rating                         & 4.913\rlap{$^{\text{\tiny ***}}$} & 4.886\rlap{$^{\text{\tiny ***}}$} & 5.788\rlap{$^{\text{\tiny ***}}$} & 2.072\rlap{$^{\text{\tiny ***}}$} \\
                                   & (0.218) & (0.220) & (0.229) & (0.233) \\[0.6ex]
    $\ln(\text{Num.\ of Reviews})$ & 0.415\rlap{$^{\text{\tiny ***}}$} & 0.431\rlap{$^{\text{\tiny ***}}$} & 0.583\rlap{$^{\text{\tiny ***}}$} & 0.248\rlap{$^{\text{\tiny ***}}$} \\
                                   & (0.023) & (0.023) & (0.024) & (0.024) \\
    \midrule
    Product Fixed Effects & \multicolumn{1}{c}{Yes} & \multicolumn{1}{c}{Yes} & \multicolumn{1}{c}{Yes} & \multicolumn{1}{c}{Yes} \\
    Observations          & \multicolumn{1}{c}{25,802} & \multicolumn{1}{c}{25,753} & \multicolumn{1}{c}{25,414} & \multicolumn{1}{c}{21,448} \\
    Choice Sets (Groups)  & \multicolumn{1}{c}{3,756} & \multicolumn{1}{c}{3,747} & \multicolumn{1}{c}{3,768} & \multicolumn{1}{c}{3,733} \\
    Pseudo R-squared      & \multicolumn{1}{c}{0.44} & \multicolumn{1}{c}{0.45} & \multicolumn{1}{c}{0.45} & \multicolumn{1}{c}{0.60} \\
    \bottomrule
  \end{tabular}
  \begin{tablenotes}[para]
    \item \tnote{*} $p < 0.05$, \tnote{**} $p < 0.01$, \tnote{***} $p < 0.001$.
  \end{tablenotes}
\end{threeparttable}
\end{table}

\newpage

\begin{table}[!htbp]
\centering
\begin{threeparttable}
  \caption{Estimates of the Conditional Logit Regression for GPT-4.1 (VLM agents)}
  \label{tab:conditional-logit-results-gpt41-aug-sep}
  \footnotesize
  \sisetup{
    input-symbols         = (), 
    table-align-text-post = false
  }
  \begin{tabular}{l
                  S[table-format=-1.3]
                  S[table-format=-1.3]
                  S[table-format=-1.3]
                  S[table-format=-1.3]}
    \toprule
    & \multicolumn{1}{c}{Default} & \multicolumn{1}{c}{Default} & \multicolumn{1}{c}{Ignore Position} & \multicolumn{1}{c}{Ignore Position \& Prioritize Price} \\
    & \multicolumn{1}{c}{\small (Fig.~\ref{fig:default-prompt-buying-agent-mock-app}; Aug'25)} &
      \multicolumn{1}{c}{\small (Fig.~\ref{fig:default-prompt-buying-agent-mock-app}; Sep'25)} &
      \multicolumn{1}{c}{\small (Fig.~\ref{fig:ignore-position-prompt-buying-agent-computer-use}; Sep'25)} &
      \multicolumn{1}{c}{\small (Fig.~\ref{fig:price-sensitive-prompt-buying-agent-computer-use}; Sep'25)} \\
    \midrule
    \multicolumn{5}{l}{\textit{Position effects}} \\
    Row 1    & 1.045\rlap{$^{\text{\tiny ***}}$} & 1.076\rlap{$^{\text{\tiny ***}}$} & 0.977\rlap{$^{\text{\tiny ***}}$} & 0.616\rlap{$^{\text{\tiny ***}}$} \\
             & (0.046) & (0.046) & (0.048) & (0.057) \\[0.6ex]
    Column 1 & 1.122\rlap{$^{\text{\tiny ***}}$} & 1.167\rlap{$^{\text{\tiny ***}}$} & 1.101\rlap{$^{\text{\tiny ***}}$} & 0.634\rlap{$^{\text{\tiny ***}}$} \\
             & (0.061) & (0.062) & (0.064) & (0.078) \\[0.6ex]
    Column 2 & 0.019 & 0.113 & 0.153\rlap{$^{\text{\tiny *}}$} & -0.113 \\
             & (0.065) & (0.065) & (0.067) & (0.080) \\[0.6ex]
    Column 3 & -0.013 & 0.051 & 0.060 & -0.102 \\
             & (0.066) & (0.066) & (0.067) & (0.080) \\
    \midrule
    \multicolumn{5}{l}{\textit{Badge effects}} \\
    Sponsored Tag    & -0.248\rlap{$^{\text{\tiny ***}}$} & -0.348\rlap{$^{\text{\tiny ***}}$} & -0.282\rlap{$^{\text{\tiny ***}}$} & -0.280\rlap{$^{\text{\tiny **}}$} \\
                     & (0.072) & (0.072) & (0.074) & (0.091) \\[0.6ex]
    Overall Pick Tag & 0.802\rlap{$^{\text{\tiny ***}}$} & 0.786\rlap{$^{\text{\tiny ***}}$} & 0.582\rlap{$^{\text{\tiny ***}}$} & 0.089 \\
                     & (0.083) & (0.083) & (0.089) & (0.117) \\[0.6ex]
    Scarcity Tag     & -0.105 & 0.007 & -0.001 & -0.060 \\
                     & (0.099) & (0.097) & (0.101) & (0.121) \\
    \midrule
    \multicolumn{5}{l}{\textit{Attribute effects}} \\
    $\ln(\text{Price})$            & -1.612\rlap{$^{\text{\tiny ***}}$} & -1.586\rlap{$^{\text{\tiny ***}}$} & -1.750\rlap{$^{\text{\tiny ***}}$} & -9.243\rlap{$^{\text{\tiny ***}}$} \\
                                   & (0.083) & (0.082) & (0.086) & (0.175) \\[0.6ex]
    Rating                         & 8.300\rlap{$^{\text{\tiny ***}}$} & 7.862\rlap{$^{\text{\tiny ***}}$} & 9.047\rlap{$^{\text{\tiny ***}}$} & 4.345\rlap{$^{\text{\tiny ***}}$} \\
                                   & (0.269) & (0.261) & (0.285) & (0.282) \\[0.6ex]
    $\ln(\text{Num.\ of Reviews})$ & 0.739\rlap{$^{\text{\tiny ***}}$} & 0.756\rlap{$^{\text{\tiny ***}}$} & 0.937\rlap{$^{\text{\tiny ***}}$} & 0.535\rlap{$^{\text{\tiny ***}}$} \\
                                   & (0.026) & (0.026) & (0.028) & (0.031) \\
    \midrule
    Product Fixed Effects & \multicolumn{1}{c}{Yes} & \multicolumn{1}{c}{Yes} & \multicolumn{1}{c}{Yes} & \multicolumn{1}{c}{Yes} \\
    Observations          & \multicolumn{1}{c}{25,066} & \multicolumn{1}{c}{26,033} & \multicolumn{1}{c}{24,157} & \multicolumn{1}{c}{22,121} \\
    Choice Sets (Groups)  & \multicolumn{1}{c}{3,931} & \multicolumn{1}{c}{3,926} & \multicolumn{1}{c}{3,942} & \multicolumn{1}{c}{3,938} \\
    Pseudo R-squared      & \multicolumn{1}{c}{0.51} & \multicolumn{1}{c}{0.51} & \multicolumn{1}{c}{0.53} & \multicolumn{1}{c}{0.68} \\
    \bottomrule
  \end{tabular}
  \begin{tablenotes}[para]
    \item \tnote{*} $p < 0.05$, \tnote{**} $p < 0.01$, \tnote{***} $p < 0.001$.
  \end{tablenotes}
\end{threeparttable}
\end{table}

\newpage

\begin{table}[!htbp]
\centering
\begin{threeparttable}
  \caption{Estimates of the Conditional Logit Regression for Gemini 2.5 Flash (VLM agents)}
  \label{tab:conditional-logit-results-gemini-aug-sep}
  \footnotesize
  \sisetup{
    input-symbols         = (), 
    table-align-text-post = false
  }
  \begin{tabular}{l
                  S[table-format=-1.3]
                  S[table-format=-1.3]
                  S[table-format=-1.3]
                  S[table-format=-1.3]}
    \toprule
    & \multicolumn{1}{c}{Default} & \multicolumn{1}{c}{Default} & \multicolumn{1}{c}{Ignore Position } & \multicolumn{1}{c}{Ignore Position \& Prioritize Price} \\
    & \multicolumn{1}{c}{\small (Fig.~\ref{fig:default-prompt-buying-agent-mock-app}; Aug'25)} &
      \multicolumn{1}{c}{\small (Fig.~\ref{fig:default-prompt-buying-agent-mock-app}; Sep'25)} &
      \multicolumn{1}{c}{\small (Fig.~\ref{fig:ignore-position-prompt-buying-agent-computer-use}; Sep'25)} &
      \multicolumn{1}{c}{\small (Fig.~\ref{fig:price-sensitive-prompt-buying-agent-computer-use}; Sep'25)} \\
    \midrule
    \multicolumn{5}{l}{\textit{Position effects}} \\
    Row 1    & 0.344\rlap{$^{\text{\tiny ***}}$} & 0.659\rlap{$^{\text{\tiny ***}}$} & 0.420\rlap{$^{\text{\tiny ***}}$} & -0.206\rlap{$^{\text{\tiny ***}}$} \\
             & (0.041) & (0.042) & (0.041) & (0.050) \\[0.6ex]
    Column 1 & -0.264\rlap{$^{\text{\tiny ***}}$} & -0.247\rlap{$^{\text{\tiny ***}}$} & -0.058 & -0.248\rlap{$^{\text{\tiny ***}}$} \\
             & (0.057) & (0.057) & (0.058) & (0.071) \\[0.6ex]
    Column 2 & -0.742\rlap{$^{\text{\tiny ***}}$} & -0.701\rlap{$^{\text{\tiny ***}}$} & -0.361\rlap{$^{\text{\tiny ***}}$} & -0.212\rlap{$^{\text{\tiny **}}$} \\
             & (0.061) & (0.060) & (0.060) & (0.070) \\[0.6ex]
    Column 3 & 0.162\rlap{$^{\text{\tiny **}}$} & 0.092 & 0.297\rlap{$^{\text{\tiny ***}}$} & -0.138\rlap{$^{\text{\tiny *}}$} \\
             & (0.054) & (0.054) & (0.055) & (0.070) \\
    \midrule
    \multicolumn{5}{l}{\textit{Badge effects}} \\
    Sponsored Tag    & -0.263\rlap{$^{\text{\tiny ***}}$} & -0.223\rlap{$^{\text{\tiny ***}}$} & -0.419\rlap{$^{\text{\tiny ***}}$} & -0.260\rlap{$^{\text{\tiny **}}$} \\
                     & (0.067) & (0.067) & (0.068) & (0.082) \\[0.6ex]
    Overall Pick Tag & 1.897\rlap{$^{\text{\tiny ***}}$} & 2.183\rlap{$^{\text{\tiny ***}}$} & 1.482\rlap{$^{\text{\tiny ***}}$} & 0.285\rlap{$^{\text{\tiny **}}$} \\
                     & (0.072) & (0.071) & (0.073) & (0.103) \\[0.6ex]
    Scarcity Tag     & -0.342\rlap{$^{\text{\tiny ***}}$} & -0.224\rlap{$^{\text{\tiny *}}$} & -0.131 & -0.207 \\
                     & (0.098) & (0.096) & (0.093) & (0.114) \\
    \midrule
    \multicolumn{5}{l}{\textit{Attribute effects}} \\
    $\ln(\text{Price})$            & -2.190\rlap{$^{\text{\tiny ***}}$} & -1.957\rlap{$^{\text{\tiny ***}}$} & -2.035\rlap{$^{\text{\tiny ***}}$} & -6.612\rlap{$^{\text{\tiny ***}}$} \\
                                   & (0.080) & (0.078) & (0.079) & (0.152) \\[0.6ex]
    Rating                         & 5.388\rlap{$^{\text{\tiny ***}}$} & 5.739\rlap{$^{\text{\tiny ***}}$} & 5.965\rlap{$^{\text{\tiny ***}}$} & 1.453\rlap{$^{\text{\tiny ***}}$} \\
                                   & (0.218) & (0.223) & (0.227) & (0.222) \\[0.6ex]
    $\ln(\text{Num.\ of Reviews})$ & 0.501\rlap{$^{\text{\tiny ***}}$} & 0.491\rlap{$^{\text{\tiny ***}}$} & 0.636\rlap{$^{\text{\tiny ***}}$} & 0.179\rlap{$^{\text{\tiny ***}}$} \\
                                   & (0.023) & (0.023) & (0.024) & (0.027) \\
    \midrule
    Product Fixed Effects & \multicolumn{1}{c}{Yes} & \multicolumn{1}{c}{Yes} & \multicolumn{1}{c}{Yes} & \multicolumn{1}{c}{Yes} \\
    Observations          & \multicolumn{1}{c}{25,215} & \multicolumn{1}{c}{26,649} & \multicolumn{1}{c}{25,272} & \multicolumn{1}{c}{20,748} \\
    Choice Sets (Groups)  & \multicolumn{1}{c}{3,953} & \multicolumn{1}{c}{3,949} & \multicolumn{1}{c}{3,964} & \multicolumn{1}{c}{3,709} \\
    Pseudo R-squared      & \multicolumn{1}{c}{0.42} & \multicolumn{1}{c}{0.43} & \multicolumn{1}{c}{0.43} & \multicolumn{1}{c}{0.58} \\
    \bottomrule
  \end{tabular}
  \begin{tablenotes}[para]
    \item \tnote{*} $p < 0.05$, \tnote{**} $p < 0.01$, \tnote{***} $p < 0.001$.
  \end{tablenotes}
\end{threeparttable}
\end{table}

\newpage
\newpage
\newpage

\subsection{Seller Response to AI-assisted shopping}


In this section, we will discuss the seller response to AI-assisted shopping. In Section \ref{app:illustrative-example-office-lamp}, we provide a case study example for the office lamp product category where the randomly selected focal product is the brand \textcolor{red}{SUNMORY}. In Section \ref{app:seller-response-category-wise-results}, we provide category-wise product-level changes in market share before and after modifying the focal product's description.

\subsubsection{Case Study of Seller's Response to AI-assisted shopping}
\label{app:illustrative-example-office-lamp}

We consider an illustrative example of the office lamp product category. Within this category, our focal product (randomly selected) is the brand {SUNMORY}. We prompt an AI seller agent with the simulated demand on the original product descriptions of the eight products in the catalog for different AI buying agents. For Claude Sonnet 4, GPT-4.1, and Gemini 2.5 Flash as buying agents, we use GPT-4.1 as the seller agent, while for Claude Opus 4.5, GPT-5.1, and Gemini 3.0 Pro Preview, we use GPT-5.1 (Thinking) as the seller agent. 

\begin{figure}[!htb]
    \centering
    \includegraphics[width=0.7\linewidth]{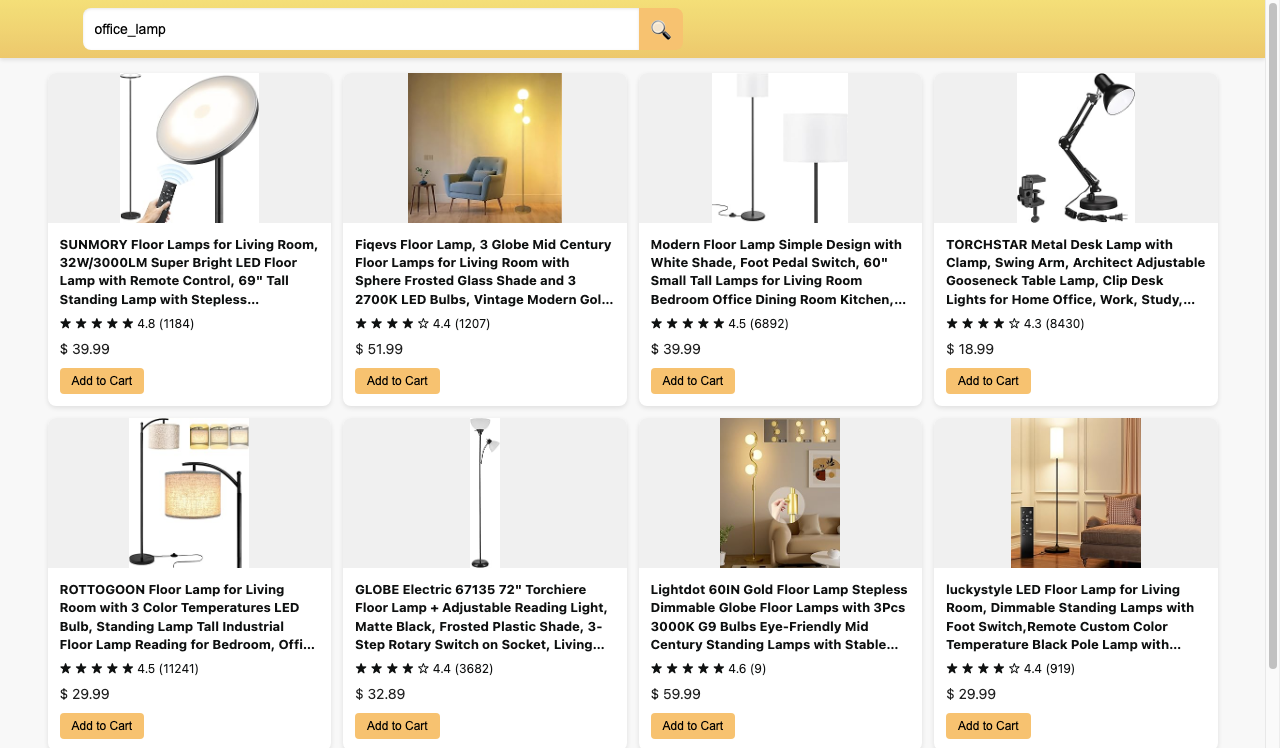}
    \caption{Screenshot provided to the seller agent GPT-5.1 providing the listing information of the focal product and its competitors}
    \label{fig:seller-response-screenshot}
\end{figure}

As an example, we provide the specific prompt used for the Claude Opus 4.5 buying model with GPT-5.1 (Thinking) as the seller agent in Figure \ref{fig:office-lamp-seller-response-prompt}. Note that similar prompts are used for different buying agents, with the only difference across the prompts being the sales data from our initial set of 200 experiments, which varies across different buying agents (as shown in \textcolor{blue}{blue}). Note that the part in \textcolor{magenta}{magenta} in Figure \ref{fig:office-lamp-seller-response-prompt} is specific to the focal product and changes across product categories. We are trying to customize the response of the seller agent to the specific simulated market shares under a given buying model; by customizing this for each buying model, we expect the seller agent to uncover the latent behavior of the buying model, which can help improve the description for the focal product. Additionally, along with the prompt, we also provide a screenshot of the entire product catalog with competitor listing information (see Figure \ref{fig:seller-response-screenshot}). The complete response of the seller agent (GPT-5.1) is provided in Figure \ref{fig:office-lamp-seller-agent-response}.

\begin{figure}[!htbp]
    \centering
    \begin{tcolorbox}[sellerbuyertemplate]
        \textbf{Context: There is an AI agent (a vision language model) which is given a screenshot of an e-commerce website selling a particular product and it decides on the products based on the given attributes. I have the sales data on all the products including mine. I want you to act as an agent on my behalf and suggest changes to the product title so that I can increase my sales by making my product more appealing to the AI agent (Claude Opus 4.5)}

        \vspace{0.5em}
        
        \textbf{My product is} \textcolor{magenta}{SUNMORY Floor Lamps for Living Room, 32W/3000LM Super Bright LED Floor Lamp with Remote Control, 69" Tall Standing Lamp with Stepless Dimmable, Modern Torchiere Corner Lamp for Bedroom, Office.} 

        \vspace{0.5em}
        \textbf{My product features are:}
        \par\smallskip
        
        \noindent
        \begin{tabularx}{\linewidth}{@{} >{\color{magenta}}l >{\color{magenta}}X @{}}
             Finish Type & Unfinished Base \\ 
             Material & Plastic  \\
             Product Dimensions & 9.8"D x 9.8"W x 69"H \\
             Item Weight & 8.9 Pounds \\
             Lamp Type & Floor Lamp \\
        \end{tabularx}

        \vspace{0.5em}
        \textcolor{magenta}{
        Super Brightness and Dimmable Floor Lamp : SUNMORY’s new LED floor lamp for living room supports stepless dimming brightness (5\%-100\%) and color temperature (3000K-6000K). And this is a super bright floor lamp with 32W/3000LM, providing sufficient lighting for your living room or bedroom. Whether you're reading, writing, standing lamp provide soft, bright light that's easier on the eyes. The 69" LED lights for bedroom has a slim profile that blends perfectly into any style of room decor. Remote Control \& Multi-Function : The floor lamp with remote control has the ability to easily manage the settings of the lamp from anywhere within 50 feet. With the remote control, you can easily switch between 4 lighting modes(reading, feeding, working and night light modes) to meet your various scene needs. This standing lamp is perfect as a corner lamp, lamps for living room, lamp for bedroom, or office lamp. It also features a memory function to remember your last lighting settings. Rotatable Lamp Head and Heavy Base : The living room lamps can be rotated 350° horizontally and 180° longitudinally to guide the light as needed. You can easily rotate the bracket of the tall lamp to the desired angle to illuminate every corner of the room. Moreover, the SUNMORY lights for bedroom has a thick base, which is strong and stable enough, so you don’t have to worry about tilting it too much, making it very safe and reliable. Long Lifespan \& Energy Saving : Modern floor lamps use high-brightness, low-heat LED advanced technology. Under the correct current and voltage, these LED lamp beads have a service life of up to 30,000 hours, and their brightness is comparable to 200W incandescent lamps. But it consumes less energy and is more efficient, saving more than 80\% of lighting electricity bills. 80+ high color rendering index brings you vivid and natural colors, better visual enjoyment and a comfortable atmosphere. Easy to Assemble : Please refer to the user manual included in our package, which explains the exact installation of the black floor lamp with a video. Simple enough to ensure you only need about 5-10 minutes to successfully complete the installation. If you have any questions or are unlikely to be dissatisfied with the lamps for living room after receiving it, please feel free to contact us. We will deal with it as soon as possible to achieve a mutually satisfactory result.
        }
        \vspace{0.5em}

        \textbf{I have data from 200 experiments and the following is my and my competitors sales data across experiments:}

        \vspace{0.5em}

        \noindent
        \begin{tabularx}{\linewidth}{@{} >{\raggedright\arraybackslash\color{blue}}X >{\color{blue}}r @{}}
            \toprule
            \textbf{Product Name} & \textbf{Times Selected} \\
            \midrule
            TORCHSTAR Metal Desk Lamp with Clamp, Swing Arm, Architect Adjustable Gooseneck Table Lamp, Clip Desk Lights for Home Office, Work, Study, Reading, E26 Base, Multi-Joint, Black & 182 \\
            \addlinespace
            ROTTOGOON Floor Lamp for Living Room with 3 Color Temperatures LED Bulb, Standing Lamp Tall Industrial Floor Lamp Reading for Bedroom, Office (9W LED Bulb, Beige Lampshade Included) -Black & 18 \\
            \addlinespace
            luckystyle LED Floor Lamp for Living Room, Dimmable Standing Lamps with Foot Switch,Remote Custom Color Temperature Black Pole Lamp with Lampshade,12W Led Bulb Tall Lamps for Bedroom Office Kid Room & 0 \\
            \addlinespace
            SUNMORY Floor Lamps for Living Room, 32W/3000LM Super Bright LED Floor Lamp with Remote Control, 69" Tall Standing Lamp with Stepless Dimmable, Modern Torchiere Corner Lamp for Bedroom, Office & 0 \\
            \addlinespace
            GLOBE Electric 67135 72" Torchiere Floor Lamp + Adjustable Reading Light, Matte Black, Frosted Plastic Shade, 3-Step Rotary Switch on Socket, Living Room Décor, Reading Light, Home Essentials & 0 \\
            \addlinespace
            Lightdot 60IN Gold Floor Lamp Stepless Dimmable Globe Floor Lamps with 3Pcs 3000K G9 Bulbs Eye-Friendly Mid Century Standing Lamps with Stable Base for Living Room/Bedroom/Corner/Office & 0 \\
            \addlinespace
            Modern Floor Lamp Simple Design with White Shade, Foot Pedal Switch, 60" Small Tall Lamps for Living Room Bedroom Office Dining Room Kitchen, Black Pole Lamp(Without Bulb) & 0 \\
            \addlinespace
            Fiqevs Floor Lamp, 3 Globe Mid Century Floor Lamps for Living Room with Sphere Frosted Glass Shade and 3 2700K LED Bulbs, Vintage Modern Gold Standing Lamp for Bedrooms, Office - Antique Brass & 0 \\
            \bottomrule
        \end{tabularx}

        \vspace{0.5em}
        \textbf{It is important that the title changes you suggest align with the product features provided to you. Do not make up product features or add spurious keywords. Just use the product feature information provided.}
    \end{tcolorbox}
    \caption{Exact prompt provided to GPT-5.1 to get AI-recommended listing description}
    \label{fig:office-lamp-seller-response-prompt}
\end{figure}

\begin{figure}[!htb]
\centering
\begin{tcolorbox}[sellerbuyertemplate]
    Here's how I'd ``talk'' to Claude Opus 4.5 on your behalf. 
    
    From your 200 experiments, Claude is basically obsessed with the TORCHSTAR ``office desk lamp'' title (182/200 picks) and almost ignores everything clearly labeled ``living room floor lamp.'' So we want to:

    \begin{itemize}
        \item Move ``Office / Office Lamp'' to the front.
        \item Keep ``Floor Lamp / Tall Standing Lamp'' so we're honest about the type.
        \item Front-load 32W / 3000LM / LED / Remote / Stepless Dimmable / 3000K--6000K / Rotatable Head / Heavy Base.
        \item Still stay fully consistent with your provided specs and description (no made-up features).
    \end{itemize}

    Here are three title variants that do that, all truthful to your product:

    \vspace{0.5em}
    \textbf{Option 1 -- Office focus + core specs up front (primary suggestion)} \\
    \textit{SUNMORY 32W 3000LM LED Office Floor Lamp with Remote Control, 69'' Tall Standing Lamp, Stepless Dimmable 3000K--6000K, Rotatable Head Corner Lamp for Living Room, Bedroom}

    \vspace{0.5em}
    \textbf{Option 2 -- Emphasize dimming + modes (for task/office use)} \\
    \textit{SUNMORY LED Floor Lamp for Office and Living Room, 32W Super Bright 3000LM, 69'' Tall Standing Lamp with Remote, Stepless Dimming Brightness \& Color Temperature (3000K--6000K), 4 Lighting Modes (Reading/Feeding/Working/Night)}

    \vspace{0.5em}
    \textbf{Option 3 -- Emphasize stability + office wording} \\
    \textit{SUNMORY LED Office Lamp, 32W/3000LM Tall Floor Lamp with Remote Control, Stepless Dimmable 3000K--6000K, Rotatable Lamp Head \& Heavy Stable Base, Modern Corner Standing Lamp for Bedroom, Living Room}

    \vspace{1em}
    \textbf{Why these should look tastier to Claude:}
    \begin{itemize}
        \item The query is ``office\_lamp''; all three now hit ``Office'' + ``Lamp'' right in the early tokens, unlike your current ``Floor Lamps for Living Room\dots'' start.
        \item We preserve all factual features: 32W/3000LM, LED, 69'' height, floor lamp, remote control, stepless dimming, 3000K--6000K, 4 modes, rotatable head, heavy base, corner/living room/bedroom/office use, energy-saving LED (implicitly via ``LED'').
        \item Structurally they now look more like the winning TORCHSTAR title (clear primary use, then details), while still honestly remaining a floor lamp.
    \end{itemize}

    You can A/B/C these three against your current title in the same experimental setup and see which wording nudges Claude away from its TORCHSTAR crush.
\end{tcolorbox}
\caption{AI Seller Agent's (GPT-5.1) Response}
\label{fig:office-lamp-seller-agent-response}
\end{figure}

The original description/title of the SUNMORY product is ``\textcolor{red}{\small \sf SUNMORY Floor Lamps for Living Room, 32W/3000LM Super Bright LED Floor Lamp with Remote Control, 69" Tall Standing Lamp with Stepless Dimmable, Modern Torchiere Corner Lamp for Bedroom, Office}''. In Table \ref{tab:seller-response-office-lamp-recommended-descriptions}, we provide the recommended description/title for different AI buying agents which we finally used in our experiments. Note that in five out of six recommended descriptions, the word ``Office'' (which gets cut off in the original description) is front loaded to appear early in the title, ensuring it is prioritized during token processing by the buying agents.

\begin{table}[!htb]
\centering
\begin{threeparttable}
    \caption{Descriptions recommended by AI seller agent for different buying models}
    \label{tab:seller-response-office-lamp-recommended-descriptions}
    \footnotesize 
    \sisetup{
      input-symbols           = (), 
      table-align-text-post = false 
    }
    \begin{tabular}{lp{12cm}}
        \toprule
        AI Buying Model & Recommended Title/Description \\
        \midrule
        \multicolumn{2}{l}{{\it AI Seller Agent is GPT-4.1}} \\
        Claude Sonnet 4 & {\scriptsize \sf SUNMORY Office Floor Lamp, 32W 3000LM Super Bright LED with Remote Control, 69" Adjustable Standing Lamp, Stepless Dimmable Lighting, Modern Slim Floor Lamp for Workspace, Study, or Home Office} \\
        GPT-4.1 & {\scriptsize \sf SUNMORY 69" LED Floor Lamp, 32W 3000LM Super Bright, Stepless Dimmable \& Color Temp, Remote Control, Rotatable Head, Stable Base, Modern Torchiere Standing Lamp for Living Room, Bedroom, Office} \\
        Gemini 2.5 Flash & {\scriptsize \sf SUNMORY LED Office Floor Lamp, Remote Control, 32W/3000LM Super Bright, Stepless Dimmable, 69" Adjustable Standing Lamp with Rotatable Head, Energy Saving, for Bedroom \& Living Room} \\
        \midrule
        \multicolumn{2}{l}{{\it AI Seller Agent is GPT-5.1}} \\
        Claude Opus 4.5 & {\scriptsize \sf SUNMORY 32W 3000LM LED Office Floor Lamp with Remote Control, 69" Tall Standing Lamp, Stepless Dimmable 3000K–6000K, Rotatable Head Corner Lamp for Living Room, Bedroom} \\ 
        GPT-5.1 & {\scriptsize \sf SUNMORY Office Floor Lamp, 32W 3000LM Super Bright 69" LED Standing Lamp with Remote Control, Stepless Dimmable 3000K–6000K, Modern Torchiere for Living Room, Bedroom, Office} \\
        Gemini 3.0 Pro Preview & {\scriptsize \sf SUNMORY Office LED Floor Lamp with Remote, 32W/3000LM Super Bright 69" Tall Standing Lamp, Stepless Dimmable 3000K–6000K for Living Room, Bedroom \& Home Office} \\
        \bottomrule
    \end{tabular}
  \end{threeparttable}
\end{table}

\paragraph{Impact on Market Share.} After running an initial set of experiments with the original product description/title, we run a new set of experiments where we just change the product description (using the titles from Table \ref{tab:seller-response-office-lamp-recommended-descriptions}) of the focal product \textcolor{red}{SUNMORY}, with different AI buying agents. In this case, we find that this rather innocuous-looking change in the product description can significantly shift the market share. 

\begin{figure}[!htbp]
    \centering
    \begin{subfigure}{1\linewidth}
        \centering
        \includegraphics[width = 0.6\linewidth]{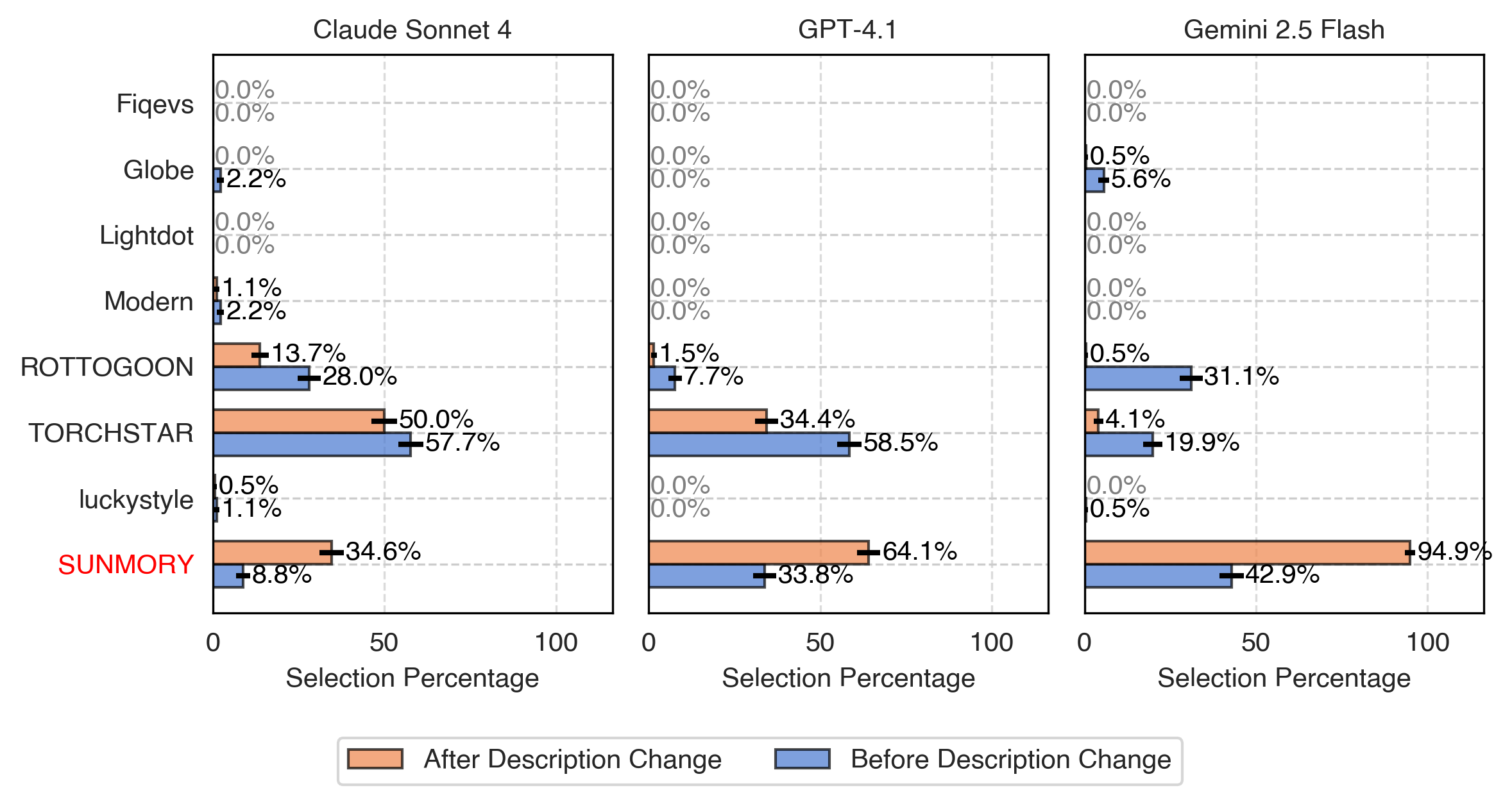}
    \end{subfigure}
    \par\bigskip 
    \begin{subfigure}{1\linewidth}
        \centering
        \includegraphics[width = 0.6\linewidth]{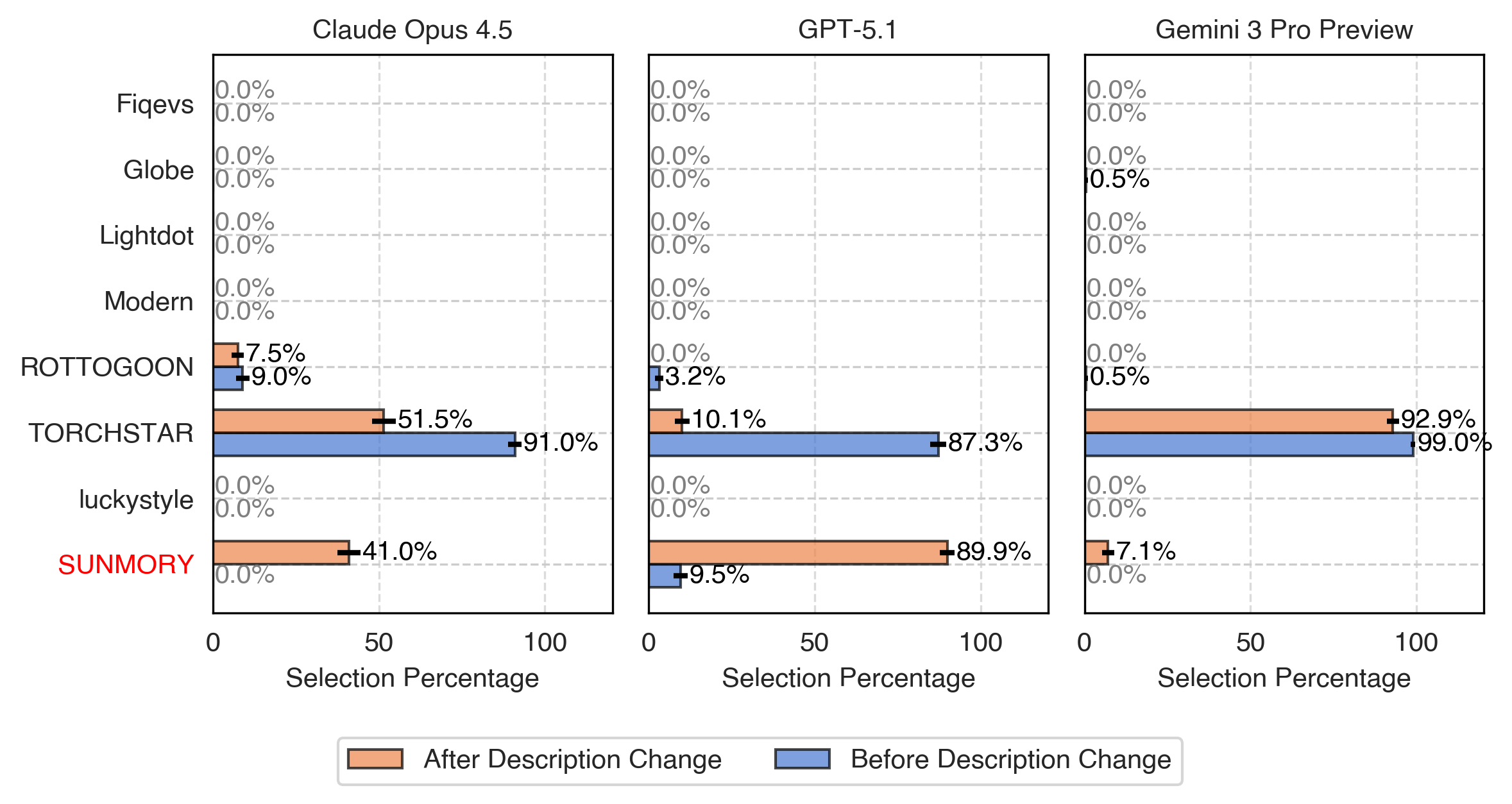}
    \end{subfigure}
    \caption{Changes in the market share for the different products in the office lamp category before and after the title/description change.}
    \label{fig:market-share-change-product-level-office-lamp}
\end{figure}

In Figure \ref{fig:market-share-change-product-level-office-lamp}, we plot the market share of different products for the office lamp category. We observe a statistically significant and large increase in the market share of the focal product across almost all buying agents. For example, when using GPT-5.1 as the AI buying agent (Figure \ref{fig:market-share-change-product-level-office-lamp} bottom panel), we observe that the market share of \textcolor{red}{SUNMORY} increases dramatically from $9.5\%$ to $89.9\%$ (an increase of over $80\%$). This comes primarily at the expense of the market leader, {TORCHSTAR}, whose share drops from $87.3\%$ to $10.1\%$. Similarly, for Claude Opus 4.5, the market share rises from $0.0\%$ to $41.0\%$. Note that the only change across the two sets of experiments is the change in the product description; hence, this impact is causal.

We posit that several factors drive this increased market share. The most prominent driver is the explicit front-loading of the keyword ``Office'' (e.g., ``SUNMORY Office Floor Lamp...''). As Large Language Models process text sequentially, tokens appearing early in the sequence often exert a disproportionate influence on the model's relevance scoring, particularly when matching against the user's query intent (``office\_lamp''). By moving ``Office'' from the end of the title (where it was previously truncated or ignored) to the first five words, the seller agent effectively reduced the semantic friction for the buying agent. Additionally, the recommended descriptions increase the density of technical specifications (``32W'', ``3000LM'', ``Stepless Dimmable'') in the title itself. This likely aligns with the buying agents' internal heuristics which may associate detailed technical metrics with higher product quality or suitability for a workspace environment.

Furthermore, the behavior of Claude Opus 4.5 highlights the exploitable sensitivities of AI agents. A shift from $0.0\%$ (essentially invisible to the buyer) to a dominant $41.0\%$ market share based solely on a title change is a magnitude of volatility rarely observed in human consumer behavior. In a traditional market, brand perception and product utility provide a buffer against such erratic swings. However, the AI buying agent exhibits a ``winner-take-all'' or ``cliff-edge'' probability distribution; once the product description crosses a specific threshold of semantic alignment, the model shifts its preference entirely. This suggests that the algorithmic market is far more brittle and susceptible to adversarial optimization (or AI-SEO) than human markets, where preferences are generally more distributed and stable.

We note, however, that the gains in market share are not uniform across different AI buying agents. For the case of Gemini 3.0 Pro Preview, the market share of \textcolor{red}{SUNMORY} only increases from $0.0\%$ to $7.1\%$ despite the optimized description, whereas for GPT-5.1 the gain is massive. This demonstrates that there can be non-uniform impact of such interventions depending on the underlying preferences and architecture of the buying model.

\subsubsection{Category-wise change in market share}
\label{app:seller-response-category-wise-results}
\begin{figure}[!htbp]
    \centering
    \begin{subfigure}[b]{0.48\linewidth}
        \includegraphics[width = \linewidth]{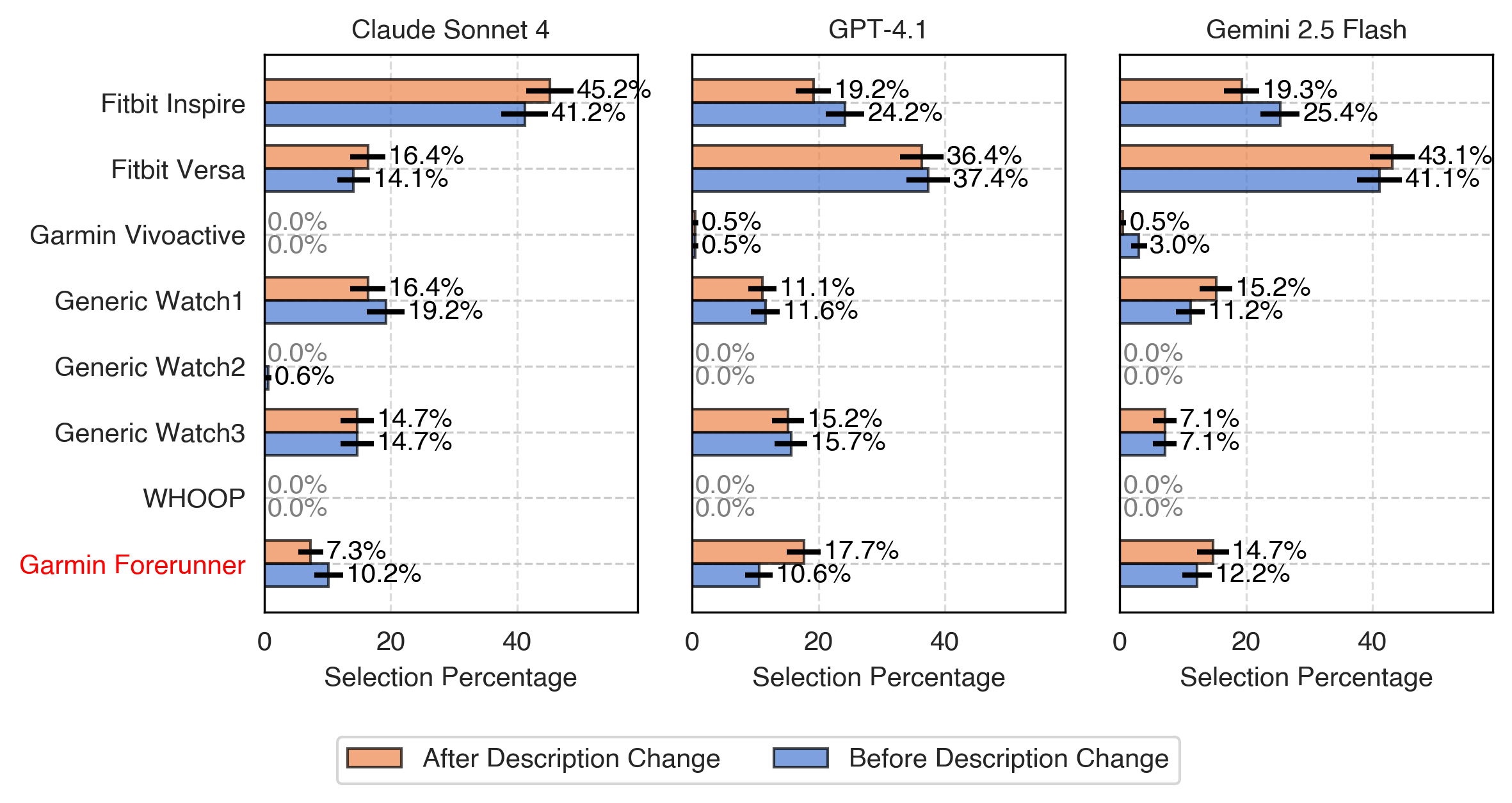}
        \caption{{fitness watch}}
        \label{fig:title-change-experiment-results-fitness-watch}
    \end{subfigure}
    \begin{subfigure}[b]{0.48\linewidth}
        \includegraphics[width = \linewidth]{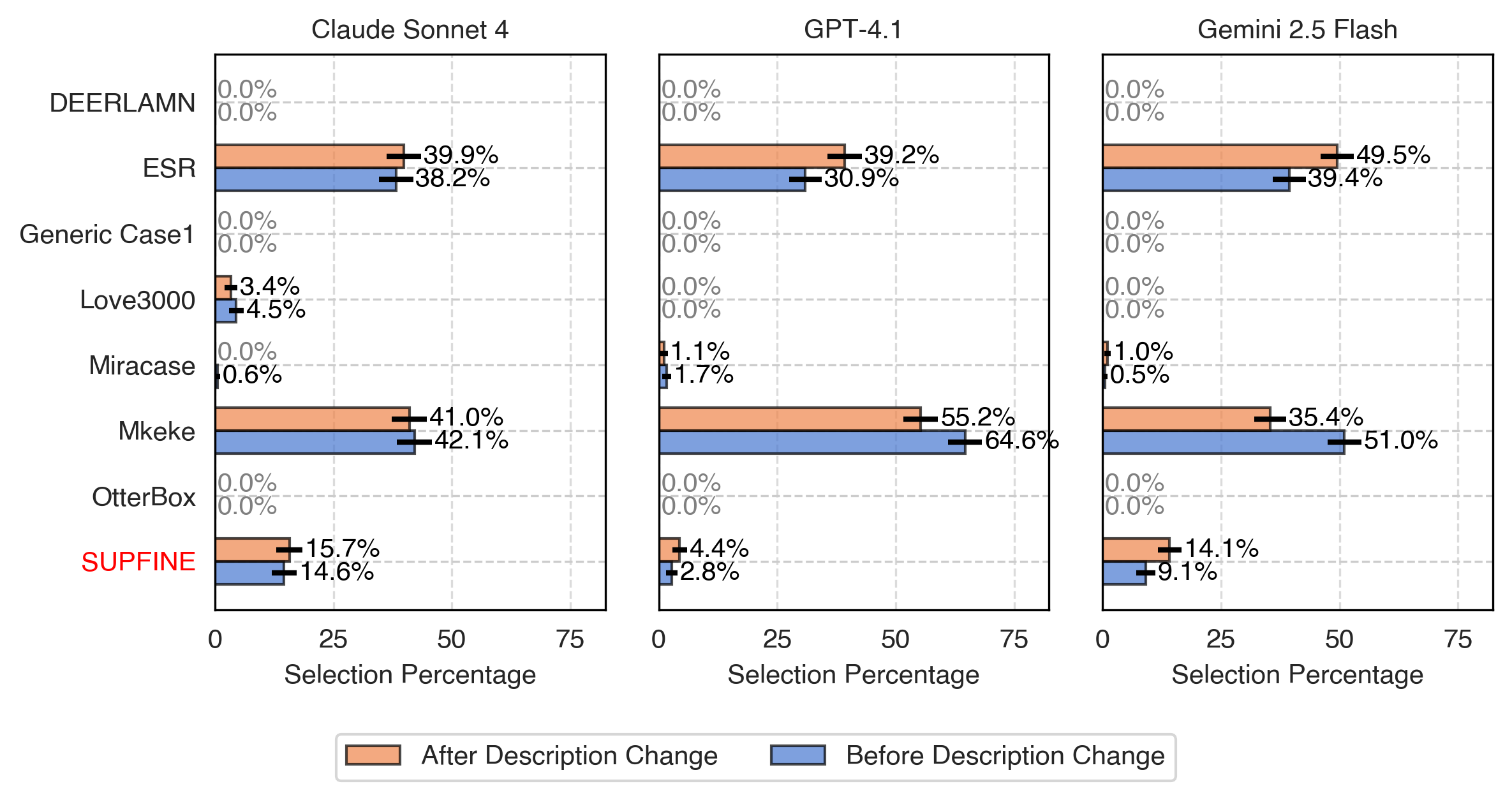}
        \caption{{iPhone 16 pro cover}}
        \label{fig:title-change-experiment-results-iphone-16-pro-cover}
    \end{subfigure}
    \begin{subfigure}[b]{0.48\linewidth}
        \includegraphics[width = \linewidth]{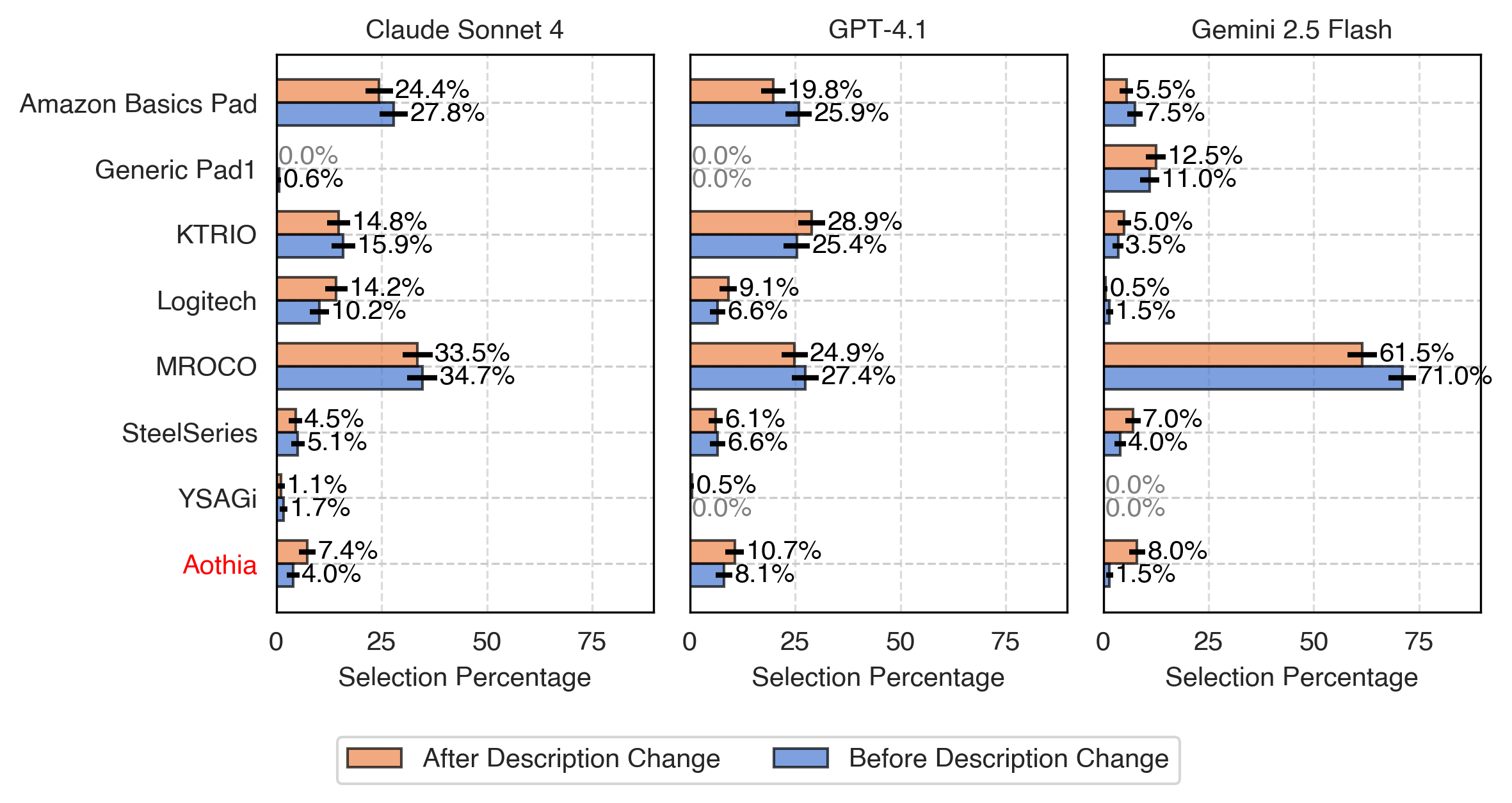}
        \caption{{ mousepad}}
        \label{fig:title-change-experiment-results-mousepad}
    \end{subfigure}
    \begin{subfigure}[b]{0.48\linewidth}
        \includegraphics[width = \linewidth]{figures/title_change_experiments/old-model-comparison/office_lamp_model_comparison.png}
        \caption{{office lamp}}
        \label{fig:title-change-experiment-results-office-lamp}
    \end{subfigure}
    \begin{subfigure}[b]{0.48\linewidth}
        \includegraphics[width = \linewidth]{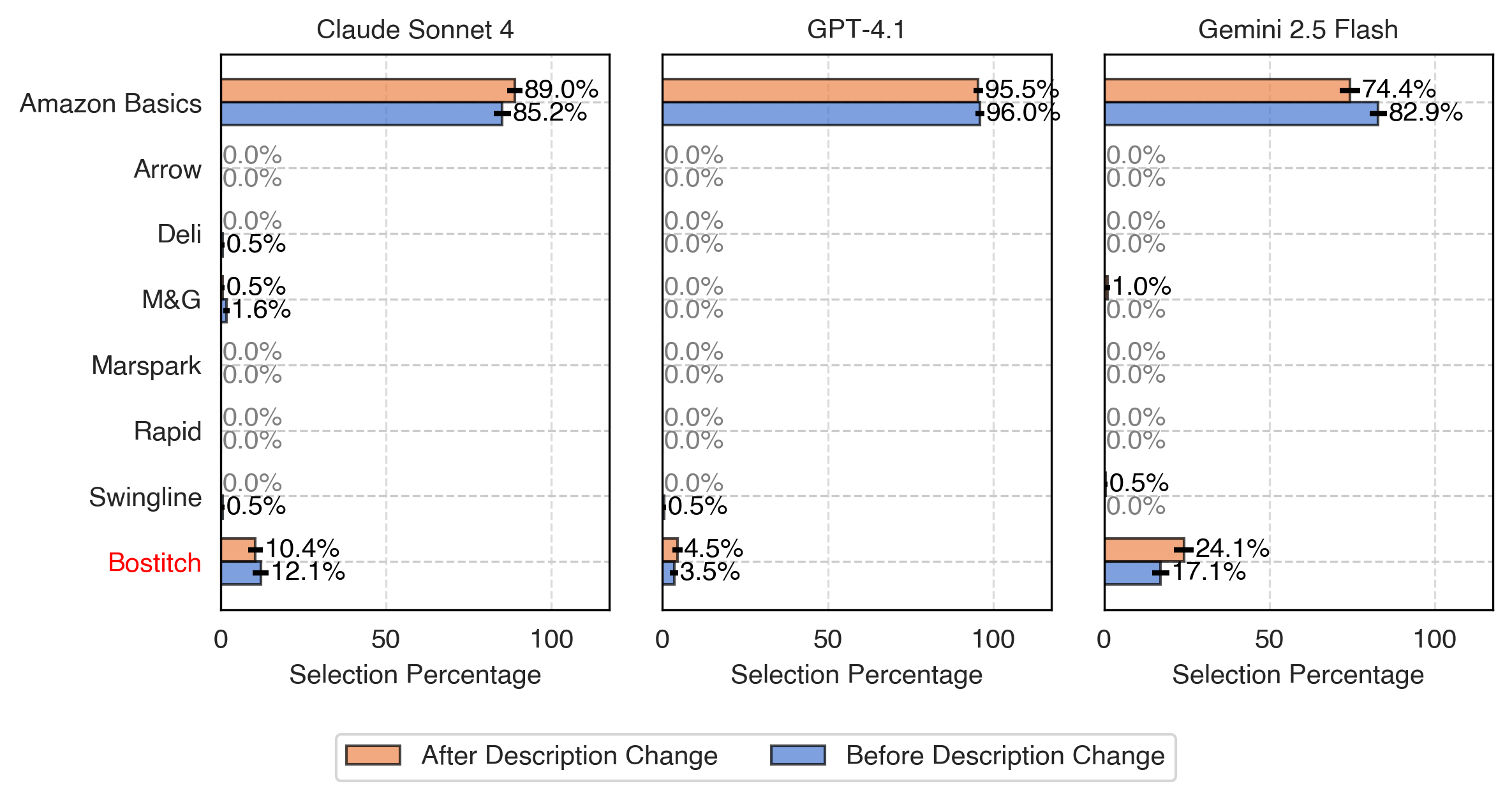}
        \caption{{stapler}}
        \label{fig:title-change-experiment-results-stapler}
    \end{subfigure}
    \begin{subfigure}[b]{0.48\linewidth}
        \includegraphics[width = \linewidth]{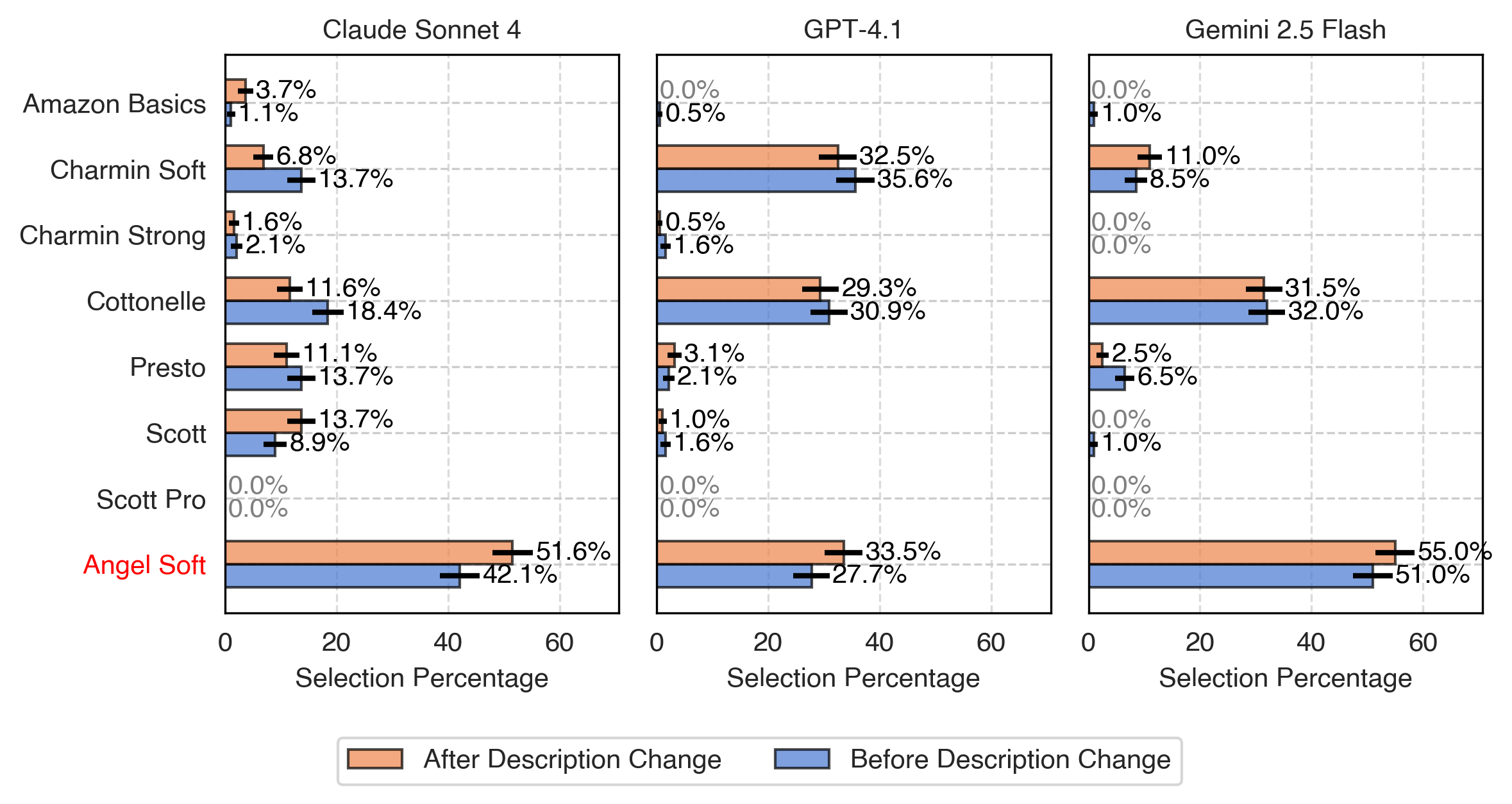}
        \caption{{toilet Paper}}
        \label{fig:title-change-experiment-results-toilet-paper}
    \end{subfigure}
    \begin{subfigure}[b]{0.48\linewidth}
        \includegraphics[width = \linewidth]{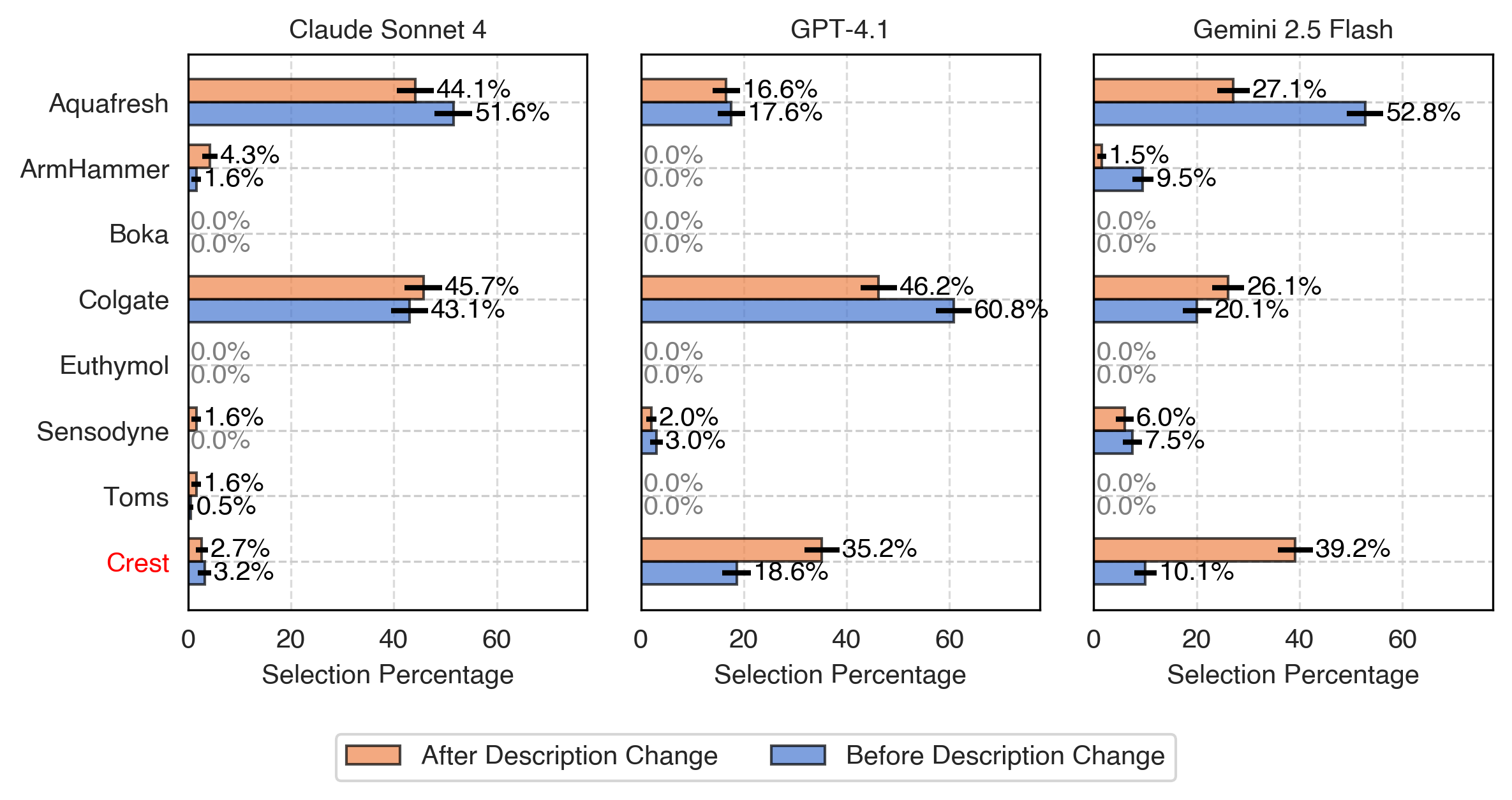}
        \caption{{toothpaste}}
        \label{fig:title-change-experiment-results-toothpaste}
    \end{subfigure}
    \begin{subfigure}[b]{0.48\linewidth}
        \includegraphics[width = \linewidth]{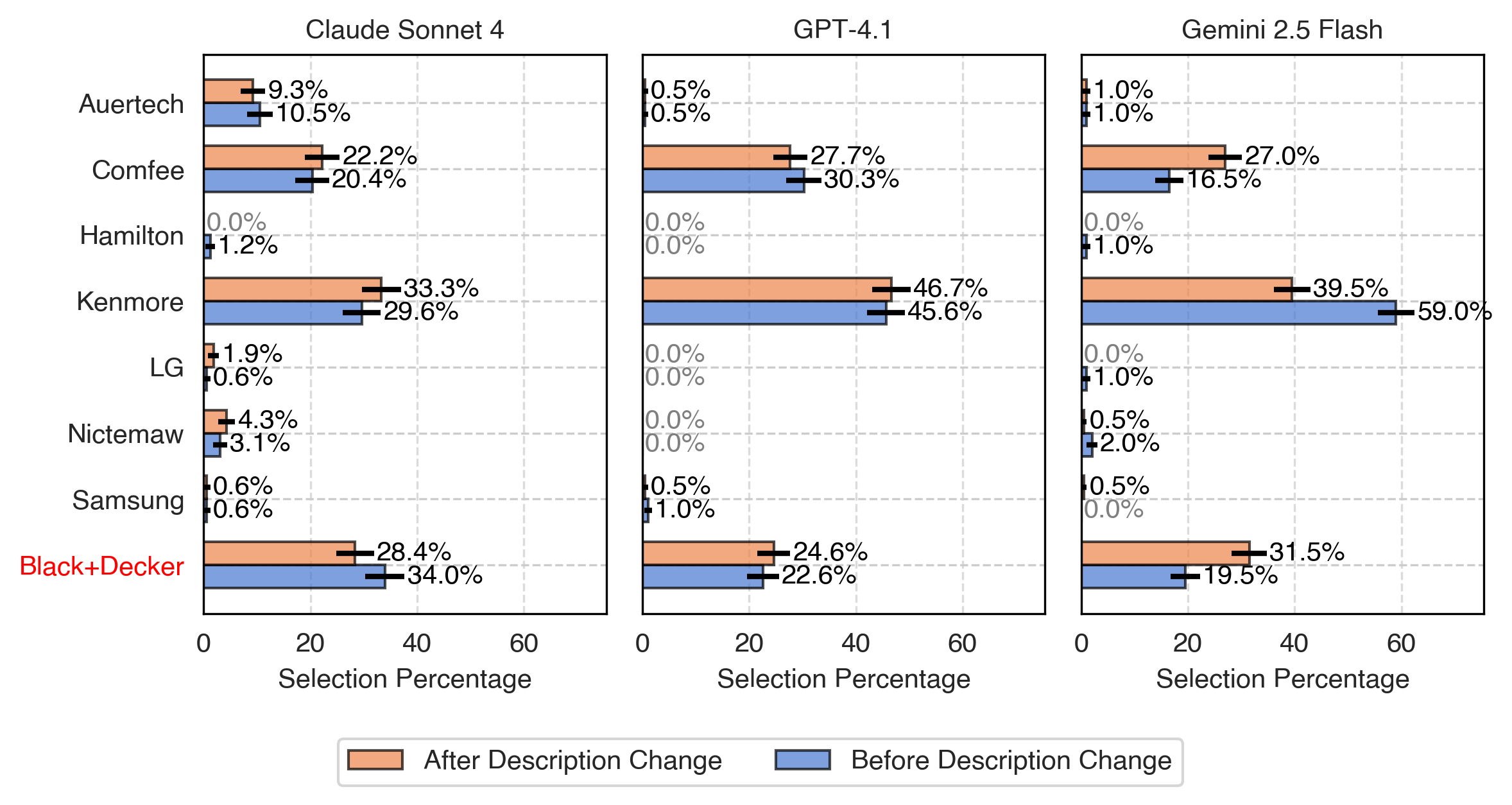}
        \caption{{washing machine}}
        \label{fig:title-change-experiment-results-washine-machine}
    \end{subfigure}
    \caption{Market shares before and after focal product description modification with the August 2025 state-of-the-art models (Claude Sonnet 4, GPT-4.1 and Gemini 2.5 Flash). The focal product is marked in \textcolor{red}{red}.}
    \label{fig:title-change-experiment-results-all-models}
\end{figure}

\begin{figure}[!htbp]
    \centering
    \begin{subfigure}[b]{0.48\linewidth}
        \includegraphics[width = \linewidth]{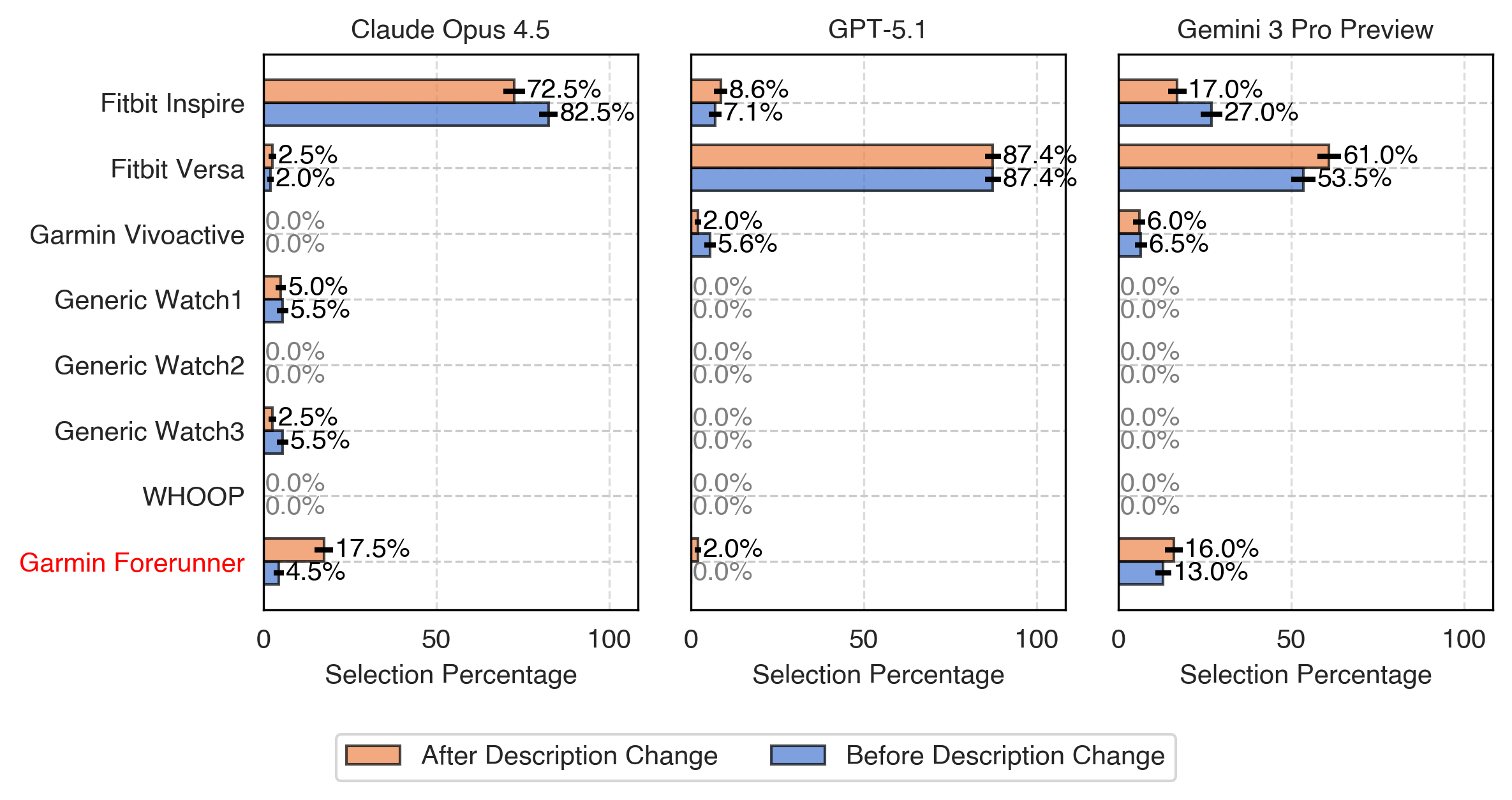}
        \caption{{fitness watch}}
        \label{fig:title-change-experiment-results-fitness-watch-new}
    \end{subfigure}
    \begin{subfigure}[b]{0.48\linewidth}
        \includegraphics[width = \linewidth]{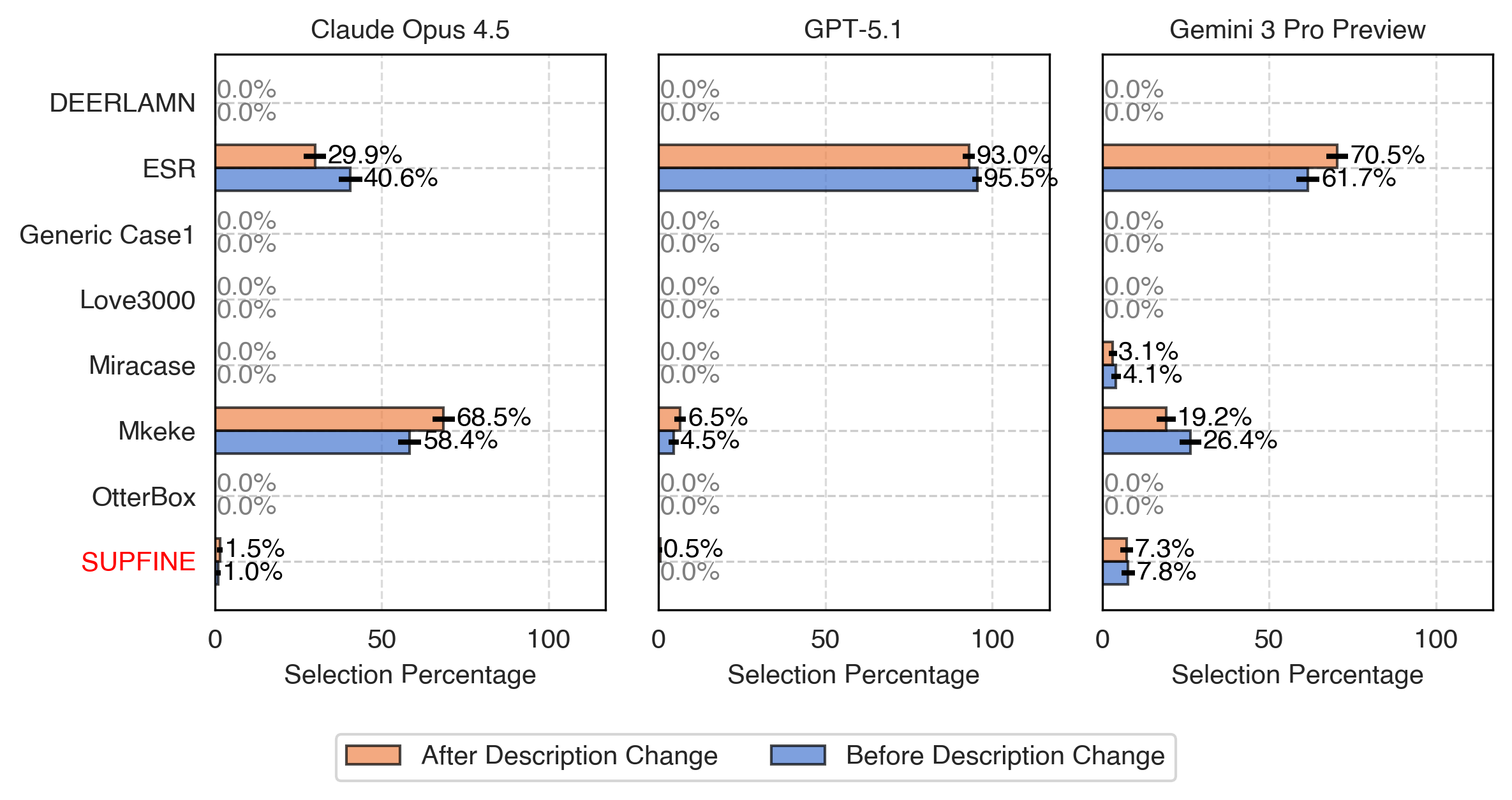}
        \caption{{iPhone 16 pro cover}}
        \label{fig:title-change-experiment-results-iphone-16-pro-cover-new}
    \end{subfigure}
    \begin{subfigure}[b]{0.48\linewidth}
        \includegraphics[width = \linewidth]{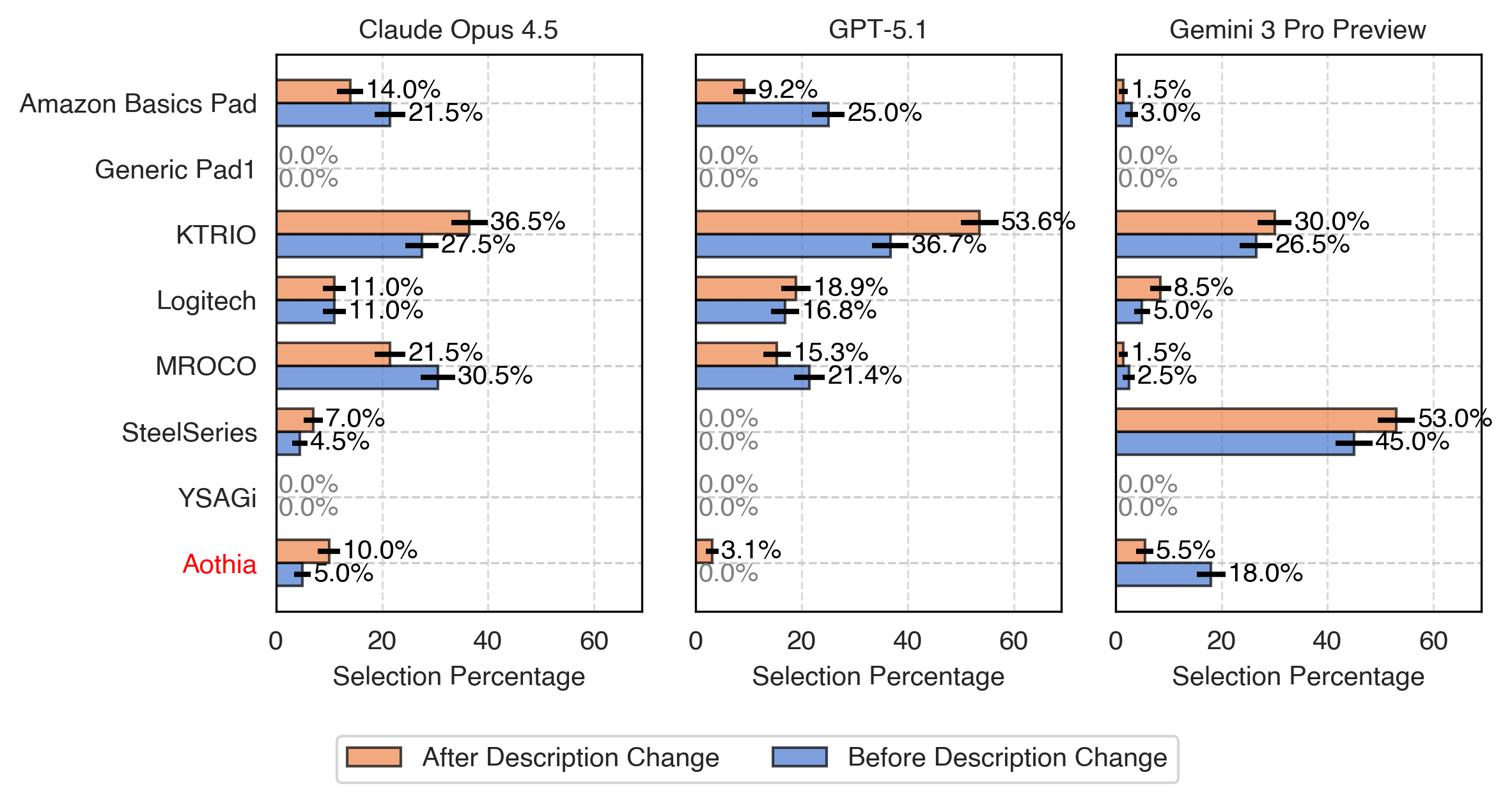}
        \caption{{ mousepad}}
        \label{fig:title-change-experiment-results-mousepad-new}
    \end{subfigure}
    \begin{subfigure}[b]{0.48\linewidth}
        \includegraphics[width = \linewidth]{figures/title_change_experiments/new-model-comparison/office_lamp_latest_model_comparison.png}
        \caption{{office lamp}}
        \label{fig:title-change-experiment-results-office-lamp-new}
    \end{subfigure}
    \begin{subfigure}[b]{0.48\linewidth}
        \includegraphics[width = \linewidth]{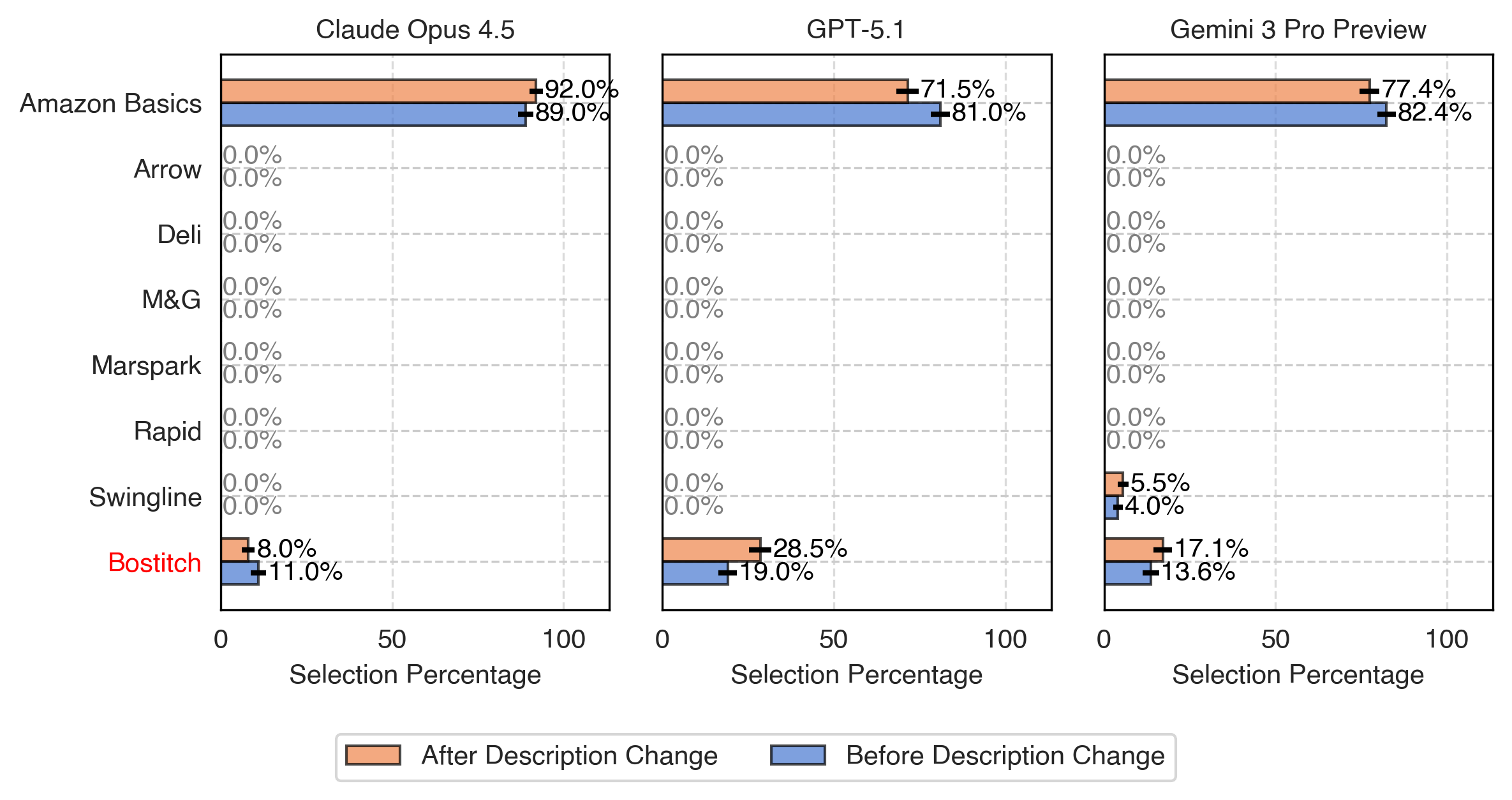}
        \caption{{stapler}}
        \label{fig:title-change-experiment-results-stapler-new}
    \end{subfigure}
    \begin{subfigure}[b]{0.48\linewidth}
        \includegraphics[width = \linewidth]{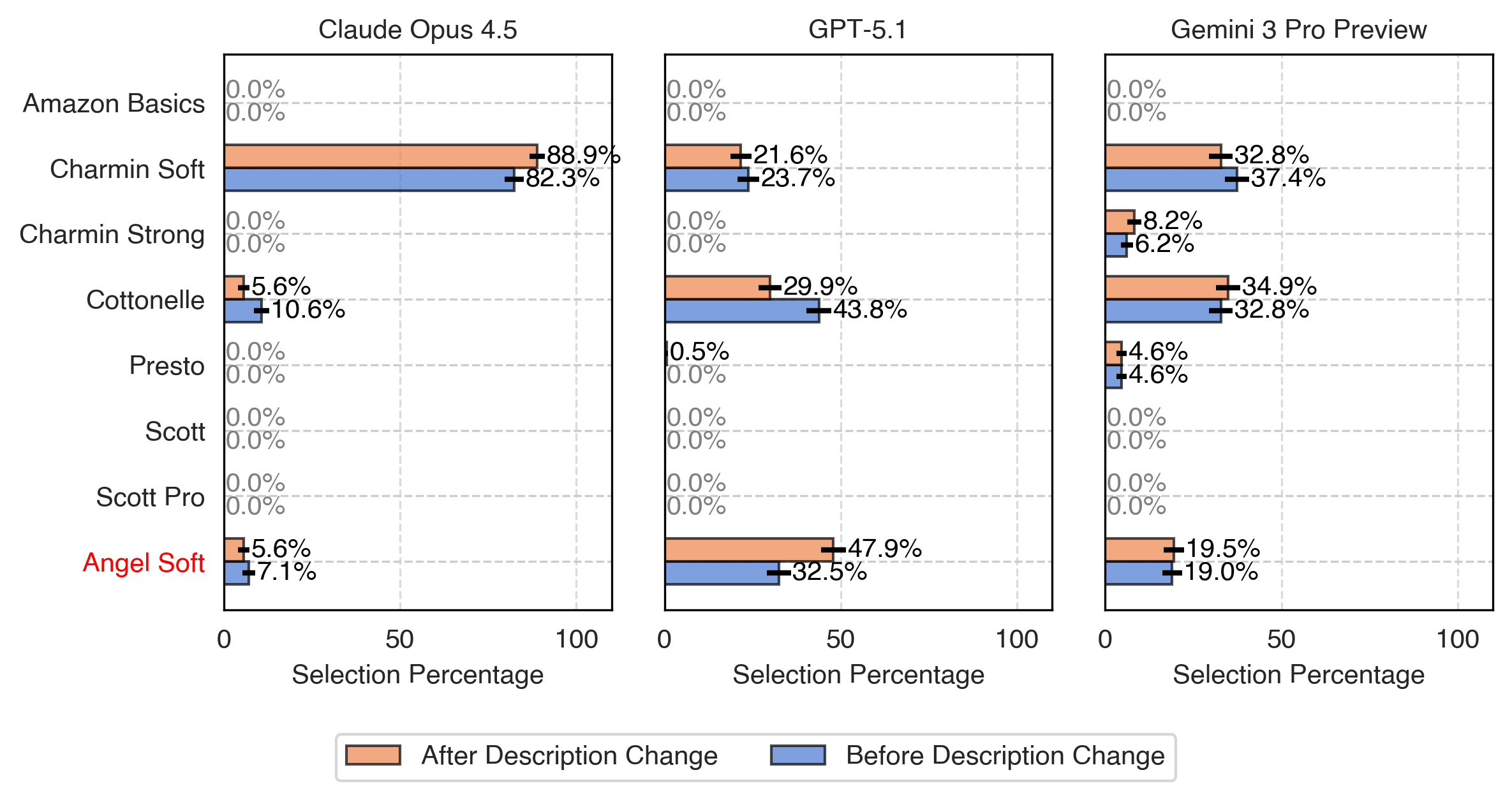}
        \caption{{toilet Paper}}
        \label{fig:title-change-experiment-results-toilet-paper-new}
    \end{subfigure}
    \begin{subfigure}[b]{0.48\linewidth}
        \includegraphics[width = \linewidth]{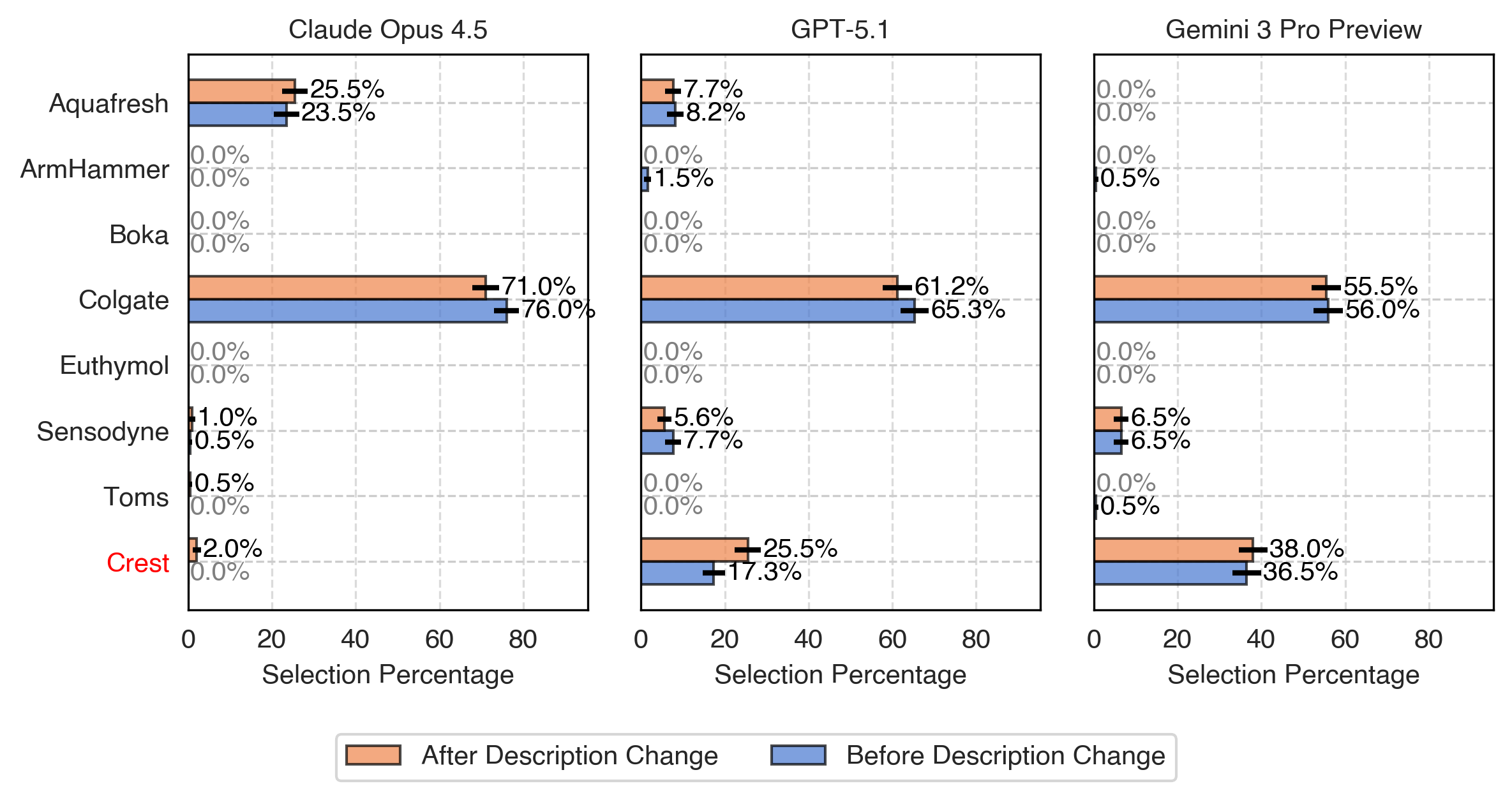}
        \caption{{toothpaste}}
        \label{fig:title-change-experiment-results-toothpaste-new}
    \end{subfigure}
    \begin{subfigure}[b]{0.48\linewidth}
        \includegraphics[width = \linewidth]{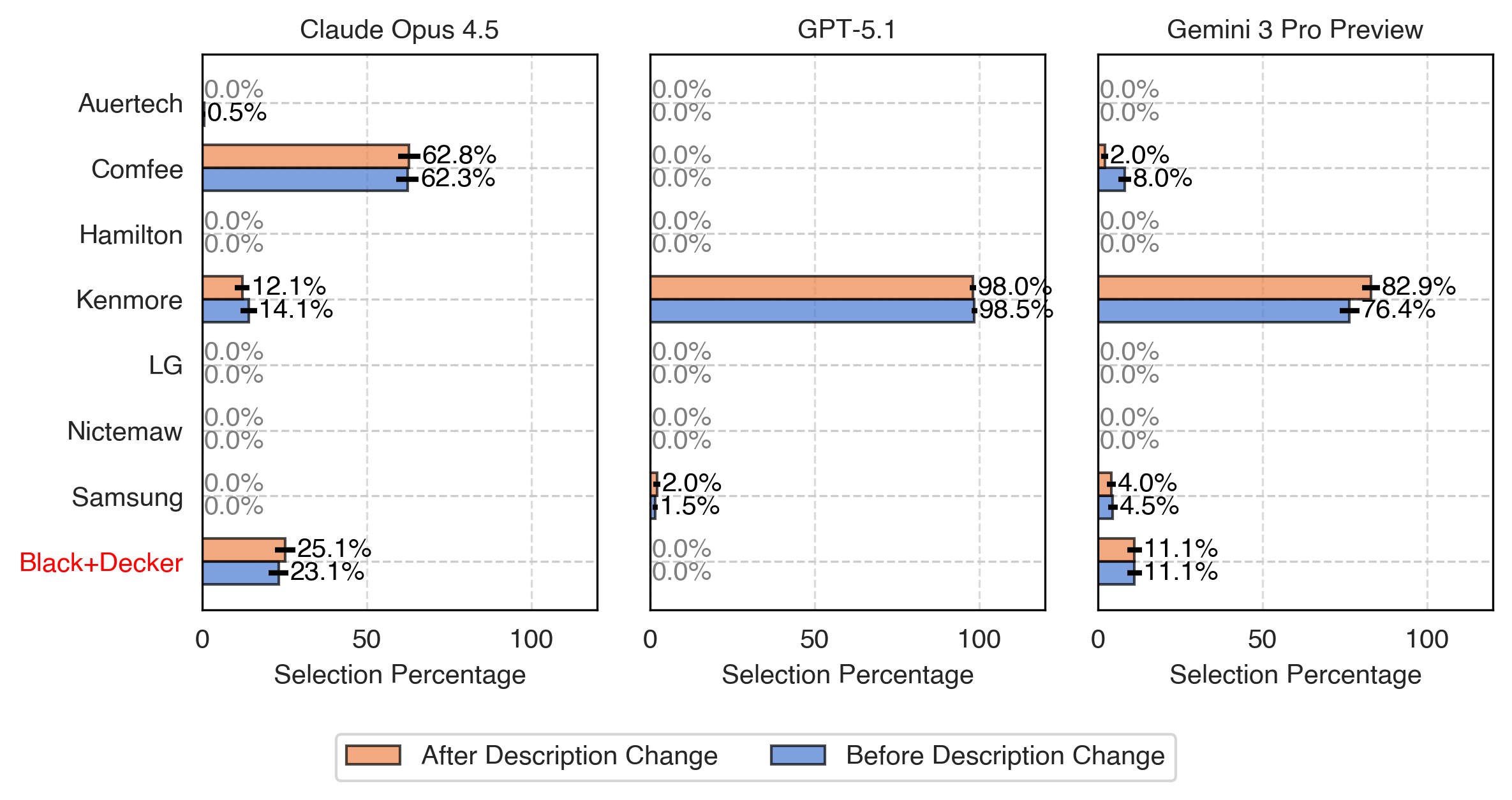}
        \caption{{washing machine}}
        \label{fig:title-change-experiment-results-washine-machine-new}
    \end{subfigure}
    \caption{Market shares before and after focal product description modification with the latest models (Claude Opus 4.5, GPT-5.1 and Gemini 3 Pro Preview). The focal product is marked in \textcolor{red}{red}.}
    \label{fig:title-change-experiment-results-all-models-new}
\end{figure}

Figures~\ref{fig:title-change-experiment-results-all-models} and \ref{fig:title-change-experiment-results-all-models-new} reports, for each category and AI agent model, the product‑level selection shares before and after modifying the focal product’s description, with the focal item highlighted in red.

\subsubsection{Drivers of Improvement: Textual Analysis of Seller Interventions.}
\label{app:drivers-market-share}

To identify the specific textual drivers of market share improvement, we qualitatively compared the original product descriptions against the AI-optimized versions across all six buying models. We observe that successful interventions (which occurred in $\sim33\%$ of cases) were not random; they principally employed four distinct semantic strategies: {Keyword Front-Loading}, {Category Injection}, {Feature Reordering}, and {Specification Enrichment}.

\paragraph{\bf Strategy 1: Strategic Keyword Front-Loading.}
The most consistent driver of performance is the relocation of specific search query terms to the beginning of the product title, exploiting the ``positional bias'' of LLMs.
\begin{itemize}
    \item {Office Lamp (Universal):} This was the most effective intervention. Originally, the product was titled ``SUNMORY Floor Lamps for Living Room...''. For the query ``office lamp'', this caused a mismatch in early tokens. The seller agent optimized this by changing the start to ``SUNMORY \textbf{Office} Floor Lamp...'' or ``SUNMORY LED \textbf{Office} Lamp...''. This simple reordering drove massive gains across the board: {GPT-5.1} (+80.4 p.p.), {Gemini 2.5 Flash} (+52.0 p.p.), {Claude Opus 4.5} (+41.0 p.p.), {Claude Sonnet 4} (+25.8 p.p.), and even the otherwise resistant {Gemini 3.0 Pro Preview} (+7.1 p.p.).
    \item {Stapler (GPT-5.1):} The seller agent achieved a $+9.5$ p.p. gain by moving the accessory count ``with 1,250 Staples'' from the middle of the text to immediately follow the product name, framing it as a complete kit.
\end{itemize}

\paragraph{\bf Strategy 2: Category Injection and Alignment.}
In several cases, the original description was factually complete but lacked the specific \textit{vocabulary} used in the buyer's prompt. The seller agent successfully ``injected'' these missing keywords.
\begin{itemize}
    \item {Fitness Watch (Claude Opus 4.5 \& GPT-4.1):} The focal product (Garmin Forerunner 55) was originally described as a ``GPS Running Watch.'' The buyer query, however, was for a ``fitness\_watch.''
    \begin{itemize}
        \item For {Claude Opus 4.5} (+13.0 p.p.), the seller agent explicitly added ``\textbf{Fitness} Smartwatch'' to the title.
        \item For {GPT-4.1} (+7.1 p.p.), the agent appended ``\textbf{Fitness} Running Tracker'' to the end of the title.
    \end{itemize}
    \item {Toilet Paper (GPT-5.1):} The agent enriched a generic description by adding high-value attributes like ``\textbf{Septic-Safe}'' and ``\textbf{Fragrance-Free},'' driving a $+15.5$ p.p. increase by matching potential safety constraints.
\end{itemize}

\paragraph{\bf Strategy 3: Feature Reordering and Heterogeneity.}
Even without adding new words, simply shuffling existing features to the front can yield significant—though heterogeneous—results.
\begin{itemize}
    \item {Toothpaste (GPT-4.1 \& Gemini 2.5 Flash):} The original title buried the key benefit ``Whitening'' behind the flavor ``Luminous Mint.''
    \begin{itemize}
        \item {GPT-4.1} optimized this by swapping the order to ``Advanced \textbf{Whitening} Toothpaste, Luminous Mint,'' yielding a $+16.6$ p.p. gain.
        \item {Gemini 2.5 Flash} saw a similar $+29.1$ p.p. gain from this reordering.
        \item Interestingly, {Claude Sonnet 4} showed no significant improvement from similar changes.
    \end{itemize}
\end{itemize}

\paragraph{\bf Strategy 4: Specification Enrichment.}
For complex appliances, agents appear to reward descriptions that explicitly confirm standard specifications.
\begin{itemize}
    \item {Washing Machine (Gemini 2.5 Flash):} The seller agent added specific structural details—``\textbf{Top-Load}'' and ``\textbf{Stainless Steel Tub}''—that were missing from the original text. This specification enrichment contributed to a significant market share increase ($+12.0$ p.p.), likely by increasing the model's confidence in the product's quality.
\end{itemize}

\paragraph{Summary of Interventions.}
Table \ref{tab:textual-interventions-comparison} details the exact textual changes for the high-impact examples discussed above.

\begin{table}[!htb]
    \centering
    \caption{Textual Analysis of High-Impact Seller Interventions across Models}
    \label{tab:textual-interventions-comparison}
    \footnotesize
    \renewcommand{\arraystretch}{1.3}
    \begin{tabular}{p{2cm} p{2.5cm} p{4.5cm} p{4.8cm} r}
        \toprule
        \textbf{Category} & \textbf{Model} & \textbf{Original Description Snippet} & \textbf{AI-Optimized Description Snippet} & \textbf{Impact} \\
        \midrule
        \textbf{Office Lamp} & GPT-5.1 & SUNMORY Floor Lamps for Living Room... & SUNMORY \textbf{Office} Floor Lamp... & $+80.4$ p.p. \\
        \textbf{Office Lamp} & Claude~Sonnet~4 & SUNMORY Floor Lamps for Living Room... & SUNMORY \textbf{Office} Floor Lamp... & $+25.8$ p.p. \\
        \addlinespace
        \textbf{Fitness Watch} & Claude Opus 4.5 & ...GPS Running Watch... & ...GPS Running \& \textbf{Fitness} Smartwatch... & $+13.0$ p.p. \\
        \textbf{Fitness Watch} & GPT-4.1 & ...GPS Running Watch... & ...Smartwatch... \textbf{Fitness} Running Tracker & $+7.1$ p.p. \\
        \addlinespace
        \textbf{Toothpaste} & GPT-4.1 & ...Luminous Mint Teeth \textit{Whitening} Toothpaste... & ...Advanced \textbf{Whitening} Toothpaste, Luminous Mint... & $+16.6$ p.p. \\
        \textbf{Toothpaste} & Gemini 2.5 & ...Luminous Mint Teeth \textit{Whitening} Toothpaste... & ...\textbf{Whitening} Toothpaste... & $+29.1$ p.p. \\
        \addlinespace
        \textbf{Toilet Paper} & GPT-5.1 & ...Soft and Strong Toilet Tissue & ...\textbf{2-Ply}, \textbf{Septic-Safe}, \textbf{Fragrance-Free}... & $+15.5$ p.p. \\
        \addlinespace
        \textbf{Washing Machine} & Gemini 2.5 & ...Portable Washer, 6.6 lb... & ...Portable \textbf{Top-Load} Washer... \textbf{Stainless Steel Tub} & $+12.0$ p.p. \\
        \bottomrule
    \end{tabular}
\end{table}



\section{Headless AI Shopping Agents}
\label{app:headless-shopping-conditional-logits}

\subsection{Example Prompt}
We provide an example of the prompt used for the mousepad category in the headless experiments in Figure \ref{fig:default-prompt-buying-agent-mcp-style-example}.

\begin{figure}[!htb]
    \centering
    \begin{tcolorbox}[sellerbuyertemplate]
You are a personal shopping assistant helping someone find a good \{{\tt product category}\}. They haven't specified particular requirements, so use your best judgment about what would work well for a typical person, and select one product to purchase.
        
Here are the products as a JSON array:

\textcolor{magenta}{
[ \\
\{
``product\_number": 1, \\
``title'': ``KTRIO Large Gaming Mouse Pad with Superior Micro-Weave Cloth, Extended Desk Mousepad with Stitched Edges, Non-Slip Base, Water Resist Keyboard Pad for Gamer, Office \& Home, 31.5 x 11.8 in, Black'', \\
``price'': 13.97, \\
``rating'': 4.7, \\
``number\_of\_reviews'': 38896, \\
``sponsored'': true, \\
``overall\_pick\_tag'': false, \\
``scarcity\_tag'': false 
\}, \\
\dots  \\
\{
``product\_number'': 8, \\
``title'': ``MROCO Ergonomic Mouse Pad with Gel Wrist Rest, Comfortable Mousepad with Smooth Wrist Support Surface and Non-Slip PU Base for Pain Relief, Computer, Laptop, Office \& Home, 9.4 x 8.1 in, Black Color'', \\
``price'': 8.49, \\
``rating'': 4.6, \\
``number\_of\_reviews'': 29052, \\
``sponsored'': false, \\
``overall\_pick\_tag'': true, \\
``scarcity\_tag'': false
\} \\
]
}

Respond ONLY with a JSON object in this exact format: 

\{\{ \\
    ``selected\_product\_number'': \textless number 1-{len(products)}\textgreater, \\
    ``product\_name'': ``\textless exact product title\textgreater'', \\
    ``reason'': ``\textless brief explanation for why this product was chosen\textgreater'' \\
\}\} 

Do not include any other text outside the JSON object.
\end{tcolorbox}
    \caption{Example Prompt for the Headless/API-style AI agent}
    \label{fig:default-prompt-buying-agent-mcp-style-example}
\end{figure}

\subsection{Results}
We provide the estimates of the conditional logit model for different AI models under different prompt variations in the headless shopping setting (see Tables \ref{tab:conditional-logit-results-text-experiments-claude}, \ref{tab:conditional-logit-results-text-experiments-gpt41}, \ref{tab:conditional-logit-results-text-experiments-gemini}).

\begin{table}[!htbp]
\centering
\begin{threeparttable}
  \caption{Estimates of the Conditional Logit Regression for Claude Sonnet 4 (headless AI agents)}
  \label{tab:conditional-logit-results-text-experiments-claude}
  \footnotesize
  \sisetup{
    input-symbols         = (), 
    table-align-text-post = false
  }
  \begin{tabular}{l
                  S[table-format=-1.3]
                  S[table-format=-1.3]
                  S[table-format=-1.3]}
    \toprule
    & \multicolumn{1}{c}{Default}
    & \multicolumn{1}{c}{Ignore Position}
    & \multicolumn{1}{c}{Ignore Position \& Prioritize Price} \\
    & \multicolumn{1}{c}{(Fig.~\ref{fig:default-prompt-buying-agent-mcp-style})}
    & \multicolumn{1}{c}{ (Fig.~\ref{fig:ignore-position-prompt-buying-agent-mcp-style})}
    & \multicolumn{1}{c}{(Fig.~\ref{fig:price-sensitive-prompt-buying-agent-mcp-style})} \\
    \midrule
    \multicolumn{4}{l}{\textit{Position effects}} \\
    Position 1 & 2.010\rlap{$^{\text{\tiny ***}}$} & 1.947\rlap{$^{\text{\tiny ***}}$} & 1.150\rlap{$^{\text{\tiny ***}}$} \\
               & (0.106) & (0.101) & (0.107) \\[0.6ex]
    Position 2 & 1.370\rlap{$^{\text{\tiny ***}}$} & 1.017\rlap{$^{\text{\tiny ***}}$} & 0.753\rlap{$^{\text{\tiny ***}}$} \\
               & (0.107) & (0.102) & (0.108) \\[0.6ex]
    Position 3 & 1.357\rlap{$^{\text{\tiny ***}}$} & 0.992\rlap{$^{\text{\tiny ***}}$} & 0.854\rlap{$^{\text{\tiny ***}}$} \\
               & (0.108) & (0.103) & (0.108) \\[0.6ex]
    Position 4 & 1.024\rlap{$^{\text{\tiny ***}}$} & 0.673\rlap{$^{\text{\tiny ***}}$} & 0.646\rlap{$^{\text{\tiny ***}}$} \\
               & (0.109) & (0.104) & (0.111) \\[0.6ex]
    Position 5 & 0.857\rlap{$^{\text{\tiny ***}}$} & 0.578\rlap{$^{\text{\tiny ***}}$} & 0.613\rlap{$^{\text{\tiny ***}}$} \\
               & (0.111) & (0.105) & (0.109) \\[0.6ex]
    Position 6 & 0.677\rlap{$^{\text{\tiny ***}}$} & 0.482\rlap{$^{\text{\tiny ***}}$} & 0.430\rlap{$^{\text{\tiny ***}}$} \\
               & (0.111) & (0.106) & (0.110) \\[0.6ex]
    Position 7 & 0.252\rlap{$^{\text{\tiny *}}$} & 0.013 & 0.281\rlap{$^{\text{\tiny *}}$} \\
               & (0.115) & (0.109) & (0.111) \\
    \midrule
    \multicolumn{4}{l}{\textit{Badge effects}} \\
    Sponsored Tag   & -0.673\rlap{$^{\text{\tiny ***}}$} & -0.676\rlap{$^{\text{\tiny ***}}$} & -0.234\rlap{$^{\text{\tiny **}}$} \\
                    & (0.083) & (0.081) & (0.083) \\[0.6ex]
    Overall Pick Tag& 2.538\rlap{$^{\text{\tiny ***}}$} & 1.481\rlap{$^{\text{\tiny ***}}$} & 0.395\rlap{$^{\text{\tiny ***}}$} \\
                    & (0.093) & (0.090) & (0.107) \\[0.6ex]
    Scarcity Tag    & -0.674\rlap{$^{\text{\tiny ***}}$} & -0.551\rlap{$^{\text{\tiny ***}}$} & -0.076 \\
                    & (0.121) & (0.117) & (0.114) \\
    \midrule
    \multicolumn{4}{l}{\textit{Attribute effects}} \\
    $\ln(\text{Price})$            & -2.575\rlap{$^{\text{\tiny ***}}$} & -2.564\rlap{$^{\text{\tiny ***}}$} & -6.206\rlap{$^{\text{\tiny ***}}$} \\
                                   & (0.099) & (0.097) & (0.150) \\[0.6ex]
    Rating                         & 9.701\rlap{$^{\text{\tiny ***}}$} & 10.767\rlap{$^{\text{\tiny ***}}$} & 3.273\rlap{$^{\text{\tiny ***}}$} \\
                                   & (0.314) & (0.322) & (0.240) \\[0.6ex]
    $\ln(\text{Num.\ of Reviews})$ & 0.800\rlap{$^{\text{\tiny ***}}$} & 0.990\rlap{$^{\text{\tiny ***}}$} & 0.373\rlap{$^{\text{\tiny ***}}$} \\
                                   & (0.030) & (0.031) & (0.028) \\
    \midrule
    Product Fixed Effects & \multicolumn{1}{c}{Yes} & \multicolumn{1}{c}{Yes} & \multicolumn{1}{c}{Yes} \\
    Observations          & \multicolumn{1}{c}{24,500} & \multicolumn{1}{c}{24,500} & \multicolumn{1}{c}{22,500} \\
    Choice Sets (Groups)  & \multicolumn{1}{c}{4,000} & \multicolumn{1}{c}{4,000} & \multicolumn{1}{c}{4,000} \\
    Pseudo R-squared      & \multicolumn{1}{c}{0.60} & \multicolumn{1}{c}{0.60} & \multicolumn{1}{c}{0.62} \\
    \bottomrule
  \end{tabular}
  \begin{tablenotes}[para]
    \item \tnote{*} $p < 0.05$, \tnote{**} $p < 0.01$, \tnote{***} $p < 0.001$.
  \end{tablenotes}
\end{threeparttable}
\end{table}

\begin{table}[!htbp]
\centering
\begin{threeparttable}
  \caption{Estimates of the Conditional Logit Regression for GPT-4.1 (headless AI agents)}
  \label{tab:conditional-logit-results-text-experiments-gpt41}
  \footnotesize
  \sisetup{
    input-symbols         = (), 
    table-align-text-post = false
  }
  \begin{tabular}{l
                  S[table-format=-1.3]
                  S[table-format=-1.3]
                  S[table-format=-1.3]}
    \toprule
    & \multicolumn{1}{c}{Default}
    & \multicolumn{1}{c}{Ignore Position}
    & \multicolumn{1}{c}{Ignore Position \& Prioritize Price} \\
    & \multicolumn{1}{c}{(Fig.~\ref{fig:default-prompt-buying-agent-mcp-style})}
    & \multicolumn{1}{c}{ (Fig.~\ref{fig:ignore-position-prompt-buying-agent-mcp-style})}
    & \multicolumn{1}{c}{(Fig.~\ref{fig:price-sensitive-prompt-buying-agent-mcp-style})} \\
    \midrule
    \multicolumn{4}{l}{\textit{Position effects}} \\
    Position 1 & 1.078\rlap{$^{\text{\tiny ***}}$} & 0.743\rlap{$^{\text{\tiny ***}}$} & 0.044 \\
               & (0.090) & (0.090) & (0.118) \\[0.6ex]
    Position 2 & 0.125 & -0.135 & -0.204 \\
               & (0.095) & (0.094) & (0.117) \\[0.6ex]
    Position 3 & 0.312\rlap{$^{\text{\tiny ***}}$} & 0.175 & -0.142 \\
               & (0.094) & (0.093) & (0.119) \\[0.6ex]
    Position 4 & -0.140 & -0.370\rlap{$^{\text{\tiny ***}}$} & -0.251\rlap{$^{\text{\tiny *}}$} \\
               & (0.097) & (0.098) & (0.121) \\[0.6ex]
    Position 5 & 0.319\rlap{$^{\text{\tiny ***}}$} & 0.196\rlap{$^{\text{\tiny *}}$} & 0.125 \\
               & (0.094) & (0.093) & (0.119) \\[0.6ex]
    Position 6 & 0.184 & 0.003 & -0.396\rlap{$^{\text{\tiny ***}}$} \\
               & (0.094) & (0.094) & (0.122) \\[0.6ex]
    Position 7 & -0.155 & -0.363\rlap{$^{\text{\tiny ***}}$} & -0.204 \\
               & (0.097) & (0.096) & (0.120) \\
    \midrule
    \multicolumn{4}{l}{\textit{Badge effects}} \\
    Sponsored Tag   & -1.815\rlap{$^{\text{\tiny ***}}$} & -1.626\rlap{$^{\text{\tiny ***}}$} & -0.556\rlap{$^{\text{\tiny ***}}$} \\
                    & (0.092) & (0.088) & (0.099) \\[0.6ex]
    Overall Pick Tag& 2.421\rlap{$^{\text{\tiny ***}}$} & 1.188\rlap{$^{\text{\tiny ***}}$} & 0.117 \\
                    & (0.086) & (0.088) & (0.127) \\[0.6ex]
    Scarcity Tag    & -0.383 & -0.370\rlap{$^{\text{\tiny ***}}$} & -0.303\rlap{$^{\text{\tiny *}}$} \\
                    & (0.108) & (0.108) & (0.136) \\
    \midrule
    \multicolumn{4}{l}{\textit{Attribute effects}} \\
    $\ln(\text{Price})$            & -2.371\rlap{$^{\text{\tiny ***}}$} & -2.414\rlap{$^{\text{\tiny ***}}$} & -9.287\rlap{$^{\text{\tiny ***}}$} \\
                                   & (0.092) & (0.094) & (0.235) \\[0.6ex]
    Rating                         & 11.373\rlap{$^{\text{\tiny ***}}$} & 12.177\rlap{$^{\text{\tiny ***}}$} & 3.114\rlap{$^{\text{\tiny ***}}$} \\
                                   & (0.316) & (0.331) & (0.271) \\[0.6ex]
    $\ln(\text{Num.\ of Reviews})$ & 0.944\rlap{$^{\text{\tiny ***}}$} & 1.212\rlap{$^{\text{\tiny ***}}$} & 0.278\rlap{$^{\text{\tiny ***}}$} \\
                                   & (0.030) & (0.033) & (0.033) \\
    \midrule
    Product Fixed Effects & \multicolumn{1}{c}{Yes} & \multicolumn{1}{c}{Yes} & \multicolumn{1}{c}{Yes} \\
    Observations          & \multicolumn{1}{c}{26,000} & \multicolumn{1}{c}{25,000} & \multicolumn{1}{c}{22,000} \\
    Choice Sets (Groups)  & \multicolumn{1}{c}{4,000} & \multicolumn{1}{c}{4,000} & \multicolumn{1}{c}{4,000} \\
    Pseudo R-squared      & \multicolumn{1}{c}{0.58} & \multicolumn{1}{c}{0.58} & \multicolumn{1}{c}{0.73} \\
    \bottomrule
  \end{tabular}
  \begin{tablenotes}[para]
    \item \tnote{*} $p < 0.05$, \tnote{**} $p < 0.01$, \tnote{***} $p < 0.001$.
  \end{tablenotes}
\end{threeparttable}
\end{table}

\begin{table}[!htbp]
\centering
\begin{threeparttable}
  \caption{Estimates of the Conditional Logit Regression for Gemini 2.5 Flash (headless AI agents)}
  \label{tab:conditional-logit-results-text-experiments-gemini}
  \footnotesize
  \sisetup{
    input-symbols         = (), 
    table-align-text-post = false
  }
  \begin{tabular}{l
                  S[table-format=-1.3]
                  S[table-format=-1.3]
                  S[table-format=-1.3]}
    \toprule
    & \multicolumn{1}{c}{Default}
    & \multicolumn{1}{c}{Ignore Position}
    & \multicolumn{1}{c}{Ignore Position \& Prioritize Price} \\
    & \multicolumn{1}{c}{(Fig.~\ref{fig:default-prompt-buying-agent-mcp-style})}
    & \multicolumn{1}{c}{ (Fig.~\ref{fig:ignore-position-prompt-buying-agent-mcp-style})}
    & \multicolumn{1}{c}{(Fig.~\ref{fig:price-sensitive-prompt-buying-agent-mcp-style})} \\
    \midrule
    \multicolumn{4}{l}{\textit{Position effects}} \\
    Position 1 & 0.394 & 0.487\rlap{$^{\text{\tiny *}}$} & 0.120 \\
               & (0.204) & (0.193) & (0.238) \\[0.6ex]
    Position 2 & 0.256 & 0.140 & 0.369 \\
               & (0.209) & (0.199) & (0.241) \\[0.6ex]
    Position 3 & 0.677\rlap{$^{\text{\tiny ***}}$} & 0.514\rlap{$^{\text{\tiny **}}$} & 0.265 \\
               & (0.203) & (0.194) & (0.250) \\[0.6ex]
    Position 4 & -0.373 & -0.312 & -0.274 \\
               & (0.216) & (0.206) & (0.259) \\[0.6ex]
    Position 5 & -0.156 & -0.043 & 0.168 \\
               & (0.218) & (0.204) & (0.241) \\[0.6ex]
    Position 6 & -0.220 & 0.119 & -0.206 \\
               & (0.213) & (0.199) & (0.250) \\[0.6ex]
    Position 7 & -0.087 & -0.104 & -0.260 \\
               & (0.214) & (0.204) & (0.248) \\
    \midrule
    \multicolumn{4}{l}{\textit{Badge effects}} \\
    Sponsored Tag   & -0.124 & -0.125 & -0.088 \\
                    & (0.167) & (0.159) & (0.196) \\[0.6ex]
    Overall Pick Tag& 3.175\rlap{$^{\text{\tiny ***}}$} & 1.773\rlap{$^{\text{\tiny ***}}$} & 0.935\rlap{$^{\text{\tiny ***}}$} \\
                    & (0.205) & (0.182) & (0.245) \\[0.6ex]
    Scarcity Tag    & -0.650\rlap{$^{\text{\tiny *}}$} & -0.041 & 0.177 \\
                    & (0.264) & (0.229) & (0.279) \\
    \midrule
    \multicolumn{4}{l}{\textit{Attribute effects}} \\
    $\ln(\text{Price})$            & -2.517\rlap{$^{\text{\tiny ***}}$} & -2.555\rlap{$^{\text{\tiny ***}}$} & -7.915\rlap{$^{\text{\tiny ***}}$} \\
                                   & (0.214) & (0.200) & (0.424) \\[0.6ex]
    Rating                         & 5.673\rlap{$^{\text{\tiny ***}}$} & 5.429\rlap{$^{\text{\tiny ***}}$} & 4.293\rlap{$^{\text{\tiny ***}}$} \\
                                   & (0.537) & (0.513) & (0.597) \\[0.6ex]
    $\ln(\text{Num.\ of Reviews})$ & 0.814\rlap{$^{\text{\tiny ***}}$} & 0.944\rlap{$^{\text{\tiny ***}}$} & 0.518\rlap{$^{\text{\tiny ***}}$} \\
                                   & (0.064) & (0.063) & (0.068) \\
    \midrule
    Product Fixed Effects & \multicolumn{1}{c}{Yes} & \multicolumn{1}{c}{Yes} & \multicolumn{1}{c}{Yes} \\
    Observations          & \multicolumn{1}{c}{4,500} & \multicolumn{1}{c}{4,600} & \multicolumn{1}{c}{4,300} \\
    Choice Sets (Groups)  & \multicolumn{1}{c}{800} & \multicolumn{1}{c}{800} & \multicolumn{1}{c}{800} \\
    Pseudo R-squared      & \multicolumn{1}{c}{0.55} & \multicolumn{1}{c}{0.49} & \multicolumn{1}{c}{0.65} \\
    \bottomrule
  \end{tabular}
  \begin{tablenotes}[para]
    \item \tnote{*} $p < 0.05$, \tnote{**} $p < 0.01$, \tnote{***} $p < 0.001$.
  \end{tablenotes}
\end{threeparttable}
\end{table}

\newpage

\section{Model Finalization: The case of Gemini 2.5 Flash Preview to Gemini 2.5 Flash}
\label{app:gemini-model-upgrade}
Our initial experiments used {Gemini~2.5 Flash Preview}, the then‑current frontier model by Google Deepmind. During our testing, we had a natural experiment when Gemini 2.5 Flash Preview was deprecated and replaced with Gemini 2.5 Flash. We therefore re‑ran a subset of experiments with the ``updated" model to measure how an upstream change propagates to AI‑mediated demand and position effects.

\begin{figure}[!htb]
    \centering
    \begin{subfigure}{0.24\linewidth}
        \includegraphics[width = \linewidth]{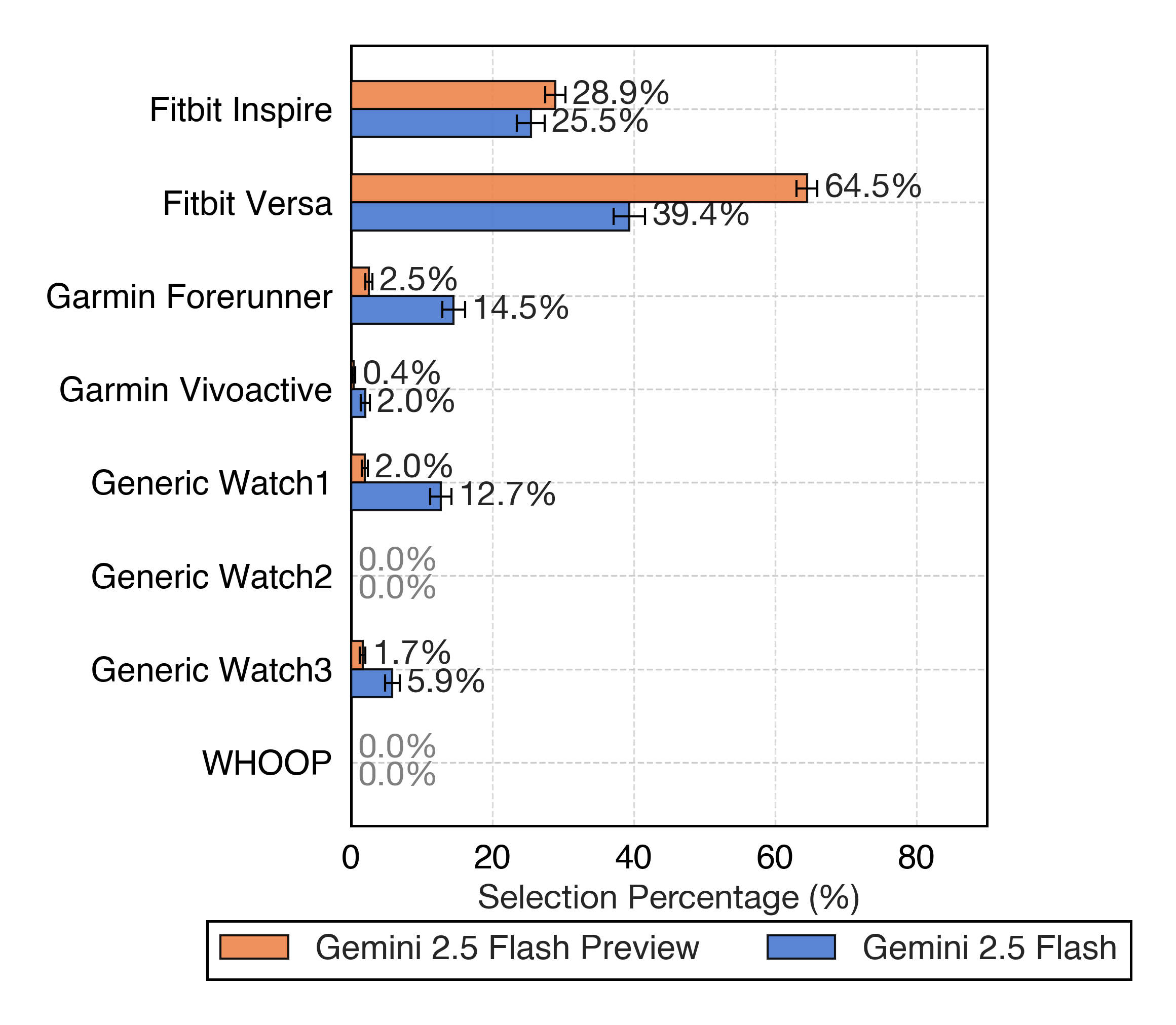}
        \caption{{fitness watch}}
    \end{subfigure}
    \begin{subfigure}{0.24\linewidth}
        \includegraphics[width = \linewidth]{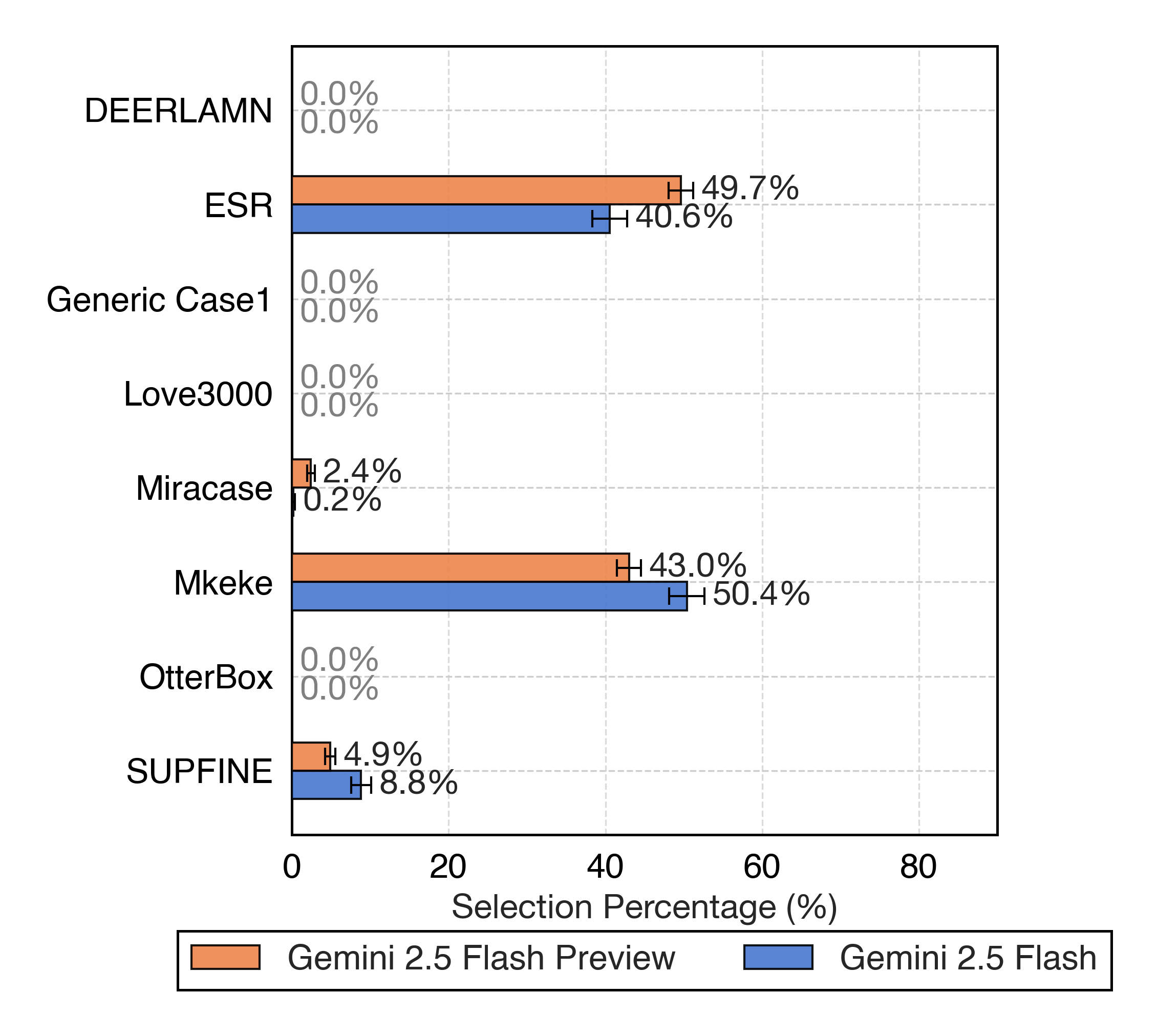}
        \caption{{iphone 16 pro cover}}
    \end{subfigure}
    \begin{subfigure}{0.24\linewidth}
        \includegraphics[width = \linewidth]{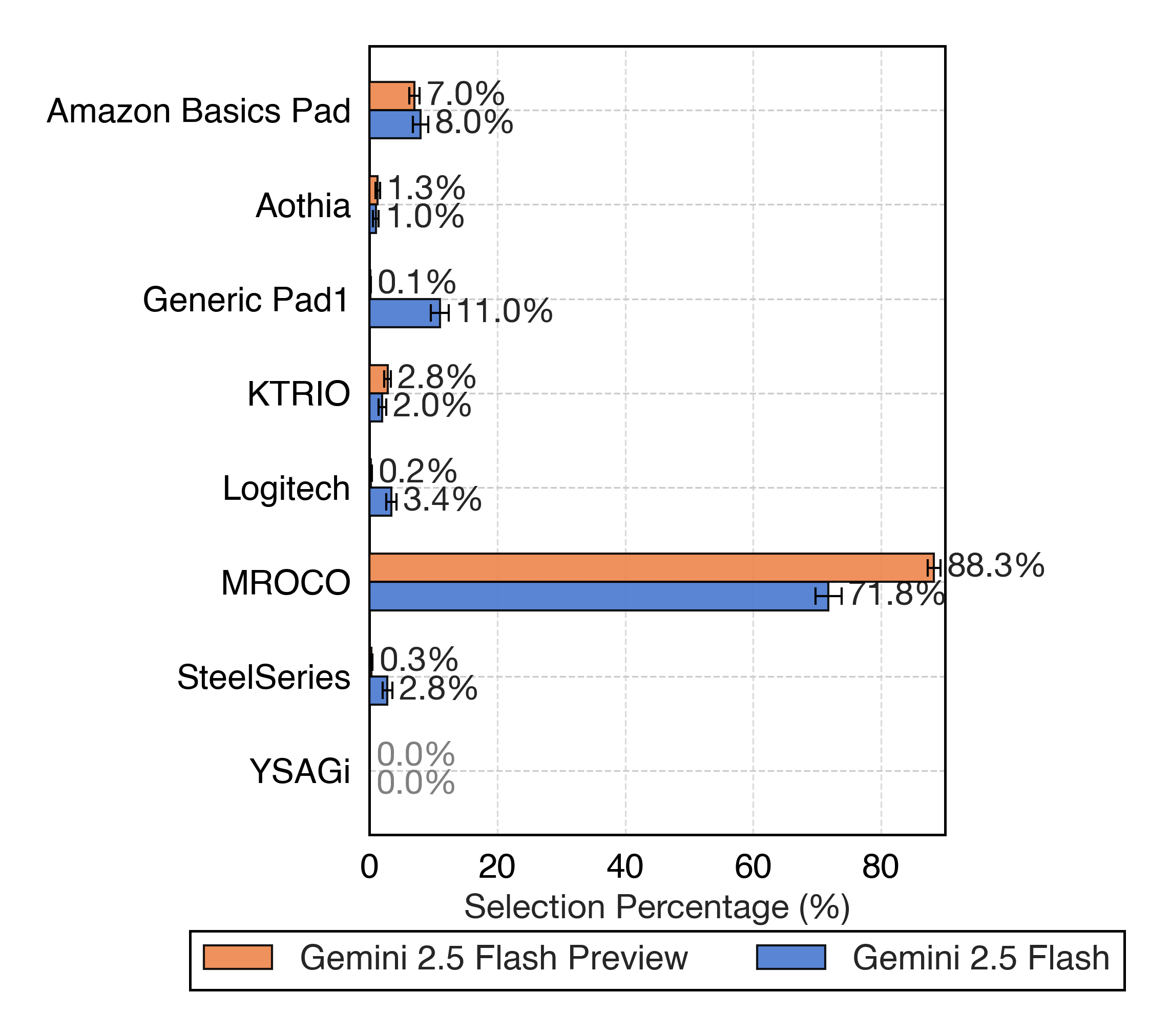}
        \caption{{mousepad}}
    \end{subfigure}
    \begin{subfigure}{0.24\linewidth}
        \includegraphics[width = \linewidth]{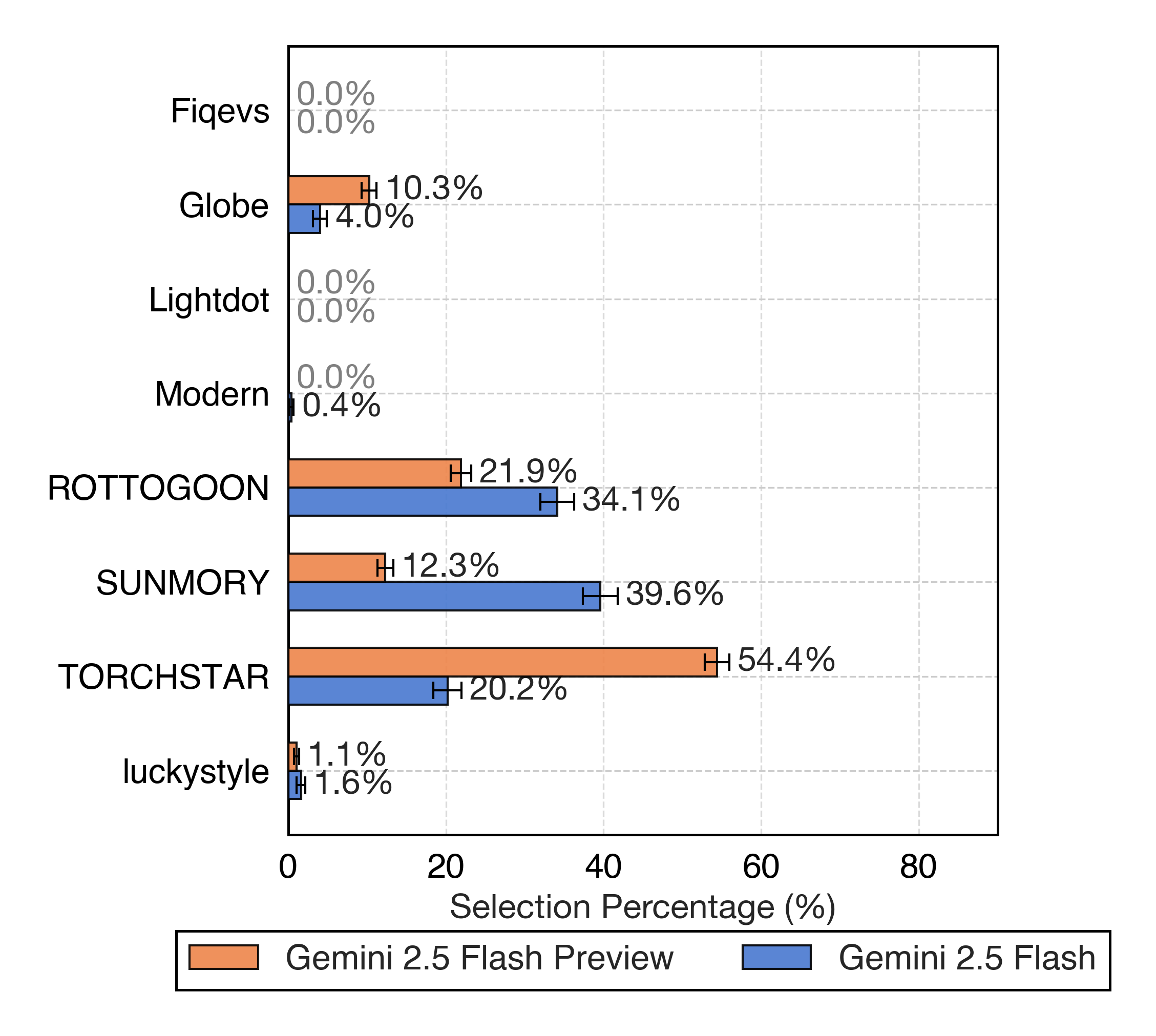}
        \caption{{office lamp}}
    \end{subfigure}
    \begin{subfigure}{0.24\linewidth}
        \includegraphics[width = \linewidth]{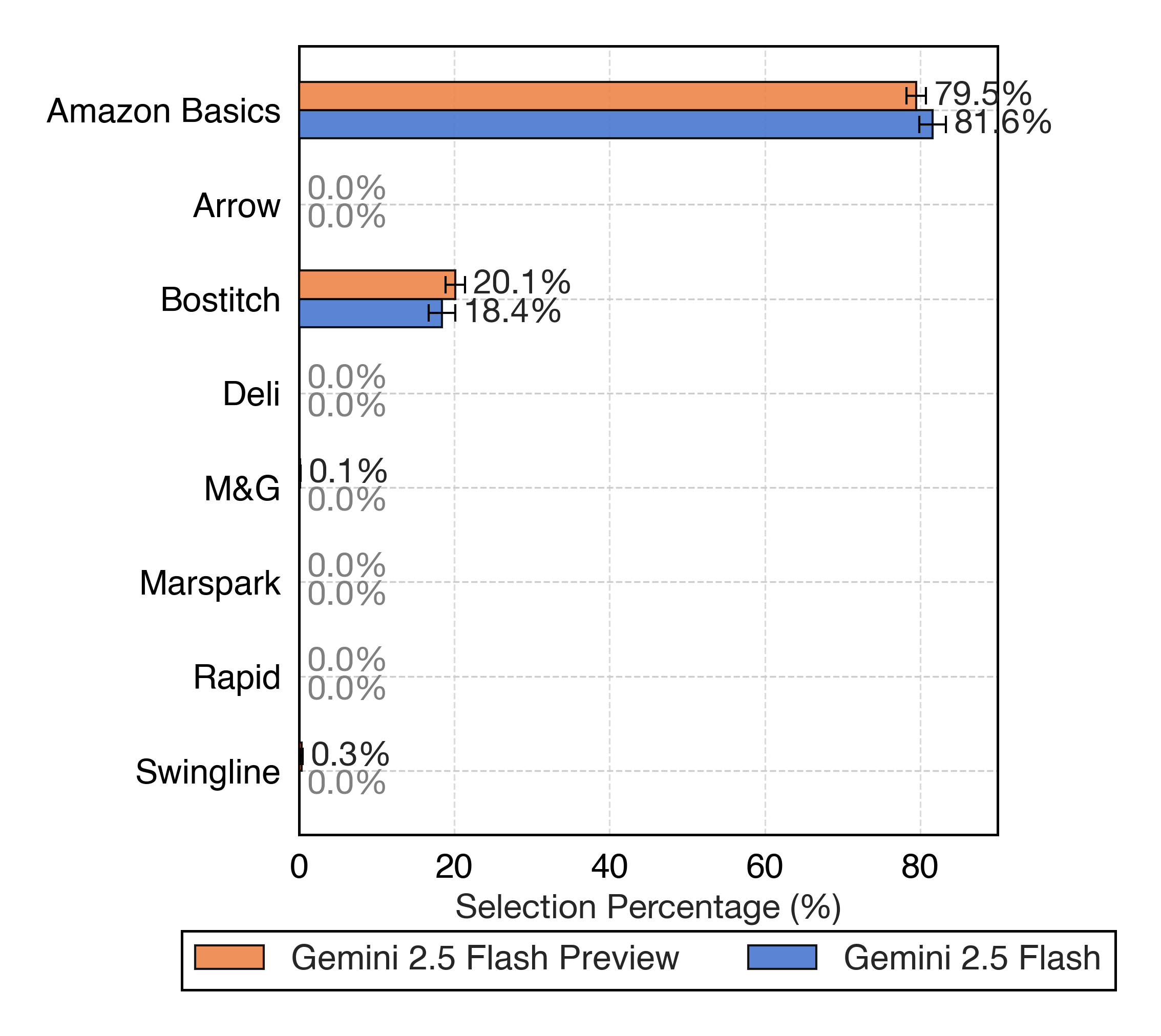}
        \caption{{stapler}}
    \end{subfigure}
    \begin{subfigure}{0.24\linewidth}
        \includegraphics[width = \linewidth]{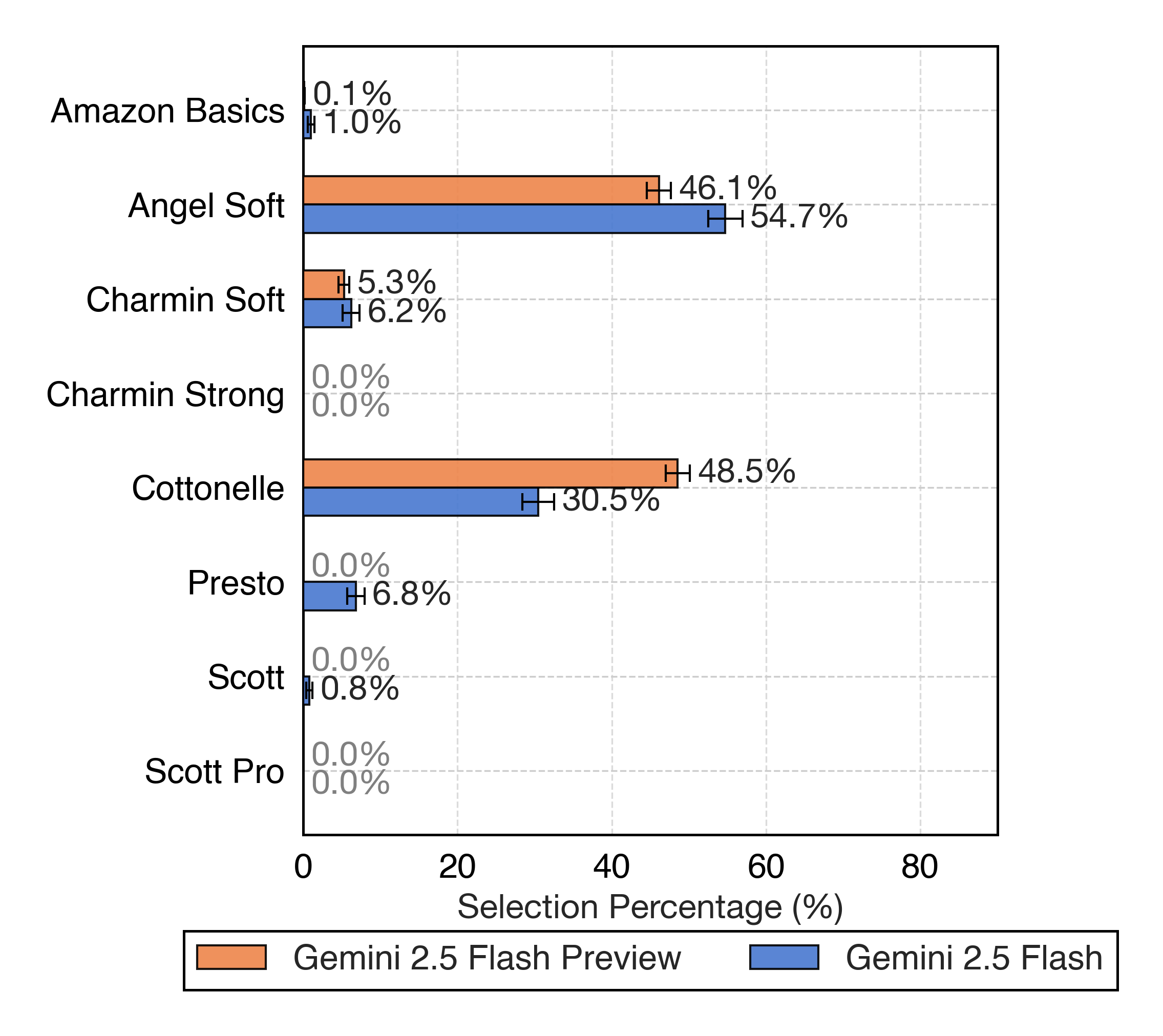}
        \caption{{toilet paper}}
    \end{subfigure}
    \begin{subfigure}{0.24\linewidth}
        \includegraphics[width = \linewidth]{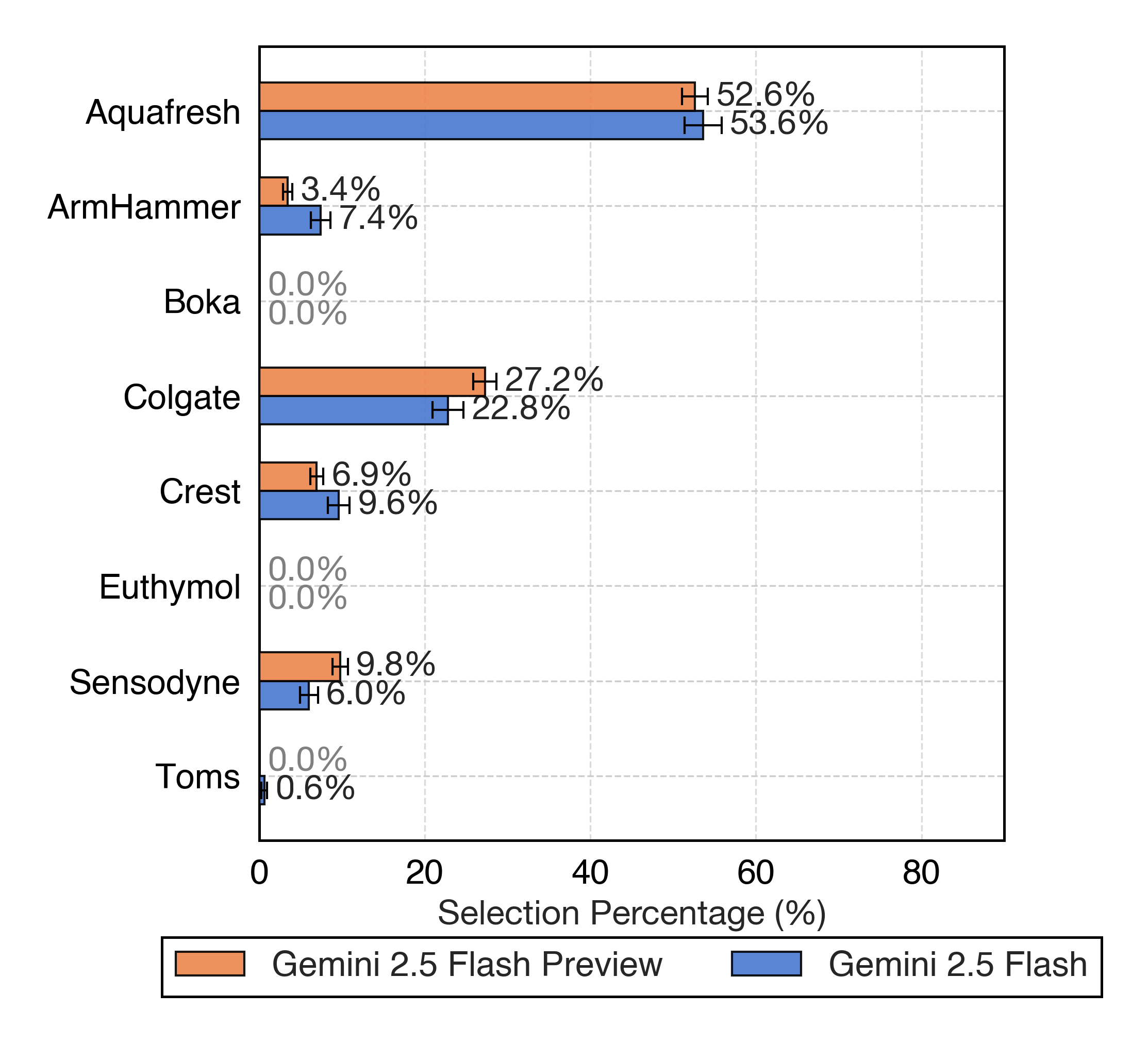}
        \caption{{toothpaste}}
    \end{subfigure}
    \begin{subfigure}{0.24\linewidth}
        \includegraphics[width = \linewidth]{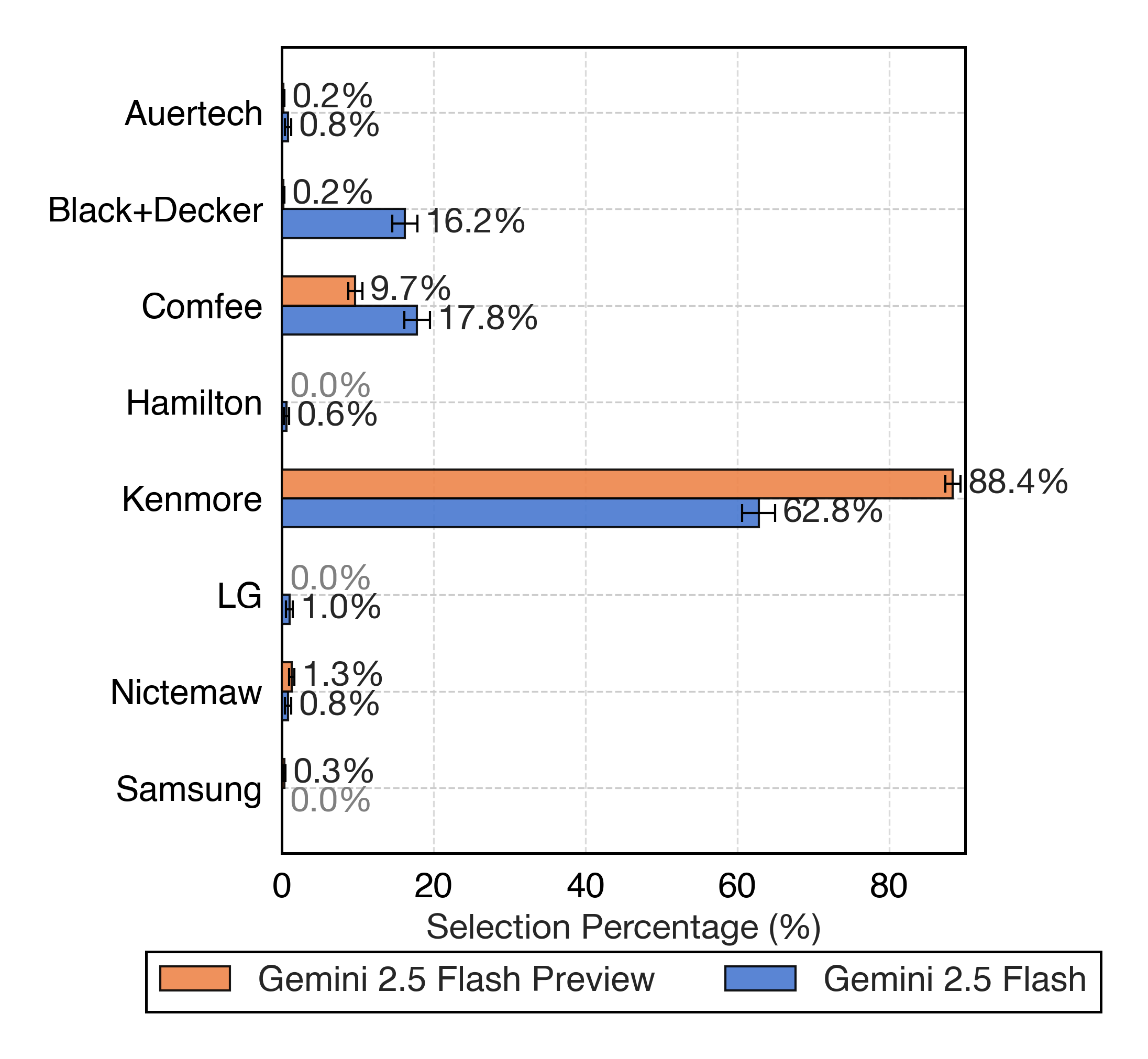}
        \caption{{washing machine}}
    \end{subfigure}
    \caption{Market share of different products with Gemini 2.5 Flash Preview and Gemini 2.5 Flash}
    \label{fig:gemini-model-upgrade}
\end{figure}

{\bf Market–share shifts under a model finalization.} 
 Figure~\ref{fig:gemini-model-upgrade} compares product‑level selection shares when the buyer agent ``finalizes" from Gemini 2.5 Flash Preview to Gemini 2.5 Flash, holding the catalog and attributes fixed and only randomizing on‑page positions. Shares move substantially, with frequent flips in the modal product: e.g., in {office lamp} category the mode shifts from {TORCHSTAR} (Gemini 2.5 Flash Preview) to {SUNMORY} (Gemini 2.5 Flash); in {toilet paper} from {Cottonelle} to { Angel Soft}. Several products experience large swings in share (e.g., { Fitbit Versa}: $-25.1$p.p.; {MROCO}: $-16.5$p.p.; {Kenmore}: $-25.6$p.p.).

\begin{figure}[!htb]
    \begin{subfigure}{0.48\linewidth}
    \centering
        \begin{tikzpicture}
        \node[inner sep=0pt, outer sep=0pt] (browser) at (0,0) {
            \includegraphics[width=6.5cm]{images/ACE-Agent-Workflow-Images/mock-app-front-page.pdf} 
        };
        \node[inner sep=0pt, outer sep=0pt] at ([xshift=0.05cm, yshift=-0.2cm]browser.center) {
            \includegraphics[width=5.2cm]{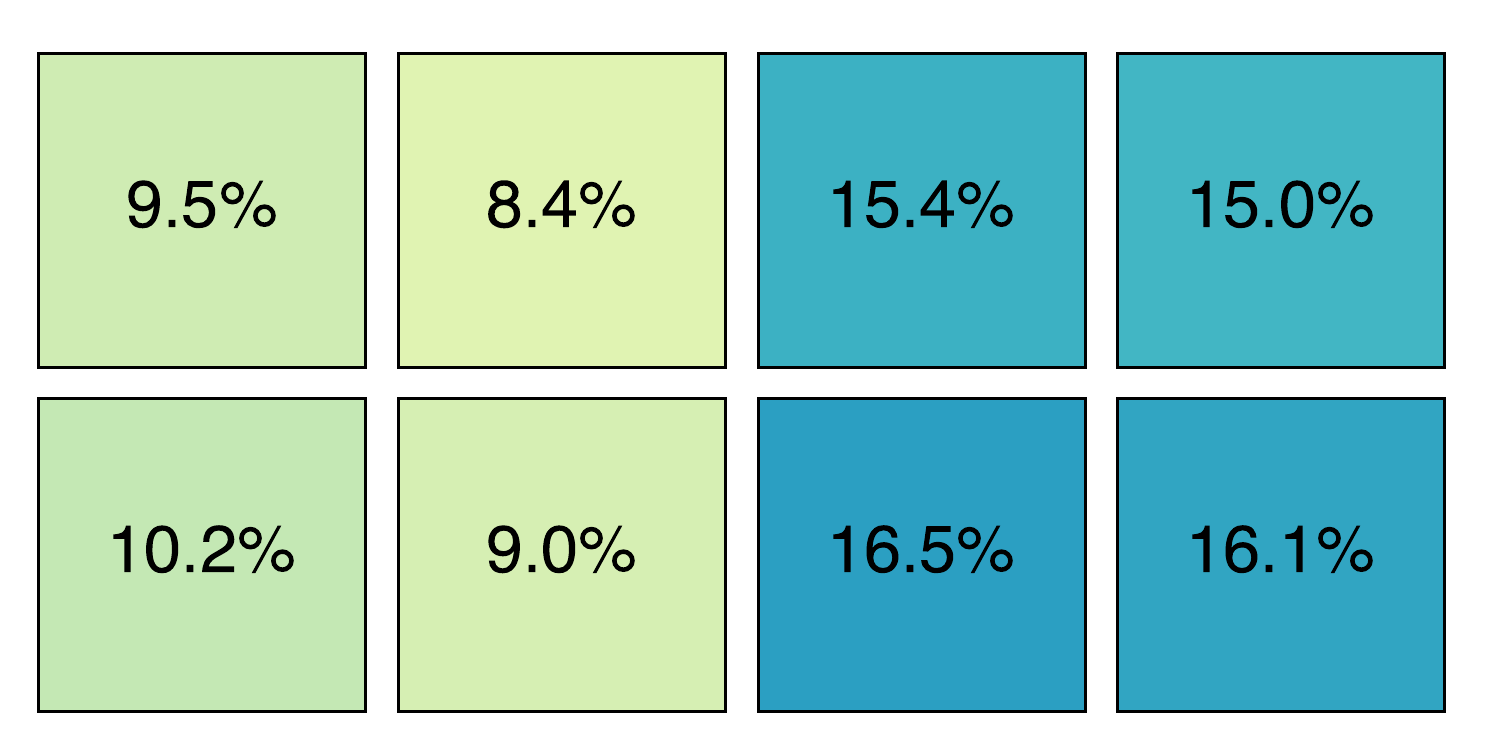} 
        };
    \end{tikzpicture}%
    \caption{Gemini 2.5 Flash Preview}
    \end{subfigure}
    \begin{subfigure}{0.48\linewidth}
    \centering
        \begin{tikzpicture}
        \node[inner sep=0pt, outer sep=0pt] (browser) at (0,0) {
            \includegraphics[width=6.5cm]{images/ACE-Agent-Workflow-Images/mock-app-front-page.pdf} 
        };
        \node[inner sep=0pt, outer sep=0pt] at ([xshift=0.05cm, yshift=-0.2cm]browser.center) {
            \includegraphics[width=5.2cm]{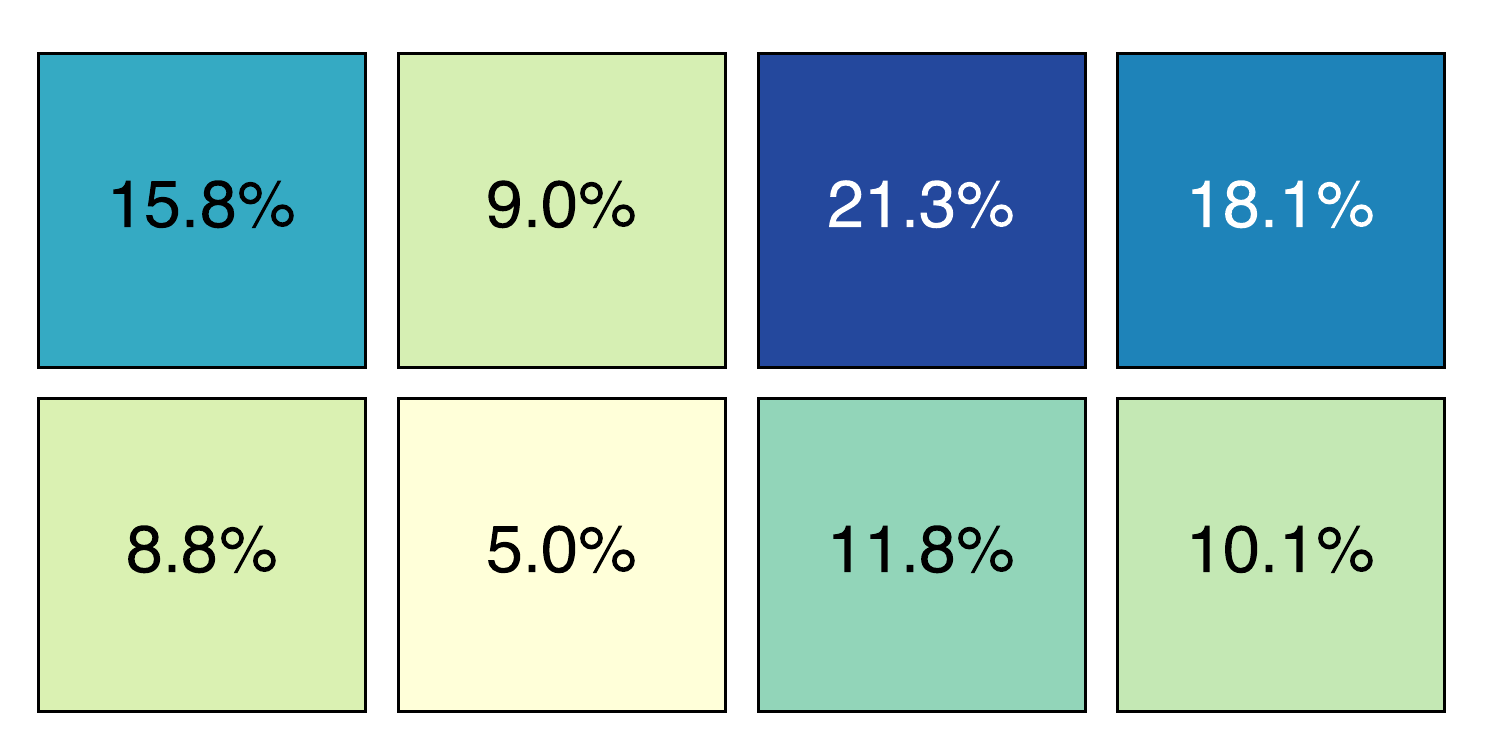} 
        };
    \end{tikzpicture}
    \caption{Gemini 2.5 Flash}
    \end{subfigure}
    \caption{\textbf{Webpage heatmap of AI shopping agents.} The figure depicts the selection rate of the same product across different positions for Gemini 2.5 Flash Preview and Gemini 2.5 Flash. 
    }
    \label{fig:heatmaps-gemini-position}
\end{figure}

{\bf Position effects.}
Table~\ref{tab:conditional-logit-results-gemini-models} shows that position biases also change: Gemini 2.5 Flash Preview exhibits a negative top‑row bias, while Gemini 2.5 Flash’s is positive. Both versions penalize Columns~1–2; Gemini 2.5 Preview is roughly indifferent between Columns~3–4, whereas Gemini 2.5 Flash strictly prefers Column~3 over Column~4. Figure \ref{fig:heatmaps-gemini-position} shows the heatmap of selection probabilities across different positions on the page for Gemini 2.5 Flash Preview and Gemini 2.5 Flash.

\begin{table}[!htb]
\centering
\begin{threeparttable}
  \caption{Estimates of the CL for Gemini models}
  \label{tab:conditional-logit-results-gemini-models}
  \footnotesize
  \sisetup{
    input-symbols         = (), 
    table-align-text-post = false
  }
  \begin{tabular}{l S[table-format=-1.3] S[table-format=-1.3]}
    \toprule
    & \multicolumn{1}{c}{Gemini 2.5 Flash Preview} & \multicolumn{1}{c}{Gemini 2.5 Flash} \\
    \midrule
    \multicolumn{3}{l}{\textit{Position effects}} \\
    Row 1     & -0.071\rlap{$^{\text{\tiny *}}$}   & 0.586\rlap{$^{\text{\tiny ***}}$} \\
              & (0.031)                            & (0.042) \\[0.6ex]
    Column 1  & -0.461\rlap{$^{\text{\tiny ***}}$} & -0.138\rlap{$^{\text{\tiny *}}$} \\
              & (0.044)                            & (0.057) \\[0.6ex]
    Column 2  & -0.581\rlap{$^{\text{\tiny ***}}$} & -0.700\rlap{$^{\text{\tiny ***}}$} \\
              & (0.044)                            & (0.061) \\[0.6ex]
    Column 3  & 0.026                               & 0.160\rlap{$^{\text{\tiny **}}$} \\
              & (0.043)                            & (0.055) \\
    \midrule
    Product Fixed Effects   & \multicolumn{1}{c}{Yes} & \multicolumn{1}{c}{Yes} \\
    Observations            & \multicolumn{1}{c}{31649} & \multicolumn{1}{c}{15723} \\
    Choice Sets (Groups)    & \multicolumn{1}{c}{8166} & \multicolumn{1}{c}{3933} \\
    Pseudo R-squared        & \multicolumn{1}{c}{0.40} & \multicolumn{1}{c}{0.28} \\
    \bottomrule
  \end{tabular}
  \begin{tablenotes}[para]
    \item \tnote{*} $p < 0.05$, \tnote{**} $p < 0.01$, \tnote{***} $p < 0.001$.
  \end{tablenotes}
\end{threeparttable}
\end{table}

{\bf Implications.} 
These results underscore that upstream model updates can function as {exogenous demand shocks}—relabeling ``best sellers'' and redistributing shares even when products are unchanged. For platforms, this volatility affects the stability of ranking and endorsement signals (e.g., badges, recommendations) and argues for {model‑aware} evaluation pipelines. For sellers, the findings motivate continuous monitoring and agile listing management: content tuned to yesterday’s model may underperform after an upgrade. More broadly, because model changes alter both attribute sensitivities and position bias magnitudes, identical designs can yield different outcomes solely due to an upstream revision---reinforcing the need for ongoing auditing, disclosure of impactful updates, and diversification of strategies across agent ecosystems.

\section{Prompt Variations}
\label{app:prompt-variations}

\subsection{Prompt Variations for UI Setting}
\label{app:prompt-variations-ui-settings}
We provide the prompt variations to the default prompt (Figure \ref{fig:default-prompt-buying-agent-mock-app}) in Figures \ref{fig:ignore-position-prompt-buying-agent-computer-use} and \ref{fig:price-sensitive-prompt-buying-agent-computer-use}.

\begin{figure}[!htb]
    \centering
    \begin{tcolorbox}[sellerbuyertemplate]
You are a personal shopping assistant helping someone find a good \{{\tt product category}\}. They haven't specified particular requirements, so use your best judgment about what would work well for a typical person, and select one product to purchase. \textcolor{red}{Do not be influenced by the position of the product. Only evaluate the products based on their title/description, price, ratings, number of reviews and perhaps tags.}

\textless instructions\textgreater \\
1. Carefully examine the entire screenshot to identify all available products and their attributes. \\
2. Use the \textasciigrave add\_to\_cart\textasciigrave{} function when you are ready to buy a product. \\
3. Before making your selection, explain your reasoning for choosing this product, including what factors influenced your decision and any assumptions you made about what would be best: \\
\hspace*{.5em}- Your primary decision criteria and why you prioritized them \\
\hspace*{.5em}- How each available product performed on these criteria \\
\hspace*{.5em}- What specific factors made your chosen product superior \\
\hspace*{.5em}- Any assumptions you made about the user's needs or preferences \\
4. If information is missing or unclear in the screenshot, explicitly mention the limitation and how it influenced your decision-making. \\
\textless/instructions\textgreater
\end{tcolorbox}
    \caption{({\sf IP-UI}) Prompt for AI buyer agent with mock-app to ignore the position of the product}
    \label{fig:ignore-position-prompt-buying-agent-computer-use}
\end{figure}

\begin{figure}[!htb]
    \centering
    \begin{tcolorbox}[sellerbuyertemplate]
You are a personal shopping assistant helping someone find a good \{{\tt product category}\}. They haven't specified particular requirements, so use your best judgment about what would work well for a typical person, and select one product to purchase. \textcolor{red}{Do not be influenced by the position of the product. Only evaluate the products based on their title/description, price, ratings, number of reviews and perhaps tags.} \textcolor{blue}{The user is price sensitive.}

\textless instructions\textgreater \\
1. Carefully examine the entire screenshot to identify all available products and their attributes. \\
2. Use the \textasciigrave add\_to\_cart\textasciigrave{} function when you are ready to buy a product. \\
3. Before making your selection, explain your reasoning for choosing this product, including what factors influenced your decision and any assumptions you made about what would be best: \\
\hspace*{.5em}- Your primary decision criteria and why you prioritized them \\
\hspace*{.5em}- How each available product performed on these criteria \\
\hspace*{.5em}- What specific factors made your chosen product superior \\
\hspace*{.5em}- Any assumptions you made about the user's needs or preferences \\
4. If information is missing or unclear in the screenshot, explicitly mention the limitation and how it influenced your decision-making. \\
\textless/instructions\textgreater
\end{tcolorbox}
    \caption{({\sf IP-PP-UI}) Prompt for AI buyer agent with mock-app to ignore the position of the product and prioritize price}
    \label{fig:price-sensitive-prompt-buying-agent-computer-use}
\end{figure}

\subsection{Prompt Variations for Headless Setting}
\label{app:prompt-variations-headless-setting}
We provide the prompt variations to the default prompt (Figure \ref{fig:default-prompt-buying-agent-mcp-style}) in Figures \ref{fig:ignore-position-prompt-buying-agent-mcp-style} and \ref{fig:price-sensitive-prompt-buying-agent-mcp-style}.

\begin{figure}[!htb]
    \centering
    \begin{tcolorbox}[sellerbuyertemplate]
You are a personal shopping assistant helping someone find a good \{{\tt product category}\}. They haven't specified particular requirements, so use your best judgment about what would work well for a typical person, and select one product to purchase. \textcolor{red}{Do not be influenced by the position of the product. Only evaluate the products based on their title/description, price, ratings, number of reviews and perhaps tags.}
        
Here are the products as a JSON array:

\{{\tt json object with product attributes}\}

Respond ONLY with a JSON object in this exact format: 

\{\{ \\
    ``selected\_product\_number'': \textless number 1-{len(products)}\textgreater, \\
    ``product\_name'': ``\textless exact product title\textgreater'', \\
    ``reason'': ``\textless brief explanation for why this product was chosen\textgreater'' \\
\}\} 

Do not include any other text outside the JSON object.
\end{tcolorbox}
    \caption{({\sf IP-HL}) Prompt for AI buyer agent with headless API-style platform to \textcolor{red}{ignore the position of the product}}
    \label{fig:ignore-position-prompt-buying-agent-mcp-style}
\end{figure}

\begin{figure}[!htb]
    \centering
    \begin{tcolorbox}[sellerbuyertemplate]
You are a personal shopping assistant helping someone find a good \{{\tt product category}\}. They haven't specified particular requirements, so use your best judgment about what would work well for a typical person, and select one product to purchase. \textcolor{red}{Do not be influenced by the position of the product. Only evaluate the products based on their title/description, price, ratings, number of reviews and perhaps tags.} \textcolor{blue}{The user is price sensitive.}
        
Here are the products as a JSON array:

\{{\tt json object with product attributes}\}

Respond ONLY with a JSON object in this exact format: 

\{\{ \\
    ``selected\_product\_number'': \textless number 1-{len(products)}\textgreater, \\
    ``product\_name'': ``\textless exact product title\textgreater'', \\
    ``reason'': ``\textless brief explanation for why this product was chosen\textgreater'' \\
\}\} 

Do not include any other text outside the JSON object.
\end{tcolorbox}
    \caption{({\sf IP-PP-HL}) Prompt for AI buyer agent with headless API-style platform to \textcolor{red}{ignore the position of the product} and \textcolor{blue}{prioritize price}}
    \label{fig:price-sensitive-prompt-buying-agent-mcp-style}
\end{figure}

\end{document}